Tri Rhena Tech

Hochschule Kaiserslautern
University of Applied Sciences

UR-AI 2021

THE UPPER-RHINE ARTIFICIAL INTELLIGENCE SYMPOSIUM

# ARTIFICIAL INTELLIGENCE

## APPLICATION IN LIFE SCIENCES AND BEYOND

EDITED BY

KARL-HERBERT SCHÄFER

FRANZ QUINT

COLLECTION OF ACCEPTED PAPERS OF THE SYMPOSIUM
KAISERSLAUTERN, 27th OCTOBER 2021

# The Upper-Rhine Artificial Intelligence Symposium
# UR-AI 2021

ARTIFICIAL INTELLIGENCE - APPLICATION IN LIFE SCIENCES AND BEYOND

Karl-Herbert Schäfer, Franz Quint (eds.)

# The Upper-Rhine Artificial Intelligence Symposium
# UR-AI 2021

ARTIFICIAL INTELLIGENCE - APPLICATION IN LIFE SCIENCES AND BEYOND

**Conference Chairs**

*Karl-Herbert Schäfer*, Kaiserslautern University of Applied Sciences
*Franz Quint*, Karlsruhe University of Applied Sciences

**Program Committee**

*Andreas Christ*, Offenburg University of Applied Sciences
*Thomas Lampert*, Télécom Physique Strasbourg
*Jörg Lohscheller*, Trier University of Applied Sciences
*Enkelejda Miho*, Universities of Applied Sciences and Arts Northwestern Switzerland
*Ulrich Mescheder*, Furtwangen University of Applied Sciences
*Christoph Reich*, Furtwangen University of Applied Sciences
*Karl-Herbert Schäfer*, Kaiserslautern University of Applied Sciences
*Franz Quint*, Karlsruhe University of Applied Sciences

**Organising Committee**

*Matthias Bächle*, Kaiserslautern University of Applied Sciences
*Anna Dister*, TriRhenaTech
*Susanne Schohl*, Kaiserslautern University of Applied Sciences
*Jessica Weyer*, Kaiserslautern University of Applied Sciences

# Table of contents

# Foreword

When people talk about breakthrough technologies today, artificial intelligence (AI) is at the forefront. More and more fields are being supported by it in their scientific, economic and social challenges. The year 2021 undoubtedly highlights two areas in particular where AI is not only urgently needed, but its development is being driven forward rapidly while being viewed with some scepticism in society: healthcare and data protection.

The COVID-19 pandemic shows that our society, despite its high level of development, remains vulnerable to ancient threats. However, it also shows that humanity is capable of finding a response, albeit imperfect, in a very short time. To this, artificial intelligence provides significant contributions. It would be unthinkable to manage the analysis and processing of the huge amounts of data needed, especially in the emerging field of medical image and data analysis, to find a solution without AI. The pandemic has also made us realize that big data is not just generated as measurement data by sensor systems. Home schooling, home office, online banking, online shopping, etc. also generate a huge amount of data, which in its nature is personal. Artificial intelligence can be used to draw automated conclusions from this data that touch on privacy. This is not desirable. As such, data protection and the legal and ethical framework for using this data are becoming increasingly important.

The TriRhenaTech alliance of universities of applied sciences from the Upper Rhine region has been addressing the topic of artificial intelligence for many years. After the focus in previous years at the conferences in Offenburg and Karlsruhe was on application-oriented AI research in the industrial sector, the focus at this year's conference in Kaiserslautern is on the aforementioned current topics from health, Life sciences, data protection and beyond. The at hand conference proceeding contains the articles on the oral and selected poster presentations. We hope you enjoy reading them and we would be pleased if you find interesting approaches, worth to be considered. We encourage you to contact the authors, jointly develop the ideas presented there further and possibly incorporate them into new products.

Kaiserslautern, October 2021

*Karl-Herbert Schäfer* *Franz Quint*

# Challenges in Live Monitoring of Machine Learning Systems

Patrick Baier[1] and Stanimir Dragiev[2]

[1] Hochschule Karlsruhe – University of Applied Sciences
patrick.baier@h-ka.de
[2] Zalando Payments
stanimir.dragiev@zalando.de

**Abstract.** A machine learning (ML) system involves multiple layers of software and therefore needs monitoring to ensure a reliable operation. As opposed to traditional software services, the quality of its predictions can only be guaranteed if the data that flows into the system follows a similar distribution as the data the ML model was trained on. This poses additional requirements on monitoring. In this paper we outline a scheme for monitoring ML services based on feature distribution comparison between the data used for training and for live prediction. To showcase this we introduce payment risk prediction as an application scenario. Its long feedback delays and real time requirements motivate monitoring and at the same time holds specific challenges which we address. In this context we discuss trade-offs for the practical implementation of the monitoring scheme and share our best practices.

## 1 Introduction

Live monitoring of software systems is an important and well studied field [1] that helps to prevent unexpected service interruption and ensures stability and reliability. Monitoring typically involves collecting metrics about a system and checking if these values lie within a range of expected values. If this is not the case, alerts are triggered to warn a system operator who checks for the healthiness of the system. Example metrics are the size of the free heap memory of a software process or the CPU utilization of the machine that it is running on. While monitoring of software systems has a long history and is widely applied, the proliferation of systems that rely on machine learning (ML) models brings a new challenge to this field.

A typical ML system consumes input data and maps it to a prediction, which is used in a downstream decision engine. For instance, a fraud detection system uses payment transaction data to predict if a transaction is fraud or eligible. This prediction is then used to decide if a warning to the card holder should be triggered or even to cancel the transactions. To ensure that the predictions of the ML system are accurate, the model is tested after training and before live deployment on a held-out test set with respect to typical quality metrics such as accuracy and area under the ROC curve. In general, the prediction quality of a ML model on unseen input data is within expectation only if the input data is similar to the training data [2]. Technically, similarity means that unseen data points are drawn from the same distribution as the data used for training the model. The assessed quality measure on the test set – e.g. accuracy – cannot be promised in live operation if the live distribution and the training distribution differ. Hence, a live monitoring system is needed that periodically checks that the data which is served to the ML model in the live system is close enough to the training data distribution.

There are two main reasons why the input data distribution in a live system may differ from the training data distribution: The calculation of an input feature in some preceding system is flawed (e.g. money values are sent in euros instead of cents) or there is a natural data drift triggered by, for instance, phenomena like inflation. Without proper monitoring such data shifts can stay unnoticed. While the technical monitoring of the software stack can look perfectly fine, the quality of the predictions and hence the decisions based upon them may already suffer substantially. This happens long before the impact is measurable, resulting in big monetary damages. Thus, ML systems need an additional layer of monitoring that is concerned with checking for sane data distributions to meet the expected quality of service.

While several deployment related concerns of ML models are already tackled [3,4] the problems that arise with live monitoring of such systems are not well studied yet. In this paper we propose a ML monitoring system which is based on the experiences of running and monitoring ML systems in production environments for almost ten years. Besides giving some basic overview of the nuts and bolts of such a monitoring system, we highlight the technical challenges and trade-offs that arise with it, e.g. windowing, seasonality and computational trade-offs.

## 2 Basic Monitoring System

Before we discuss the technical challenges of ML monitoring, we start with a basic overview of the monitoring system we propose. To make this more tangible, we first introduce an example application scenario which will be used to lay out the subsequent concepts.

### 2.1 Application Scenario

Given is an online payment provider that handles the complexity of payment transactions on behalf of an online merchant. The provider strives for the best customer experience which includes seamless transaction processing and convenient payment options, e.g. credit/debit card, cash-on-delivery, etc., and most notably deferred payment. For each transaction, the provider needs to decide if a consumer can be offered a deferred payment option. For instance, if a person wants to buy shoes online, will the person have the option to pay only after receiving the shoes? In general, deferred payment options make the online buying experience closer to the offline shopping and thus increase conversion rates of online shops, but also bear the risk of not receiving the money from the customer (also known as *payment default*). To make sure that deferred payment options are offered only to the right customers, the payment provider needs to predict the risk of a payment default for every customer. Moreover, this prediction has to be finished before the customer arrives at the payment selection page of its shopping session. To tackle this problem payment providers typically employ ML systems that use features about the customer and the current shopping session. The output is a prediction of the payment default likelihood of a customer for this purchase. Such a model can be trained on historic payment data and is widely employed among online payment providers.

This scenario has two special characteristics which make monitoring especially challenging: (1) The model has to provide its prediction in *real-time*, i.e. a payment default probability has to be provided in a matter of a few hundred milliseconds. Models with such a requirement are typically running encapsulated in web services that are deployed in the cloud. (2) The scenario has a *delayed feedback loop*. This means that we can only evaluate the decision of the model after a long delay. In the above scenario, if the model predicts that deferred payment should be granted to a customer it can take weeks until the money comes in. Only then a label for the data point (customer paid or defaulted) is available and we know if the model's prediction was right. Other examples for systems with such a delayed feedback loop are order return prediction or customer churn prediction.

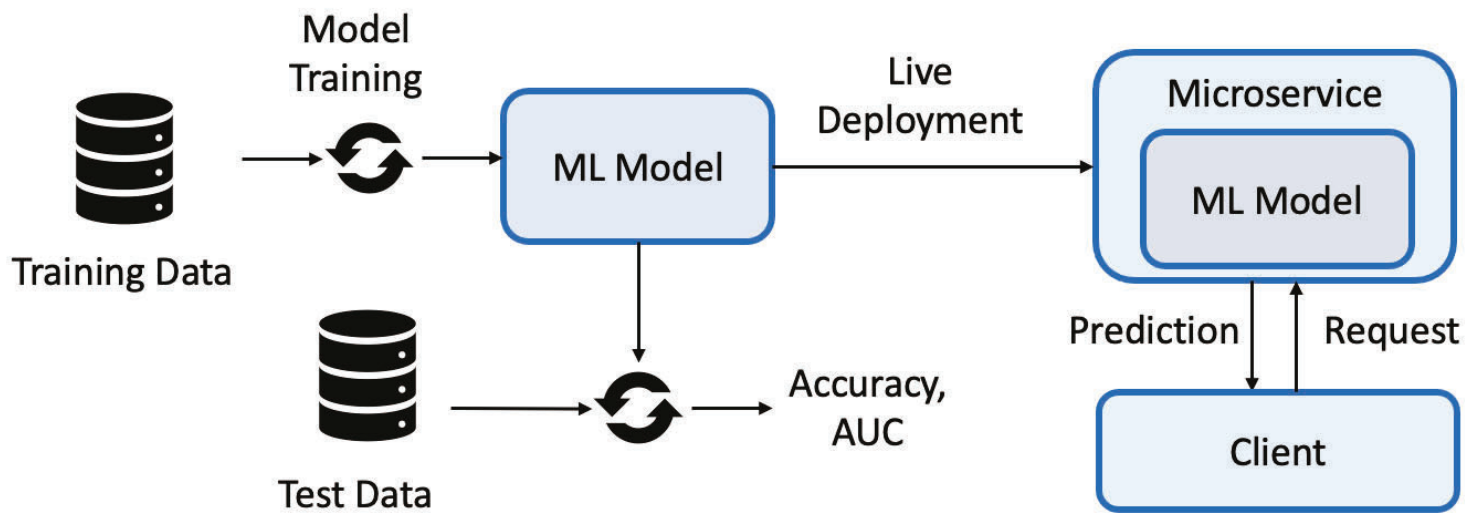

Fig. 1: Model Life Cycle

Especially the combination of these two properties highlights the importance of monitoring: We need to ensure that the model is working as expected, otherwise we may only find about it weeks later when the quality of the system's decisions can finally be assessed. In worst case, several weeks of wrong predictions may result in a complete financial fiasco for a company.

### 2.2 Model Life Cycle

To make the contextual dependencies of model monitoring visible, we shortly sketch the typical ML workflow that is preceding the monitoring (see Figure 1). What is left aside here, are all the phases that precede model training (e.g. label definition, data acquisition, feature engineering, etc).

The model life cycle starts by training the model on the training data. To check the models performance it is evaluated on a hold-out test data set and performance metrics like accuracy or area under the curve are calculated. If everything looks fine, the model is deployed to the live system where it receives requests from a client application. A request contains a data point to predict on, i.e. it contains all features that the model needs for a prediction. The model returns its output probability to the client within a few hundred milliseconds. Typically only a few requests are routed at this stage to the newly deployed model in order to do a final check on its technical readiness. After that more and more requests are gradually routed to the new model until the full traffic arrives there. That is the point when model monitoring kicks in to make sure that the model runs reliably and that performance observed on the test data can be expected under real conditions.

### 2.3 Monitoring Metrics

The most important question in monitoring is *what* exactly to monitor to ensure that the system is working as expected. In systems with a delayed feedback loop we cannot just monitor a metric like accuracy, since a feedback about the model's decision is not immediately observable. However, as a proxy we can monitor the sanity of the data that is flowing into the model, which are the input feature values.

It is widely understood that ML models only perform as expected if the data fed to the live system is similar to the data used to train the model [2]. As a result, the primary candidate metric to monitor is the difference between the data distributions of the training data and the live data that is currently flowing into the model. Both are empirical data distributions that can be compared by statistical metrics like the Wasserstein distance (also known as Earth mover's distance) or the Kullback–Leibler divergence. Such a distance score quantifies the similarity of the two distributions. Figure 2 shows an example with different empirical distributions and the corresponding Wasserstein distance. Note that for the rest of this paper, we assume that the Wasserstein distance is used for comparing distributions.

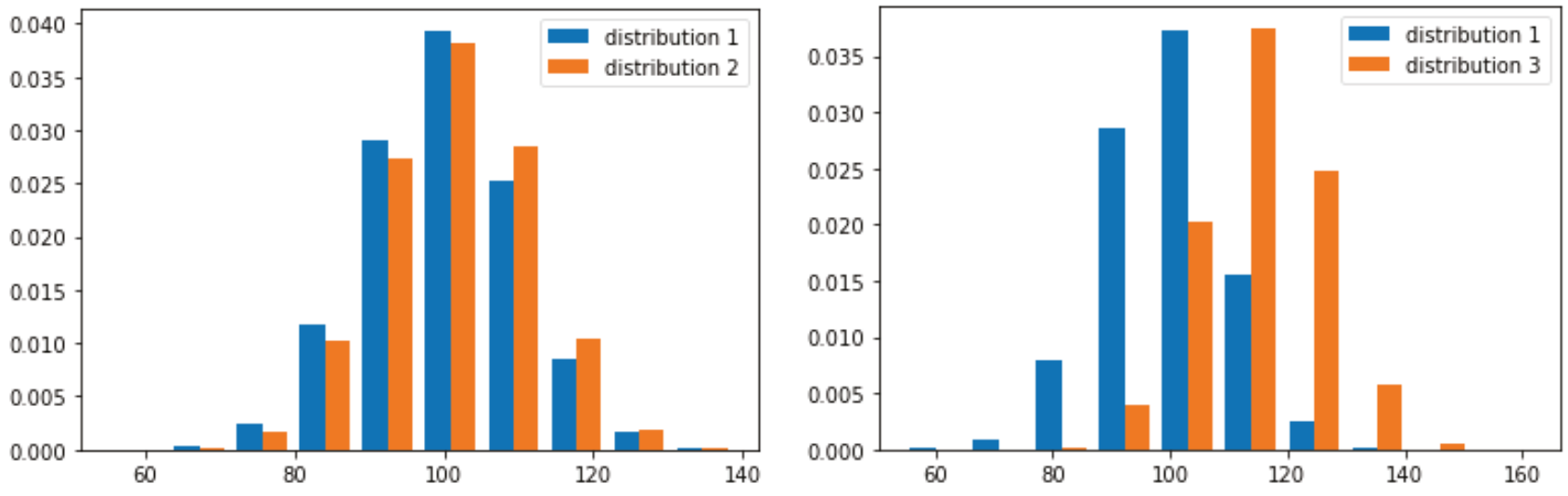

(a) Similar distributions, WD=0.9 (b) Different distributions, WD=15

Fig. 2: Wasserstein distances (WD) when comparing two empirical distributions.

### 2.4 Monitoring System

The proposed monitoring system works as follows: Every $t$ seconds the system computes for every input feature of the model its current live distribution considering the last $n$ requests that were sent to the model. For every feature, this distribution is compared to the training data distribution using the Wasserstein distance. As a result, the monitoring outputs every $t$ seconds the Wasserstein distance for every feature of the model. If one of the computed scores lies above a preselected alerting threshold the monitoring system triggers an alert to a system operator to investigates the cause for the distribution shift. We will quickly discuss possible reasons for this in the following subsection. The whole monitoring process in summarized in Figure 3 that shows an example flow for one feature.

Finding the right alerting threshold is crucial: If the chosen threshold is too low, false positive alerts are triggered, which means that the monitoring alerts even though the feature distributions did not significantly change. This can for instance happen due to some recent outlier data points in the live system. On the other hand, if the alerting threshold is too high, real distribution changes can stay unnoticed since no alerts are triggered. Problems may then not be detected fast enough and the whole purpose of monitoring becomes obsolete. A typical way to find a good threshold is too start with a rather low value and then slightly increase the threshold if the resulting alerts can safely be identified as false positives.

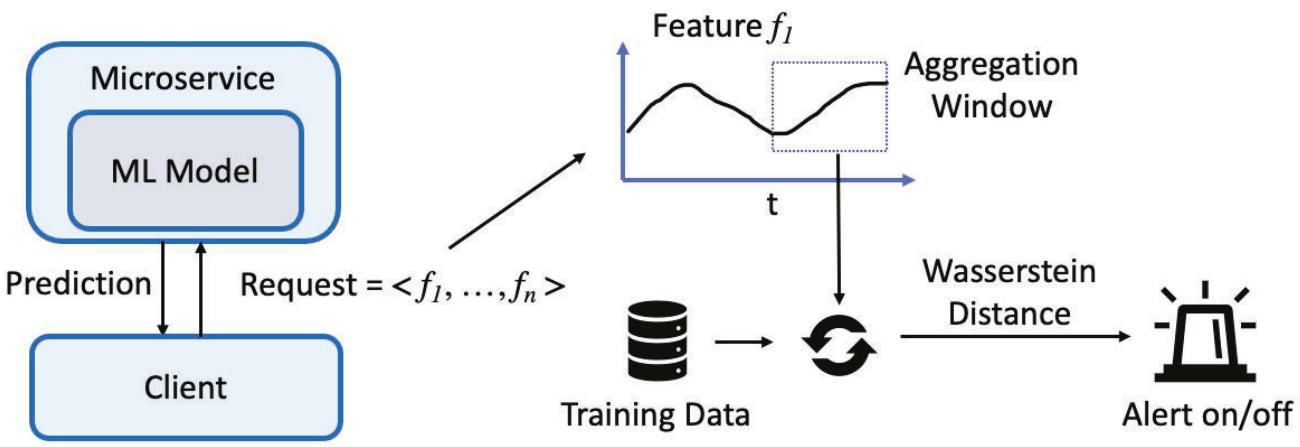

Fig. 3: Overview of monitoring process for one feature.

The description of the system above contains two parameters that must be set before the monitoring system can operate: (1) Time $t$ that determines how often a distribution comparison is triggered. (2) Window size $n$ which determines how many recent live requests are considered for calculating the live data distribution. Finding good parameter settings is again not straightforward but crucial. Both are discussed in more detail in the technical challenges in Section 3.

### 2.5 Sources of Errors

An alarm triggered by the monitoring system means that there is a change in the data distribution for at least one of the features. The cause of such a shift typically comes from one of the following sources: (1) There is a technical problem with delivering the correct feature value to the model, i.e. the client sends the wrong data. (2) The input feature data suffers from a natural distribution shift.

In the first case, the model receives wrong data from the preceding system. This can have several causes. For instance, a downtime in a database or a unstable network connection may lead to missing feature values which over time lead to a changed data distribution. Another problem could be a newly introduced bug in the preceding system that alters a feature value. For instance, money values are sent in cents instead of euros after a new software deployment. It is crucial to detect such cases since such a tiny bug can have immense effects on the output of a ML model.

In contrast, a natural shift in the input data does not stem from a technical problem but some underlying phenomena in the data itself. This can typically be observed by a gradual shift in the live data over time. The resolution to such a problem could be to analyse the data shift and remove any trend before the data goes into the model. Another alternative is to re-train the model in short time intervals to always include the freshest available data.

## 3 Technical Challenges

In this section we look into the technical challenges that are inherent to the presented monitoring system. To solve them, one has to decide for certain trade-offs that are specific to the available data and the application scenario.

### 3.1 Aggregation Window Size

One important parameter for the system is the size of the aggregation window which is used to determine the distribution of the live data. In the previous section this window size was denoted as parameter $n$. The choice of this parameter has trade-offs in both directions: If $n$ is chosen rather small, there is only very recent data in the aggregation window. This is on the one hand desirable since we prefer to build the live data distribution from relatively fresh data. For instance, consider the extreme case in which the aggregation window contains all live data seen until time $t$. If there is a change in a feature at time $t+1$, it would take a long time to be visible in the live data distribution since the aggregation window is dominated by the old data.

On the other hand, choosing a small $n$ could detect such distribution shifts very quickly but comes with problems regarding the representation of the distribution. Building the empirical distribution from only a small aggregation window suffers from uncertainty due to the small sample size. Only if we build the distribution from a large enough number of empirical data points, we can be sure to approximate the underlying distribution. Hence, if we choose $n$ too small we will derive a wrong empirical distribution and the alerting system will kick in since the distance to the training data is too big.

To show this trade-off we conducted a small experiment. For a sample feature, we created 10k training data points drawn from a Gaussian distribution with $\mu = 100$ and $\sigma^2 = 10$. Going back to the sample scenario, this feature could for instance represent the summed price of items in a customer order. To simulate live data we created a new data point at every time step which was drawn from the same distribution. In the two plots in Figure 4 at every 50 time steps the Wasserstein distance between the training and live distribution with window size $n \in \{10, 100, 1000\}$ is plotted.

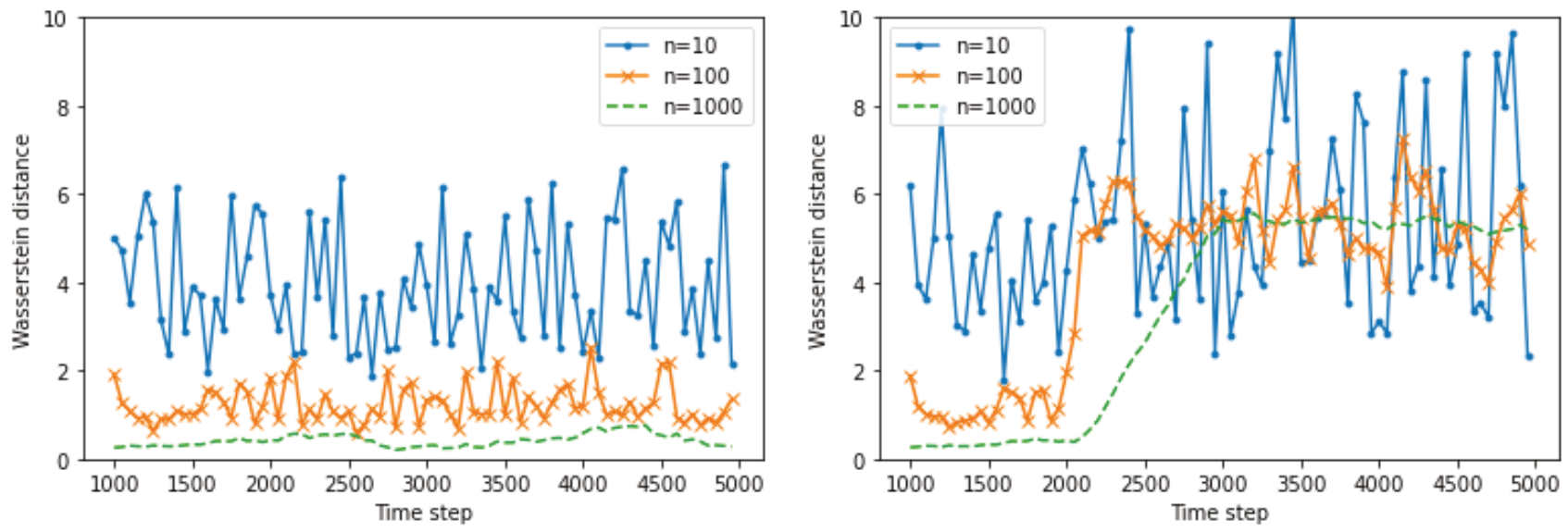

(a) No data shift (b) Data shift at $t = 2000$

Fig. 4: Effect of different window sizes on the Wasserstein distance

From Figure 4a we see that a small window size results in rather high Wasserstein distances even though the data comes from the same distribution. As discussed before, if the window size is chosen too small, the live distribution is not representative for the underlying distribution. Only if we increase $n$ to 1000, the Wasserstein distance is small enough such that the two distributions can be considered equal. Hence, we have to choose $n$ big enough to avoid false-positive alerts. In Figure 4b, we changed the distribution of the live data at time $t = 2000$ by reducing the mean of the Gaussian to 90. Here we can see that choosing a big $n$ can lead to a rather late detection of a distribution shift, which may be very costly for the business.

To find a good window size in practice, we recommend to choose $n$ as small as possible but at the same time to make sure that the Wasserstein distance stays low for data from the same distribution.

### 3.2 Seasonality

The second problem that can arise when monitoring distribution shifts is seasonality in input features. To illustrate this we assume that *the amount of payment transactions in the last hour* is a feature in the ML model of the aforementioned payment provider. The distribution of such a feature naturally fluctuates across one day. For an experiment we assumed the feature to fluctuate within one day as shown in Figure 5a. The curve in this figure shows at every time of the day the mean of all data points within the last five minutes. If we compare the live distribution of this feature against the distribution of the training data, we face the problem that the current live distribution contains a seasonal shift, while the training data is aggregated over all training data points and is static with $\mu = 1000$. As a result, the Wasserstein distance can increase significantly during the day even when using a big window size of $n = 1000$ (see the upper curve in Figure 5b).

To overcome this problem, we can limit the comparison of live data to only the set of training data that comes from the same time window. For instance, if we want to compare live vs. train distribution at 2pm, we compare the current live distributions only with training data points for which the data was also collected in that time frame, i.e. every day in the corresponding window before 2pm. In this way, the seasonality can be factored out and the distribution comparison results in a low Wasserstein scores (see the lower curve in Figure 5b).

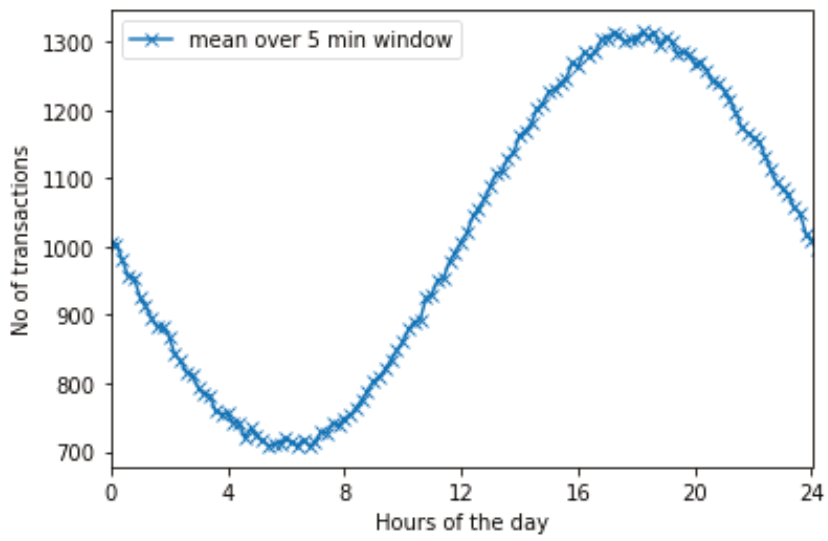

(a) Feature with seasonality

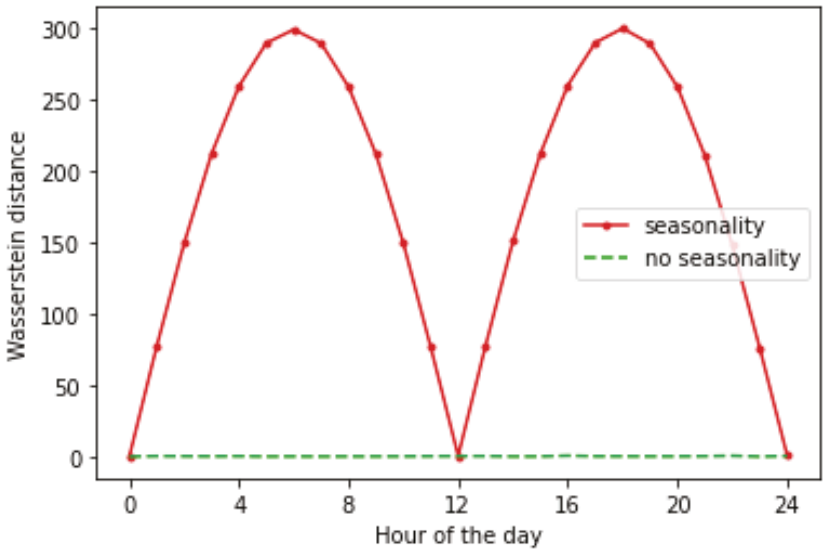

(b) Distribution Monitoring

Fig. 5: Effect of seasonality on distribution monitoring

### 3.3 Computational Aspects

One last consideration is about the computational effort that is required for model monitoring. The Wasserstein distance is computationally complex but has linear approximations [5] that are usually good enough for model monitoring.

Here we face again a trade-off: If the distribution difference is calculated frequently, computational costs increase and, hence, also increase monetary efforts for running the monitoring system. On the other hand, if the distribution difference is calculated only occasionally there can be a significant delay in detecting problems in the ML system. To find a good trade-off one has to consider the amount of traffic in the live system and determine how much computation is actually spend for calculating the Wasserstein distance. Based on this number a meaningful trade-off between computation costs and monitoring delay can be found.

## 4 Related Work

While monitoring in software systems is a field that is studied quite well, the additional complexity that ML systems introduce to monitoring is only addressed so far in few publications. Breck et al. [3] were among the first that summarized the challenges that arise when running ML models in live systems. While they list model monitoring as one important aspect, they do not go into technical details on how to implement this. Klaise et al. [6] discuss in their work aspects of monitoring and explainability of deployed models. In this context, they discuss statistics measure for detecting a drift in the live distribution but do not cover the comparison between live and training data distribution. The authors of this articles also provide an open source system that helps to automatically detect distribution drifts in the live system [7]. Finally, Paleyes et al. [8] review reports of deployed ML solutions and also shortly discuss the aspect of monitoring. However, none of the reported system has a delayed feedback loop and hence lacks a concept of distribution comparison which is necessary in this case.

## 5 Discussion

### 5.1 Monitoring adaptive systems

A question that may arise in Section 2.3 is why, in the first place, we should monitor changing feature distributions instead of building a model that can adapt to these changes? Indeed, some models can accommodate drifts. If a drift is predictable, it can be modeled explicitly. Unexpected drifts, on the other hand, can be accounted for by updating the model by online training on the new data points. This is only possible by limiting the "memory" in the training process. Training data which does not look far in the past makes a model more sensible to recent changes. The smaller the training data horizon back in time, the faster the new models will pick up the new situation. However, a short training data period poses the risk that the new model "forgets" patterns available in the old data but not in the recent data. This introduces another trade-off for how to deal with drifts and is rather an argument for monitoring than an alternative: each setting favours a particular case and imposes a risk which is better discovered sooner than later.

### 5.2 Limitations

The approach described in this paper is based on monitoring individual features to catch a change in order to maintain a promised quality of the ML system. Examples can be constructed, however, that show a distribution change in a higher-dimensional feature space while the univariate distributions of the individual features remain stable. The ML system may not generalize as well as expected to some examples from the changed distribution and the performance can degrade. With the current approach, such a change may stay unnoticed. A remedy would be to extend the distribution comparisons to pairs, triples or higher tuples of features or even to the whole feature space. However, enumerating all such tuples introduces a huge overhead for a questionable benefit: knowing that, say, 20 features collectively deviate is not very actionable in general. In special cases, though, the collective behaviour of subsets of features may well be of interest.

While monitoring distributions does a good job to uncover changes, there are a class of problems which are rooted in the opposite. A popular write-up with practical advices for ML engineers [9] features ”stale tables” as a particular pitfall which occurs more for ML systems than for others. Let us assume a table owned by another team has not been updated for a while. If a feature aggregates some counts from the table and these counts are missing for the recent past, the feature will slightly drift towards zero. For example, consider a feature that counts how often a customer visited a page in the past week. If a table with the daily visits of customers per page freezes, the aggregations will decrease. Sooner or later this will appear in the distribution comparison. However, if a feature does not slide a window, the distribution will not change. Here should be noted that monitoring feature distributions may help discover bugs opaque to traditional monitoring, however it mainly aims at detecting shifts innate to the environment.

## 6 Conclusion

The monitoring of an ML service is a prerequisite for the reliable operation and for maintaining service levels promised at the time of development of the ML models. Especially in a complex environment like payment risk prediction, without ML monitoring the enterprise is put at risk. Against the background of delayed feedback, not fully observable decision effects and non-stationary features, we outline the main traits of a monitoring system based on comparing the observed feature distributions. We discuss ways to aggregate and compare the distributions avoiding misalignment by making trade-offs needed in the practical implementation.

While monitoring ML services can discover a range of internal and external hazards, it is not a panacea, in particular it is not a substitute to the traditional software service monitoring; it is rather an augmentation to the existing well maintained monitoring practices.

# Prediction of Activators for Pathogen Sensing Receptors using Machine Learning

Pyaree Mohan Dash[1,2], Pratiti Bhadra[1], Volkhard Helms[1] , Bernd Bufe[2]

[1] Center for Bioinformatics, Saarland Informatics Campus, Saarland University, D-66041 Saarbrücken, Germany.

[2] Department of Informatics and Microsystems Technology, University of Applied Sciences Kaiserslautern, Germany.

**bernd.bufe@hs-kl.de**

**Abstract.** Formyl peptide receptors (FPRs) are G protein-coupled receptors (GPCRs) that are predominantly expressed in the immune system, where they play a critical role in detecting bacterial invasion and inflammatory responses [1] through detection of pathogen-derived formylated peptides [2]. Recent studies highlighted an involvement of FPRs in various diseases [1, 3], such as bacterial and viral infections, Alzheimer's and prion diseases, immunodeficiency, diabetes, and cancer. Given the sheer importance of FPRs, there is an immediate need for a better understanding of the mode of action of these receptors. A current challenge in FPR research is their well-documented capability to intact with an extremely vast number of structurally diverse ligands such as bacterial and virus-derived peptides, various small non-peptide molecules, and even some lipid-derivatives, that lack any obvious common structural motifs [1]. Because of the high potential of FPRs as a therapeutic target, we developed a computational method to predict FPR ligands using machine learning. Moreover, we can provide experimental evidence that our computation models are promising data mining tools that are useful tools to identify FPR activators from a vast amount of bacterial amino-acid sequence information that is contained in public databases.

The human genome encodes the three FPR genes FPR1, FPR2, and FPR3. In this study, we focused on FPR1 and FPR2. The proposed agonist prediction classifiers utilize amino-acid composition and physicochemical properties as features. Our optimized prediction models showed high test accuracy (FPR1: 82% and FPR2: 90%), Matthew's correlation coefficient (MCC) of 0.5 (FPR1) and 0.6 (FPR2), and area under the receiver operating characteristic curve (AUC-ROC) score of 0.76 (FPR1) and 0.90 (FPR2). To demonstrate the performance of the proposed prediction models in the real world, we screened the Escherichia coli K12 proteome and selected 30 novel peptides (20 predicted as activators and 10 as non-activators) for experimental validation. Human embryonic kidney (HEK-293T) cells were used to perform a cell-based calcium flux assay using Molecular Devices' Flex station. The experimental validation showed a true negative rate of 90% (9/10 non-activators) and a true positive rate of 80% (18/20 activators). Furthermore, our study also sheds light on the physio-chemical properties of FPR agonists and antagonists. A feature descriptor analysis revealed that FPR1 is activated by peptides with higher aromaticity, low hydrophobicity, low volume, and high density when compared to the peptide activators of FPR2. Moreover, the gene set annotation analysis of the predicted FPR agonists indicated that FPR1 and FPR2 activators are involved in different metabolic processes and transport systems related to bacterial stress responses. This indicates that our models can be used to mine novel information on the biological function of FPRs, which is potentially helpful for the rational design of therapeutic approaches.

# Deep Learning associated with Computational Fluid Dynamics to predict pollution concentration fields in urban areas

Xavier Jurado[1,2], Nicolas Reiminger[1], Marouane Benmoussa[1], José Vazquez[1,2], and Cédric Wemmert[2]

[1] AIR&D, Strasbourg, France
[2] University of Strasbourg, ICube Laboratory, France

**Abstract.** air quality is a worldwide major health issue, as an increasing number of people are living in densified cities. Several methods exist to monitor pollution levels in a city, either physical models or sensors. Computational Fluid Dynamics (CFD) is a popular and reliable approach to resolve locally pollutant dispersion in urban context for its capacity to consider complex phenomenon at local scale. Nevertheless, this method is computationally expensive and is not suitable for real time monitoring over large areas and city shape that evolves permanently. To overcome this issue, a deep learning model based on the MultiResUNetarchitecture have been trained to learn pollutant dispersion from precalculated computational fluid dynamics. This model has been used in situ on an area spanning $1\text{km}^2$ with real values from traffic and meteorological sensors in the surroundings of Strasbourg (France) and compared against the equivalent CFD results. Classic air quality metrics shows that the Deep Learning model manages to have satisfying results against the CFD model. The similarity index used in the study shows a 62% similarity for a result obtained in minutes against the CFD result obtained in tenth of hours.

## 1 Introduction

Air pollution is a critical worldwide health issue with about 8 million death related to it yearly, according to the World Health Organization (WHO) [1,2]. To tackle this issue, WHO provided pollution concentration values that should not be exceeded. In European Union, regulation has been enforced on the main air pollutant such as particulate matter or nitrogen dioxide [3]. To check if these values are respected, several measures have been implemented in France:

- New real estate project near pollutant sources such as heavy traffic roads, plants, or central heating system must study thoroughly air quality in the wanted area. However, these regulations are only applied at some particular timestamps and specific places.
- Sensor monitoring. But reliable sensors are expensive to acquire and maintain. For the entirety of Strasbourg city (around $80\text{km}^2$), only 4 sensors are deployed to date.
- Simulation of the annual pollution dispersion on the entire city. However, models that allow large area to be simulated may not be adapted for urban areas because of buildings not taken into account.

Among the possible models of the third point, a popular approach in the scientific community is to create airborne pollutant dispersion maps in urban areas is to use Computational Fluid Dynamics (CFD) [4,5]. It allows to accurately consider a lot of different physical phenomena from building impact on the flow to solar radiation or chemical reaction. Indeed, pollutant dispersion concentration field error can reach less than 10% when compared to experimental data [6] and about 30% when compared to real life in situ experiments [7]. Nevertheless, the counterbalance of this method is that it is computationally expensive. For instance, to cover 1km², the method roughly needs around 30 million cells and can require a week of computation to converge on 96 CPUs. Furthermore, each time the building layout changes, it would require starting new simulations again. CFD is therefore not adapted for real time simulation, despite its great accuracy and detailed description of physical phenomena.
To accelerate the computation, an innovative solution based on deep learning was developed. The idea consists in training a neural network with pre-calculated CFD simulations, to create a new air quality model that can determine pollutant dispersion in a matter of minutes over a large area. Indeed, recent advances in deep learning for spatial information treatment with convolutional based architectures have proved to be able to solve issues, notably in semantic segmentation that was impossible before. A popular model, the MultiResUNet[8], heir of UNet[9], has proved to be particularly capable at handling spatial information. This model has been trained with about 5,000 examples of CFD results of pollutant dispersion from different urban areas. The input of the model is the 3D shape of the buildings, the wind force and direction, and the position of the roads, considered as the sources of pollution.
This deep learning model is then included in a wider system that uses real time meteorological, traffic and sensor data to map the concentration field in real time on an entire urban district.

# 2 Material and method

## 2.1 CFD air quality modeling

To train the Deep Learning architecture examples of pollutant dispersion were obtained using Computational Fluid Dynamics (CFD). The software to compute the simulation is OpenFoam 5.0 which is an open source software for numerical simulations of different kind such as fluid mechanics or radiation. The approach elected here to solve the air flow is a Reynold Averaged Navier Stokes (RANS) with a k-epsilon renormalization group (RNG) [10] performing unsteady simulation. For the pollutant dispersion a transport equation coupled with the air flow is used.

The boundary conditions for the upper and lateral boundaries are symmetry conditions, the ground as a wall with a rugosity of $z_0 = 0.1m$, the building as a wall condition, the outlet as a freestream, the inlet as a logarithmic wind profile law as proposed by [11].

For the meshing, the guidelines from [12] are respected with the top and lateral boundaries situated at 5H from the closest building including with H the height the highest building. The mesh is insensitive with cells of 0.5m nearest to the buildings. The model, equations and validation have been detailed in previous published paper [13] where the same approach has been described and properly validated.

## 2.2 Deep learning network

The Deep Learning network used to learn the CFD is the MultiResUNet from [8]. This network is first designed to be applied for segmentation. In this work, it has been converted to solve pollutant dispersion from fluid mechanics. The input are the distance from the pollutant source and the height of the buildings in the area and the output is the pollutant dispersion field. The final results covers an area of $100 \times 100m^2$ by AI predictions as showed in Figure 1. The details of the MultiResUNet architecture are presented in Figure 2.

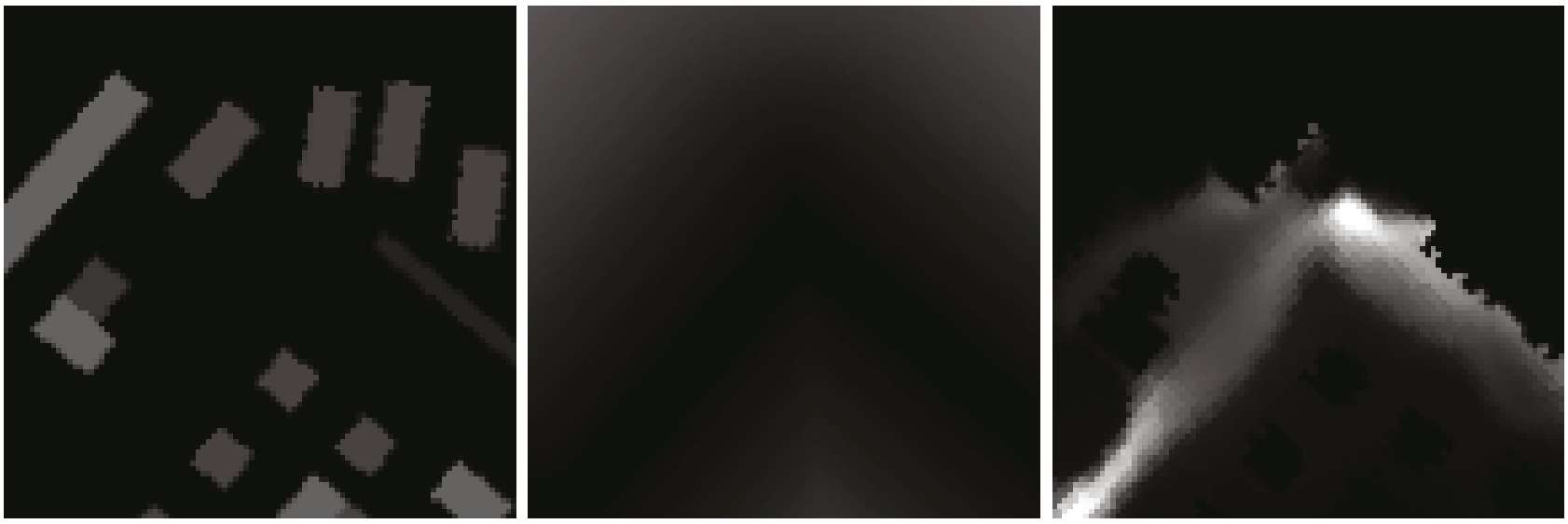

Fig. 1: Input/output images for the Deep Learning model

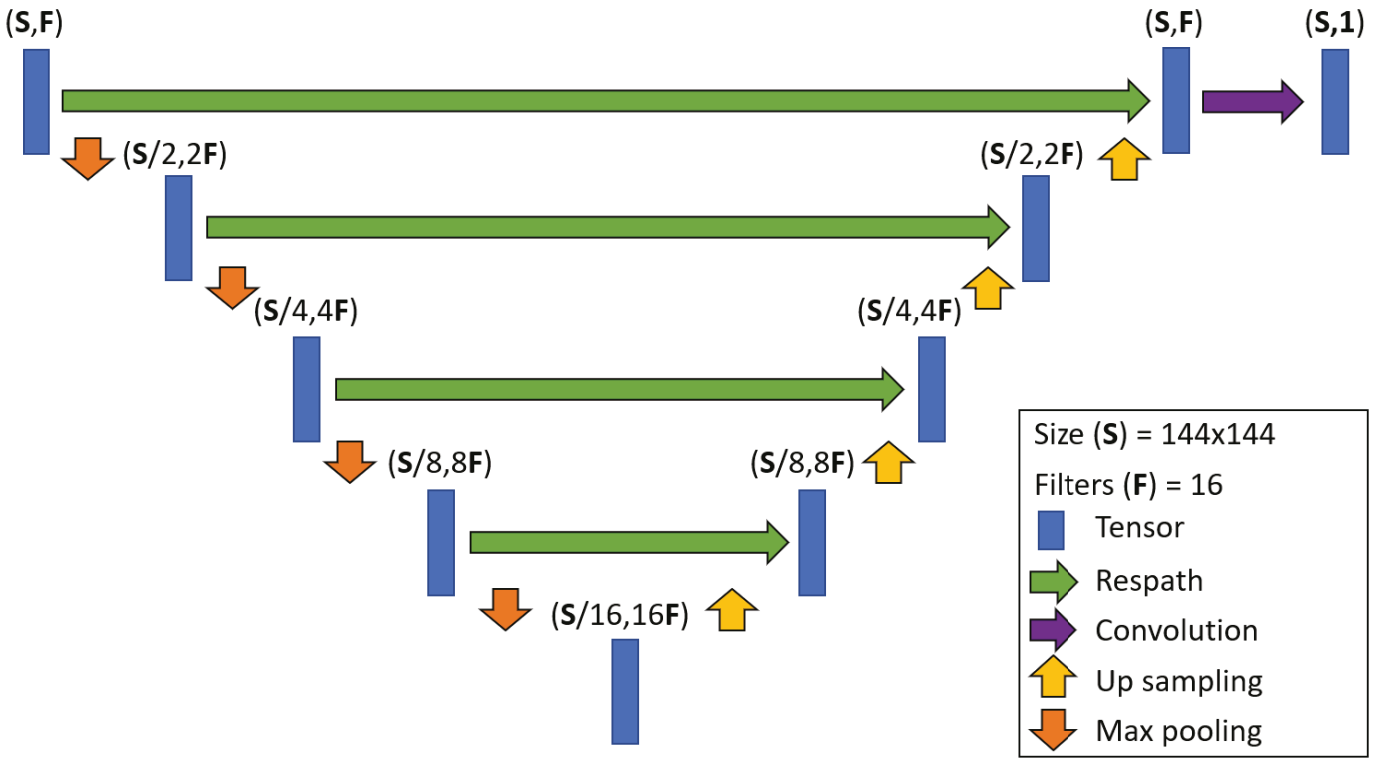

Fig. 2: Architecture details of the MultiResUNet

The loss function used is a custom loss called $J_{3D}$ and defined as followed:

$$J_{3D} = 1 - \frac{V_{pred} \bigcap V_{true}}{V_{pred} \bigcup V_{true}} \simeq 1 - \frac{min(y_i, \hat{y}_i)}{max(y_i, \hat{y}_i)} \qquad (1)$$

where $V_{pred}$ and $V_{true}$ is the volume represented by the grayscale value of respectively the ground truth and the predicted result, $y_i$ and $\hat{y}_i$ are respectively the ground truth image and the predict deep learning result.

The dataset for the training and validation are made of around 5,000 examples of different CFD simulations with varying building layouts and pollution sources. 20% are used for the validation and 80% for the training. For the test to check on the AI capability of predicting pollutant dispersion field on unseen neighborhood, it will be compared with a real neighborhood presented in Section 2.3 that will be modelled in CFD. The training was made on 25 epochs with a patience of 5 epochs on the validation data.

### 2.3 Case study

The site is located in the surrounding of Strasbourg (GPS coordinates: 48.603468, 7.743355). The building layouts of the case study is obtained thanks to the open data of the city of Strasbourg which provide digital model of the whole city (https://data.strasbourg.eu). For the test case, a real life situation is used, the first of April of 2021 at the traffic peak which happens around 08:30 AM (to have the highest concentration related to road traffic in the area). The wind speed and directions were obtained using the API openWeatherMap with a wind speed of 2m/s and a wind direction 200°N.

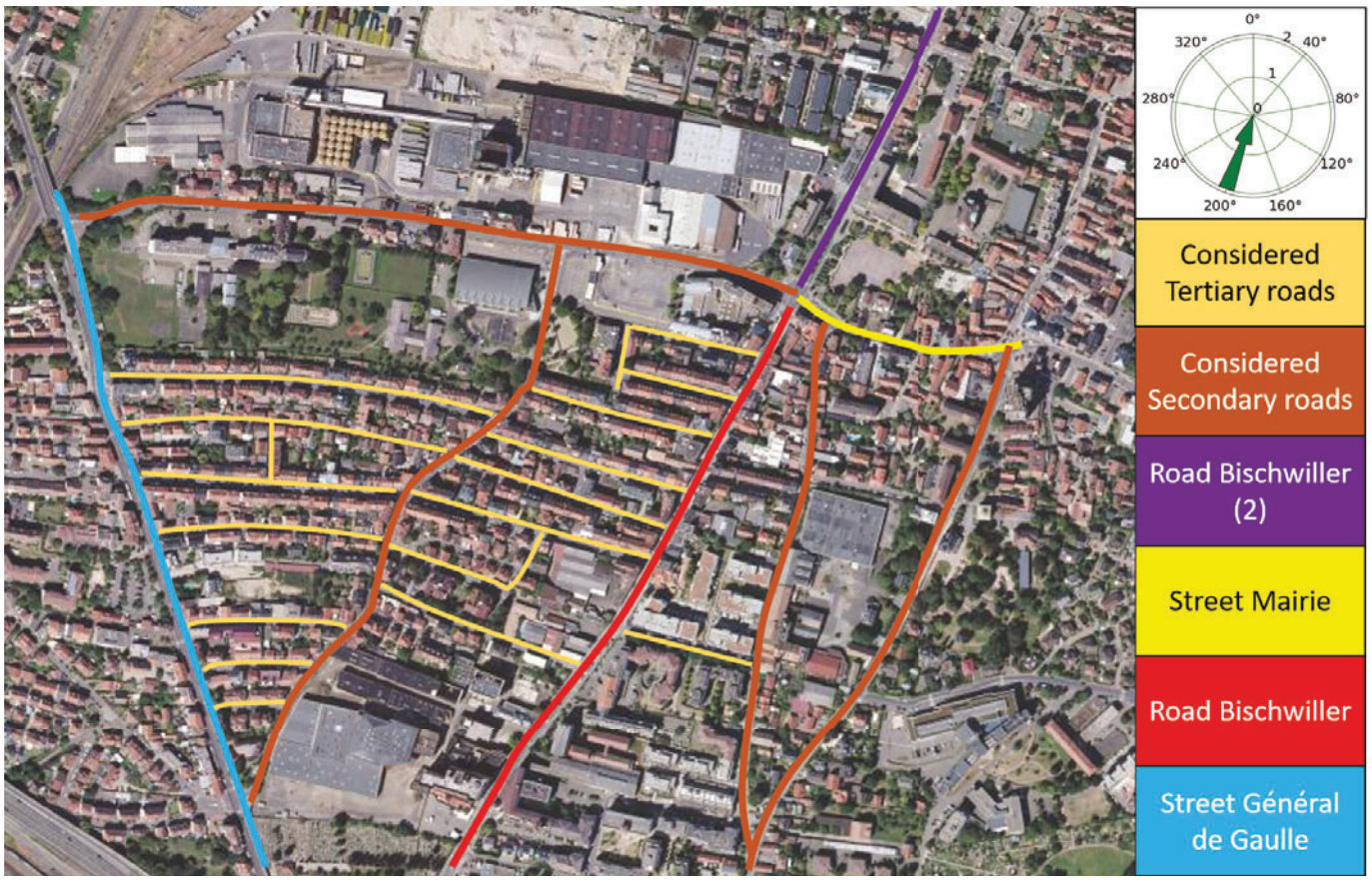

Fig. 3: Map of the Schiltigheim district with the 3 main roads used in this study

There are 27 different roads in the area. The data on traffic were obtained through the open data of the city of Strasbourg for the 4 available roads (https://data.strasbourg.eu):

– Road Bischwiller (part 1): 560 vehicles in 30 min (18.7 veh/min) with a mean velocity of 37.9km/h,
– Road Bischwiller (part 2): 784 vehicles in 30 min (26.1 veh/min) with a mean velocity of 15.5km/h,
– Street Mairie: 488 vehicles in 30 min (16.3 veh/min) with a mean velocity of 17.8km/h,
– Street General de Gaulle: 654 vehicles in 30 min (21.8 veh/min) with a mean velocity of 16.3km/h.

For other roads in the area, traffic information is lacking, thus they have been classified as secondary that will have 30% of the traffic of closest main road and tertiary that will have 5% of the closest main road. Figure 5 shows the map of the district of the study, with the three main roads and the secondary and tertiary roads. The choice of 30% and 5% is arbitrary for the sake of the example since there is no study on this traffic either with sensors or models.

Emissions are calculated based on methods proposed by the European Environment Agency (EEA) in their "EMEP/EEA Air pollutant emission inventory guidebook 2016", Tier 3 method for engine-related NOX, PM10 and PM2.5 emissions (hot and cold emissions); 2017 metropolitan fleet data found in the "OMINEA" databases provided by the Centre Interprofessionnel Technique d'Études de la Pollution Atmosphérique (share of different vehicle types, fuels and EURO standards in France).

The whole neighborhood have been modeled at once with CFD spanning an area of 1 $km^2$ made of 28 million cells. The buildings as well as the velocity magnitude field at an height of 1.5m is shown on Fig. 4.

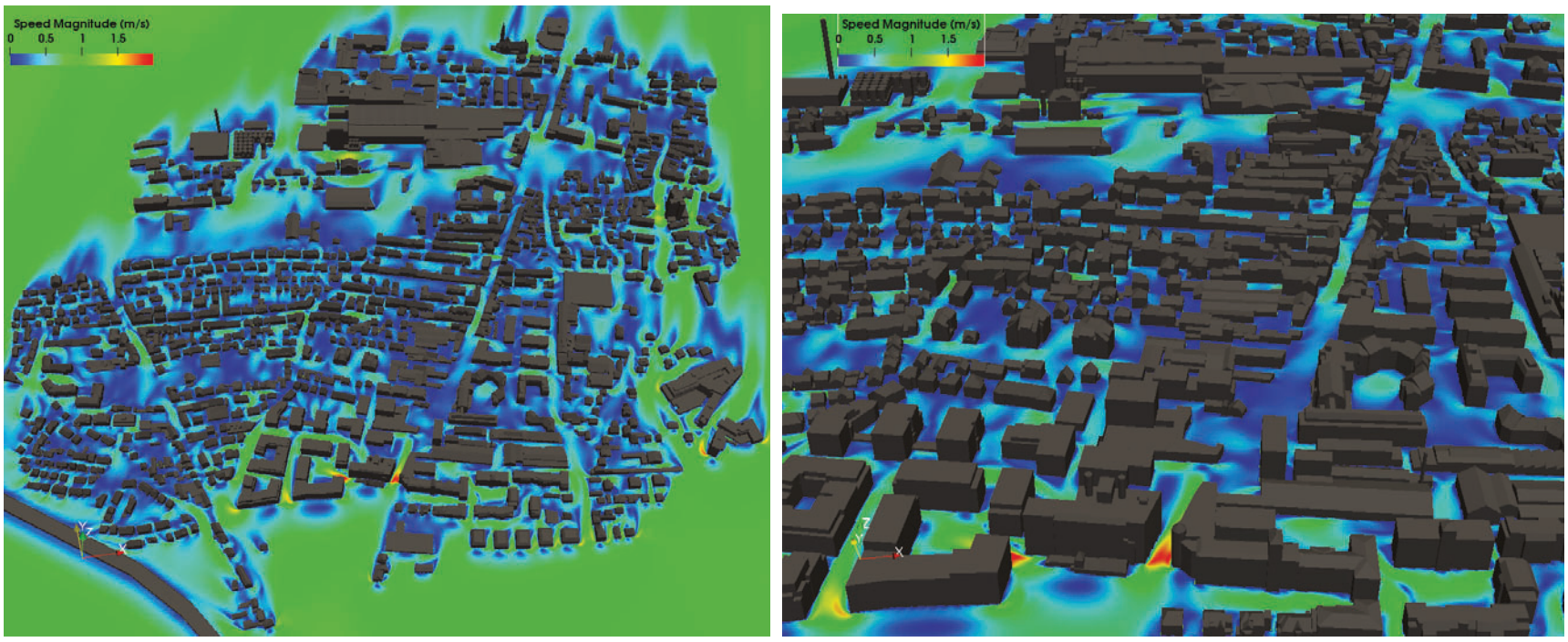

Fig. 4: Building layouts and flow field at an height of 1.5m

### 2.4 Evaluation

Seven metrics will be used, 4 from the air quality domain and three others from the computer vision. The air quality criteria have been chosen according to [14] in which the authors present several metrics with some overlapping since they evaluate the same aspect of the model. They also provides empirical threshold to consider a model as making good predictions:

- Fraction of predictions within a factor of two of observation, noted $FAC2$, a good model should respect $\simeq> 0.5$,

$$FAC2 = \text{fraction of data that satisfy } 0.5 < \frac{C_{pred}}{C_{ref}} < 2 \quad (2)$$

- Normalised Mean Squared Error, noted NMSE, a good model should respect NMSE $\simeq< 1.5$,

$$NMSE = \frac{\overline{(C_{ref} - C_{pred})^2}}{\overline{C_{pred}C_{ref}}}, \quad (3)$$

- Fraction Bias noted FB, |FB| < 0.3,

$$FB = \frac{(\overline{C_{ref}} - \overline{C_{pred}})}{0.5(\overline{C_{pred}} + \overline{C_{ref}})}, \quad (4)$$

- Correlation coefficient, noted $R$ (no threshold is given for this parameter),

$$R = \frac{\overline{(C_{ref} - \overline{C_{ref}})(C_{pred} - \overline{C_{pred}})}}{\sigma_{C_{pred}}\sigma_{C_{ref}}}, \quad (5)$$

The three other metrics are:

– $J_{3D}$

$$J_{3D} \simeq \frac{min(C_{ref}, C_{pred})}{max(C_{ref}, C_{pred})} \tag{6}$$

– Relative mean absolute error $MAE_{rel}$

$$MAE_{rel} = \frac{|C_{ref} - C_{pred}|}{\overline{C_{pred}}} \tag{7}$$

– Structural similarity $SSIM$

$$SSIM(A,B) = \frac{(2\mu_A\mu_B + c_1)(2\sigma_{AB} + c2)}{(\mu_A^2 + \mu_B^2 + c1)(\sigma_A^2 + \sigma_B^2 + c2)} \tag{8}$$

$$c_1 = (k_1L)^2 \quad c_2 = (k_2L)^2 \tag{9}$$

with $C_{pred}$ the model prediction concentration, $C_{ref}$ the reference concentration (ground truth), $\mu_A$ and $\mu_B$ are the respective average of A and B, $\sigma_A^2$ and $\sigma_B^2$ are the respective variances of A and B, $\sigma_{AB}$ is the covariance of $A$ and $B$, $L$ is the dynamic range of the pixel values and $k_1$ and $k_2$ are two constants respectively 0.01 and 0.03 (by default).

## 3 Results

To evaluate the deep learning capabilities to be applied in real life situation, a comparison has been made with real world data at the traffic at 08:30AM in the south of Schiltigheim, France the first of April 2021 between results from a CFD simulation and our deep learning approach on the $NO_x$ dispersion from traffic emissions. The results proposed respectively by the CFD and MultiResUNet for the whole neighborhood are shown on Fig.5

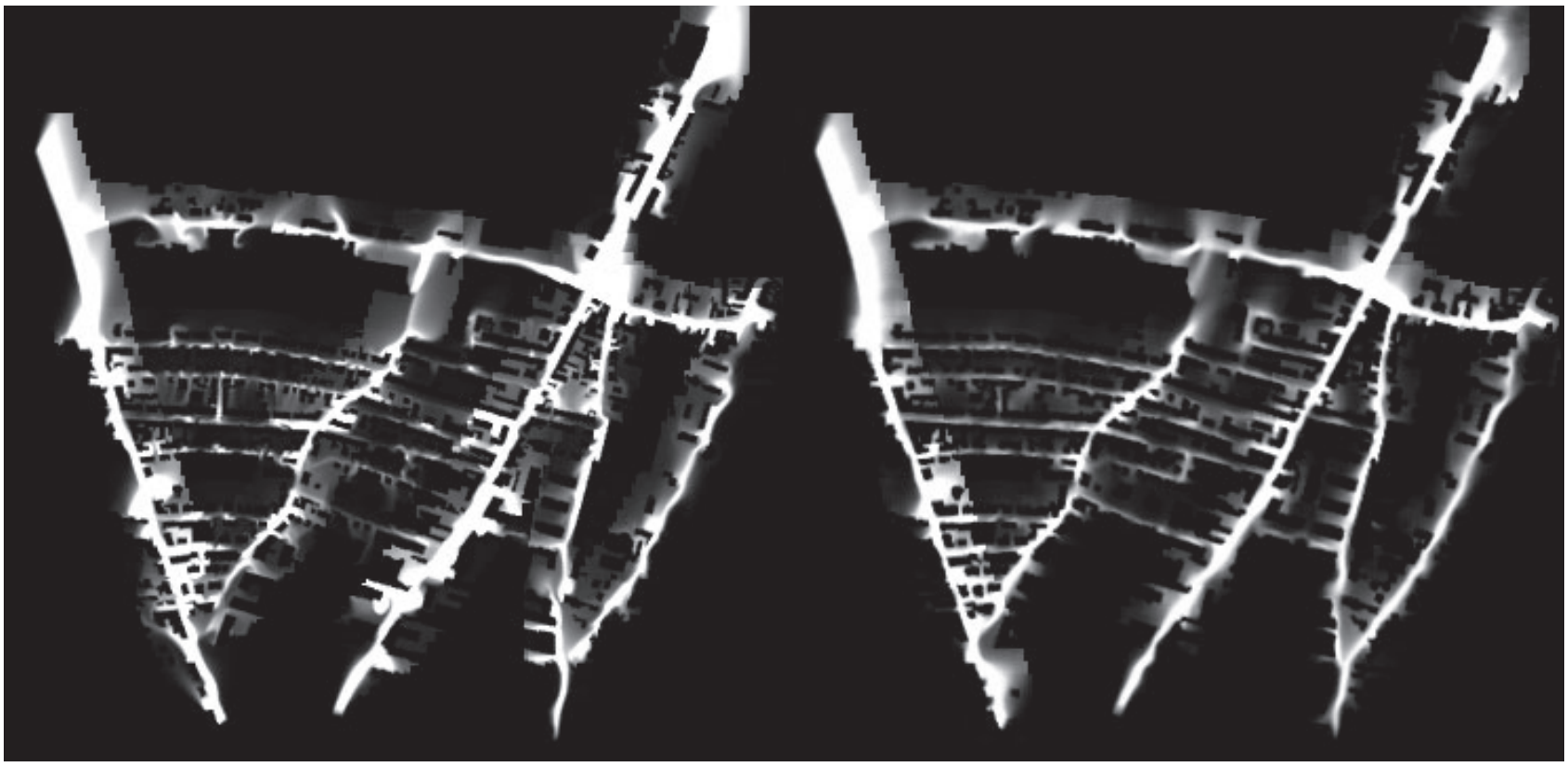

(a) CFD result (b) MultiResUNet result

Fig. 5: Maps of the studied district and comparison of the two results)

It can be tedious to compare the results between the CFD and the deep learning network since the CFD determines the dispersion in 3D while the deep learning approach works in 2D only at a given height. Nonetheless, the CFD needed one week of computation on 96 CPU while the deep learning network needed around 3 minutes on a GTX 1080Ti GPU, representing a speed up by x3000. To evaluate the accuracy of the predictions, the metrics presented above were computed between the prediction and the CFD considered as the ground truth and are presented below on Table 1.

| **Metrics** | $FAC2$ | $NMSE$ | $FB$ | $R$ | $MAE_{rel}$ | $J_{3D}$ | $SSIM$ |
|---|---|---|---|---|---|---|---|
| **Score** | **0.818** | 1.565 | **0.176** | 0.851 | 0.431 | 0.620 | 0.768 |
| **Expected values** | $> 0.5$ | $< 1.5$ | $< 0.3$ | 1 | 0 | 1 | 1 |

Table 1: Evaluation of the quality of the dispersion model given by the deep learning approach.

# 4 Conclusion

As demonstrated by our work, deep learning has proved to be able to predict results close to CFD for air pollutant dispersion. Moreover, the MultiResUNet architecture was able to compute the dispersion in a matter of minutes over a wide area against several days for the CFD. This makes the Deep Learning approach a potential model to predict in real time over large scale the pollutant dispersion from traffic related pollution.

# Improving COVID-19 CXR Detection with Synthetic Data Augmentation

Daniel Schaudt[1], Christopher Kloth[2], Christian Späte[1], Andreas Hinteregger[2], Meinrad Beer[2], and Reinhold von Schwerin[1]

[1] Technische Hochschule Ulm - Ulm University of Applied Sciences
daniel.schaudt@thu.de, spaete@mail.hs-ulm.de, reinhold.vonschwerin@thu.de
[2] Universitätsklinikum Ulm - Ulm University Medical Center
christopher.kloth@uniklinik-ulm.de, andreas.hinteregger@uni-ulm.de,
meinrad.beer@uniklinik-ulm.de

**Abstract.** Since the beginning of the COVID-19 pandemic, researchers have developed deep learning models to classify COVID-19 induced pneumonia. As with many medical imaging tasks, the quality and quantity of the available data is often limited. In this work we train a deep learning model on publicly available COVID-19 image data and evaluate the model on local hospital chest X-ray data. The data has been reviewed and labeled by two radiologists to ensure a high-quality estimation of the generalization capabilities of the model. Furthermore, we are using a Generative Adversarial Network to generate synthetic X-ray images based on this data. Our results show that using those synthetic images for data augmentation can improve the model's performance significantly. This can be a promising approach for many sparse data domains.

## 1 Introduction

The ongoing COVID-19 pandemic brings many challenges for societies all around the globe. For the healthcare sector, it is important to screen infected patients in an effective and reliable manner. This is especially true in an emergency setting, where patients already experience advanced symptoms. The prevalent test used for COVID-19 detection is the reverse transcription polymerase chain reaction (RT-PCR) [1–3]. This method has a high false negative rate and the processing requires dedicated personnel and can take hours to days [4].

Since chest X-ray (CXR) images of COVID-19 patients show typical findings including peripheral opacities and ground class patterns in the absence of pleural effusion [4, 5], they can be used as a first-line triage tool [6]. This could speed up the identification process, as CXR images are easy to obtain and rather inexpensive with a lower radiation dose than computed tomography (CT) images. Using deep learning models for detection of COVID-19 prevalence in CXR images is promising, because it eliminates the need for specialized medical staff in an emergency setting. This can further help to alleviate the challenges to the healthcare systems around the world and has the potential to save lives.

In this retrospective study, we are training a deep convolutional neural network (CNN) on the openly available COVIDx V8b dataset [7] and evaluate the model on local hospital CXR data. We specifically choose this learning framework to assess the generalization abilities of a CNN in the medical imaging context. Since high quality CXR image data is sparse, we see this as the most common use case for models in production.

Furthermore, we are using a modified version of the StyleGAN architecture [8] to generate synthetic COVID-19 positive and COVID-19 negative CXR images for data augmentation. This is done to offset some negative side effects encountered by a distributional shift between the training and testing data.

## 2 Related Work

There has been a lot of previous work on applying deep learning to CXR images to detect a COVID-19 pulmonary disease [7, 9–12]. However, most of the existing work is using publicly available CXR and COVID-19 image data. Most of those images are collected from heterogeneous sources with varying image and label quality, which raises concerns about the quality and valid evaluation of deep learning models [13, 14].

Generative Adversarial Networks (GANs) [15] have been used for many applications in the medical imaging domain [16–18]. Some studies show promising results specifically for the CXR and COVID-19 domain [19, 20]. In contrast to existing work, we integrate differentiable augmentation [21] into our GAN architecture. This enables us to train on a very small dataset and still get meaningful results.

## 3 Materials and Methods

Our goal for this work is to correctly detect a COVID-19 pulmonary disease in chest X-ray images on local university hospital study data. Therefore, we train a deep learning model on publicly available COVID-19 image data and evaluate the model based on our study data. We further enhance the amount of available training data by generating synthetic X-ray images. In this section we explain the origin and distribution of the data, as well as the deep learning model and training process.

### 3.1 Data

In this work we analyze chest X-ray images in posteroanterior (PA) and anteroposterior (AP) front view. Typically the AP view is encountered for cases where the patient is bedridden. Figure 1 shows two male patient example CXR images from our study data.

**Training data** We use two different training datasets, see Table 1. As a first step, we use the COVIDx V8b dataset [7] to train our model. This dataset is one of the biggest curated and publicly available COVID-19 CXR datasets. We use the training split of the dataset, which contains 13.794 COVID-19 *negative* and 2.158 COVID-19 *positive* frontal view X-ray images of 14.978 unique patients.

In a second step we enhance this training data by using 20.000 additional synthetic CXR images that we generated based on our study data. With that, we can add 10.000 COVID-19 *positive* and 10.000 COVID-19 *negative* images to our existing COVIDx V8b training data. This synthetic data is used to further augment the training of the classification model and increase image diversity. A sample of the generated images has been reviewed by a radiologist to ensure that the model produces meaningful data.

**Validation data** We validate the model by calculating loss metrics on the so called *test* split of the COVIDx V8B dataset. This dataset contains 200 COVID-19 positive and 200 COVID-19 negative images. We used this dataset to tune model parameters. This is to avoid overfitting our model to the testing data.

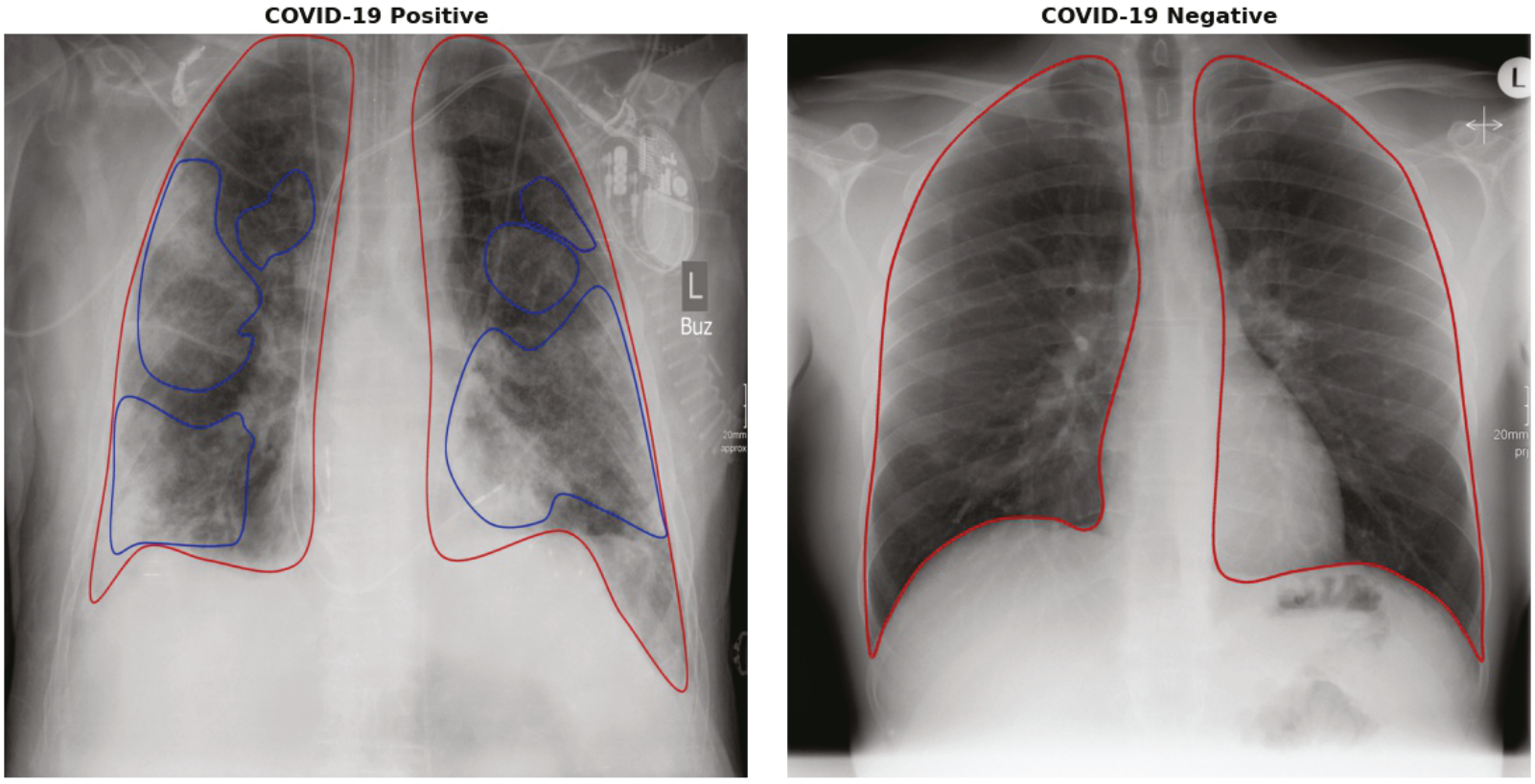

**Fig. 1.** Chest X-ray images with lungs marked in red. (Left) COVID-19 positive image, typical Ground-glass opacification marked in blue. (Right) COVID-19 negative image.

**Testing data** The central data in this work comes from a single center retrospective study of the Universitätsklinikum Ulm. For this study 566 patients (average age 51.12y +/- 18.73y; range 23-82y, 315 women) of a single institution (11/2019-05/2020) were included. The data has been carefully reviewed and labeled by two radiologists after dedicated training into *COVID-19 positive* and *COVID-19 negative*. The senior radiologist (CK) has 8 years of experience in thoracic imaging. This resulted in 110 positive images and 223 negative images, as seen in Table 1.

This testing data is used as a holdout set for final model evaluation. With this method we make sure to avoid any patient overlap between the training and testing data. Furthermore, we get a high-quality estimation of the generalization capabilities of the model when evaluating on the testing data. This is because the testing images come from a different data source, which leads to a *distributional shift*.

**Table 1.** Distribution of images for all datasets

| Dataset | Split | COVID-19 positive | COVID-19 negative |
|---|---|---|---|
| COVIDx V8B | Training | 2.158 | 13.794 |
| COVIDx V8B + Synthetic | Training | 12.158 | 23.794 |
| COVIDx V8B | Validation | 200 | 200 |
| Uniklinik Ulm Study | Test | 110 | 223 |

### 3.2 Network Architecture

For classification we use the ResNet50 architecture [22]. The network has been pretrained on the ImageNet [23] database. We replace the final fully connected layer with a linear layer of two outputs, one for each class. To get the predictions we apply a softmax activation function. Since training was very stable, we did not use any additional dropout layers or regularization methods.

The generative model is based on a modified version of StyleGAN [8]. We specifically integrate differentiable augmentation [21] into our StyleGAN architecture. This is to prevent memorization of the training data and helps to stabilize the training process. This combined architecture enables us to generate meaningful synthetic images based on our small study dataset [24]. Figure 2 shows one example image along a classification by the COVIDx+Synth model.

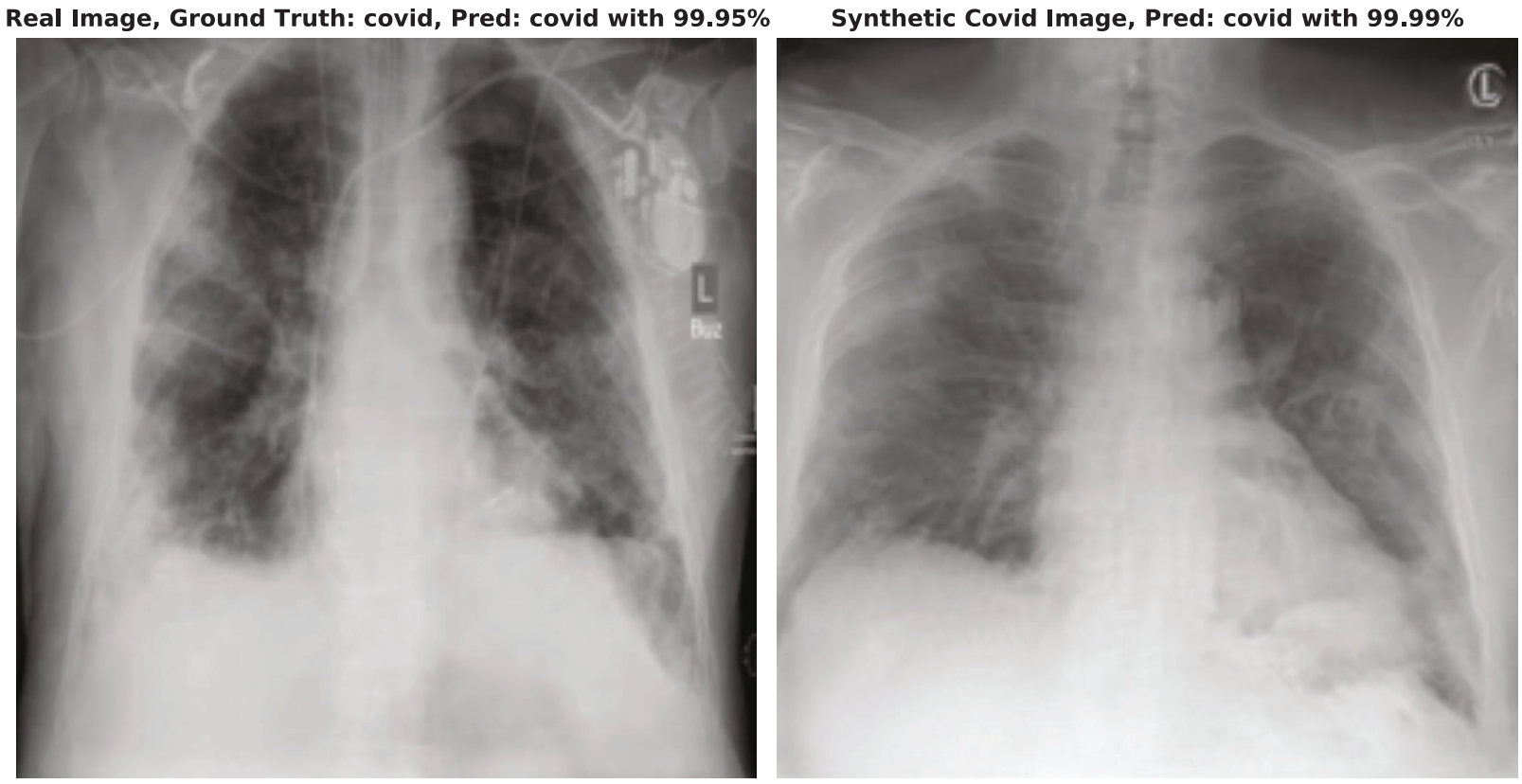

**Fig. 2.** Example images for classification by the COVIDx+Synth model. (Left) Real COVID-19 positive image, correctly predicted as *positive*. (Right) Synthetic COVID-19 positive image, correctly predicted as *positive*.

### 3.3 Training details

To train the ResNet classifier we use the Adam solver [25] with default parameters ($\beta_1 = 0.9$ and $\beta_2 = 0.999$) and a cross-entropy loss. We train the model using minibatches of size 16. We use an initial learning rate of 0.001 and apply the One-cycle learning rate scheduler [26] with a maximum learning rate of 0.006. We initially freeze all but the new last network layer for 5 epochs of training. After those 5 epochs all network parameters are trained for 30 additional epochs. The One-cycle learning rate scheduler is only applied after the initial freeze period. Increasing the amount of training showed no further improvement empirically.

All images are being scaled down to $224 \times 224$ and normalized with the mean and standard deviation of images in the COVIDx V8B dataset [7] before feeding them into the network. During training, we augment the images with random horizontal flipping and random rotation ($\pm 5°$).

Since we use two different training datasets (see Section 3.1) we get two different models: *COVIDx* and *COVIDx+Synth*. Both classification models use the exact same hyperparameters and training procedures as described. This is to evaluate the effect of using the synthetic data and make the results comparable.

For the StyleGAN generator, we train two different models: one for each class of COVID-19 positive and negative images. This is a simple method to ensure that we can generate a specific image class. For further details regarding the training process of the StyleGAN generator see Späte 2021 [24].

## 4 Results

To investigate our models in a quantitative manner, we computed the accuracy, as well as F1-score, precision and recall for each class on the validation and testing data. The metrics for the validation data are shown in Table 2. Both models perform quite well on the validation data with an accuracy of 96 % and 95.5 % respectively. The results are in line with Wang et al. 2020 [7] and their COVIDNet-CXR-2 model. Interestingly, the Covidx+Synth model falls behind the other models, despite having a lot more training data. This could be another indication of a distributional shift between the COVIDx dataset and the study data of the Universitätsklinikum Ulm.

The results for the testing data are also shown in Table 2. The table shows that a model trained on the COVIDx dataset can adapt quite well to the testing data, with an accuracy of 89.49 %. The model achieves a decent precision for COVID-19 cases (90.32 %), which is good since too many false positives would increase the burden for the healthcare system due to the need for additional PCR testing. With a rather low recall of 76.36 % the model does miss quite a lot of COVID-19 cases. This can be especially problematic in this sensitive medical setting, since false negatives lead to undetected cases of COVID-19.

This drawback can be controlled by using additional synthetic data to train the model. Table 2 shows an increase in accuracy (92.49 %) and most evaluation metrics. Especially the improved recall of 95.45 % is very desirable. This comes with the cost of a slight reduction in precision (-6.32 %). Based on those results, it can be seen that our models perform quite well, especially when incorporating the synthetic data, but there are still several areas for improvement.

**Table 2.** Evaluation metrics for models COVIDx and COVIDx+Synth on validation data (with reported metrics from Wang et al. [7] for comparison) and on testing data.

| | **Accuracy** | **F1-Score** | | **Precision** | | **Recall** | |
|---|---|---|---|---|---|---|---|
| **Model** | | C19 pos. | C19 neg. | C19 pos. | C19 neg. | C19 pos. | C19 neg. |
| *Validation Data* | | | | | | | |
| COVIDx | **0.9600** | **0.9583** | **0.9615** | **1.0000** | 0.9259 | 0.9200 | **1.0000** |
| COVIDx+Synth | 0.9550 | 0.9548 | 0.9552 | 0.9596 | 0.9505 | 0.9500 | 0.9600 |
| COVIDNet-CXR-2 [7] | - | - | - | 0.9700 | **0.9560** | **0.9550** | 0.9700 |
| *Testing Data* | | | | | | | |
| COVIDx | 0.8949 | 0.8276 | 0.9244 | **0.9032** | 0.8917 | 0.7636 | **0.9596** |
| COVIDx+Synth | **0.9249** | **0.8936** | **0.9420** | 0.8400 | **0.9760** | **0.9545** | 0.9103 |

## 5 Limitations and Discussion

In this work we showed that a deep learning model trained with a comparatively large volume of publicly available data for COVID-19 detection is able to generalize well to single source, local hospital data with patient demographics and technical parameters independent of the training data. This is not without limitations, since the distributional shift between the training and testing data can lead to some undesirable results, especially for important metrics like low recall values.

We show that this can be improved by using synthetically generated data to augment the training data. Although this works quite well, one of the reasons could be a rebalancing effect, that could have been achieved with various resampling methods as well. Another reason could be a light form of *data leakage*, since the synthetic data was generated based on the testing data. This is not fully clear, since the StyleGAN generator has no direct access to the ground truth data and just learns based on the feedback of a discriminator. Despite these concerns, the model shows promising first results and using such a model in an emergency setting could give a fast estimation for the prevalence of pulmonary infiltrates and therefore improve clinical decision-making and resource allocation.

# Detection of Driver Drowsiness by Calculating the Speed of Eye Blinking

Muhammad Fawwaz Yusri, Patrick Mangat, and Oliver Wasenmüller

Mannheim University of Applied Science, Germany
muhammadfawwaz.yusri@stud.hs-mannheim.de
p.mangat@hs-mannheim.de
o.wasenmueller@hs-mannheim.de

**Abstract.** Many road accidents are caused by drowsiness of the driver. While there are methods to detect closed eyes, it is a non-trivial task to detect the gradual process of a driver becoming drowsy. We consider a simple real-time detection system for drowsiness merely based on the eye blinking rate derived from the eye aspect ratio. For the eye detection we use HOG and a linear SVM. If the speed of the eye blinking drops below some empirically determined threshold, the system triggers an alarm, hence preventing the driver from falling into microsleep. In this paper, we extensively evaluate the minimal requirements for the proposed system. We find that this system works well if the face is directed to the camera, but it becomes less reliable once the head is tilted significantly. The results of our evaluations provide the foundation for further developments of our drowsiness detection system.

## 1 Introduction

Around 74% of European road users mostly agree that tired driving or microsleep is a frequent crash cause. The statistics were gained in 2018 by E-Survey of road users' attitudes from more than 35,000 respondents across 32 countries [1].

Thus, driver monitoring becomes of increased importance [2], since the consequence of drowsiness can be recognized distinctively during driving. This behavior can be seen as the driver slowly starts losing consciousness. Furthermore, one of the important characteristics of drowsiness is slow eye movement [3,4]. In this paper, the movement of the eyes will be the key criterion to distinguish between wakeful and drowsy drivers. We implement and evaluate a practical and simple drowsiness detection algorithm that can be easily integrated into driver-assistance systems. The system is merely based on the eye aspect ratio and eye blinking rate, where we combine Histograms of Oriented Gradients (HOG) and linear support vector machines for reliable and accurate eye detection. Upon extensive experiments, we determine a threshold for the eye blinking rate, below which our algorithm triggers an alarm. We conducted extensive evaluations based various test cases, which challenge our system. While our drowsiness detection algorithm works in principle, we identify circumstances, in which our system is less accurate. In this way we systematically elaborate the next steps to further improve our simple drowsiness detection system.

## 2 Related Work

There are several methods to detect the features of the eyes as well as drowsiness. For instance, some of the researches apply Viola-Jones cascade classifier to differentiate the eyes from other facial parts [5,6]. By determining the number of pixels on the iris, cornea and eyelid, the number of blinking and duration of the closed eyes can be calculated. While comparing the number of blinking with the blink rate set (normal = 8-10 blinks per minute, sleepy = 4-6 blinks per minute), drowsiness can be identified [5]. Islam et al. [6] calculate the eye aspect ratio to determine the eye closure time and total blink per minute. These values can then be compared to appropriate thresholds and an alarm is activated if the value exceeds or falls below the corresponding thresholds (depending on the critical variable to consider).

Picot et al. [7] developed a more advanced method to determine the drowsiness state of a driver. Their idea is analogous to the use of electro-oculograms (EOG) [8], where electrodes are placed near the eyes and the voltage signals are measured. They record visual signs from 60 hours of driving from different drivers. Based on data-mining techniques their algorithm then identifies patterns of drowsiness. Moreover, another similar approach uses a head gear to record the pupils on a driving simulation [9]. The algorithm computes the vertical length of dark pupils and is able to detect drowsiness from this variable.

A development in drowsiness detection uses binarization in combination with image filters. The system proposed by Ueno et al. [10] is able to detect the vertical position of the eyes. Therein the algorithm takes into consideration the size of the eyes to calculate the ratio of opened and closed eyes.

The drowsiness of the driver is addressed by detecting the state of the eyes when they are closed for a certain period of time [11,12,13]. Therein, the relevant parts of the eyes are detected using Haar-Cascade classifiers [11,12]. This approach seems particularly suitable to be easily integrated into a driver-assistance system, since it is merely based on eye detection. However, it only detects whether the driver has already fallen into the microsleep state, which may be too late for the successful prevention of road accidents.

The goal of our work is to set up a comparably simple real-time drowsiness detection system with minimal requirements and to challenge it in an extensive evaluation by executing various test cases. In our work we follow Haq et al. [5] and use the eye blinking rate to decide if a driver becomes drowsy. However, we determine the eye blinking rate differently. We measure the eye aspect ratio (used by Islam et al. [6]) and derive the eye blinking speed from it. In order to set the threshold for the eye blinking rate below which the driver is considered to be drowsy, we follow Picot et al. [7] and Hayami et al. [9] by simulating a scenario that imitates sleepiness of a driver. The threshold will also be dependent on the individual size of the eyes [10]. After fixing the threshold experimentally, we challenge our drowsiness detection method in a series of test cases (partly inspired by Suhaiman et al. [13]). While the proposed method works in principle, our extensive evaluation reveals a reduction of the reliability in certain scenarios (e.g. tilting the head by larger angles). In this way we can systematically elaborate the next research steps to increase the accuracy of our simple drowsiness detection system in a broader range of scenarios.

## 3 Methods

The drowsiness of a driver can be anticipated by analyzing the movement of the eyelids. The eyelids move slower than a normal blink. In this paper, we implement an algorithm

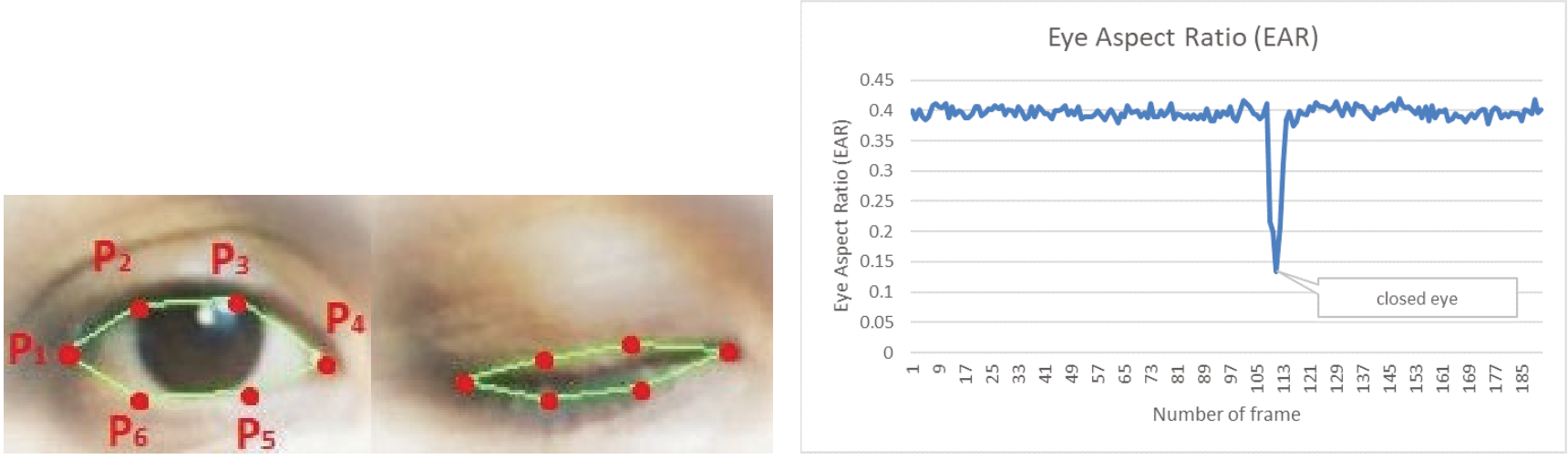

(a) 6 points eye landmarks

(b) EAR over time for an eye blinking around frame 110

Fig. 1: Comparison between opened and closed eye [18]

that allows us to determine the speed of the blinking eye. Moreover, we use Histogram of Oriented Gradients (HOG) and a Linear Support Vector Machine (SVM) method to improve eye detection (4.89% higher accuracy compared to Haar-Cascade, see Rahmad et al. [14]).

### 3.1 Eyes Detection

The algorithm is trained to detect the landmarks of the facial features in the dlib library by using an ensemble of regression tress [15]. HOG image descriptor and SVM are the method for the process training of an object [16]. There are many datasets available to detect these landmarks and we are using the dataset from IBUG which has 68 points of facial landmarks [17].

### 3.2 Eye Blinking Speed

Having detected the eyes of the driver, the next step is to determine the eye blinking speed. Firstly, we have to detect whether the state of the eye is opened or closed. A suitable measure to derive the state of the eye is the eye aspect ratio (EAR). We follow the definition of Soukupová et al. [18]:

$$\mathrm{EAR} = \frac{|p_2 - p_6| + |p_3 - p_5|}{2|p_1 - p_4|}\ , \tag{1}$$

where $p_1$ to $p_6$ are the facial landmarks as depicted in Figure 1a. When the eye is opened, the EAR is above 0.35, but when it is closed, the value rapidly dropped below 0.15 (see Figure 1b and Section 4 for the corresponding experiments).

Based on the flow diagram in Figure 2, firstly, the algorithm will find the average eye size (AES) of the driver defined by

$$\text{Average Eye Size (AES)} = \frac{\text{Max EAR1} + \text{Max EAR2} + \text{Max EAR3}}{3}\ , \tag{2}$$

i.e. we take the arithmetic mean of three measured maximum EARs. (Notice that the accuracy of the AES can be improved by measuring more maximum EARs, but this is at the expense of a higher computational effort.)

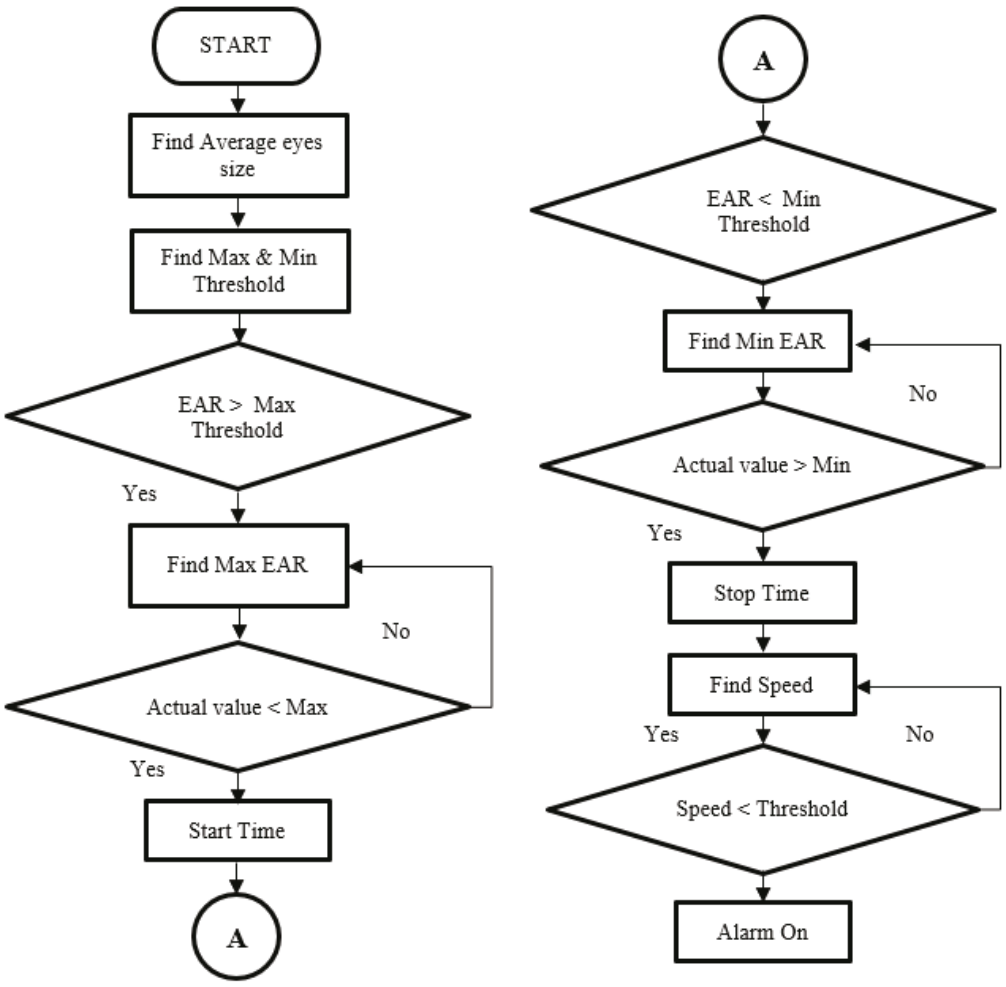

Fig. 2: Flow diagram to determine the eye blinking speed

Afterwards, Max and Min Threshold will be calculated based on the average eye's size value. Max and Min Threshold are defined by

$$\text{Max Threshold} = \frac{2}{3}\text{AES} + 0.0467 \ , \tag{3}$$

$$\text{Min Threshold} = \text{Max Threshold} - 0.05 \ . \tag{4}$$

The numerical values in these equations were found empirically. After Max and Min Threshold have been determined, the algorithm will search for the maximum value of the EAR (denoted by Max EAR) while capturing the images frame by frame. When the current EAR is less than the current maximum value, it will start the timer and at the same time find the minimum value of the EAR (denoted by Min EAR). The final minimum value is determined, when the current EAR eventually becomes larger than the minimum value, thus the timer will stop. The blinking speed for each blink can be calculated by

$$\text{Blinking Speed} = \frac{\text{Max EAR} - \text{Min EAR}}{\text{Start Time} - \text{Stop Time}} \ . \tag{5}$$

If the blinking speed becomes sufficiently low, the algorithm will activate an alarm system. For this purpose we introduce an empirically determined drowsiness threshold. Whenever the eye blinking speed is below this drowsiness threshold, the algorithm identifies the driver as being in a drowsy state.

## 4 Evaluation

For the evaluation of our drowsiness detection system we proceed as follows. We first show experimentally that changing the distance between eyes and camera leaves the EAR invariant. Then, we determine the speed of the eye blinking in the wakeful and sleepy state, respectively. Finally, we evaluate the impact of changes in the head positions on our system.

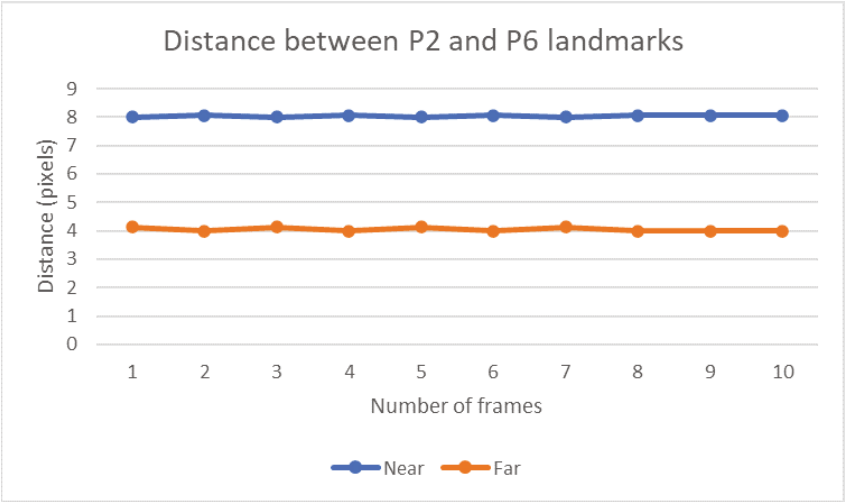

(a) The distance between $p_2$ and $p_6$

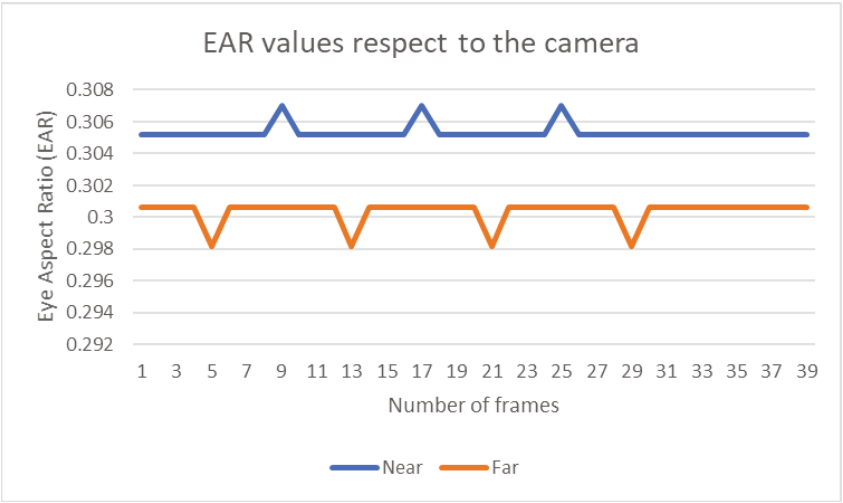

(b) EAR from both images

Fig. 3: Result for constant EAR's evaluation

### 4.1 Evaluation of the Impact of Distance Variations on the EAR

When alternating the distance between eyes and camera, the apparent size of the eyes will change. However, based on eq. (1) we expect the EAR to remain invariant.

The following evaluation shows experimentally that the EAR is indeed invariant under modification of the distance of camera and eyes. For this purpose, two similar images with different sizes are used instead of a live stream video. The reason is to have a fixed EAR reference from a static image, enabling a comparison of the EAR from both images that have different eye's sizes. The sizes of the eyes in both images are relatively different because one image is close to the camera and the other is far from it. The size can be measured by calculating the difference distance between two points of facial landmarks (e.g. $p_2$ and $p_6$ in Figure 3a, which are the upper and lower eyelid) in both images.

The distance between these points in Image 1 (Near, blue line in Figure 3a and Figure 3b) is two times bigger than Image 2 (Far, orange line in Figure 3a and Figure 3b) which is approximately 8 and 4 pixels (see Figure 3a), respectively. It shows that the position of the eyes in Image 1 is nearer to the camera than Image 2.

The ratio between both distances is approximately 1:2. The result in Figure 3b shows that the measured EARs for Images 1 and 2 are approximately 0.3052 and 0.3006, respectively. The deviation is only 1.5%, which is acceptable for our purposes.

### 4.2 Evaluation for Normal Blinking

The purpose of this evaluation is to check whether Max EAR, Min EAR, and hence, the blinking speed in eq. (5) are calculated correctly. Moreover, the average blinking speed in the wakeful state can be determined from this test.

The participants were asked to blink normally for 8 times. The first three blinks were analyzed by the algorithm to obtain the average size of the eyes. The other five blinks were necessary to evaluate the EAR and the speed of the blinking. Three tests from three participants were conducted thoroughly and the results are as follows.

**Participant 1:** Figure 4a shows that the maximum of the EAR, which is the highest value of the EAR before the eyes start to close, is calculated correctly above the Max Threshold (defined in (3)). Moreover, we see that there is only one data point below the minimum threshold (orange line). The minimum threshold can be determined using (4). We experimentally obtain Max EAR = 0.4572 and Min EAR = 0.1098. Figure 4b shows that the values of the blinking speed are above 1 pixel/second and the average speed is 2.0418 pixel/second.

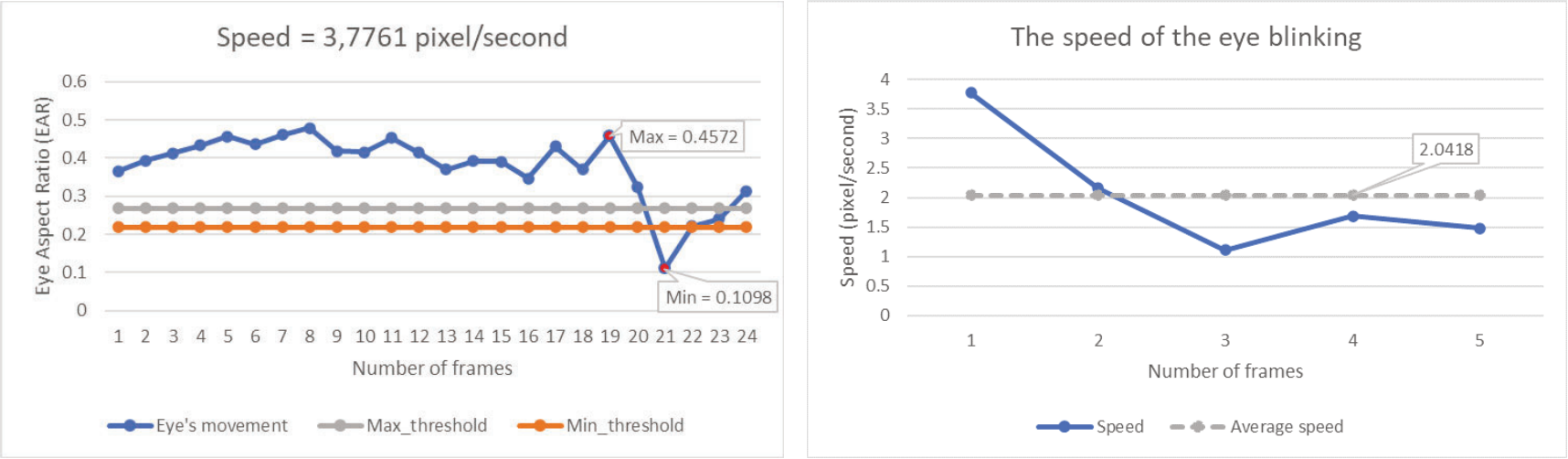

(a) Speed of normal blinking (b) Speed values from five blinks

Fig. 4: Result from Participant 1 (normal blink)

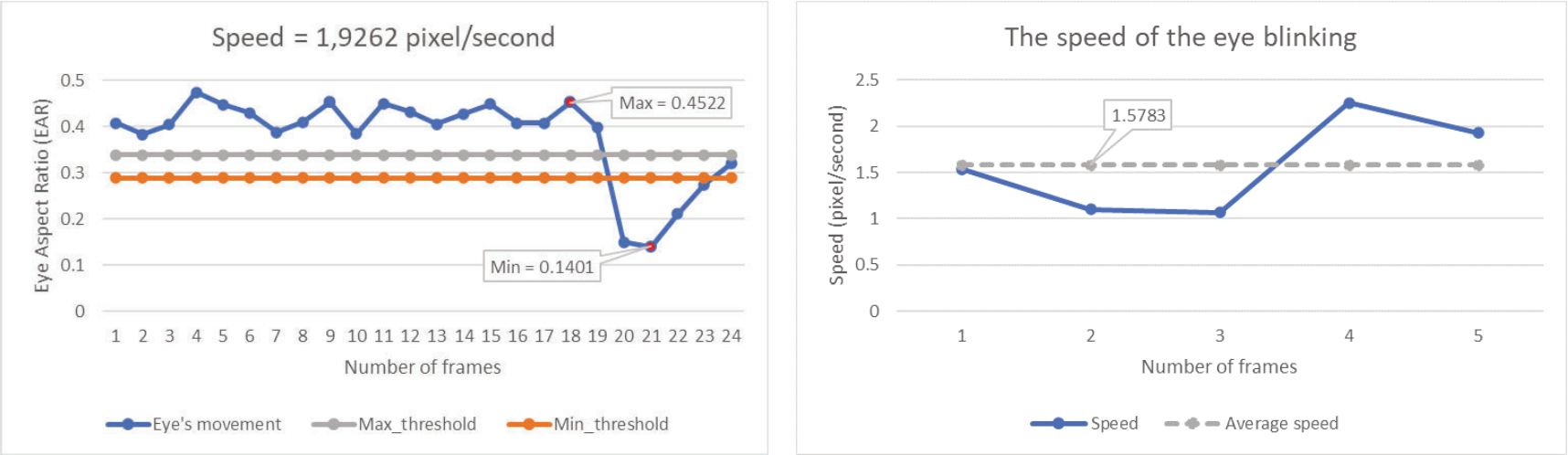

(a) Normal blinking's speed (b) Speed values from five blinks

Fig. 5: Result from Participant 2 (normal blink)

**Participant 2:** Figure 5a shows that Max EAR = 0.4522. However, there are two data points below the minimum threshold, whose difference in the EAR is 0.01 pixels. The algorithm needs to determine which value to choose as a minimum. The final minimum value is Min EAR = 0.1401 which is the correct value because this is the value when the eyes are completely closed. If the differences between the values below minimum threshold is less than 0.01, then the final minimum will be the first value because this is the point where the eye is completely shut.
We can see that Figure 5b is similar to Figure 4b where the speed values are above 1 pixel/second and the average speed is 1.5783 pixel/second.

**Participant 3:** In this case the EAR reaches a rather flat maximum at Max EAR = 0.3558 before dropping sharply below the minimum threshold (see Figure 6a). The minimum EAR is measured to be Min EAR = 0.1049. Again, the blinking speed never drops below 1 pixel/second (see Figure 6b). The average eye blinking rate is approximately at 2 pixels/second.

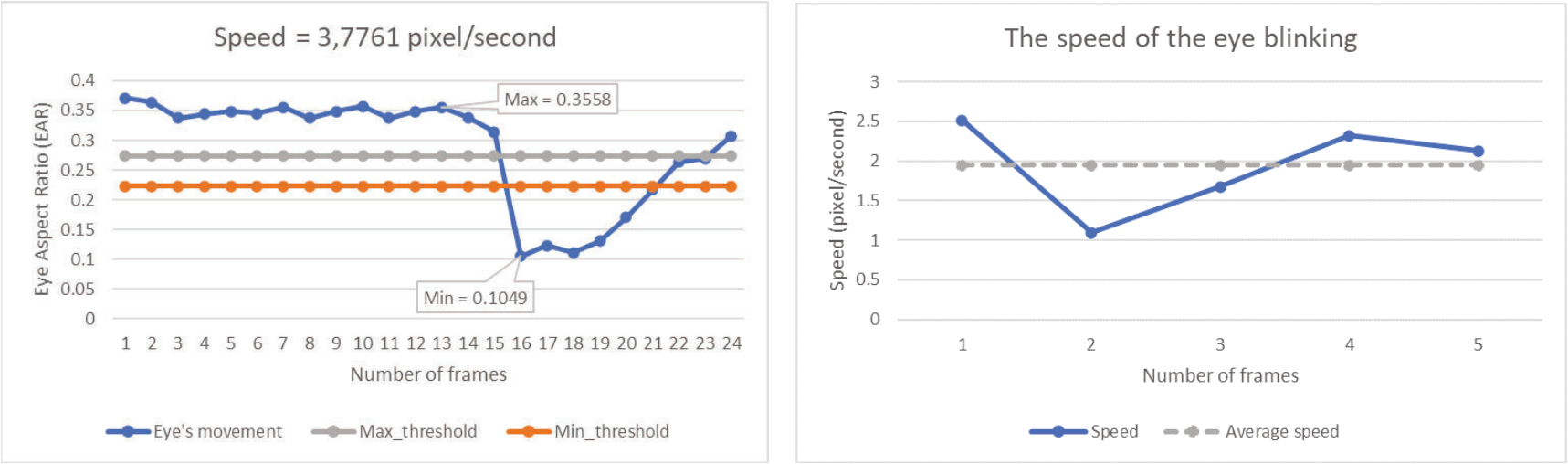

(a) Normal blinking's speed (b) Speed values from five blinks

Fig. 6: Result from Participant 3 (normal blink)

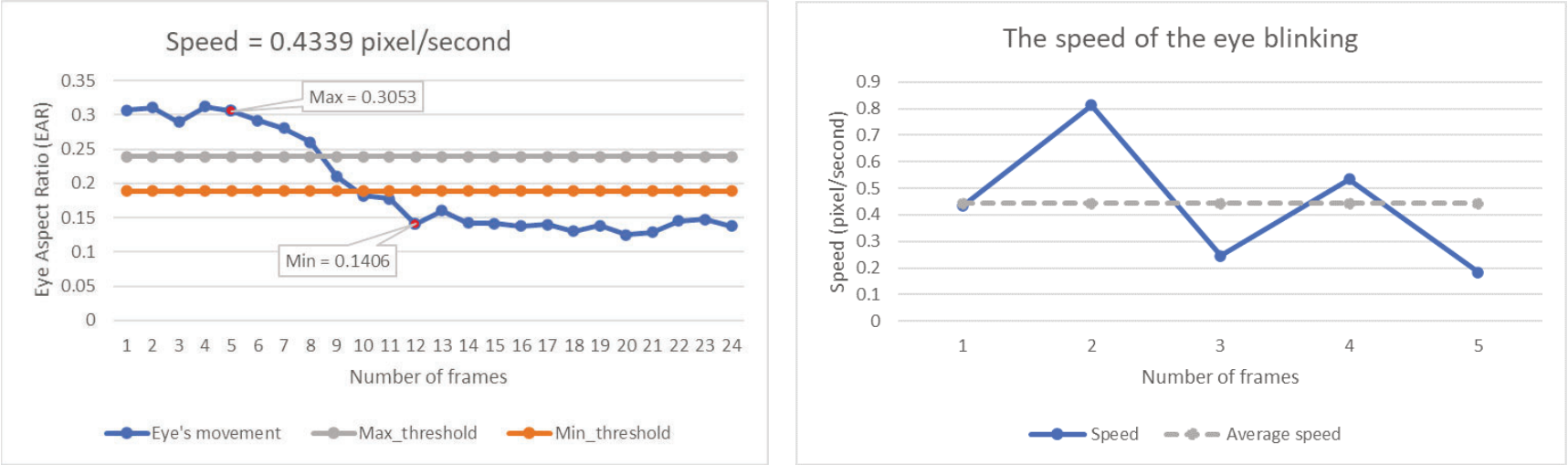

(a) Sleepy blinking's speed (b) Speed values from five blinks

Fig. 7: Result from Participant 1 (sleepy blink)

To summarize the results of the three tests, the average speed for normal blinking has a threshold value of 1 pixel/second. However, there are certain cases where the eye blinking speed can slightly drop below this threshold.

### 4.3 Evaluation for Sleepy Blinking

For the measurement of the EARs and the eye blinking rate in the drowsy state, we essentially repeat the previous experiments. The evaluation was conducted by testing three participants with 5 trials. They were asked to imitate the behavior of a sleepy driver in front of the camera by closing their eyes slowly. The corresponding data are shown and discussed below.

**Participant 1:** In Figure 7a it can be seen that the number of frames from Max EAR = 0.3053 down to Min EAR = 0.1406 is larger than in the wakeful state, indicating a lower eye blinking rate. Indeed, Figure 7b shows the smaller speed values for all five blinks. The average speed is 0.4421 pixel/second.

**Participant 2:** Figure 8a has a slightly different result from the first test (see Figure 7a), specifically regarding the values below the minimum threshold. It can be seen that the EAR remains below the minimum threshold after Min EAR = 0.1056 has been reached.

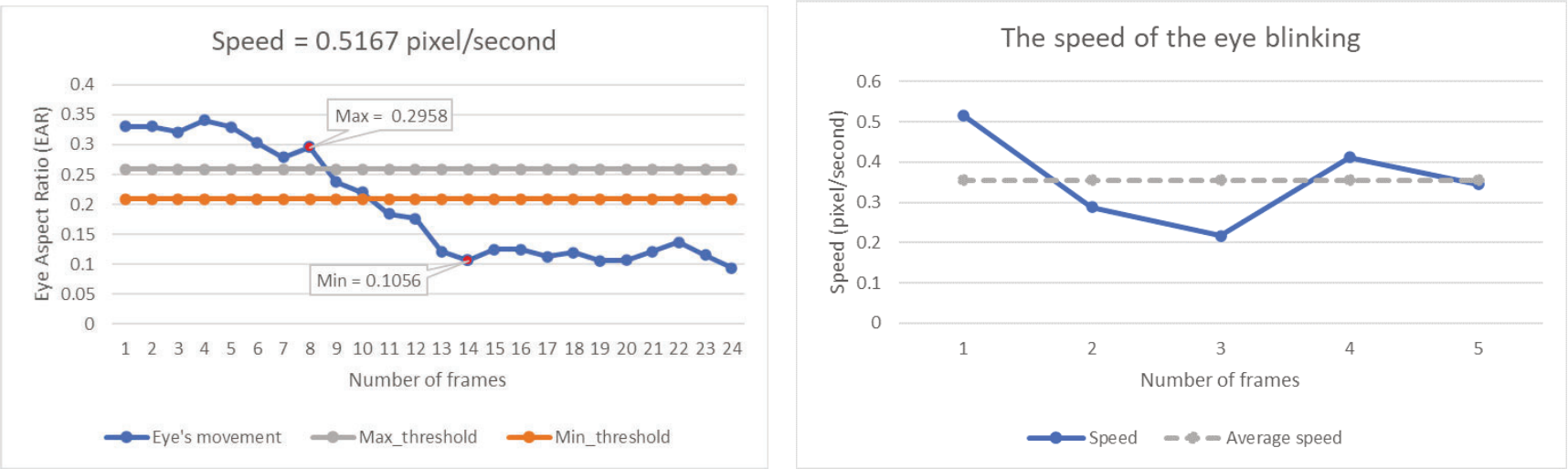

(a) Sleepy blinking's speed (b) Speed values from five blinks

Fig. 8: Result from Participant 2 (sleepy blink)

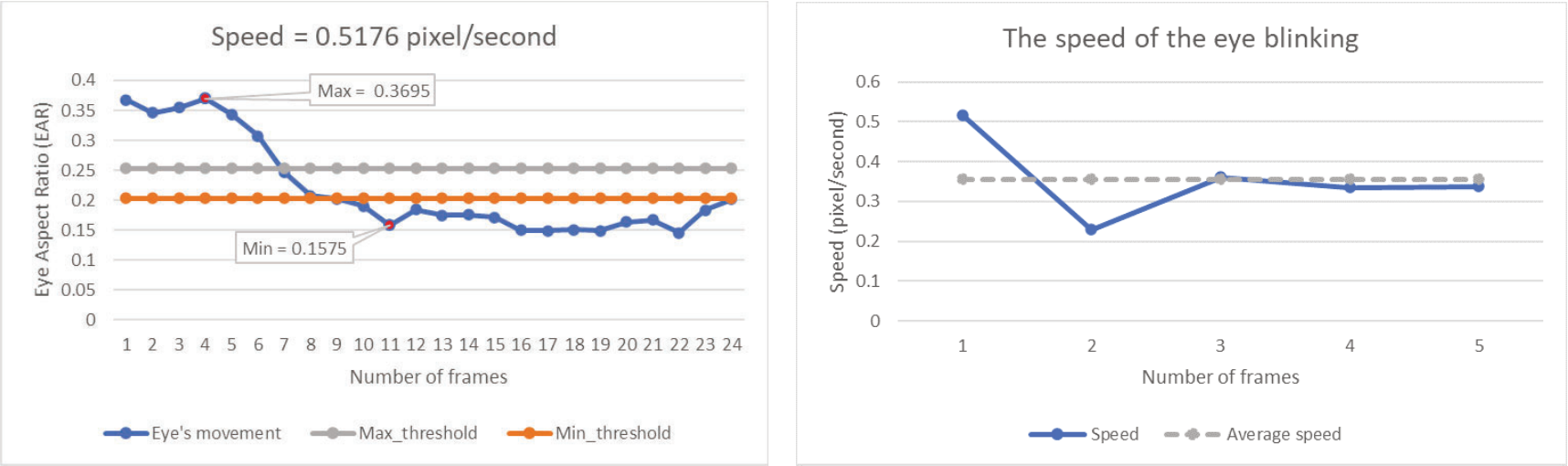

(a) Sleepy blinking's speed (b) Speed values from five blinks

Fig. 9: Result from Participant 3 (sleepy blink)

The participant was closing his eyes longer than usual which mimics one of the main behaviors of a sleepy person. The values are in the range between 0.5167 pixel/second and 0.2172 pixel/second. The average speed value is 0.3557 pixel/second.

**Participant 3:** The graph in Figure 9a has a similar pattern as in the second test (Figure 8a), where there are many values below the minimum threshold. The results from the third test (Figure 9b) came out as expected and the speed values are in the range between 0.5176 pixel/second and 0.2292 pixel/second. The average speed of the eye blinking is 0.35586 pixel/second.

As a result of these tests, it can be concluded that the average speed value for sleepy eye blink is below 0.5 pixel/second and hence the drowsiness threshold to activate the alarm will be set to 0.55 pixel/second (taking into account a safety buffer). Thus, this is the threshold value which will activate the alarm if the speed drops below it.

### 4.4 Evaluation for Different Head Positions

In this last evaluation, the participant went through different situations to test the reliability of the system. These test cases, inspired by Suhaiman et al. [13], simulate the scenario when the driver moves his head in different directions.

In the first test case, the participant was instructed to move his head upwards, downwards, left and right while looking in front. Eventually, in the second test case, the participant also move his head in the same direction as in the first situation but the eyes also follow the direction of the head. For example, if the participant tilts his head upwards, his eyes should look upwards. Table 1 shows the different situations of the head movement and the results.

Table 1: Evaluation of different head positions

<table>
<tr><th colspan="2">Eyes look in front</th></tr>
<tr><th>Test case</th><th>Result</th></tr>
<tr><td>Head faces upward and downward</td><td>• Able to detect eyes up to a certain degree<br>• Able to detect eye blinking<br>• EAR becomes smaller<br>• Smaller EAR affects finding the correct Max EAR and Min EAR</td></tr>
<tr><td>Head turns to the left and right</td><td>• Able to detect eyes as long both eyes are visible to the camera<br>• Able to detect eye blinking but the speed is inaccurate<br>• Some speeds cannot be calculated for certain head poses</td></tr>
<tr><th colspan="2">Eyes follow head's movement</th></tr>
<tr><th>Test case</th><th>Result</th></tr>
<tr><td>Head faces upward and downward</td><td>• Able to detect eyes up to a certain degree<br>• EAR becomes smaller (bigger than when eyes look in front)<br>• Able to detect eye blinking<br>• Speed is less accurate</td></tr>
<tr><td>Head turns to the left and right</td><td>• Able to detect eyes as long both eyes are visible to the camera<br>• Able to detect eye blinking but the speed is inaccurate<br>• Some speeds cannot be calculated for certain head poses</td></tr>
</table>

## 5 Conclusion

In this paper, we have shown that by calculating the speed of the eye blinking, we are able to distinguish between a wakeful and a drowsy blink of a driver in real-time. In particular, we can also detect the gradual process of a driver becoming drowsy. Such a real-time drowsiness detection system plays a key role in preventing car accidents due to microsleep.

However, our extensive evaluations also revealed some deficits, which should be addressed in future developments of our simple algorithm. Firstly, the problem where the algorithm cannot detect the eyes in a certain angle when the head is in a certain position (such as tilted upwards or downwards) can be improved either by identifying the rotation of the head and give conditions in the program or by including additional cameras positioned in different angles [19]. Secondly, facial expression such as smiling has an impact on the measured EAR. Therefore an additional algorithm is needed to detect facial expression, which can then be used to adapt the maximum and minimum thresholds defined in (3) and (4). Another improvement of our algorithm regards the inclusion of optical effects that can occur due to eye glasses.

# MFmap: A semi-supervised generative model matching cell lines to cancer subtypes

Xiaoxiao Zhang[1,2], Maik Kschischo[1]

[1] Department of Mathematics and Technology, RheinAhrCampus, University of Applied Sciences Koblenz, 53424 Remagen, Germany
**kschischo@rheinahrcampus.de**
[2] Department of Informatics, Technical University of Munich, 81675 Munich, Germany

**Abstract.** Cell lines are widely used experimental models in cancer research. However, translating preclinical findingsinto clinical applications is limited by the discordance between cell lines used and tumours. We developedthe model fidelity map (MFmap), a semi-supervised generative model to integrate high-dimensional geneexpression, copy number variation and somatic mutation data of both tumours and cell lines into a small set of features that are highly associated with cancer subtypes, and predict the cell line subtypes simultaneously. These low-dimensional features are biologically interpretable and can be used for matching a given cell line to individual tumours. This enables cancer researchers to select the best cell line model for their experiments.The high accuracy (test set F 1 score > 90%) of MFmap cancer subtype prediction is validated in ten different cancer datasets. From an application perspective, we demonstrate how the predicted cancer subtype for cell lines can be exploited for discovering drug sensitivity differences among subtypes in glioblastoma and breast cancer. This is helpful for guiding personalised treatment decisions and could facilitate drug repurposing for cancer treatments. Thanks to its generative nature, MFmap enables the analysis of cellular status transitions during cancer progression. In particular, we show that artificially perturbing cancer samples from a baseline subtype to an aggressive subtype indeed acquires marker features unique to the targeted subtype in glioblastoma. From a methodological perspective, the newly derived loss function of our MFmap allows to jointly train a classification model and a generative model on both labelled (tumours) and unlabelled (cell lines) data in an one-step-optimisation manner. We further empirically show that the MFmap can achieve excellent classification accuracy and good generative performance simultaneously. These results show that the MFmap will be useful for many semi-supervised prediction tasks in the biomedical sciences and beyond.

This work was supported by the FOR2800 research unit funded by the Deutsche Forschungsgemeinschaft.

# Systematic investigation of Basic Data Augmentation Strategies on Histopathology Images

Jonas Annuscheit, Benjamin Voigt, Oliver Fischer, Patrick Baumann, Sebastian Lohmann, Christian Krumnow and Christian Herta

University of Applied Sciences (HTW) Berlin, Centrum für Biomedizinische Bild- und Informationsverarbeitung (CBMI), Ostendstraße 25, 12459 Berlin, Germany
{first name}.{last name}@htw-berlin.de

**Abstract.** Recent years have witnessed the rapid progress of deep neural networks. However, in supervised learning, the success of the models hinges on a large amount of training data. Therefore, data augmentation techniques were developed to increase the effective size of the training data. Using such techniques is especially important for domains where the amount of available data is limited. In digital pathology, data augmentation is therefore often applied to improve the performance of classifications. This work systematically investigates single data augmentation techniques on different datasets using multiple network architectures. Furthermore, it proposes guidelines on using data augmentation when training deep neural networks on histopathological data.

## 1 Introduction

The prediction quality of supervised learning models relies on the available data's quantity, quality, and heterogeneity. In training a deep neural network, these factors are essential to create a robust and generalizing model. Different transformation techniques can be utilized on the available data to synthesize new samples if a dataset lacks some of these aspects. Such techniques are summarized under the term data augmentation.

Nowadays, there are a variety of different augmentation methods to synthesize new data. These range from classic image manipulation approaches to more contemporary methods like training with adversarial examples [1] or generating entirely new datasets using generative adversarial networks [2]. Several studies investigated the effect of such data transformations on traditional machine learning datasets and proved their benefit [3, 4]. We applied and reviewed some of these transformations to the domain of histopathological datasets.

In recent years, the medical field of pathology has been subject to digital change. Part of this change is to aid the traditional diagnostics, i.e., inspecting extracted tissue sections under a light microscope with computer algorithms [5, 6]. A promising option is to let machine learning or deep learning support pathologists' diagnostic work. Therefore, numerous research studies attempt to answer specific pathological questions using neural networks. Since these questions are usually image classification problems, the approaches use the supervised learning regime, utilizing convolutional neural networks (CNN) [7–9].

Although there are examples of publicly available digitized tissue samples [10], there is a lack of well-curated datasets useful for the supervised learning approach. In addition, highlighting the need for data augmentation methods in this domain, most public datasets are relatively small. Collecting suitable images for a given medical problem is challenging due to the non-uniformity of manifestations and the need to consider patient rights. Labeling these images requires the highly specialized expertise of a pathologist, adding to an already busy workload.

In this work, we build a pipeline to systematically investigate basic data augmentation techniques on different classification datasets and network architectures. For this purpose, we selected two public histopathological datasets for different medical problems: *classification of mitosis candidates* and *tissue type classification*. We trained three contemporary CNN architectures for all of these data sets, examining different types of augmentation methods. This paper describes the experimental setup to measure the influence of a single data augmentation technique on the model's performance. In addition, it proposes guidelines for using data transformations in the supervised learning setting regarding different types of histopathological data. Finally, it discusses under which circumstances data augmentation has a reliable benefit for a model's training process.

## 2 Related Work

Due to its regularization effect, data augmentation is a popular method used in deep learning pipelines to reduce overfitting and increase the robustness of a model, especially concerning an image classification problem. In fact, the method is so established that several tools exist to make standard techniques more accessible [11–13] or even automate the augmentation process [14, 15].

Several studies examined the actual influence of different data augmentation techniques and showed its beneficial effects in the context of natural images [3, 16]. A widely used taxonomy to divide the common techniques into categories is *basic image manipulations*, e.g., geometric transformations, cropping, occlusion, noise injection, filtering, color transformations, and *deep learning approaches*, e.g., adversarial training, style transfer, synthetic image generation via generative adversarial networks [3, 17].

Unlike in the natural image domain, where datasets can provide millions of images, far fewer qualitatively annotated samples are available in the field of histopathology. Hence, data augmentation has established itself as an integral part of the training pipelines in this area as well. Interestingly, it is almost exclusively the use of newer augmentation techniques from the deep learning approaches that have been broadly reviewed thus far. Generative adversarial networks (GANs) were investigated to solve the stain normalization problem using style transfer methods. Color differences and disturbances are a considerable challenge through various tissue staining protocols and the varying digitization processes. Style transfer can homogenize the color distribution in a data set and thus the distributional shift in a dataset [18–20]. Some studies even explored the transfer of staining protocols utilizing GANs; e.g., Mercan et al. trained a model that converts images obtained from H&E stained tissue into virtual PHH3 staining [21]. In addition, GANs are used to synthesize completely artificial samples to enrich small data sets [22–24].

Concerning *basic image manipulation*, many approaches use several techniques to augment their datasets but do not evaluate the influence of augmentations; see, e.g. [25, 26]. Primarily, basic manipulation techniques are used intensively in conjunction with semi-supervised learning methods, which are becoming increasingly popular in this domain [27–29]. Color transformations, in particular, are one of the most widely used techniques due to the nature of histopathological images [26]. Tellez et al.[30] and Karimi et al.[31]

examined the stain normalization problem more closely and developed custom augmentation techniques for it.

However, in-depth studies which comprehensively evaluate the effect of basic image manipulation techniques can only be found for radiology images in the medical domain [17, 32–34].

## 3 Method

We have developed a pipeline to measure the effect of data augmentation techniques in supervised learning on histopathology datasets.[1] We can configure experiments as a triple $(d, m, t) \in D \times M \times T$ where $D$ is the set of possible datasets, $M$ denotes the set of considered deep neural network architectures, and $T$ corresponds to the set of different data augmentations. A specific transformation $t$ manipulates online the batches drawn from the dataset $d$ to create transformed batches, that are used to train a deep neural network $m$. Keeping the pair $(d, m)$ constant, the influence of $t$ on the trained model can be measured by comparing its parameters and performance. This section describes the sets $D$, $M$, $T$ in more detail and contains information about the training and evaluation protocol used.

### 3.1 Datasets $D$

Set $D$ consists of the publicly available datasets MIDOG[2] and BACH[3]. Both sets were pre-processed to fit a classification problem.

The task of the BACH dataset is to distinguish between four different tissue types. It is a tiny dataset with images of 2048x1536 pixels and 100 samples per class, i.e., 400 samples overall. Therefore, we cropped patches from the original images using a 512x512 window with a 256-pixel step to increase the dataset size. In addition, we discarded patches not containing any H&E-stained tissue during the process by removing tiles with less than 3% tissue. Finally, we split the dataset into three subsets using random sampling for training (4801 samples), validation (1655 samples), and test (1647 samples). Patches with overlapping pixels in the subsets were removed. The classes of the dataset are nearly balanced.

The MItosis DOmain Generalization Challenge [35] published a dataset of human breast cancer tissue samples. However, we note that up to this point, only the training set is publicly available, consisting of 1721 mitotic figures and 2714 non-mitotic examples. The samples were acquired using three different whole slide image scanners and annotated by trained pathologists with a multi-expert blind annotation pipeline. We pre-processed the dataset by cropping a 250x250 patch around each annotation center. We sampled three distinct subsets keeping the class balance: training (2219), validation (1071), and test (1145).

### 3.2 Models $M$

The model set M consists of the networks VGG[36], Inception[37] and Densenet[38]. These networks form a cross-section over the development of CNN architecture and, therefore, have distinct structural elements. We intend to investigate whether these structures react differently to data augmentation. We use a pre-trained PyTorch model with reinitialized classification head, i.e. *vgg11_bn*, *inception_v3* and *densenet121*, as baseline for each network type. These models require as input square images of a model-dependent size $n_m$, i.e. the input images have shape $n_m \times n_m$.

[1] The source code is available via Github: https://github.com/CBMI-HTW/Data-Augmentation-Histology

[2] https://imi.thi.de/midog/the-dataset/

[3] https://iciar2018-challenge.grand-challenge.org/Dataset/

### 3.3 Transformations $T$

The transformations in $T$ fall in the categories: color-based, geometric-based, filter-based transformations, and erasing. All transformations are realized by using the implementation of the torchvision library.

In our setup, transformations are applied with a certain probability $p$, with $p = 1$ if not stated differently. For most of these transformations $t$ an additional parameter $s$ controls the strength of the transformation on the input $x$, i.e. the output of the transformation is $t_s(x)$. For each corresponding transformation, $s$ is sampled from a certain interval, where the size of this interval is a hyperparameter in our setting.

The hyperparameters of the geometric-based transformations are determined by $n_m$ and the maximal distortion without getting blacked borders. For all other transformations we perform a hyperparameter optimization on the BACH dataset and VGG model to identify the best parameter ranges. We trained for each configuration of parameters a minimum of 7 models and choose the setting with the highest mean validation accuracy as the best performing one. These models were only used to determine the hyperparamters and not for the final test results.

**Color-Based Transformation** As color-based transformations we use the standard transformation for brightness, contrast, gamma value, hue angle and saturation. The strength $s$ is sampled randomly from the interval $[s_0 - s_1, s_0 + s_1]$ where, the center $s_0$ is defined by $t_{s_0}$ = identity. Here, we sample $s$ indirectly by drawing $r \in [0, 1]$ according to the beta distribution Beta($\alpha = 8, \beta = 8$) and computing $s = s_0 + (2r - 1)s_1$. The half-width $s_1$ was determined in a hyperparameter optimization. The parameters for the hyperparameter optimization are summarized in Tab. 1, where the intervals for choosing $s_1$ where found iteratively by hand over multiple trials, ensuring that the chosen value does not lie on the boundary.

**Geometric-Based Transformation** As geometric transformations, we investigate flips, rotations, random cropping to size $n_m$, shearing and scaling. For scaling, rotation, and shearing, we sample $s$ uniformly from an interval $[s_0 - s_1, s_0 + s_1]$ as summarized in Tab. 1, where we consider two scaling scenarios. In the case of the flip transformation, we apply horizontal and/or vertical flips to the input, each with a probability of 0.5. Both scaling transformations, as well as shearing, is done with $p = 0.9$.

**Filter- and Erasing-Based Transformation** As filter-based transformations, we study Gaussian blurring as well as a sharpness adaption.

For Gaussian blurring, we pick uniformly an odd kernel size between 3 and 15 and use a minimum and maximum sigma (these are direct inputs to the trochvision implementation) of 0.001 and 0.5, respectively and set $p = 0.5$. We did a hyperparameter search for the maximal kernel-size in $\{7, 11, 15\}$, the maximum sigma value in $[0.5, 8]$ and $p$ in $\{0.5, 0.75\}$.

**Table 1.** Parameters used for the color- and geometric-based transformations. The center $s_0$ is fixed where the interval half-width $s_1$ was either optimized with a hyperparameter optimization within the given intervals or chosen as stated.

| transformation | $s_0$ | $s_1$ | choice of $s_1$ | transformation | $s_0$ | $s_1$ | choice of $s_1$ |
|---|---|---|---|---|---|---|---|
| brightness | 1.0 | 0.0175 | $[0.005, 0.3]$ | rotation | 0 | 180° | fixed geometrically |
| contrast | 1.0 | 0.1 | $[0.025, 0.6]$ | scale I | 1 | 0.15 | chosen by hand |
| hue | 0.0 | 0.00625 | $[0.00025, 0.6]$ | scale II | 1 | 0.29 | fixed geometrically |
| saturation | 1.0 | 0.025 | $[0.01, 0.6]$ | shear | 0 | 22° | fixed geometrically |
| gamma | 1.0 | 0.05 | $[0.00625, 0.6]$ | | | | |

The sharpness adaption depends on a parameter $s \geq 0$ where for $s < 1$, the image is blurred and for $s > 1$ sharpened and $s = 1$ corresponds to the identity. To define $s$ we sample $r \in [0, 1]$ according to Beta($\alpha = 8, \beta = 8$) and set $s = 2r$ if $r \leq 0.5$ and $s = 4.5(2r - 1) + 1$ if $r > 0.5$. The factor 4.5 determines the maximal sharpening factor and was chosen from a hyperparameter search in the interval $[1.0, 8.0]$.

For erasing-based transformations, we select 3 potentially overlapping erasing rectangles with a size ranging from $0.01n_m$ to $0.2n_m$ and aspect ratio ranging from 0.5 to 2. The trochvision implementation selects the corresponding parameters uniformly within these ranges. Each rectangle is then applied with $p = 0.75$. The applied rectangles are then filled with zeros (erasing (black)) or with white noise (erasing (random)). All three parameters, the number of rectangles, the maximal size and the appliance probability were hyperparameter optimized for the erasing (black) scenario over the sets $\{1, 3\}$, $[0.025n_m, 0.4n_m]$ and $\{0.5, 0.75\}$, respectively.

### 3.4 Training Protocol

For the data augmentation experiments we calculated all triplet combinations $(d, m, t)$. We implemented equal training settings to ensure maximum comparability with the baseline models, i.e., optimizer, scheduler, learning rates, weight initialization, and fixed data loading. The only difference in the pipeline was using the investigated transformation $t$ in the default transformation sequence.

To implement one training cycle, we followed Chollet's recommendations [39] on fine-tuning. First, we trained reinitialized layers for a warm-up period before updating all network parameters' to prevent the negative effect of a possible large error signal on previously learned features. We implemented this by training 10 epochs with a 1-cycle learning rate scheduler [40], a LAMB optimizer [41] and a L2 loss regularization. Then, followed by another training over 50 epochs, all weights are updated in the network using the same optimizer and scheduler policies. We always selected the last model of the training process for the evaluation to ensure that the data was seen equally.

When loading a sample $x$, we apply a fixed transformation sequence. We perform this sequence to avoid artificial padded black borders in arbitrary rotation transformations. However, to maintain comparability, the sequence is always used. (1) Resize the $x$ to $\sqrt{2} * n_m$ with $n_m$ being the network input size. (2) Apply the investigated transformation $t$ to the resized sample with probability $p$. (3) Center crop the transformed sample with the size $n_m$. (4) Normalize the final input by the mean and standard deviation of each color channel. Exceptions to this sequence are erasing-based transformations, which are applied at the end of the sequence to not corrupt the normalization process.

We performed a hyperparameter optimization for the training settings of each baseline model and dataset $(d, m)$ using grid search. These final settings are reported in table 2 and used for the training of each triplet $(d, m, t)$.

**Table 2.** Final parameter used for the training of each triplet $(d, m, t)$ where the parameters are optimized for the baseline model of every pair $(d, m) \in D \times M$.

| | BACH | | | MIDOG | | |
|---|---|---|---|---|---|---|
| Parameter | VGG | Inception | Densenet | VGG | Inception | Densenet |
| Epochs (total) | 60 | 60 | 60 | 60 | 60 | 60 |
| Epochs (warm-up) | 10 | 10 | 10 | 10 | 10 | 10 |
| Learning Rates | 0.0016 | 0.0016 | 0.0016 | 0.0064 | 0.0032 | 0.0032 |
| Weight Decay | 0.01 | 0.01 | 0.01 | 0.01 | 0.01 | 0.01 |
| Batch Size | 64 | 64 | 64 | 64 | 64 | 64 |

### 3.5 Evaluation Protocol

To account for the randomness during training given due to weight initialization of the classification head, dropout layer and random application of augmentations, that results in a spread of experiment outcomes, we conducted 10 experiment runs per configuration $(d, m, t)$. The 10 experiments only differ by a random seed given at the start of training. This allows gathering a statistic, which makes a more accurate statement about the effectiveness of an augmentation and also enables an expressive comparison between them. As metric, we use the accuracy value on the plain test set for testing the model.

## 4 Results and Discussion

We summarize our findings for each triplet in a boxplot.[4] We interpret an augmentation technique as effective if its median is above the mean of the baseline model (continuous horizontal line) and their overlap in the interquartile range (IQR $= Q_3$ - $Q_1$) is minimal. We show the results for each dataset in Figure 1 and Figure 2 grouped by the augmentation strategies and the network architecture. For a clean plot appearance, a few extreme outliers were discarded from the visualization.

For the BACH dataset, we observe an expected behavior of the baseline models. With the rising complexity of the network architecture, the accuracy increases slightly. However, comparing the baseline with models trained using data augmentations shows that the geometric transformations stand out. Since the histopathological data is rotation invariant and the importance of morphological structures, we anticipated such results. Furthermore, erasing-based augmentations also provide a beneficial effect. On the other hand, filter-based and color-based transformations do not seem to have a positive influence, nor do they harm the model's performance for the chosen hyperparameters. Especially regarding the color-based augmentations methods, that finding was surprising since we presumed the manipulation of the color space to be a critical factor in the context of histopathological data. Our hypothesis that the color distortions caused by the hyper-optimized transformations were too close to the identity could not be verified. Additional experiments with artificial parameter values for the transformations to change the input drastically lead to similar results. The results for the MIDOG dataset are essentially alike. Geometric-based augmentations raise the performance, whereas color-based and filter-based transformations have no significant effect. Indeed, the scatter and the overlap with the baseline is so tremendous that no positive effect can be ascribed to these transformations. Erasing-based augmentation harms the model performance significantly. We assume that the augmentation occasionally covers the cell nucleus, which is essential for distinguishing between mitosis and non-mitosis. However, we did not investigate this further and leave this for future work.

[4] Interactive versions of the charts and more details of the results can be viewed at https://cbmi-htw.github.io/Data-Augmentation-Histology-Website/

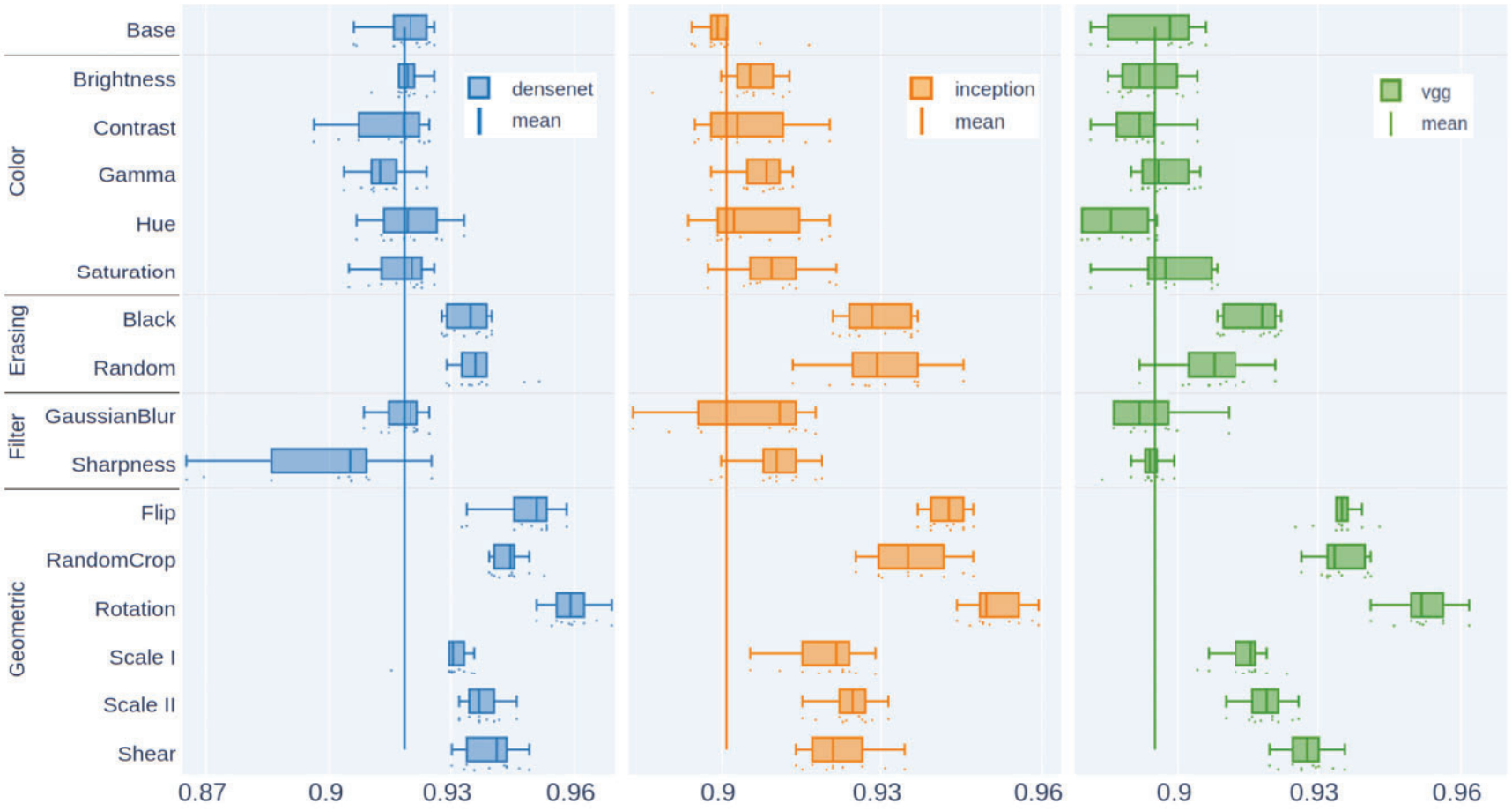

**Fig. 1.** Results for the BACH dataset. We show the individual mean accuracies on the test set as well as resulting boxplots for the 10 runs for the baseline configuration and the individual transformations for all three network architectures.

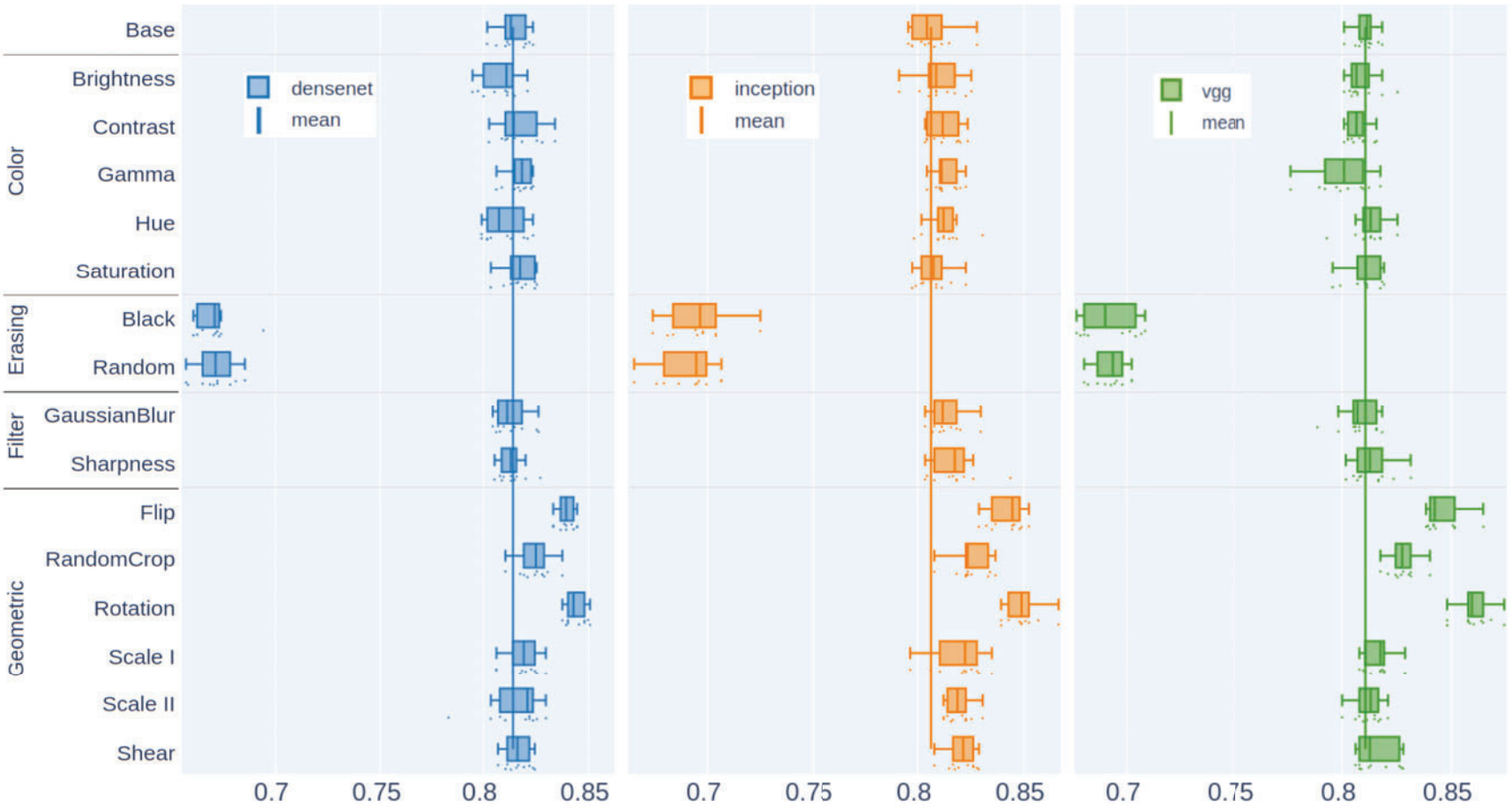

**Fig. 2.** Results for the MIDOG dataset. We show the individual mean accuracies on the test set as well as resulting boxplots for the 10 runs for the baseline configuration and the individual transformations for all three network architectures.

Additionally, we have to mention that it took much computational effort to get the results reliable and robust due to intensive hyperparameter optimization. We observed a strong sensitivity towards the hyperparameters of the learning process. Tiny changes in the training settings, e.g., learning rate or batch size, let the benefits of data augmentation vanish in noise. Therefore, we advise using augmentation techniques in combination with well-optimized training hyperparameters to profit from the method. We even suggest tuning the transformation parameters for an optimal result.

## 5 Conclusion

We examined basic data augmentation techniques frequently used in deep learning classifier training pipelines on two histopathological data. Overall geometric-based techniques increase the model performance on such datasets. However, surprisingly, color-based augmentations do not have the expected impact and are costly due to the required parameter optimization. Next to supervised learning settings, we expect this work to improve contemporary semi-supervised learning methods, e.g., contrastive learning, and assume such methods will considerably impact the training of deep learning models in the histopathological domain.

**Acknowledgments**
The authors thank Dr. Tim-Rasmus Kiehl, M.D. (Charité, Institute of Pathology) for helpful comments and acknowledge the financial support by the Federal Ministry of Education and Research of Germany (BMBF) in the project deep.HEALTH (13FH770IX6).

# Online extraction of functional data from video recordings of gut movements using AI features

Pervaiz Khan[1], Manuela Gries[2], Ahmed Sheraz[1], Steven Schulte[2], Anne Christmann[2], Marko Baller[2], Karl-Herbert Schäfer[2], Andreas Dengel[1]

[1] DFKI (German Research Centre for Artificial Intelligence, Kaiserslautern Germany),
[2] University of Applied Sciences Kaiserslautern, Campus Zweibrücken, Germany

**Abstract.** The gut is an often underestimated organ which contributes significantly to our health condition 1 . It is also one of the main entry gates for pathogens, toxins or drugs, thus influencing the whole body. The gut harbors an intrinsic and autonomeously working nervous system, the socalled enteric nervous system (ENS) that regulates blood flow, resorption, mucosal barrier function and gastrointestinal motility. The gastrointestinal motility is an appropriate readout to evaluate the health status of the whole organ, since all kind of compromising or challenging agents, either orally or systemically administered will affect the ENS 2,3,4 . Moreover the gut can be compromised in various diseases. During the last years, the role of the gut in neurodegenerative processes and diseases came more and more into focus and it could i.e. be demonstrated that gut motility changes also in models of Alzheimer and Parkinsons disease (PD) 5,6 . In PD patients, gastrointestinal problems often appear long before the disease is diagnosed. In a recent study it could nicely be demonstrated that in very young (2 month) mice that overexpress the alpha synuclein pathogenic peptide, the alteration of colonic motility was related to molecular alterations of the ENS 7 . While the use of muscle strips with the included ENS to investigate isometric contractions is a rather artificial approach, using intact gut segments with intact mucous layers and even if necessary, mesenterial perfusion, can deliver much more in vivo equivalent data to analyse gut activity under the influence of external or systemic factors.

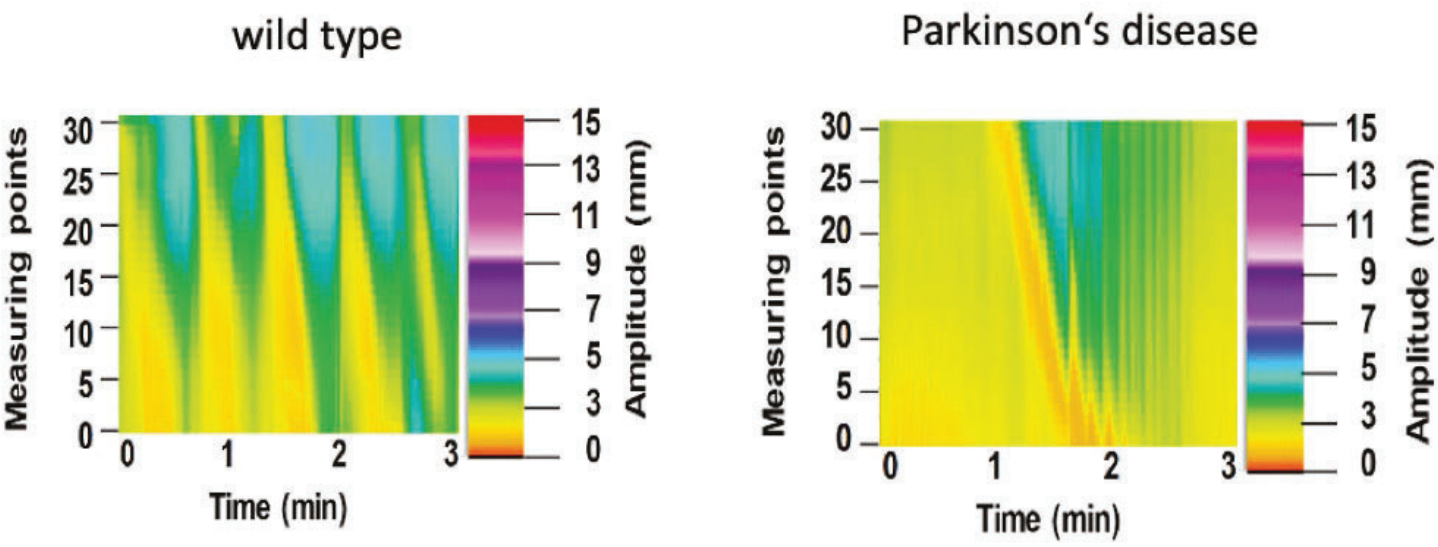

Figure 1 Heat maps that represent the movement of the colon in a PD model (right) compred to the wild type mouse colon. While the WT shows a continuous series of propagating contractions, the PD gut is rather slow.

Gut segments from the colon of adult mice were fixed in a tubing system and placed in an organ bath under continuous luminal perfusion. The perfusion was performed using an efflux resistance of 3cm H 2 O to induce gut movements. The organ bath chamber was equipped with a frontal and bottom glass plate, so that a continuous video recording from front and bottom could be realized. The gut was perfused and superfused with a 37°C warm Tyrode buffer solution at an pH of 7,4. The buffer was oxygenized prior to perfusion to obtain a sufficient oxygen saturation. The gut was allowed to equilibrate for 10 min after fixaton in the organ bath under physiological conditions. Then the experiment was started. Two cameras, positioned either in front or at the bottom window of the organ bath chamber, were started at the same time and the spontaneous activity ot the gut recorde for 10 min. Then the gut was challenged with individual drug compounds to stimulate its activity for another 10 min. At the end of the experiment, a maximal stimulation was achieved using Acetylecholine. To analyze the movement of the gut, initially 33 virtual dots positions are selected on the boundary of the gut manually. Then, in each frame of the video, vertical movement of the dots is tracked using distance transforms while considering the fixed horizontal position. Then, the distance of every dot is measured to the corresponding dot in the previous frame. In this way, gut movement is tracked in the complete video. The heat map of the movement is presented in the Figure below:

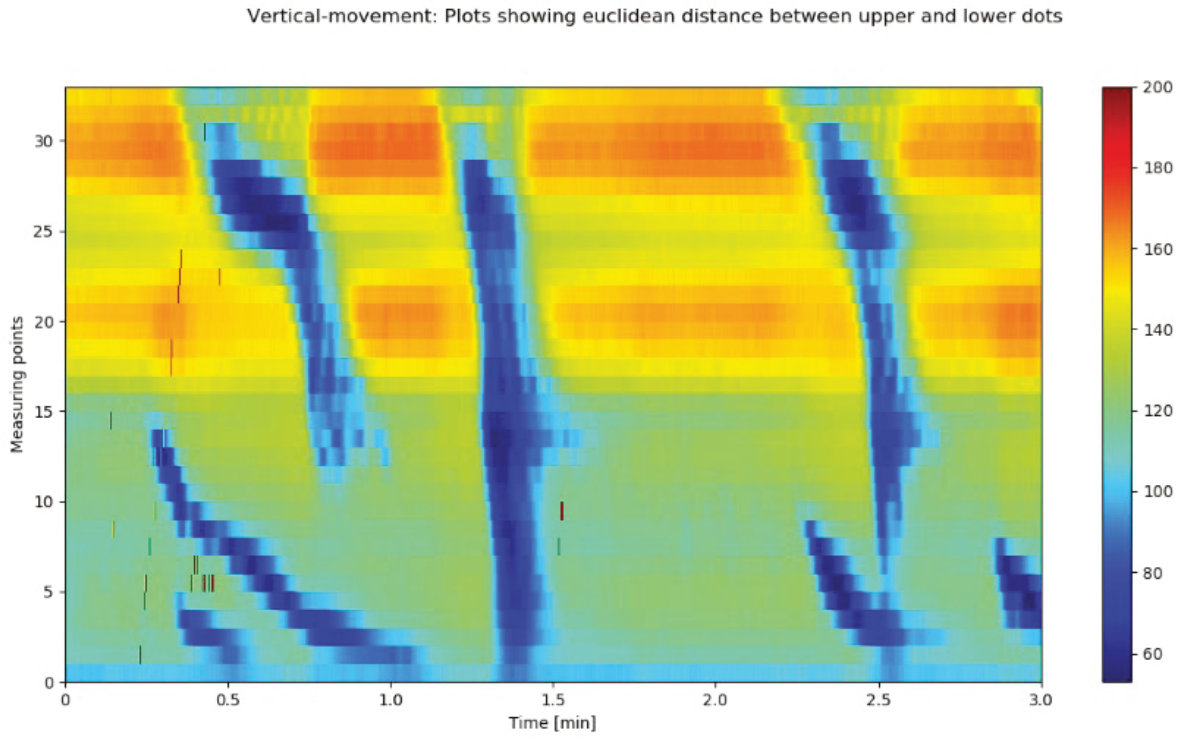

Figure 2 Heat map presents the vertical movement of guts in a video.

The approach demonstrates the use of AI algorithms to extract valuable and quantifiable data from gut movements online in real time. This allows to investigate the impact of either drugs, toxins, nutritional components or even diseases upon gastrointestinal motility. The method delivers timely and spatial resolution of gut movements, so that a detailed analysis of functional distinct entities of the gut (i.e. proximal and distal colon) can be evaluated and compared.

# An artificial neural network-based toolbox for the orphological analysis of red blood cells in flow

Marcelle Lopes, Stephan Quint

Cysmic GmbH, Germany
**marcelle.lopes@cysmic.de, stephan.quint@cysmic.de**

**Abstract.** We present a toolbox that combines image processing techniques with artificial intelligence to enable single-cell the detection and characterization of red blood cells observed in microfluidic flow. In healthy subjects, red blood cells show a smooth transition between an axis-symmetric ("croissant") and non-axis-symmetric ("slipper") shape depending on their flow velocity. However, in subjects with blood diseases this shape dynamics is disturbed and results in deviating blood flow properties. Current diagnostic methods rely on the identification of genetic mutations in addition to functional tests, including the manual evaluation of red blood cells in stasis. Although the latter technique is considered a clinical standard, it is not sufficient to discriminate between blood diseases and their severities. The automation of the characterization of images of single red blood cells in flow is an unbiased technique that could set new standards in blood disease clinical diagnostics. Considering the large variety of red blood cells shape deformations, we developed a semi-supervised neural network for a reliable and reproducible cell shape evaluation. By arranging ideal shapes as cornerstones of the training data set, cell shape transitions are self-learned during the training process of the neural network. This highly reduces the amount of required training data as well as the need for a manual pre-classification. Our approach , in addition to avoiding errors due to manually selected training data (supervised training), also enables the definition of custom thresholds and metrics for further discrimination and statistical analysis. The technique will be tested on blood of patients with inherited rare anemias, e.g., sickle-cell disease, as well as on transfusion blood and chronic and infectious diseases, such as COVID-19.

# Comparing a deterministic and a Bayesian classification neural network for chest diseases in radiological images

Jonas Nolde and Ruxandra Lasowski

Hochschule Furtwangen University
jonas.nolde@hs-furtwangen.de,
ruxandra.lasowski@hs-furtwangen.de

**Abstract.** A common mantra for automated decision systems is that a system should know when it doesn't know. Bayesian neural networks are designed to capture uncertainties over the network weights and in theory, they perform better predictions and output uncertainties. To this end, we compare in this paper a deterministic neural network and a Bayesian neural network for the classification of chest diseases in radiological images. We use the ChestX-ray14 data set [1] involving 14 respiratory diseases like pneumonia and atelectasis. We found that the deterministic network similar to CheXNet [2] outperformed the Bayesian version in this task, whereas, employed on the more simplistic MNIST dataset it did not. Our experiments suggest that there is a gap between theory and practical use of BNNs for very deep networks and real clinical data.

## 1 Introduction

In 2016 almost 2.38 million people died from lower respiratory infections worldwide. This was the sixth leading cause of mortality for all ages that year [3]. For the last three decades, pneumonia was the most common cause of death for children under 5 years [4]. While the treatment of a diagnosed pneumonia patient can be done efficiently with low-cost, low-tech medication and care [5], the detection of lower respiratory infections leaves room for improvement. The diagnosis of such diseases with the help of chest X-rays is very effective and currently the best available method [6]. Unfortunately, this task is challenging and requires expert radiologists which are rare in impoverished regions.

Past work about the detection and classification of lung diseases on chest X-rays with convolutional neural networks (CNNs) could achieve promising results that match or exceed the performance of radiologists [2]. These deterministic neural networks, albeit being right most of the time, won't tell you how certain they are about their decisions. In critical applications, where the cost of error is high, an indication of confidence can be extremely valuable, especially in uncertain edge cases.

While deterministic neural networks contain a specific set of weights, Bayesian neural networks (BNNs) assign probabilities to all possible sets of weights, allowing for uncertainty quantification. In this work, we will assess the applicability of Bayesian deep learning in the field of medical diagnosis. We, therefore, implement a version of the deterministic neural network CheXNet [2] and a Bayesian version of CheXNet using recent advances in Bayesian deep learning. We then evaluate and compare their performance on the *ChestX-ray14* data set [1]. Furthermore, we efficiently measure aleatoric and epistemic uncertainties in the Bayesian model's predictions.

## 2 Related Work

### 2.1 Bayes' Theorem

Bayesian deep learning utilizes *Bayes' Theorem* to calculate conditional probabilities. In the context of deep learning, Bayes' Theorem can be rewritten as

$$p(w|\mathcal{D}) \;=\; \frac{p(\mathcal{D}|w)\,p(w)}{p(\mathcal{D})}, \tag{1}$$

with a neural network's parameters $w$ (the *hypothesis*) and data $\mathcal{D}$ (the *evidence*). Bayesian deep learning aims to calculate the *posterior* distribution $p(w|\mathcal{D})$, which "captures the set of plausible model parameters, given the data" [7]. This is done by multiplying the *likelihood* $p(\mathcal{D}|w)$ of data $\mathcal{D}$ occurring given parameters $w$ with the *prior* distribution $p(w)$ and normalizing it by the data distribution $p(\mathcal{D})$.

We can implement a Bayesian neural network by replacing a "deterministic network's weight parameters with distributions over these parameters, and instead of optimising the network weights directly we average over all possible weights (referred to as *marginalisation*)" [7]. With *Bayesian inference* we can calculate the posterior distribution $p(w|\mathcal{D})$ and predict probabilities $y^*$ given new data $x^*$:

$$p(y^*|x^*,\mathcal{D}) \;=\; \int p(y^*|x^*,w)\,p(w|\mathcal{D})\,dw\;. \tag{2}$$

### 2.2 Variational Inference

Hinton and Van Camp initially proposed *variational inference* for neural networks in 1993 [8] as an alternative to methods involving expensive Monte Carlo sampling. In 2011, Graves [9] published an improved approach, suitable for more complex neural networks. Variational inference solves the intractability of the integral over the true posterior distribution $p$ in eq. 2 by integrating over a simplified posterior distribution $q$ with variational parameters $\theta$ instead. This variational posterior $q_\theta$ is an approximation of the true posterior $p$ but is easier to sample from. In most cases, a Gaussian distribution $\mathcal{N}(\mu,\sigma^2)$ with parameters $\theta=(\mu,\sigma^2)$ for the mean and the variance respectively is used. Graves' approach optimizes the posterior approximation $q_\theta(w|\mathcal{D}) \approx p(w|\mathcal{D})$ by minimizing the variational free energy $\mathcal{F}$, also referred to as the negative variational lower bound or negative evidence lower bound (ELBO). For deep learning, $\mathcal{F}$ can be reinterpreted as *minimum description length cost function* [9]:

$$\mathcal{F}(\mathcal{D},\theta) \;=\; KL[q_\theta(w)||p(w)] - \mathbb{E}_{q_\theta(w)}[log\,p(D|w)], \tag{3}$$

where $KL[q_\theta(w)||p(w)]$ is the *Kullback-Leibler (KL) divergence* between both distributions. It consists of a data-dependent part (the *likelihood cost* or *error loss*) and a prior-dependent part (the *complexity loss*). The function embodies a trade-off between satisfying the complexity of the data $\mathcal{D}$ and the simplicity of the prior $p(w)$ [10].

Graves' approach for a tractable approximation of the Bayesian neural network's posterior distribution was the cornerstone for more efficient and stable estimation methods like Stochastic Gradient Variational Bayes (SGVB) [11], the local reparameterization trick [12] and the flipout estimator [13].

### 2.3 Uncertainty Estimation in Bayesian Deep Learning

Kendall and Gal [7] describe the two types of uncertainty in the context of Bayesian deep learning as follows:

*Epistemic* uncertainty is often referred to as *model uncertainty*, as it "captures our ignorance about which model generated our collected data" [7].

*Aleatoric* uncertainty, on the other hand, "captures noise inherent in the observations" [7] and thus is often referred to as *data uncertainty*. This type of uncertainty can further be categorized into *homoscedastic* uncertainty and *heteroscedastic* uncertainty. While homoscedastic uncertainty stays constant for different inputs, heteroscedastic uncertainty varies from input to input, with some potentially having more noisy outputs than others. They note that "heteroscedastic uncertainty is especially important for computer vision applications" [7].

The paper concludes, that measuring both types of uncertainty is crucial for the safety and reliability of models. Aleatoric uncertainty is important for "large data situations, where epistemic uncertainty is explained away" and "real-time applications, because we can form aleatoric models without expensive Monte Carlo samples" [7]. Epistemic uncertainty is important for "safety-critical applications, because epistemic uncertainty is required to understand examples which are different from training data" and "small datasets where the training data is sparse" [7].

Kendall and Gal [7] proposed a method to estimate both the aleatoric and the epistemic uncertainty, which was later refined for classification by Kwon et al. [14]:

$$\underbrace{\frac{1}{T}\sum_{t=1}^{T}\mathrm{diag}(\hat{p}_t) - \hat{p}_t\,\hat{p}_t^T}_{\text{aleatoric}} + \underbrace{\frac{1}{T}\sum_{t=1}^{T}(\hat{p}_t - \overline{p}_t)(\hat{p}_t - \overline{p}_t)^T}_{\text{epistemic}}, \tag{4}$$

with the predicted probability vector $\hat{p}_t = p(\hat{w}_t) = \mathrm{Softmax}\{f^{\hat{w}_t}(x^*)\}$, that is sampled $T$ times, the diagonal matrix $\mathrm{diag}(\hat{p}_t)$ with elements of vector $\hat{p}_t$, and the mean predicted probability $\overline{p} = \sum_{t=1}^{T}\hat{p}_t/T$. We later use the formula of 4 in our experiment to measure our model's uncertainties.

## 3 Methods

### 3.1 First Tests with TensorFlow Probability and MNIST

To gain our first practical experience in implementing a Bayesian convolutional neural network for disease classification on X-rays with TensorFlow 2.0 and its probabilistic programming library TensorFlow Probability, we first looked at a simpler image classification task. We implemented the small LeNet-5 network [15]. It consists of only 7 layers (3 convolutional layers, 2 subsampling layers, 2 fully connected layers) making it a relatively simple and small network in today's time. The original LeNet-5 architecture is a deterministic neural network and does not use probabilistic methods. To make it Bayesian we replaced the deterministic convolutional and fully connected layers from TensorFlow with the probabilistic layers from TensorFlow Probability. The model's task is to classify the images of the *MNIST* data set [15] and tell which digit is depicted. The data set contains 28×28 pixel small, grayscale images of a handwritten digit (0-9), normalized in size and centered in the image and is split into 60,000 training samples and 10,000 test samples. We trained the model in mini-batches with 128 normalized images each, where the Adam optimizer [16] minimizes the categorical cross-entropy loss.

**Results and Conclusion** After training for 75 epochs, the Bayesian model achieved a validation accuracy of 0.984, which was good enough for us to stop the training. We then performed Monte Carlo sampling by asking the trained model to predict the labels of the same unseen images 50 times. The resulting outputs were different at each prediction as figure 1(b) shows. For comparison, we trained a non-Bayesian, deterministic version of the network. 10 epochs of training already yielded a training accuracy of 0.997. Figure 1(a) shows the resulting test prediction plots. Note that a deterministic model always outputs the same values for the same input, requiring only one prediction per sample at inference. To show how the models perform on unusual data, we tested with "fake" MNIST images from the *notMNIST* database, which contains images that look similar to those in the MNIST database but show letters instead of numbers. The resulting prediction plots can be seen in the last samples of figure 1.

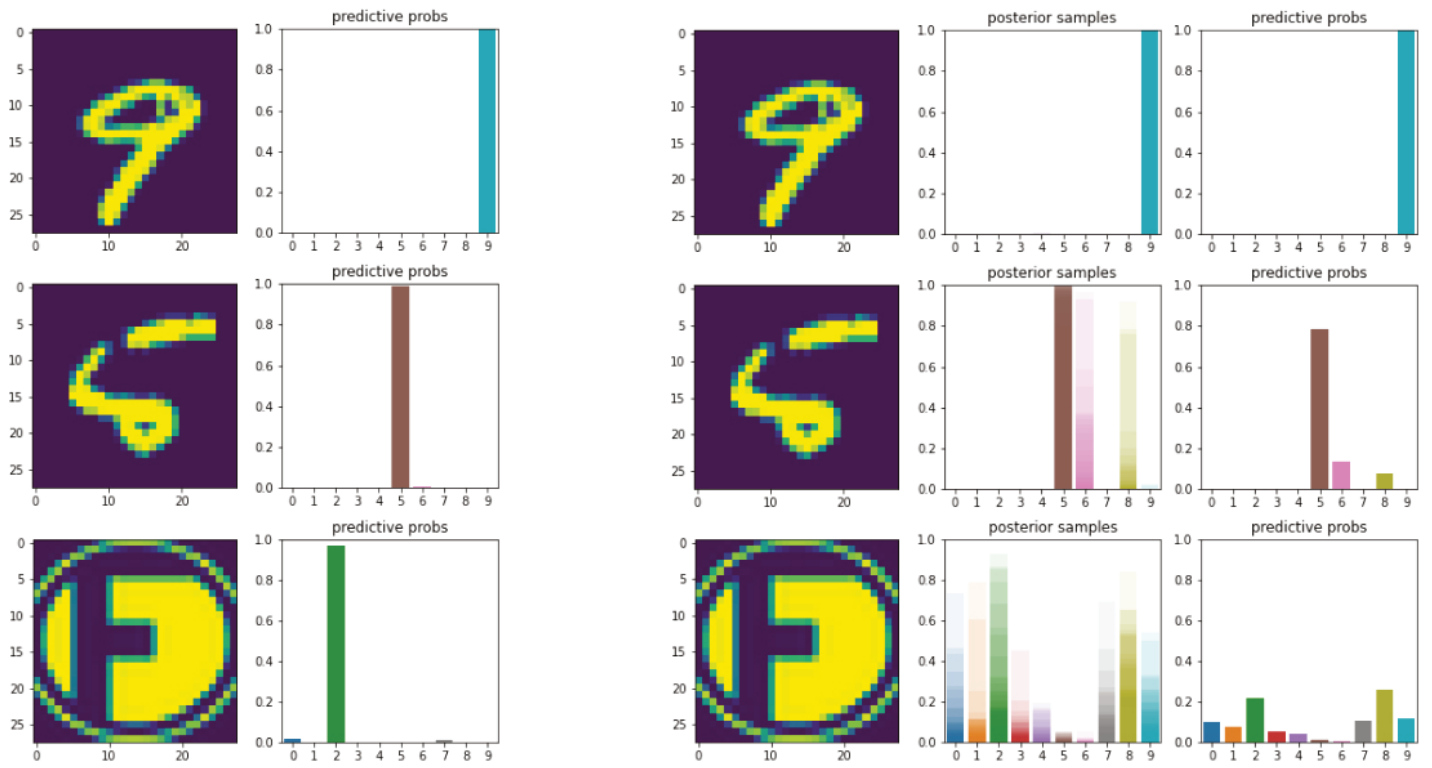

(a) Deterministic model (b) Bayesian model

Fig. 1: Predictions of the deterministic and the Bayesian model on the MNIST data set.

Concluding, the first thing to note is that training the Bayesian neural network takes considerably longer than training the deterministic one. Figure 1 shows that both models correctly classified the first sample. However, the Bayesian model predicted wrong classes several times during the Monte Carlo inference on the second sample. Looking at the input image, we can argue that the "5" looks like a "6" or an "8" to some degree. Thus, having the Bayesian model predict those digits a few times is a justifiable and possibly desirable sign of uncertainty. The third sample depicts the letter "F" in a circle and is also inverted. Hence, the sample neither belongs to a class that can be predicted by the model nor did the model see comparable images during training. While the deterministic model was sure it saw the digit "2" (figure 1(a)), the Bayesian model was highly uncertain in its decision (figure 1(b)). The plot shows that during Monte Carlo sampling the model predicted different classes every time, causing the average prediction to have a low, almost similar probability for every class. This can be interpreted as an indication of high uncertainty and the model saying "I don't know". In critical applications like medical diagnosis, this behavior is highly desirable.

### 3.2 CheXNet and Bayesian CheXNet

The main goal of this work is to show the advantages of Bayesian deep learning over traditional, deterministic deep learning for medical image classification. We compare performances of the deterministic CheXNet [2] and our Bayesian neural network with a similar architecture.

CheXNet achieves state-of-the-art results on all 14 diseases of the publicly available *ChestX-ray14* data set [1]. The data consists of 112,120 single grayscale image files and CSV files with metadata like the images' disease labels and bounding boxes indicating the location of the disease. We pre-processed the images by down-scaling them to 224×224 pixels and normalizing the 8-bit pixel values (0 - 255) to float values between $-1$ and 1. Finally, we split the data into train, validation, and test set containing 98,656 (93.5%), 6,336 (6%), and 432 (0.5%) data points respectively.

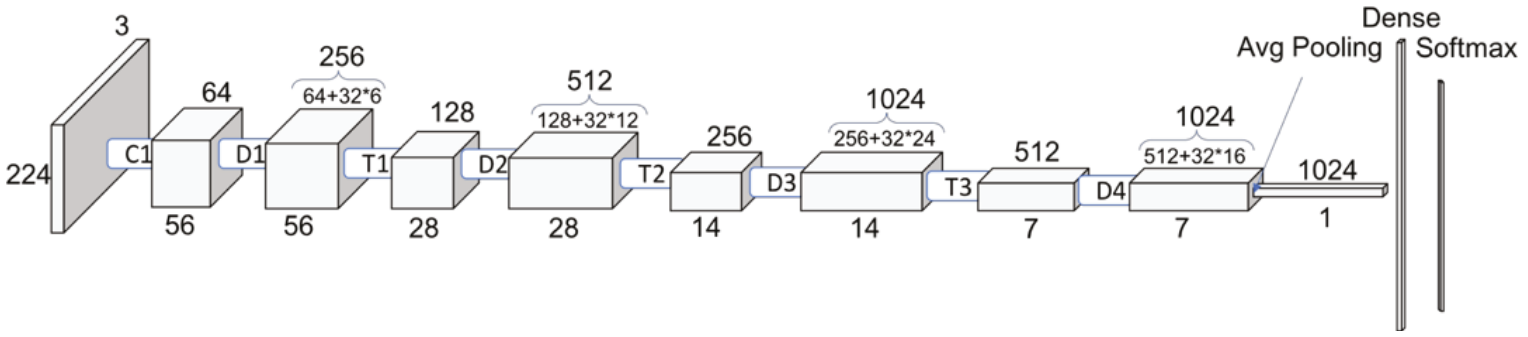

Fig. 2: DenseNet121 architecture with Dense Blocks D and Transition Blocks T; Source: [17]

To replicate CheXNets model architecture, we used a copy of the *DenseNet121* model (depicted in figure 2) coming with TensorFlow's *Keras* API. We changed the input shape of the model to match the monochromatic 224×224 image matrices. Furthermore, we changed the output layer to return 14 values and used the *sigmoid* non-linearity activation function instead of *softmax*, as we want the model to predict independent continuous values to indicate each disease. For our Bayesian version of CheXNet, we replace the deterministic convolutional and fully connected layers with probabilistic layers from the TensorFlow Probability library.

**Training Results** We trained both models with mini-batches by minimizing the binary cross-entropy loss with the Adam optimizer. After 70 epochs, the deterministic model achieved a training AUC of 0.9006 and a validation AUC of 0.8582. We outperform the original CheXNet with an average per-class AUC score of 0.8414 by a small margin. Our Bayesian model failed to achieve good performance just by introducing the probabilistic layers. We stopped the training after 17 epochs as the validation loss started to increase while the AUC decreased to 0.5 which is the random baseline.

### 3.3 Model Improvement Approaches

As our first results show, we couldn't achieve good model performance in this task simply by using Bayesian layers. Although we could get additional benefit from implementing uncertainty measures, a bad performing model isn't practically useful in any real-world application, let alone in disease detection. In the following, we will discuss and test several approaches we took to increase the model's performance.

**Choosing Hyperparameters** In our first approach, we searched for better hyperparameters by training the neural network with different handpicked sets of *learning rates*, *mini-batch sizes*, and *optimization algorithms*. The model that performed best was trained with the Adagrad optimizer, a learning rate of 0.1, and a mini-batch size of 64. An automated search for parameters with *Hyperopt* [18] led to similar results.

**Dealing with Imbalanced Data** As our data set is extremely imbalanced (with significantly more negatives than positives), we introduce a *weighted loss function* that penalizes misclassified positive samples more. This way we can nudge the model towards looking at positive training samples more carefully. This approach, which was also proposed in the CheXNet paper, significantly improved our model's learning and performance on the F1-score and the AUC value.

**Initialization with Pre-trained Weights** The deterministic CheXNet is initialized with weight parameters pre-trained on the *ImageNet* data set. In an attempt to make use of parameter transfer learning in our Bayesian model, we initialize the model's priors with normal distributions with variance 1, mean-shifted towards the single-point parameter values from the pre-trained weights. After training for a while, we concluded that the initialization with weights pre-trained on *ImageNet*, neither improved model performance nor sped up the training.

## 4 Results and Conclusion

| Test set | AUC | F1-score | F2-score | Epistemic | Aleatoric |
|---|---|---|---|---|---|
| Deterministic | | | | | |
| Full | 0.8339 | 0.1444 | 0.1019 | - | - |
| 1 | 0.7552 | 0.2858 | 0.2000 | - | - |
| 2 | 0.6940 | 0.2858 | 0.2000 | - | - |
| 3 | 0.9502 | 0.0000 | 0.0000 | - | - |
| 4 | 0.8523 | 0.3333 | 0.2778 | - | - |
| 5 | 0.9091 | 0.4000 | 0.2941 | - | - |
| 6 | 0.7614 | 0.0000 | 0.0000 | - | - |
| Bayesian | | | | | |
| Full | 0.6579 | 0.1298 | 0.2551 | - | - |
| 1 | 0.5378 | 0.1176 | 0.1923 | 0.0111 | 0.2098 |
| 2 | 0.4544 | 0.1176 | 0.1923 | 0.0143 | 0.2088 |
| 3 | 0.4851 | 0.0571 | 0.1135 | 0.0033 | 0.2149 |
| 4 | 0.6098 | 0.0851 | 0.1695 | 0.0076 | 0.2129 |
| 5 | 0.5739 | 0.0976 | 0.1888 | 0.0109 | 0.2080 |
| 6 | 0.6686 | 0.1499 | 0.2884 | 0.0037 | 0.2166 |

Table 1: Test results of the deterministic and the Bayesian model on each test set.

We assessed the deterministic and the Bayesian models' performance on the test set and additional samples with Gaussian noise. Our deterministic model achieved an AUC of 0.8339, similar to the 0.8414 stated in the original CheXNet paper. This verifies that our implementation of the model and the rest of our deep learning pipeline work as expected. The Bayesian model achieved a lower AUC of 0.6579, as well as lower F1-, and F2-scores. We measured mean *epistemic* uncertainties (”model uncertainty” [7]) of $0,0085$ and mean *aleatoric* uncertainties (”data uncertainty” [7]) of $0,2118$.

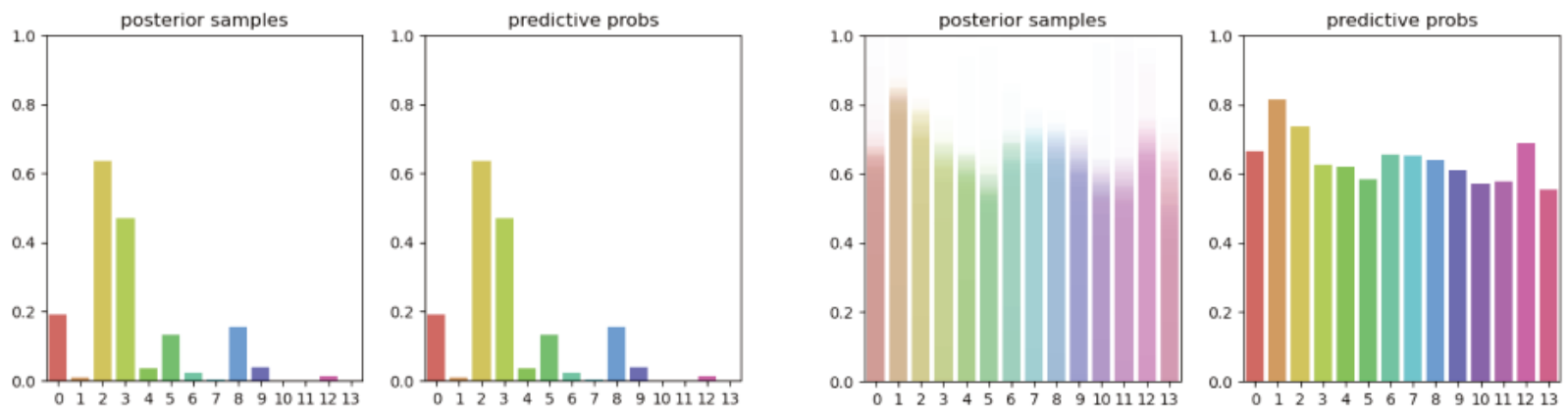

(a) Deterministic model (b) Bayesian model

Fig. 3: The models’ predictions for a test sample.

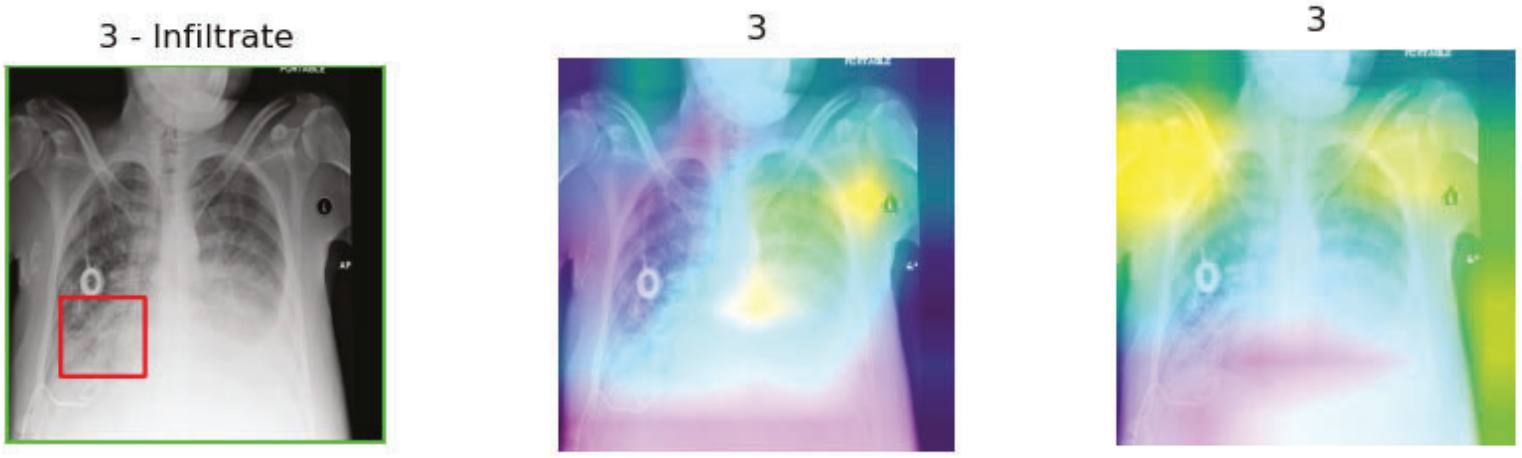

(a) Ground Truth (b) Deterministic model (c) Bayesian model

Fig. 4: The models’ CAMs for class 3 of a test sample and the actual location of the disease.

Figure 3 shows an example of the models’ predicted probabilities for a test sample with true labels 2 and 3. The deterministic model predicted classes 2 and 3 with the threshold of 0.5, while the Bayesian model’s averaged predictions of the 50 predictions sampled show all classes to be true. The generated class activation maps (CAMs) of figure 4 show that both models didn’t look at the right location for their prediction of class 3.

We interpret the much higher aleatoric uncertainty as a result of the nature of radiological images, which can also contain pacemakers and/or other patient-specific aids and the low resolution of the images that were fed into the network. The epistemic uncertainty suggests that the prior for the model should be adjusted. The sampled posterior probabilities range between 0.6 and 0.8 for each class on most of the test samples, which could imply that the model’s weight distributions were initialized with a too high variance that couldn’t be reduced during training. So far, state-of-the-art results with Bayesian neural networks were achieved on simplistic and carefully curated data sets like MNIST and CIFAR-10 moderate deep networks [19]. Our experiments with radiological images and very deep networks didn't achieve state-of-the-art results. This suggests that the complexity of the data, the model size, and/or the initialized variance are the most important factors that can be further analyzed.

# Gaussian Process Inspired Neural Networks for Spectral Unmixing Dataset Augmentation

Johannes Anastasiadis and Michael Heizmann

Institute of Industrial Information Technology (IIIT),
Karlsruhe Institute of Technology (KIT),
Hertzstr. 16, 76187 Karlsruhe, Germany
anastasiadis@kit.edu

**Abstract.** Hyperspectral imaging is increasingly used for product monitoring in industrial processes. Spectral unmixing is an important task in this context. As in many other areas of signal processing, neural networks also provide promising results for spectral unmixing. Unfortunately, it is very time-consuming to prepare labelled training data for the neural networks. To address this problem, this paper presents a method where small training datasets are augmented to improve spectral unmixing performance. Inspired by Gaussian processes, simple neural networks are trained which are capable of generating additional training data. These are similar to the original training data but cover areas in the continuous label space that are not covered by the original data.

## 1 Introduction

Since they are non-contact and non-destructive, optical measurement methods are often used for monitoring industrial processes. This also includes checking for the correct product composition. For this task hyperspectral images are often used because they have a finely sampled spectrum in each pixel characterizing the materials involved [1]. In contrast, conventional colour images are usually not able to solve this problem sufficiently because these only contain three colour channels and different spectra can result in the same colour channel values. Spectral unmixing is needed if more than one material is contained in a pixel and therefore only a mixed material spectrum is available. The aim is to get the relative proportions, the abundances, of the pure materials covered by the pixel [2]. This is often done using mixing models, such as the linear mixing model (LMM), which has proven to be a good approximation. However, depending on the problem, more complex mixing models can provide better results but are also more difficult to use [3]. In addition, there is spectral variability, which can be taken into account by the models using additional parameters.

Instead of a model-based, a data-based approach is also feasible. Artificial neural networks in particular have achieved great success in recent years. This is also true for spectral unmixing and comes with additional advantages [4]. One of them is that the non-negativity and the sum-to-one constraints can be enforced by output layer design. The other advantage is that spectral variability can be taken into account if it is contained in the training data [5]. Ideally, the trained neural networks are robust to spectral variability. To achieve this and a good spectral unmixing performance, lots of significant training data are needed, which are often not available in this domain.

Mainly for classification problems, augmentation strategies are widely used to increase the size of training datasets synthetically and improve performance [6, 7]. Data augmentation can also be used with spectral unmixing, which can be considered as a regression problem. Here it appears useful to generate spectra based on abundances that do not occur in the original training dataset. Ideally, spectral variability is also taken into account by generating many spectra for each abundance set. We have shown in a previous paper that this improves spectral unmixing performance of convolutional neural networks (CNNs) [8]. There we used a generative convolutional neural network with additional random inputs for spectral variability to learn the relationship between abundances and mixed spectra.

In this paper we model the spectra as Gaussian processes with the wavelength as the index and the abundances as parameters. Gaussian processes are defined by the mean and covariance functions. Here we are dealing with functions which depend on the wavelength and are parametrised by the abundances. To learn these functions, we use simple neural networks. To generate the additional training data, further abundances are given to the neural networks. Spectral variability is taken into account by the generation of multiple spectra.

In a previous paper we have already modelled the spectra as Gaussian random vectors [9]. However, that paper was not about dataset augmentation, but about model-based data generation taking spectral variability into account. For the model-based approach, only a set of spectra of each pure material is needed, whereas here we need additional sets of spectra of material mixtures. The additional information should lead to better spectral unmixing performance. Another approach exists where spectral unmixing is achieved by direct application of Gaussian process regression [10], however, not for the augmentation of training data.

The rest of the paper is organized as follows: Section 2 summarises the necessary basics regarding spectral unmixing. Afterwards in Section 3 the proposed approach is described in detail. The evaluation of the approach is given in Section 4. The paper is summarized and a conclusion is drawn in Section 5.

## 2 Spectral Unmixing

This paper deals with supervised spectral unmixing, which assumes that the spectra of the pure substances involved are known [2]. Common spectral unmixing methods are model-based, with the LMM, representing a good approximation in many cases, being the most commonly used [2, 11–14]. There also exist non-linear mixing models [3], which are not considered in this paper. The objective, the estimation of abundances $\hat{\mathbf{a}} \in \mathbb{R}^P$, is achieved using the LMM by

$$\hat{\mathbf{a}} = \arg\min_{\mathbf{a}} \|\mathbf{y} - \mathbf{M}\,\mathbf{a}\|_2^2 \,. \tag{1}$$

Here $\mathbf{y} \in \mathbb{R}^\Lambda$ is a measured spectrum, i.e. a pixel of a hyperspectral image, sampled at $\Lambda$ wavelength channels, $\mathbf{M} = [\mathbf{m}_1, ..., \mathbf{m}_P] \in \mathbb{R}^{\Lambda \times P}$ are the spectra of the up to $P$ involved pure materials, and $\mathbf{a} = [a_1, ..., a_P]^\mathrm{T} \in \mathbb{R}^P$ are the corresponding abundances. The optimisation can be done by calculating the pseudo-inverse. However, constraints must be fulfilled for the abundances in order to remain physically plausible. Those constraints are the non-negativity constraint (2) and the sum-to-one constraint (3).

$$a_p \geq 0 \quad \forall p \tag{2}$$

$$\sum_{p=1}^{P} a_p = 1 \tag{3}$$

The consideration of these constraints counteracts model errors caused by the assumption of a linear mixing behaviour. A well established approach that optimises the LMM considering (2) and (3) is the Fully Constrained Least Squares (FCLS) algorithm [15]. Instead of (1), the Lagrangian $L : \mathbb{R}^{P+1} \to \mathbb{R}$ with the Lagrange multiplier $l \in \mathbb{R}$ is optimized:

$$L(\mathbf{a}, l) = \|\mathbf{y} - \mathbf{M}\,\mathbf{a}\|_2^2 - l\left(\sum_{p=1}^{P} a_p - 1\right). \tag{4}$$

The second part of (4) forces constraint (3). Additionally, negative $\hat{a}_p$ and the corresponding spectra in $\mathbf{M}$ are removed in an iterative procedure to ensure (2) as well.

Until now, the assumption has been made that the pure substances involved can be represented by a single spectrum. However, there is so-called spectral variability. It is caused, among other things, by changing surface conditions and the resulting variation in the angle of illumination [5]. Extended mixing models are available that take spectral variability into account by using additional parameters, such as the extended linear mixing model (ELMM) [16] or the generalized linear mixing model [17]. The ELMM uses the diagonal matrix $\mathbf{B} \in \mathbb{R}^{P \times P}$ to extend the LMM optimization problem to

$$\hat{\mathbf{a}} = \arg\min_{\mathbf{a},\mathbf{B}} \|\mathbf{y} - \mathbf{M}\,\mathbf{B}\,\mathbf{a}\|_2^2\,. \tag{5}$$

After presenting the basics, the next section describes the approach used to augment training datasets. The aim is to improve the performance of spectral unmixing for data-based methods.

## 3 Proposed Approach

The prerequisites for this approach are a set of available spectra for different abundance vectors $\mathbf{a}$. This is quite reasonable in an industrial environment, e.g. in a calibration dataset. The measured spectra are available as vectors $\mathbf{y_a} \in \mathbb{R}^{\Lambda}$ in which each entry corresponds to the reflectance of light for a specific wavelength. For each abundance vector $\mathbf{a}$ there are different measured spectra, which differ due to spectral variability. These are now to be modelled as one Gaussian process $Y(\lambda|\mathbf{a})$ with the wavelength index $\lambda \in \mathbb{N}$ as the index and parametrised with the abundance vector $\mathbf{a}$.

Gaussian processes are completely defined by a mean function and an (auto-)covariance function [18, p. 13]. In this case, the mean value function is

$$m_Y(\lambda|\mathbf{a}) \tag{6}$$

and the covariance function with the second wavelength index $\lambda^* \in \mathbb{N}$

$$k_Y(\lambda, \lambda^*|\mathbf{a})\,. \tag{7}$$

### 3.1 Data Preparation

In order to be able to represent this model with neural networks, the data are prepared. First, for all abundance vectors $\mathbf{a}$, the mean vector (8) and auto-covariance matrix (9) are calculated.

$$\mathbf{m_a} = \frac{1}{N_\mathbf{a}} \sum_{n=1}^{N_\mathbf{a}} \mathbf{y}_{\mathbf{a}n} \tag{8}$$

$$\mathbf{K}_{\mathbf{a}} = \frac{1}{N_{\mathbf{a}} - 1} \sum_{n=1}^{N_{\mathbf{a}}} (\mathbf{y}_{\mathbf{a}n} - \mathbf{m}_{\mathbf{a}}) (\mathbf{y}_{\mathbf{a}n} - \mathbf{m}_{\mathbf{a}})^{\mathrm{T}} \tag{9}$$

Here $N_{\mathbf{a}} \in \mathbb{N}$ denotes the number of measured spectra for a given abundance vector $\mathbf{a}$. The elements of $\mathbf{m}_{\mathbf{a}}$ and $\mathbf{K}_{\mathbf{a}}$ for all available $\mathbf{a}$ can now be used as training data for the neural networks $\mathcal{N}_m$ and $\mathcal{N}_k$ that are supposed to learn (6) and (7).

The neural network $\mathcal{N}_m$ has the abundance vector $\mathbf{a}$ and the wavelength index $\lambda$ as input variables and as the output variable the corresponding value of $\mathbf{m}_{\mathbf{a}}$. The neural network $\mathcal{N}_k$ has the abundance vector $\mathbf{a}$ and the transformed indices $\lambda_1' = \lambda + \lambda^*$ and $\lambda_2' = \max(\lambda) - |\lambda - \lambda^*|$ as input variables and as the output variable the corresponding value of $\mathbf{K}_{\mathbf{a}}$. The indices $\lambda_1'$ and $\lambda_2'$ are used because of two properties that the covariance matrices have. Firstly, there are higher values on the main diagonal, and secondly, they are symmetrical. This results in the neural network being kept quite simple later on, as it has to learn fewer changes in monotonicity (see Fig. 1).

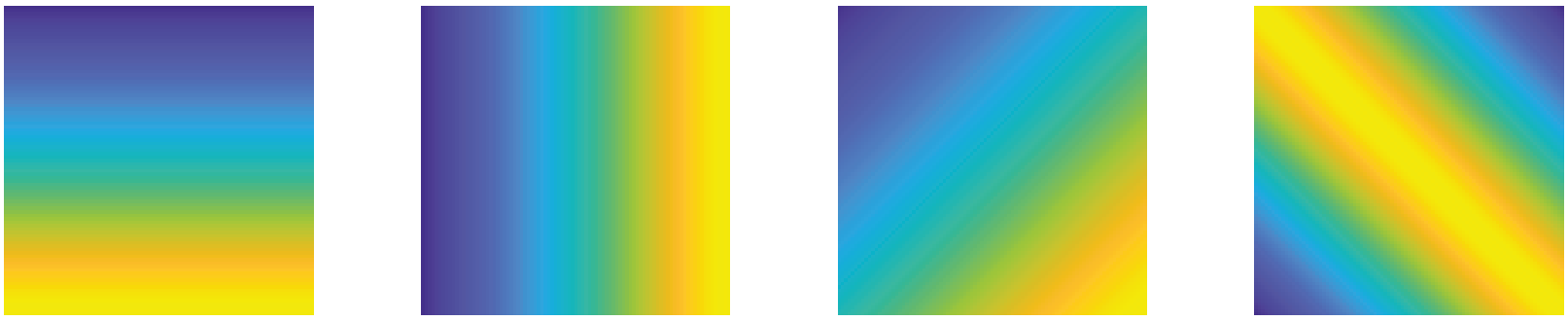

**Fig. 1.** Illustration of the values of the indices in the auto-covariance matrix (from left to right): $\lambda$, $\lambda^*$, $\lambda_1'$ and $\lambda_2'$. Dark blue denotes a low value, yellow a high value

### 3.2 Neural Networks for Data Augmentation

The neural networks can now be trained with the prepared training data as described above. The neural networks each have $P+1$ ($\mathbf{a}$ and $\lambda$) or $P+2$ ($\mathbf{a}$, $\lambda_1'$, and $\lambda_2'$) inputs and only one output. To learn the desired relation a quite simple neural network is sufficient.

The networks consist of fully connected layers, i.e. layers in which all neurons of one layer are connected to all neurons of the neighbouring layers. The rectified linear unit (ReLU) $f_{\mathrm{relu}}(z) = \max(0, z)$ is used as the activation function in all layers, with the exception of the last layer, where the logistic function

$$f_{\log}(z) = \frac{1}{1 + \mathrm{e}^{-z}} \tag{10}$$

is used. Batch normalisation is carried out prior to the rectified linear units [19]. The networks $\mathcal{N}_m$ and $\mathcal{N}_k$ have the same structure, which is shown in Fig. 2. The logistic loss function is used as objective function:

$$-\frac{1}{B} \sum_{b=1}^{B} o_b \cdot \log(\hat{o}_b) + (1 - o_b) \cdot \log(1 - \hat{o}_b)\,. \tag{11}$$

It is evaluated for each output value $\hat{o}_b \in (0, 1)$ and corresponding label $o_b \in (0, 1)$ of a training batch of size $B \in \mathbb{N}$. The logistic loss function is often used for a two-class

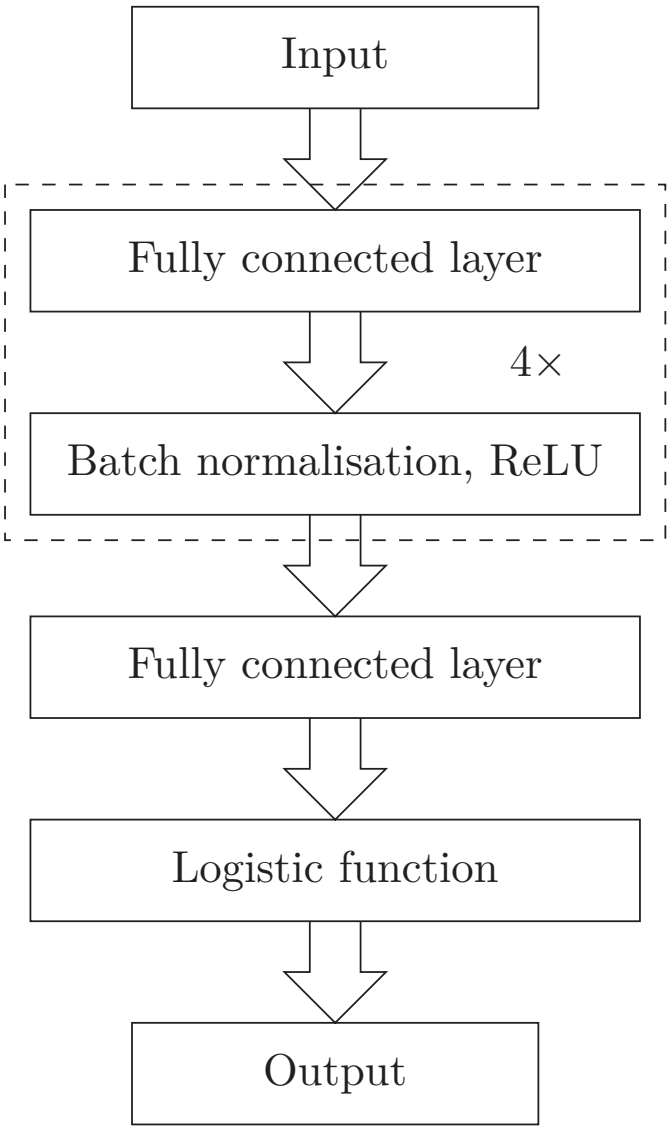

**Fig. 2.** Schematic representation of the neural network: There are four blocks consisting of a fully connected layer, batch normalisation and a ReLU activation function. This is followed by a fully connected layer with the logistic activation function

classification problem (cross-entropy loss). However, it also works with continuous labels and is suitable here because the values of the spectra range between 0 and 1.

Using the trained neural networks $\mathcal{N}_m$ and $\mathcal{N}_k$, an augmentation of the original training dataset can now be performed.

### 3.3 Augmentation Strategy

For the augmentation, additional mean value vectors and covariance matrices can now be generated by specifying abundance vectors for $\mathcal{N}_m$ and $\mathcal{N}_k$ that do not occur in the original training dataset. This allows pseudo-random generators to be used to produce spectra that complement the original training dataset. The spectra generated in this way also show spectral variability.

In order to augment the datasets at a lower effort a second strategy is used, where the original training datasets are only augmented by the mean value spectra. The neural network, which is later used for spectral unmixing (see Section 4), then has to learn the spectral variability on basis of the already existing training data.

The spectral unmixing performance of the augmented datasets is compared with that of the non-augmented datasets.

## 4 Experimental Results

Preceding the evaluation, the parameters used for $\mathcal{N}_m$ and $\mathcal{N}_k$ and the evaluation datasets are presented. The number of neurons was determined to be 32 for all layers and in both cases ($\mathcal{N}_m$ and $\mathcal{N}_k$). The neural networks were trained with the Adam optimizer [20]. The parameters from [20] were used, except for the learning rate, which was set to 0.01. The number of epochs was set to 2000 (both networks) for the datasets containing mixtures of quartz sand (see below) and to 3000 ($\mathcal{N}_m$) and 4000 ($\mathcal{N}_k$) for the colour powder dataset.

### 4.1 Datasets

Three datasets are used, which were recorded in our image processing laboratory. This ensures that we know the abundances as accurately as possible. All datasets consist of fine powders. These were mixed according to the specified abundances until the mixtures were homogeneous. A white balance with a reflectance standard was carried out after the recordings of the hyperspectral images, which compensates both spatial and spectral inhomogeneities of illumination and measurement setup. All datasets were acquired in 91 wavelength channels, ranging from 450 nm to 810 nm. For each mixture, 400 samples were acquired.

Two of the datasets contain mixtures of coloured quartz sand. The first of them (quartz-3) contains 45 mixtures of at maximum 3 components varying in abundance steps of 0.125 . The other one (quartz-4) includes 56 mixtures of at most 4 components, varying in abundance steps of 0.2 . The quartz sand datasets have a lower spectral variability and the non-linearity in the mixing behaviour is less significant compared to the following dataset. The third dataset consists of 56 mixtures of colour powders (colour-4), which also have up to 4 components. Again, the components are varied in abundance steps of 0.2 . The colour-4 dataset shows a high non-linearity between mixed spectra and the spectra of the pure substances and a high spectral variability. Hence, its spectra are more difficult to unmix.

All three datasets are divided into a test and a training dataset according to the abundances. For the datasets with four components, the samples with no abundance of value 0.2 or 0.8 are included in the training dataset. All other samples are included in the test dataset. This yields 16 abundance vectors in the training and 40 in the test dataset. The quartz-3 test dataset includes those where at least one abundance has the value 0.125, 0.375, 0.625 or 0.875. In consequence, there are 30 abundance vectors in the test dataset and 15 in the training dataset.

### 4.2 Evaluation of Generated Data

Using the neural networks from Section 3, new data are generated. The abundance vectors used as inputs are exactly the same as those in the test dataset. For each given abundance vector 400 spectra are generated, which correspond to the number of spectra per abundance vector in the test dataset.

As a measure of performance, the average minimum norm $\Delta_{\mathrm{AMN}}$ is used between $I$ measured spectra $\mathbf{y}_i$ and $H$ generated spectra $\hat{\mathbf{y}}_h$ corresponding to an abundance vector:

$$\Delta_{\mathrm{AMN}} = \frac{1}{I} \sum_{i=1}^{I} \min_h \|\mathbf{y}_i, \hat{\mathbf{y}}_h\|_2 \,. \tag{12}$$

This performance measure was chosen because it tests whether a spectrum was generated as similar as possible to each spectrum in the test dataset. The calculation is done separately for each abundance vector and the corresponding spectra. The mean value of all $\Delta_{\mathrm{AMN}}$ over all abundance vectors in the test dataset is called global average minimum norm $\Delta_{\mathrm{GAMN}}$ (see Table 1). The results of both proposed augmentation strategies are compared with the performance of the generative convolutional neural network (Gen. CNN) with and without covariance matrix regularisation (-CovR) we presented in [8].

**Table 1.** Comparison of $\Delta_{\mathrm{GAMN}}$ for all test datasets. In the first two columns the results from [8] are listed for comparison. The third column presents the values for the proposed method and the last column uses only the generated mean vectors $\hat{\mathbf{m}}_{\mathbf{a}}$ as generated spectra

| $\Delta_{\mathrm{GAMN}}$ | Gen. CNN [8] | Gen. CNN-CovR [8] | Proposed (normal) | Proposed (mean only) |
|---|---|---|---|---|
| quartz-3 | 0.1219 | 0.0812 | 0.0993 | 0.1371 |
| quartz-4 | 0.1113 | 0.0787 | 0.0903 | 0.1298 |
| colour-4 | 0.1242 | 0.0967 | 0.1016 | 0.1470 |

Table 1 shows that the inclusion of the covariance matrices results in lower $\Delta_{\mathrm{GAMN}}$ values for all datasets compared to only using the mean vectors. This is because spectral variability is taken into account. The results of the proposed method are better than those of the unmodified generative CNN. However, the best results were achieved with the generative CNN with covariance matrix regularisation. It is also noticeable that within a method, the order of the datasets regarding $\Delta_{\mathrm{GAMN}}$ always remains the same, which is due to the difficulty of the datasets.

In the next subsection, it will be investigated whether these results are consistent with those of spectral unmixing using augmented training datasets.

### 4.3 Spectral Unmixing Performance

For evaluation of the spectral unmixing performance, we use the same CNN as in [8], of which we have already presented the three-dimensional version in [4]. The CNN is trained with the original training dataset as well as with different augmented training datasets. The performance with respect to the test dataset is compared below. The CNN for spectral unmixing consists of three convolutional layers with a convolutional kernel length of 3. Then two fully connected layers follow. The numbers of feature maps from the input layer to the output layer are 1, 16, 32, 64, 64 and 1. We use the root-mean-square error over all $N$ samples of a test dataset

$$\Delta_{\mathrm{RMSE}} = \sqrt{\frac{1}{N}\sum_{n=1}^{N}\frac{1}{P}\sum_{p=1}^{P}(\hat{a}_{pn} - a_{pn})^2} \tag{13}$$

as a performance measure. For the methods that are not based on neural networks (see Section 2), the results are shown in Table 2 for the sake of clarity. For the remaining methods $\Delta_{\mathrm{RMSE}}$ is displayed in Figure 3.

To obtain the results below, the network was trained with different numbers of epochs depending on the dataset and method. The quartz-3 dataset was trained for 251, the quartz-4 dataset for 41 and the colour-4 dataset for 31 epochs for the proposed method. When only mean vectors are used for augmentation, the number of epochs reduces to 31 (quartz-4) and 21 (colour-4). As a reference, we use the non-augmented training dataset, that was trained for 81 (quartz-3), 21 (quartz-4) and 21 (colour-4) epochs. The different numbers of epochs are chosen to avoid overfitting.

The original training datasets were augmented with a different number of spectra. Figure 3 shows the step size s $\in [0, 1]$ in which the additional abundance vectors were varied to generate the new data. All possible abundance vectors corresponding to the step size s are used in spectra generation, except for those already contained in the original training dataset.

**Table 2.** Comparison of $\Delta_{\mathrm{RMSE}}$ for all test datasets for FCLS and ELMM based spectral unmixing

| $\Delta_{\mathrm{RMSE}}$ | quartz-3 | quartz-4 | colour-4 |
|---|---|---|---|
| FCLS | 0.1608 | 0.1115 | 0.2987 |
| ELMM | 0.1555 | 0.1056 | 0.2990 |

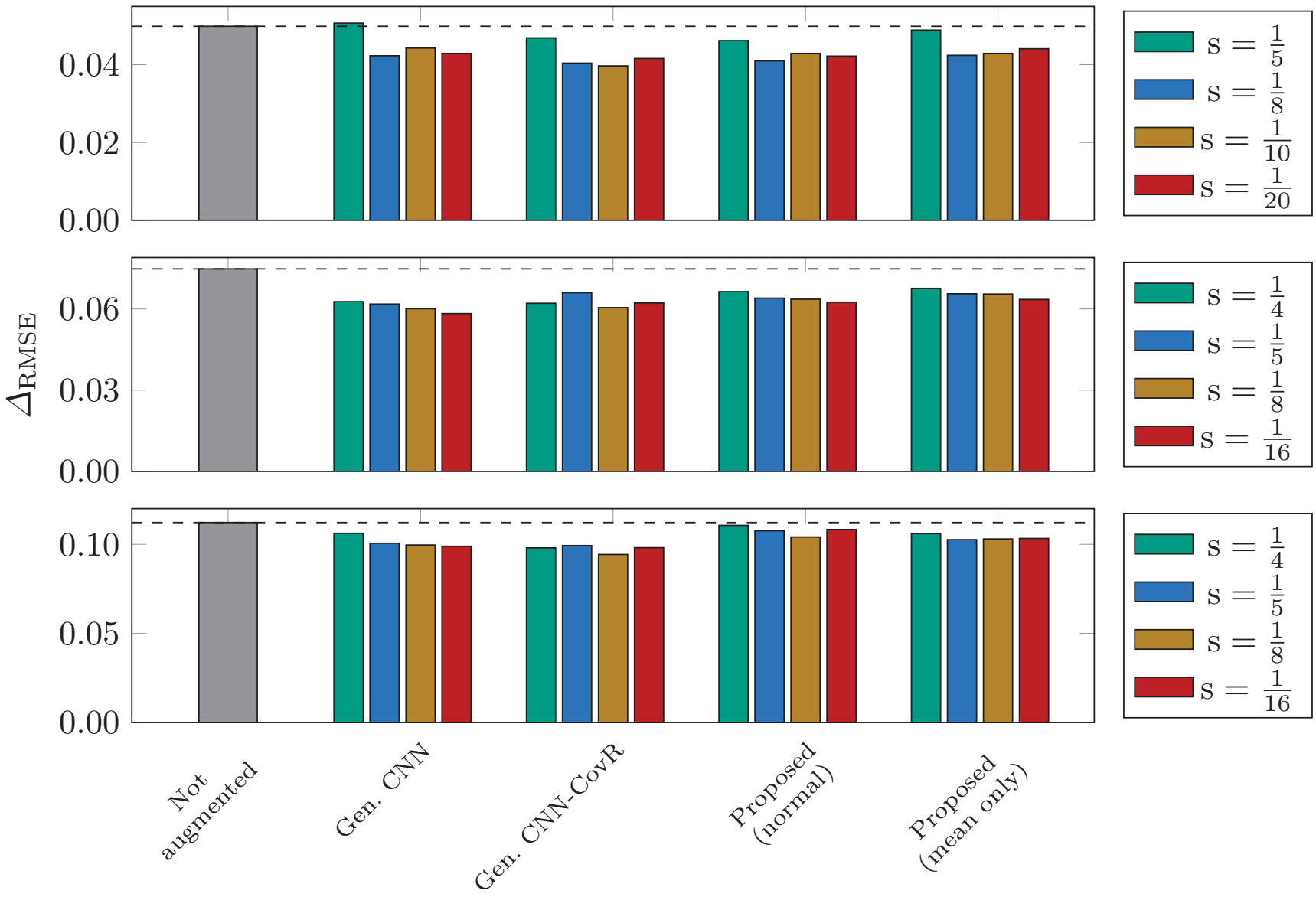

**Fig. 3.** Comparison of $\Delta_{\mathrm{RMSE}}$ for the test datasets of the quartz-3, quartz-4, and colour-4 datasets (top to bottom) for CNN-based spectral unmixing using different augmentation strategies. The dashed lines are used for a better visual comparability with the non-augmented case

It is shown that the data-based spectral unmixing methods (Figure 3) perform better than the model-based methods (Table 2). Gaussian process based augmentation leads to an improvement compared to the non-augmented training dataset for all datasets and all step sizes s. The size of s does not have a major influence, unless it is chosen too large, in which case the performance deteriorates. If only the mean spectra are used for augmentation, the results are comparable. This is probably due to the fact that the spectral variability does not depend too much on the abundances and is already well represented by the spectra available in the original training dataset. For the colour-4 dataset, the result is worsened by adding the information from the covariance matrices. In this case, the assumption of a Gaussian process is likely to be an oversimplification.

The results from [8] cannot be reached with this approach. However, the training time of the neural networks for augmentation is approximately 9 times[1] shorter. On the one hand, this is due to the lower dimensional data points and therefore simpler neural networks. On the other hand, the size of the training dataset for augmentation is reduced if only the described moments are used as training data. The latter is especially true if only the mean spectra are used. In this case it is only one spectrum per abundance vector instead of $N_{\mathbf{a}}$. This leads to an approximately 120 times[1] shorter training time compared to [8].

## 5 Conclusion

In this work, an approach to augment training datasets for spectral unmixing was presented. For this purpose, inspired by Gaussian processes, a mean and a covariance function are learned by two neural networks. These networks are then used to generate additional training data.

It was shown that the performance of spectral unmixing with a CNN can be improved by the additional training data generated by these neural networks. It depends on the dataset how significant the improvement is. The improvement is slightly lower as with an existing method that uses a generative CNN for augmentation. However, the training time is an order of magnitude shorter. If only the neural network for the mean value function is used, where a similar increase in performance was observed depending on the dataset, the training time decreases by another order of magnitude.

In the future, something in between the two approaches presented would also be feasible. There, only the more relevant parts of the covariance functions would be used.

[1] Training performed on NVIDIA Quadro P5000.

# R&D of a Multisensory System for Excavation Machines for the Real-time Generation of AI/ML classified, Georeferenced and BIM compliant Voxel Models of Soil (ZIM Project HOBA)

Almagboul, M., Anantha, P. C., Jäger, R. and G. P. Sridhar

Karlsruhe University of Applied Sciences, Center of Applied Research (CAR)
**reiner.jaeger@web.de**

**Abstract.** The R&D project "Homogeneous soils assistant for the automatic, construction site-specific recording of soil classes according to the new VOB 2016", shortly HOBA, deals with the development of automatic classification, detection & segmentation of construction site specific soil types (http://www.navka.de/index.php/de/weitere-projekte/hoba-project-overview-1). HOBA is financed as a so-called ZIM (Central Innovation Program for SMEs) research and development project by the Federal Ministry for Economic Affairs and Energy (BMWI). Research and development are carried out in the GNSS & Navigation Laboratory (http://goca.info/Labor.GNSS.und.Navigation/index.php) in collaboration with the main industry partners MTS Schrode AG (www.mts-online.de) and VEMCON GmbH (www.vemcon.de).

The aim of the R&D at the HKA is the development of the hardware and software of a compact sensor- and computing system unit, mounted on the excavator, briefly called HKA HOBA-Box (fig. 1). The hardware and software development of the HKA HOBA-Box is an innovative contribution to the BIM-compliant digital real-time documentation of excavation work in civil engineering.

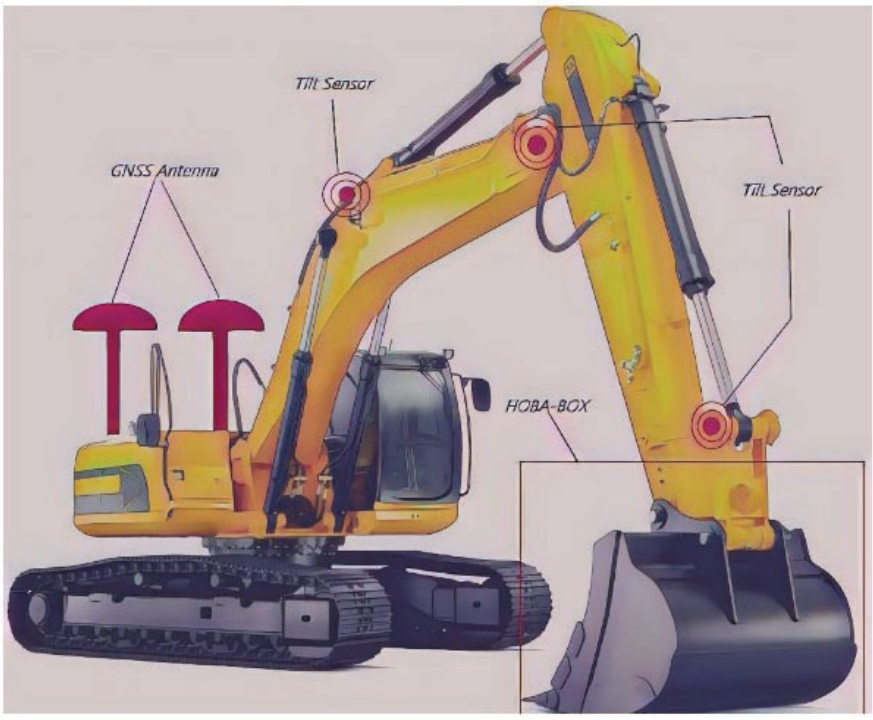

Figure 1 Excavator with distributed sensors and HKA HOBA-Box located in the area of the shovel

The HKA HOBA-Box enables a multi-sensory 3D geo-referencing of the excavation in the ETRF89 / ITRF, based on GNSS/MEMS/Optics (RGB/ToF-camera) sensors. The complete geo-referencing steps of the box are based on a Bayesian sensor-fusion algorithmic fusion in the general NAVKA (www.navka.de) multisensory-multiplatform lever arm design, leading to the full navigation state vector

$$y(t) = (x^e y^e z^e | \dot{x}^e \dot{y}^e \dot{z}^e | \ddot{x}^e \ddot{y}^e \ddot{z}^e | r^e p^e y^e | \omega^b_{eb,x} \omega^b_{eb,y} \omega^b_{eb,z} | \dot{\omega}^b_{eb,x} \dot{\omega}^b_{eb,y} \dot{\omega}^b_{eb,z})^T$$

The Bayesian SLAM (Simultaneous Localization and Mapping) means an extension of $y\ (t)$,, which is based - in the case of HOBA - on the Gauß-Markov-model's information of the optical sensor data of the ToF and the RGB camera. The extension of the parameter space for SALM is on the 3D map $m\ (t)$, and so it leads to $y(t)_{SLAM} = (y(t), m\ (t))$.

The SLAM parameters $y(t)_{SLAM} = (y(t), m\ (t))$ are used for the computation of a so-called ETRF89 / ITRF "voxel"-based 3D model of the excavation volume, based on an Octree-based representation of the environment, using the ToF-camera data.

Furthermore, the HKA HOBA-box will allow the classification of the soil types on the site using image-based AI/ML algorithms, and finally the re-calculation of the classified and geo-referenced 2D images onto the geo-referenced 3D voxel model according to the soil types. Here pixels to pixels in the image have to be assigned to the 3D surface on the model by calculating correspondent points (P, P $'$ ) between the two representations, i.e. 2D and 3D. The part of Machine Learning (ML) introduces a real-time soil texture classification and segmentation on the construction site. The R&D target of the abovementioned BIM-compliant soils assistant, according to the new VOB 2016 - realized as a physical system by HKA HOBA-box - enables civil engineers an insight into the geo-technical properties of the excavated soil, as well as to the borders between these soil types, on the construction site. The mathematical and algorithmic approach for soil classification is the Fully Convolutional Network-Based Semantic Segmentation (FCN). This is a pixel-level classification, which achieves a pixel fine-grained inference. Therefore, it is then the best practice for such application of soil texture labelling on the site.

The improved FCN-based approaches e.g., SegNet & DeepLap, are used as encoders for the training of the soil texture-model via Transfer Learning (TL). The TL is utilized, as it is a well-known method and regarded to shorten the processing time of model training, and therefore decrease the computational efforts considerably. Furthermore, the HKA HOBA-box trains the model on a specific dataset already, as provided by the HKA industry partner i.e. MTS Schrode AG. The dataset i.e., images of the construction site with different present soil types, provides - through a high-precision 2D annotation process - i.e. instance-aware segmentation and detailed precise masks for a corresponding 15 classes of the soil types, which are used for training, testing, and validation.

The respective AI/KL developments are based on Python & C/C++, using both machine learning frameworks, namely TensorFlow 1.x and PyTorch. The re-trained model is converted into an ONNX (Open Neural Network Exchange), which is an open format for ML models allowing interchanging between different ML frameworks.

# Predictive prognostic for Li-Ion batteries in electric vehicles

Inès Jorge, Ahmed Samet, Tedjani Mesbahi, and Romuald Boné

ICube, CNRS (UMR 7357)
INSA Strasbourg – University of Strasbourg
ines.jorge@insa-strasbourg.fr

**Abstract.** The development of clean vehicles and more specifically electric and hybrid vehicles relies on the performances of Lithium Ion batteries. More efficient than all the other battery chemistry in terms of energy density and output power, these batteries bring hybrid and electric vehicle in line with thermal vehicles. However, they still suffer from a limited driving range and lifespan, and their performances can be affected by numerous factors, one of the most important one being the driving profile imposed by a user.
Prognostics and health management strategies make use of operating data in order to better understand the ageing mechanisms of Lithium Ion batteries and to forecast their future degradation trend. In this article, we introduce our method to predict the Remaining Useful Life of Lithium Ion batteries based on the dataset published by the Massachusetts Institute of Technology, through the use of low computational cost machine learning algorithms. Our artificial neural networks take both historical data and time series representing the driving profile of a battery as input, and predict with accuracy the Remaining Useful Life of a battery. Compared to previous approaches in the literature, we obtain reliable and accurate predictions of the Remaining Useful Life of any battery at any moment in its life from the observation of only charge and discharge cycle. The importance of driving data in prognostics and health management strategies of Lithium Ion batteries is shown throughout this article.

## 1 Introduction

In the case of hybrid vehicles (HEV), and even more so in the case of all-electric power-trains, the on-board energy storage system remains the weak link: very expensive, limited in driving range, slow to recharge, main cause of over-costs... The challenge for any car manufacturer wishing to develop a HEV or an Electric Vehicle (EV) is therefore not only to optimise the electric power-train, both in terms of cost and range, but also to bring the battery into line with the life of the vehicle. Battery lifetime is therefore a crucial element for the development of EVs under acceptable cost conditions. Indeed, the battery is the key component and the most expensive one in a HEV or EV. In this context, the failure of battery could lead to serious inconvenience, performance deterioration, accelerated ageing and costly maintenance.
Therefore, the prognostics and health management (PHM) of on-board energy storage systems, which aims to monitor their health and to predict their degradation trend, appears to be a crucial element in the development of new battery powered vehicles.

PHM strategies make it possible to forecast the evolution of the storage capacity of a battery and to predict its Remaining Useful Life (RUL), which correspond to the number of charge and discharge cycles it can withstand before reaching its end of life. That allows to perform maintenance service in advance if necessary, using the past and current information about battery usage and capacity degradation trend.
The aim of this article is to present a method based on machine learning for the predictive prognostics of Li-Ion batteries in EV applications. The challenge is to use ageing data of Li-Ion batteries in order to extract knowledge on the state of health (SOH) of the batteries. In this paper, we focus specifically on the dataset published in [1] as it is the largest available and contains extremely valuable data that apply very well to machine learning techniques. The key contributions are (*i*) the development of low computational models based on Artificial Neural Networks (ANN), (*ii*) an online forecasting of the RUL of batteries, (*iii*) the use of driving data in the predictive model.

The remainder of this article is structured as follows : section 2 is a brief introduction to related work in the fields of predictive prognostics of Li-Ion batteries, section 3 provides a detailed presentation of the dataset on which we based our work, section 4 focuses on the developed architecture for RUL prediction based on ANN and feature extraction, section 5 presents all experimental results with a comparison with other approaches found in the literature and section 6 offers a brief conclusion with a presentation of future works.

## 2 Background

Concerning Li-Ion batteries, prognostics and health management strategies (PHM) aim at determining how and when a failure will occur and to give a long term image of the state of health (SOH) of the battery [2]. This can be done either by observing previous data acquired through various sensors or by simulating the behaviour of a battery in its environment thanks to physical models.

In a great majority of papers, PHM of Li-Ion batteries consists in determining their RUL, which is the number of charge and discharge cycles it can go through before reaching the End of Life (EoL) criteria. A battery is considered out of use for an electric vehicle when it has reached a SOH of 80%. The SOH of a battery represents the storage capacity at a given time compared to its initial storage capacity :

$$SOH = \frac{Q_{actual}}{Q_{nominal}} \quad and \quad SOH\% = \frac{Q_{actual}}{Q_{nominal}} * 100 \tag{1}$$

Most approaches deal with the prediction of the RUL in terms of cycles. This can be done either by designing a complete physical model and simulating the behaviour of a battery, or by focusing on real data taken as input of machine learning models. This latter type of models makes it possible to forecast the temporal evolution of the battery SOH using a sliding window approach, or to predict the RUL directly from the observation of ageing features.

### 2.1 Model based approaches

Model-based techniques were the first ones to be developed, before massive data acquisition and challenges linked to big data appeared. A model-based approach for the PHM of a system relies on the establishment of a simulation model according to physical rules and functioning equations. The aim is to understand and reproduce the behaviour of a system in order to obtain simulated data that could be exploited, in particular with the introduction of disturbances. It implies a complete understanding of the system and gives a global representation of the different answers to solicitations. Downey **et al.** in [3] have modeled the degradation phenomena of active materiel loss in Li-Ion batteries in order to estimate the battery capacity. The battery was represented by an electrochemical model that takes into account heat generation equations in [4]. Zhang *et al.* [5] elaborated a comprehensive lead-acid battery model made of seven sub-models each modeling a physical phenomena. The model estimates internal resistance, terminal voltage, internal temperature, SOC and battery capacity using the load current and ambient temperature.

### 2.2 Data driven approaches

Data driven approaches for PHM have emerged with the development of industry 4.0 and massive data acquisition strategies. Real operating data is collected and given as input to a black box model, that uses past data to forecast the evolution of a system. Operating data most of the time consists in physical features observed according to time through different sensors linked to a battery or a battery cell during ageing tests (observation of current, voltage, internal resistance, temperature...). This results in very large sequences of data as the cycle life of a Li-Ion battery can reach more than 2000 cycles. All data driven strategies require a data prepossessing step in order to make operating data compatible with data driven models. However, there can be a great variety of approaches when building a data driven predictive model, mainly due to the development of machine learning algorithms that apply very well to large amounts of data and PHM problematic. This quick state of the art of data driven approaches separates the different models found in the literature into two categories : window-based models and early cycle models.

**Window approach** As explained earlier, operating data of Li-Ion batteries can result in very larges sequences of data due to their very high lifespan. A common approach for simplifying the problem is to use a window approach. There are two types of data sequences in Li-Ion battery ageing data. The first one are historical data sequences, which are represented as a function of the number of cycles. For example, SOH is computed at each cycle, just as internal resistance or charging time. For each of these data sequences, there is one value for each cycle, and a sequence window is therefore composed of several consecutive cycles. The second type of data sequence are temporal data, which are represented as a function of time. Here, operating data is directly acquired though sensors, and for each cycle, the temporal evolution of several features such as current, voltage or temperature can be observed. Temporal data represent the real use of the battery. A window of time sequences can then either be a sample of time series from one cycle, or a succession of time series that corresponds to several consecutive cycles. Most approaches deal only with the observation of window of historical data, and especially the SOH. The evolution of SOH contained in one window makes it possible to forecast the future degradation trend and therefore to predict the RUL according to the predicted SOH fade. This method has proved very effective and can be applied to a great variety of ML techniques [6–10]. However, the main drawback is that the accuracy of the prediction depends on the size of the window. The larger the amount of historical data, the better the accuracy. Moreover, very few approaches take advantage of time series.

**Early cycle prediction** Some approaches mention other RUL prediction techniques based on features calculated from early cycles data. Severson *et al.* have computed several features from cycle 1 to cycle 100 and applied a linear regression as a supervised learning technique to predict the cycle life of a given cell. This method also removes the problem of dealing with temporal or sequential values but requires to use only brand-new batteries, after cycling them 100 times.

# 3 Battery ageing data

Throughout the literature, several datasets are often cited and used for data driven approaches concerning PHM of Li-Ion batteries. The NASA Prognostics Center of Excellence (PCoE) published a massively used dataset for SOH prediction [11]. It consists in 34 batteries tested under different charge and discharge conditions until the EoL criteria is reached. More batteries were tested in this dataset without reaching end of life though, which limits the applications.

The NASA PCoE also published a dataset that consists in testing 4 batteries with random charge and discharge currents during 1500 cycles [12]. After 1500 cycles, characterisation cycles are performed in order to evaluate the evolution of the batteries' SOH.

An other dataset published by the Sandia National laboratories aims at studying the effect of Depth of Discharge (DoD), load current and temperature on battery degradation. 86 cells of three different chemistries (LFP, NMC and NCA) are tested in this study.

Some papers also mention custom battery ageing datasets [13]. The main drawback when testing batteries for health prognostics is that EoL criteria need to be reached. Considering the performances of Li-Ion batteries, this may take a long time and require a lot of resources for testing a representative number of cells. To overcome the resource and time problem, a paper also shows the use of training data generated with a physic-based model of Li-Ion battery [14].

Even though all this data is of great interest, we decided to base our approach on a new dataset described in [1] (with supplementary information in [15]). This dataset gathers more information than all earlier mentioned datasets, to our knowledge, as it offers complete operating data of 124 cells tested from beginning to EoL.

## 3.1 MIT dataset

In [1], the department of chemical engineering of the Massachusetts Institute of Technology, in collaboration with Toyota engineering and with the Department of Materials Science and engineering of Stanford university, have built the largest available dataset regarding Li-Ion battery ageing. This dataset is a highly valuable source of information as very few public data can provide that much resource. The cells that were used for testing are LFP/graphite cells from A123 manufacturer, model APR18650M1A. These cells have a 3.3V nominal voltage and a 1.1Ah nominal capacity. They can provide discharge currents up to 30A.

The cells were tested in a 30°C chamber and cycled with an battery tester from Arbin manufacturer. The batteries are always discharged at a constant current of 4.4A. The most important factor in the tests is the charging policy. Batteries are charged following a multi-steps constant-current/constant-voltage (CC-CV) policy which makes it possible to reduce the charging time. By applying a fast charging policy, batteries are tested under conditions that are close to the real use of batteries in an EV. Indeed, one of the main challenges that EV are facing is the charging time, which should be as short as possible without damaging the cells.

As explained in section 2.2, there are two different kind of data sequences in this dataset : historical data sequences and times series.

Figures 1, 2 and 3 are representations of time series for one given cycle (the charging pattern, evolution of external temperature during one cycle, discharge voltage...) and figures 4, 5 and 6 show the global evolution of a given historical data sequence over the full life cycle of a battery.

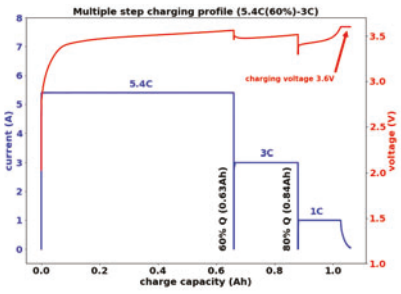

**Fig. 1.** Charging pattern

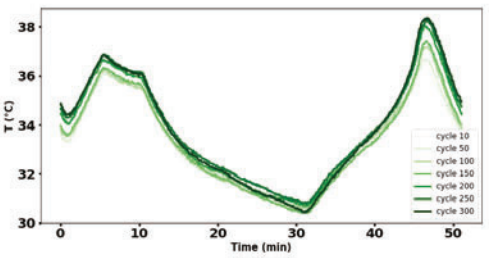

**Fig. 2.** Cell T°

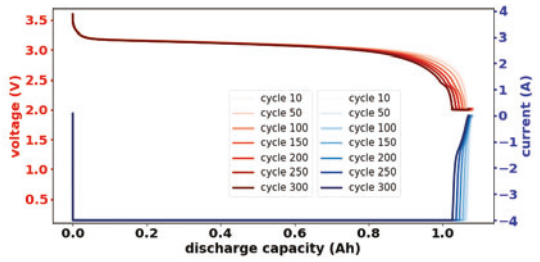

**Fig. 3.** Discharge V and I

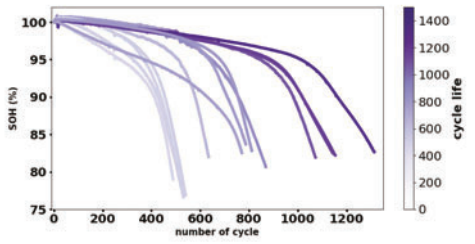

**Fig. 4.** Capacity fade

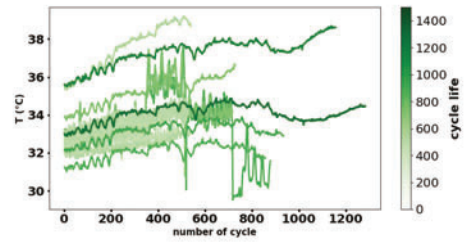

**Fig. 5.** Average cell T°

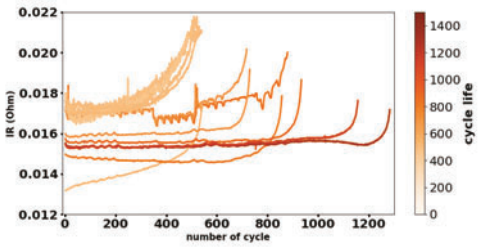

**Fig. 6.** Internal resistance

The available dataset offers a considerable amount of ageing data from the first cycle of each cell to the EoL. Every cycle gives information about ageing signs and SOH that should be taken into account. Both historical data and time series contain information about the RUL and SOH of the battery at a given time, but they need to be pre-processed and combined in order to highlight the factors that most represent the degradation trend of a battery.

### 3.2 Exploitation of driving data

As detailed in section 2.2, most approaches are based on the exploitation of SOH historical curves only to forecast the future SOH degradation trend [7, 16, 13]. We see two major drawbacks in designing a SOH forecasting model based on previous SOH data only. First, as all available data consists of experimental data built from laboratory cell tests, the degradation trend is quite steady. Indeed, in the MIT dataset, cells are discharged at a constant current rate, identically throughout their whole cycle life. Similarly, the charging protocol does not vary from beginning to end of life. This results in SOH degradation patterns that are very similar from one battery to another, as can be seen in fig 4. Therefore, forecasting future SOH degradation trend from past SOH data is simplified and can be implemented with most machine learning algorithms.

Secondly, studying the global trend of SOH can give a good idea of the long term degradation but could not make it possible to catch local variations due to a specific use of the battery. Current (I), voltage (V) and temperature (T°) time series reflect the real use of the battery : I and V curves represent the driver solicitations (acceleration, speed, breaks...) and T°brings information about the environment in which the battery is used (cold or warm weather, night or day etc . . . ). Therefore, we believe that using driving data as input to our PHM model is crucial in understanding all possible causes of deterioration.

### 3.3 Training dataset

The RUL of a battery decreases at each cycle. Our target is to predict the RUL of a battery at any given cycle, focusing on only one cycle. That means that in this paper, a window based approach is used, where the window size is of one cycle. RUL can be calculated for each cycle following equation 2 :

$$RUL = k_i - n \tag{2}$$

where $k_i$ is the cycle life of cell number $i$ and $n$ is the observed current cycle of the cell.

Our approach mixes the use of historical data and temporal series. These two types of data can't be used directly together as historical data have one scalar value per cycle and time series have one vector of varying length per cycle. Therefore, before using time series in our model, a feature extraction technique is used to condense the information contained in each vector into one scalar value. For example, several features are computed from each time series such as the root mean square value, area under the curve or average value. Computed features from time series can then be exploited as input to any given model in the same way as historical data. The training dataset as used in our approach is represented in figure 7.

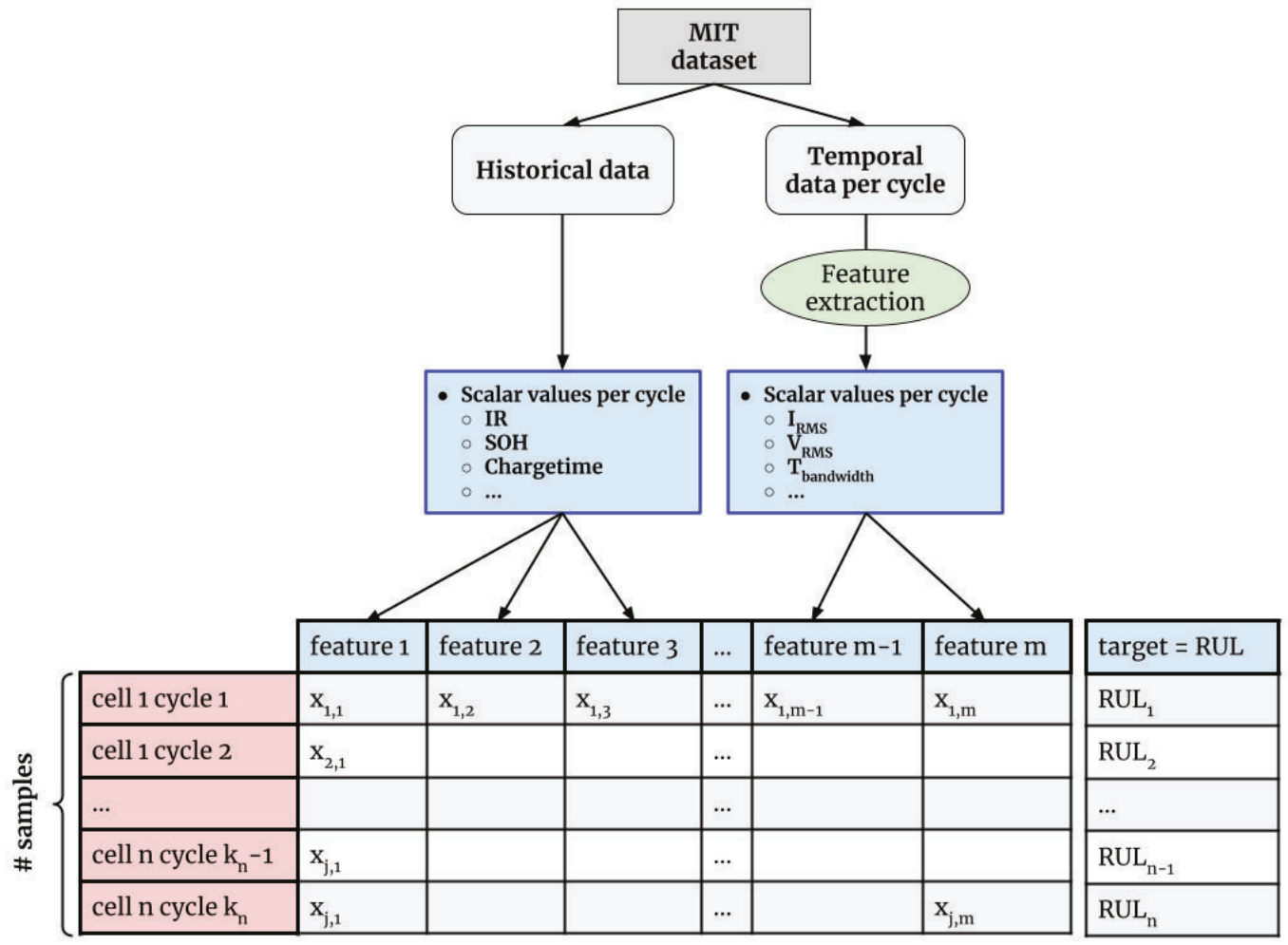

**Fig. 7.** Training dataset composed of historical data and time series features

In order to compare the performances of different combinations of features, two models were developed : one using only historical data, and another one using a selection of historical data and time series features.

# 4 Proposed architecture

As the dataset is quite recent, only little work has been based on it. In this paper, the architecture described consists in extracting features from temporal series and use them along with historical data to predict the RUL of a battery. Our predictive model is based on a well known machine learning regression tool : Artificial Neural Networks (ANN). Our goal here is not to investigate new predicting approaches but to prove that the available data combined with low computational models can lead to very efficient prediction performances. We only used ANN in this work because they can adapt to a great variety of data types and size. Moreover, we based our approach on a prediction of RUL as a scalar value. No sequential prediction of SOH or ageing features is made.

Two different ANN were built according to the number of features they take as input. The first one takes as input features extracted from time series and SOH, and will be referred to as TSF ANN (Time Series Features ANN). The second one takes as input only historical features and will be referred to as HF ANN (Historical Features ANN).

As explained is the previous section, this article also studies the impact of input features on the prediction performances. In the dataset described in section 3.3, each cycle is considered as one training sample, and the target is the RUL of the battery. As each cycle of each cell is considered, the dataset results in more than 99000 samples and a varying number of features according to the model. As there is a great amount of available data, the structure of the ANN can extend to several layers, each with a great number of neurons. Dropout was added after each layer in order to avoid over fitting during training. In order to find the best combination of hyper parameters (number of layers, number of units per layers, activation function, dropout rate...), several configurations where tested following a Bayesian optimisation procedure. In all cases, the output layer of our model only contains one unit and no activation function as a regression is made on the RUL, directly in terms of number of cycles.

# 5 Experiments

The following section describes how each model was tested on the different feature selections and compares the performances of our different models between them and in comparison with other RUL prediction approaches on the same dataset as us.

## 5.1 Training process

The first step of the training process is to perform the optimisation of the hyper parameters as explained earlier. After having completed the setup of hyper-parameters for different models, optimised models are completely re-trained. The dataset described in section 3.3 is randomly separated into three distinct ensembles : a training, a validation and a test set. In order to obtain reliable results, the process is repeated several times. The error measure is computed as the mean of all obtained measures during successive training.

## 5.2 Error metrics

During training, the back propagation process for weight optimisation is carried out with the Adam optimiser. The loss is calculated with Mean Square Error and performance is judged with the Mean Absolute Error metrics. We used mini-batch gradient descent in order to obtain an efficient and relatively short training time combined with an accurate convergence towards the minimum loss.

In order to compare the performances of our models between them and with other approaches in the literature, several scoring measures are used. In a vast majority of works, the evaluation of models is based on the Root Mean Square Error (RMSE) and Mean Absolute Error (MAE). We also add the Normalised Mean Square Error (NMSE) in order to compare the performances of our models with future works, and the Standard Deviation of the MAE ($\sigma_{MAE}$) in order to evaluate the reliability of the model. These quality measures are expressed as follows:

$$RMSE = \sqrt{\frac{1}{N}\sum_{i=1}^{N}(y_{pred,i} - y_i)^2} \tag{3}$$

$$MAE = \frac{1}{N}\sum_{i=1}^{N}|y_{pred,i} - y_i| \tag{4}$$

$$\sigma_{MAE} = \sqrt{\frac{1}{N}\sum_{i=1}^{N}(a_i - MAE)^2} \tag{5}$$

$$NMSE = \frac{\sum_{i=1}^{N}(y_{pred,i} - y_i)^2}{N * V} \tag{6}$$

In all these formulas, $y_{pred,i}$ is the RUL predicted by the model, $y_i$ is the real RUL and N is the number of samples on which error is calculated.

In the equation of standard deviation, $a$ is the absolute error of sample $i$.

$V$ is the variance of $y$. For example, the use of the mean of $y$ as the predicted values would give an NMSE of 1.

### 5.3 Prediction performances

In this section the predicting results of our ANNs will be compared between them. Our two models are built to predict one single value of RUL. A 2D dataset is fed to the networks and a 1D output is given, which corresponds to the predicted RUL in terms of cycle. The output can take any possible positive value. Figures 8 and 9 represent the predicting performances of the different networks. The predicted RUL is plotted as a function of the real RUL.

Table 1 details the predicting performances of the two developed models. The best prediction performances are obtained with the TSF ANN, which proves that the information contained in time series is highly valuable when designing a PHM strategy for Li-Ion batteries. Not only is the MAE lower with the TSF ANN (11.44 cycle compared to 15.08 with the HF ANN), but the predictions are more reliable. Indeed, the standard deviation of absolute error is lower with the TSF ANN, which means that there are less aberrant predictions and that more prediction errors are closer to the MAE. Histograms of the absolute error are represented in figures 10 and 11 show that a greater number of prediction with the TSF ANN have an error between 0 and the MAE.

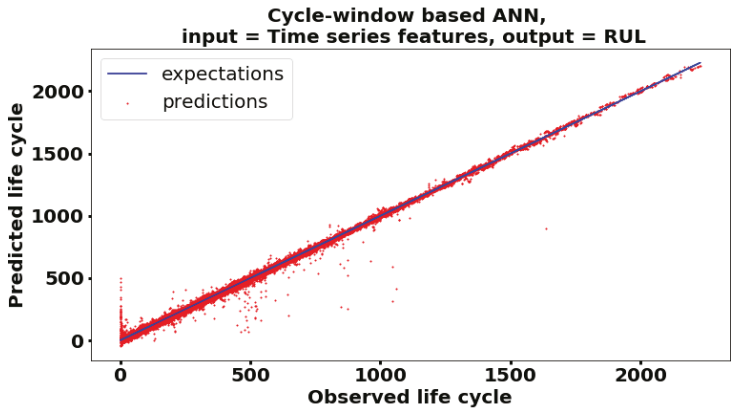

**Fig. 8.** TSF ANN predicting performances

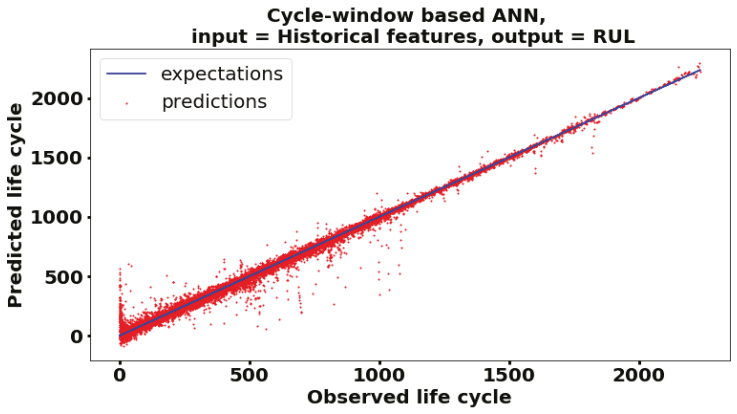

**Fig. 9.** HF ANN predicting performances

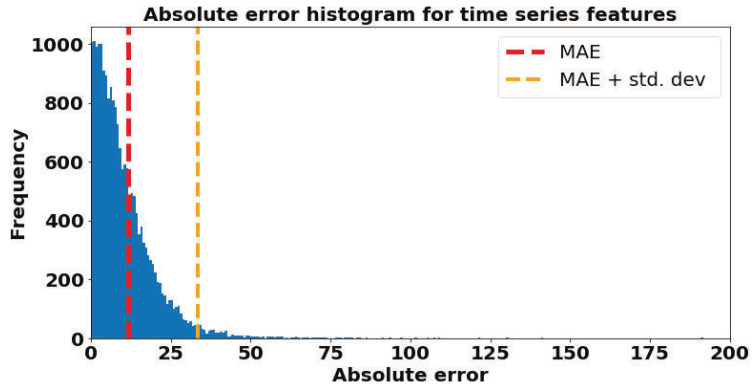

**Fig. 10.** TSF ANN absolute error histogram

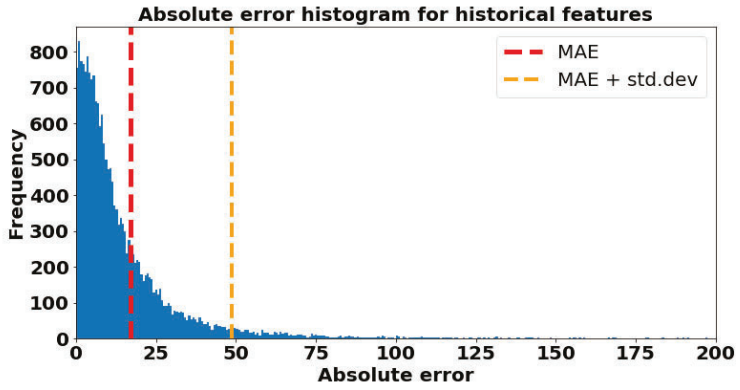

**Fig. 11.** HF ANN absolute error histogram

**Table 1.** Performance of the cycle-window based ANN according to the type of features

| | MAE | $\sigma_{MAE}$ | RMSE | NMSE |
|---|---|---|---|---|
| Historical features | 15.08 | 31.45 | 34.88 | $8.4*10^{-3}$ |
| Time series features | **11.44** | **26.78** | **29.16** | $\mathbf{5.9*}10^{-3}$ |

### 5.4 Comparison with other approaches

Although many papers in the literature mention their performances in the prediction of RUL, we can only compare our results with others that were obtained using the same dataset. For now, very few papers have based their approach on this dataset. The original paper [1] proposed a feature-based approach using a linear combination of the selected features. The only other approach we have found using this dataset was proposed by a research group in an online application designed to predict the RUL and current cycle of any battery [10]. They have based their approach on a CNN.

**Table 2.** Comparison of different approaches in the literature

| | RMSE | MAE |
|---|---|---|
| Historical Cycle based ANN | 34.88 | 15.08 |
| TSF Cycle based ANN | **29.16** | **11.44** |
| LR from [15] | 173 | N/A |
| CNN from [10] | N/A | 115 |

Table 2 compares the results obtained by all existing approaches with our best performing model. Although not all the same scoring measures were used in the two comparative works, the available scores show that our approach outperforms the prediction performances of the linear model developed by [1] and CNN developed by [10]. These results illustrate the fact that accurate prediction through machine learning needs a great number of training samples and a good feature extraction strategy. Designing a window based approach at the scale of one cycle, and extracting features from driving curves is more efficient than building features from early cycles or from a temporal window over several consecutive cycles for a use in ANN.

## 6 Conclusions

This paper is a description of our work on an innovative dataset published by the MIT, dealing with the ageing of Li-Ion batteries. Building a performing data-driven model relies essentially on the quality of data. With this work, we have proved that the dataset that had triggered our attention contains highly valuable information, with features representing the ageing phenomenon both in the historical domain and time series domain (driving data). We propose a low computational cost technique with well-known machine learning models such as artificial neural networks combined with features extraction techniques based on the exploitation of driving curves.
Our results show that a basic approach can outperform more complex models such as CNN. With our one cycle window based approach, we take advantage of all the information contained in the dataset. The prediction of RUL can be made at any cycle when testing a battery, and can above all be applied to cells whose current cycle is not known. For future work, we plan to dig further in the same direction. The use of driving data appears to be crucial, and we believe that employing Recurrent Neural Networks that are particularly adapted to the study of temporal series and forecasting problems could improve the performances of our models.

# An Architecture to Quantify the Risk of AI-Models

Alexander Melde[1], Astrid Laubenheimer[1], Norbert Link[2], and Christoph Schauer[2]

[1] Karlsruhe University of Applied Sciences
alexander.melde@h-ka.de astrid.laubenheimer@h-ka.de
[2] Inferics GmbH
norbert.links@inferics.com christoph.schauer@inferics.com

**Abstract.** In this paper we propose a multi-step approach to quantify the risk of AI-models. To evaluate the quality of a learned AI-model for image classification, a previously unseen part of a dataset is classified and the predictions are compared with their groundtruth to measure the accuracy of a model. In contrary, we first split the test dataset into two parts based on how unambiguous each sample can be assigned to a class. Samples that are close to the class decision boundary of multiple learned models are considered particularly difficult to classify. Second, we create a quantification of the model's ability to extrapolate on hard-to-classify or unseen data by training the model on "easy" data and evaluating it on the "difficult" split. Inside our models, we project the data into a 3-dimensional space using a neural network. We analyze this projection using the histogram of mutual distances, the silhouette measure [1] and the entropy of it to assess the extrapolation quality and thus robustness of the model. Subsequently, we apply our approach to the MNIST dataset [2] to prove its effectiveness. We see that models trained only on "easy" data are less robust than models trained on mixed data, which includes "difficult" data that lies in-between classes. This behavior is evident in both our quantitative measurements and qualitative evaluation In this paper, after an introduction to the topic and scope, related work is presented and the approach is explained in general terms. Subsequently, the application of the approach to the MNIST dataset is described and the results of these experiments are presented. Finally, a conclusion is drawn and options for future work are given.

## 1 Introduction

In recent years, there has been increasing research on artificial intelligence (AI) methods in order to make our everyday lives easier and safer. For instance, assisted living in combination with outpatient care services has become increasingly popular as an alternative to nursing homes. In addition to the established emergency call systems, more and more sensor-based AI systems are entering the market. These systems can inform nursing staff or trigger an alarm when they detect dangerous situations or unusual activities involving residents. Further increases in popularity are to be expected for these systems as new AI-based technologies support safety and security for self-determined living in familiar surroundings. In general, these technologies are based on machine learning (ML) models trained on datasets whose quality of being representative for real world scenarios is unknown.

For providers of such systems the introduction of new ML models into their products is of a high risk. The set of data on which the model can be evaluated in advance of the product launch commonly is not numerous enough and often generated under laboratory conditions which do not represent the true conditions for the product in real use. Furthermore, the performance of the models (from sight of economic efficiency) can not securely be predicted in advance because well introduced measures such as precision and recall [3, 415] do not lead to a reliable estimation of the risk of failures such as false alarms or missed detections and the associated financial cost.

In the context of systems that rely on activity recognition in domestic environments, this leads to problems, when closely related activities have to be distinguished. As one example "drinking a glass of water" and "brushing your teeth" are hard to distinguish, especially when the decision has to be made in the absence of a semantic context. In practice, classifiers usually are trained and tested on data that represents both activities in a clearly distinguishable way, which often leads to models with high accuracy.

In the wild however, one has to expect situations that are positioned fuzzy between the two classes and therefore hard to recognize. If such situations are not only underrepresented in the training data but also in the test data, they are not sufficiently considered in the quality assessment as well. Nevertheless, the behavior of the model on exactly these situations defines the risk of the use of the model in a real-life system.

In this paper, we propose a framework to evaluate how models which are trained on "easy" data only perform on "difficult" data which lies in-between classes. We demonstrate its effectiveness on the MNIST dataset [2].

## 2 Related Work

Approaches to assess model quality regardless of the distribution of the test data are provided by the field of Explainable Artificial Intelligence (XAI), which is a term to describe methods that explain decisions made by AI algorithms. Many deep neural network architectures are fundamentally black boxes whose decision-making is not comprehensible. Explainable AI attempts to make individual model executions or the entire model decision strategy more transparent. By understanding the decisions of a model, its quality can be assessed more independently from the particular test data and misconceptions arising from a limited scope of test data can be avoided. In recent years, numerous methods for XAI have been published. So far, these usually follow one of the following three approaches: First, the model can be built in a way that it is explainable naturally [4] (or by design, e.g. by using decision trees [5]), second original models can be replaced by fitted surrogate models that allow local or global interpretations [6, 7] and third, explanations can be generated using a direct process for either local or global explanations by putting a model into a more explainable state during training or by explaining single predictions, e.g. by determining the most important features for a certain decision [8, 9]. An extensive survey of XAI methods is given by Burkart and Huber in 2021 [9]. XAI tries to identify the risk of an AI model's decision being wrong and questions decisions made by the model. However, our approach does not explain the model itself, but instead attempts to quantify this risk. It describes the set of training data and how well it can be used to describe a real-world problem, and therefore does not fall into any of these categories.

Another related field of research is coverage testing, a technique determining whether the test cases used are actually covering the application area. Mani et. al. [10] explain the necessity to measure the quality of a dataset beyond the standard accuracy measure (proposing a set of four metrics to measure the coverage quality of a test dataset in the feature space of a model) and propose and demonstrate the effectiveness of a systematic test case generation system (samples additional test cases from the feature space) [10, 1]. They propose four different test quality dimensions that (1) measure the distribution of test data across individual classes, (2) measure the percentage of test data for each class that lies close to the centroid of the trained cluster or (3) near the boundary with respect to every other class of trained class clusters and (4) measure for each pair of classes the percentage of the boundary-condition (3) [10, 2]. This proposed system is evaluated on the MNIST dataset as well. This approach is closely related to active learning, a term to describe methods that find areas of the feature space that are not sufficiently sampled and ask the user to add test data for this specific areas [11]. This might also be done by evaluating the density function in the feature space. In contrary to these approaches, we do not measure the quality of coverage in the feature space but the quality of data distribution in the latent space.

A related approach that can be applied in both the latent and feature space is outlier detection, which looks for data points outside the distribution of the dataset [12]. In contrast, we evaluate quality using only the data points that lie between classes.

## 3 Approach

Our proposed approach starts by splitting the test split of a dataset into two parts: one of them representing data samples that can clearly be assigned to one class ("easy data") and the other one representing data samples that cannot be assigned to a class unambiguously due to its proximity to the class decision border in the latent space of the network ("difficult data"). The split can be found by means of a majority voting approach across several proven model architectures.

In the second step, we quantify the ability of arbitrary models to extrapolate from "easy" to "difficult" data, by training the model on "easy" data and evaluating it on the "difficult" split, which then leads to a quantification of the models ability to extrapolate onto hard to classify or unseen data. For our approach we assume that the model consists of an embedder backbone network which projects the data into the latent space and a classifier head evaluating the projections. The extrapolation quality of the model is assessed by analyzing the projections in the latent space, where the histogram of mutual distances is analyzed. The silhouette measure [1] and the entropy are used to compare the model performance on the "easy" and "difficult" datasets as well as to measure the extrapolation power of the model.

The steps are visually summarized in Fig. 1 and described in detail below.

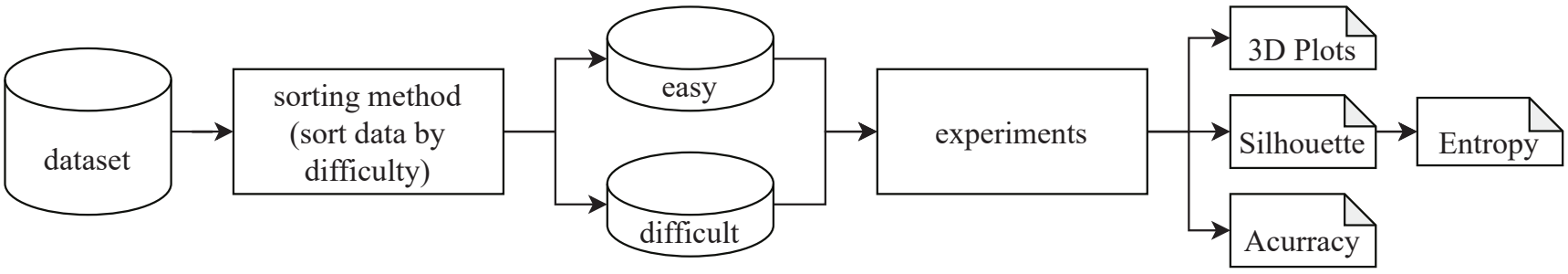

Fig. 1: Simple Overview of our architecture

To split the dataset by difficulty, we train seven different models using the same original training split but each time a different network architecture. We are using convolutional neural networks (CNNs) [13] of various complexity.

For each learned model, the test set is predicted and compared with the ground truth. We count the number of incorrect predictions per sample image of the test set to quantify the difficulty of each sample. The higher this number is, the more difficult this sample was to predict. We then group all of these samples into the two datasets which we consider "easy" and "difficult".

We train and test different combinations of dataset splits for test and training, loss functions, network architectures and number of epochs to measure the accuracy [3, 101] of different combinations to finally be able to find the "best working" combination in the traditional sense.

In order to measure robustness, we first create a visual representation of the model. For this we train multiple deep neural networks (with different dataset difficulties). We analyze the latent space of a model, which represents the training quality of a model and will serve as an important indicator for its robustness. The samples of each class are expected to form clusters in this space. We use both this visual representation of the 3-dimensional latent space and the silhouette histogram [1] and its entropy to measure the separability of these clusters.

The silhouette value of an object ranges between $-1$ and 1. Higher values mean the object is positioned better in the clustered space. The distribution of these values for a certain set of objects can serve as an indicator of the quality of the clustering. The silhouettes are calculated as

$$s(\vec{x}) = \frac{b(\vec{x}) - a(\vec{x})}{\max\{a\left(\vec{x}\right), b\left(\vec{x}\right)\}} \tag{1}$$

with $a(\vec{x})$ being the average dissimilarity (distance) of an object $\vec{x}$ to all other objects of its own cluster $X$ and $b(\vec{x})$ being the minimum average dissimilarity of $\vec{x}$ to all objects of all other clusters $C_x$ [1, 55]. To quantize this primarily visual measurement, we calculate the entropy of the frequency distribution of the calculated silhouette using the following formula:

$$S = \sum_i (p_i * log_2(p_i)) \tag{2}$$

with each $p_i$ being a bin of the histogram representing the silhouette values. In our examples we used 500 bins because it provided well distinguishable visual results, neither being too abstract nor too detailed to visually assess separation quality.

A well separated cluster will lead to a more robust implementation and a lower risk when applied to a real-world use case. These "easy" situations are characterized by a silhouette diagram where most values are concentrated in a peak on the right hand side (see Fig. 2a), leading to a low entropy value. In contrast, for "difficult" situations the silhouettes values are spread across the histogram (see Fig. 2b) leading to a higher entropy.

## 4 Application to MNIST

In this paper our method is applied on MNIST, a dataset of 70.000 fixed-size images of size-normalized and centered handwritten digits. It was created in 1998 by LeCun et. al. [2] based on a subset of the NIST handprinted forms and characters dataset [14].

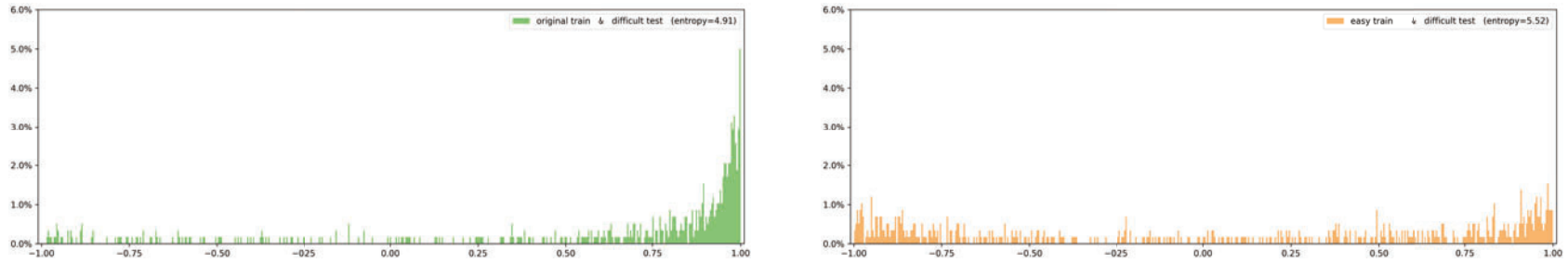

(a) The model is trained with both “easy” and “difficult” data, the silhouette has a peak on the right side.

(b) The model is trained with “easy” data, the silhouette values are spreaded across the cosine similarity spectrum.

Fig. 2: Visual distinction of training dataset difficulty based on the silhouette. The same network was trained using differently difficult dataset splits and each time tested with the same previously unseen “difficult” data.

The complete process of our approach applied to the MNIST dataset is shown in the architecture diagram (see Fig. 3).

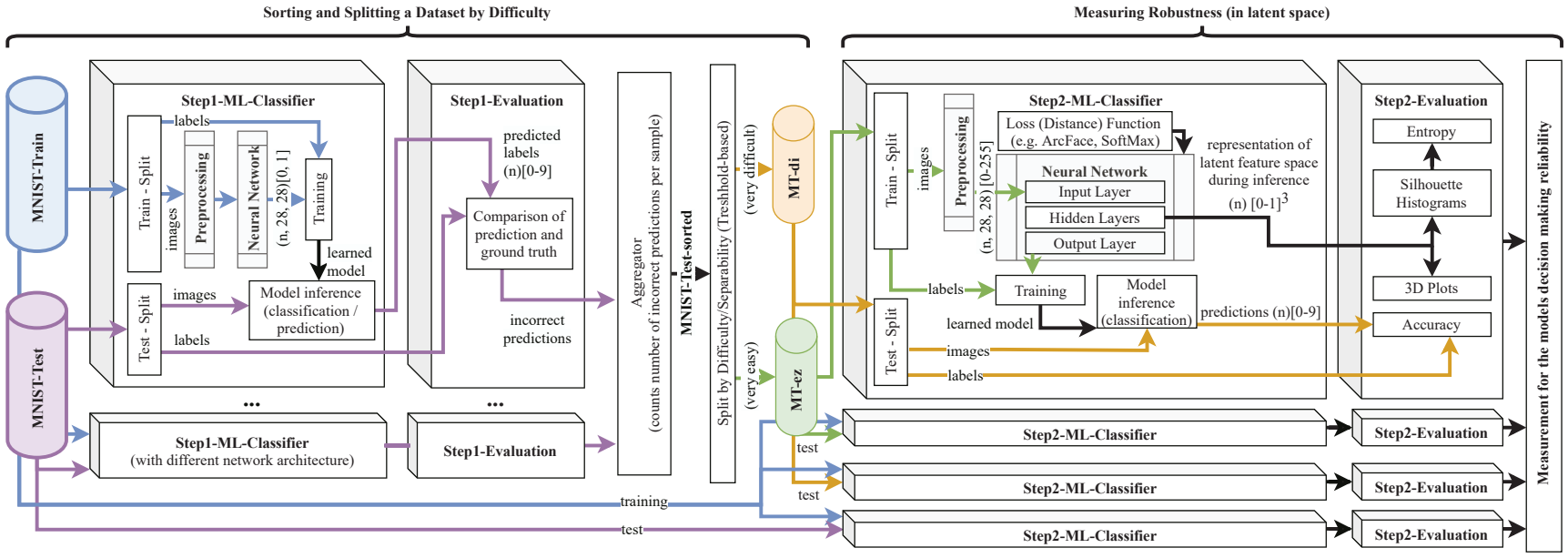

Fig. 3: Detailed overview of our architecture applied to the MNIST dataset

We use the original MNIST training and test split from the ‘keras.datasets‘ library and apply common preprocessing steps to reshape the list of pixels to a multidimensional array of shape (28,28,1), then rescale all values between 0 and 1 and finally one-hot encode the class labels.

For splitting the dataset by difficulty we use various architectures taken from popular publications and blog posts for image classification [15–17]. An exemplary selection of these is shown in Fig. 4 to visualize the differences in their approaches and complexity (different amount, selection, arrangement and sizes of layers and filters).

The trained deep neural networks for determining robustness are based on a version of the established VGG architecture [17]. We use the VGG8 architecture shown in Fig. 4c, which, unlike VGG16, allows us to process images smaller than 32x32px [17], which is important when applying this method to the MNIST dataset, of which the images have a size of 28x28px [2]. The architecture is retrieved from a repository containing an implementation of the ArcFace loss function [18], which we will use to visualize what the model has learned. By removing the output layer of the tested deep neural network architecture (SoftMax/ArcFace Layer in Fig. 4c), we can directly access the last dense layer representing the latent space of the model.

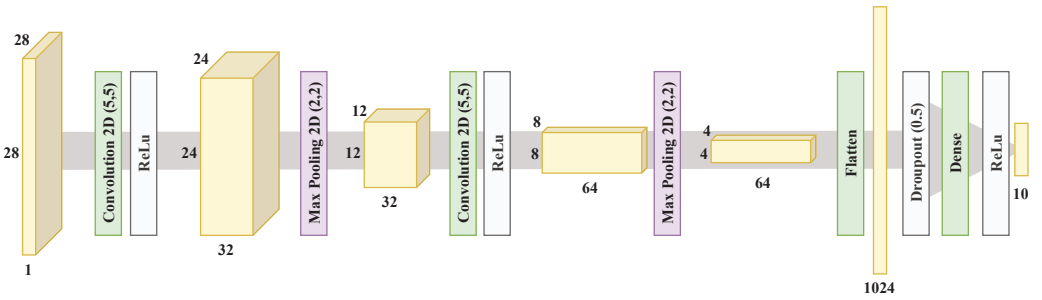

(a) CNN1 [15]

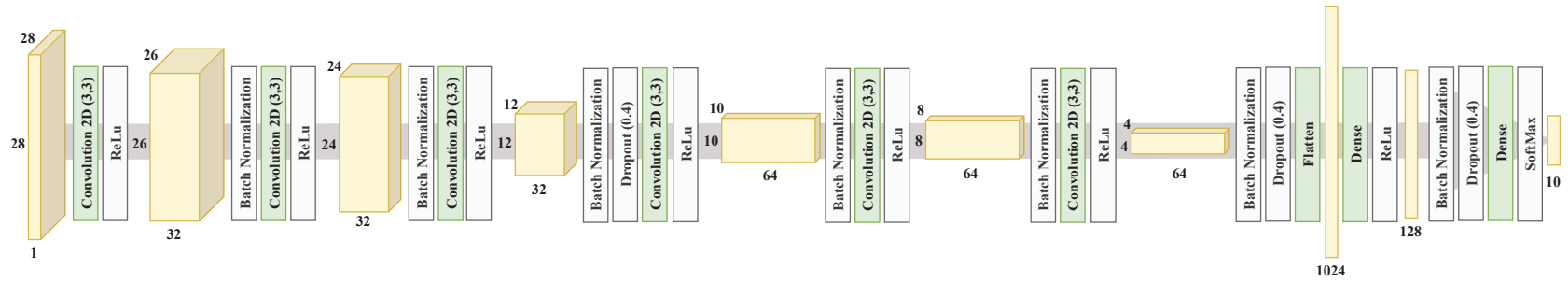

(b) CNN2 [16]

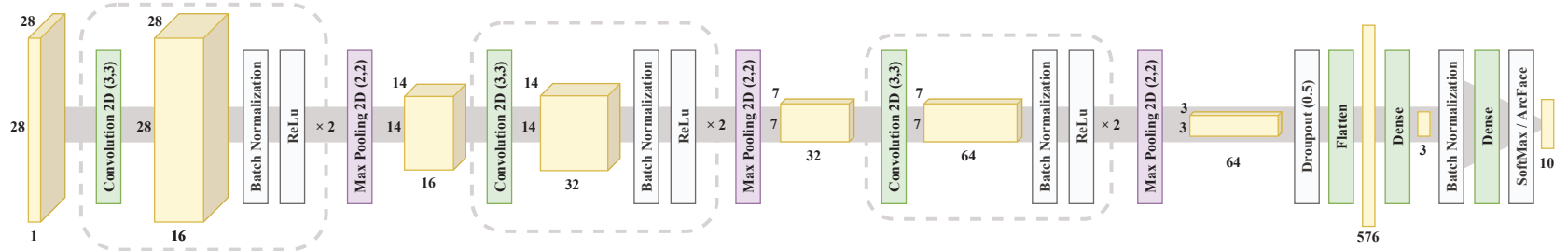

(c) VGG8 (VGG16: [17])

Fig. 4: Selected network architectures used in our experiments. Network layers are colored green, the dimension of data between layers is visualized by yellow boxes. The size of the convolutional filters is given in parentheses after the layer name (width, height), similar to the percentage of the drop-out in the corresponding layers.

In our experiments, we are measuring the dissimilarity of two objects $d(\vec{x}, \vec{y})$ using the cosine function where $\vec{x} \cdot \vec{y}$ is the dot product of $\vec{x}$ and $\vec{y}$:

$$d(\vec{x}, \vec{y}) = 1 - \frac{\vec{x} \cdot \vec{y}}{||\vec{x}||_2 \, ||\vec{y}||_2} \,. \tag{3}$$

Consisting of very basic mathematical functions, it is computationally simple and in our experiments provided well distinguishable results in the silhouette graphs.

# 5 Results

We apply the proposed method on the well-known MNIST dataset of handwritten digits [2] to demonstrate the effectiveness of our proposed approach.

To determine the difficulty of the image samples, seven different CNN-based classificators were trained and used to classify the test dataset. We consider every element that was incorrectly classified at least once as part of the "difficult" data split, leading to a final split of 9632 "easy" and 368 "difficult" images (colored green and orange in Fig. 3).

Next, the classifiers shown in Fig. 4 are trained with the "easy" data and then were validated using the "difficult" data split and the standard accuracy measurement. The results in Table 1 indicate that the different network architectures have different learning curves and in general perform differently well, pointing out they differ enough to ensure a reliable selection of network architectures to measure dataset sample difficulty.

Table 1: Testing accuracy when training with “easy” and testing with “difficult” data.

| Network | Validation Accuracy after 7 Epochs | Validation Accuracy after 100 Epochs | Training Accuracy after 7 Epochs | Training Accuracy after 100 Epochs |
|---|---|---|---|---|
| CNN1 | 0.56739 | 0.78190 | 0.99686 | 0.99956 |
| CNN2 | 0.37174 | 0.84733 | 0.99476 | 0.99967 |
| VGG8 | 0.45435 | 0.78408 | 0.97935 | 0.99978 |

To visualize the classification confusion, the predicted classes of the VGG8 classifier are grouped by their groundtruth classes in Fig. 5a) in the form of a confusion matrix. A confusion matrix representing a classifier with high accuracy shows high values on the main diagonal and small values outside of it. We consider training with the “easy” data and testing the “difficult” data as a difficult task, which explains why in this case high numbers appear outside the diagonal. The highest number of confusions is given for the case that a handwritten digit 9 is classified as 5, but we also see that the number 9 is generally over-represented in the “difficult” dataset. Example images of these confused MNIST digits can be seen in the other images of Fig. 5, labeled with the groundtruth followed by predictions of two classificators used in our experiments.

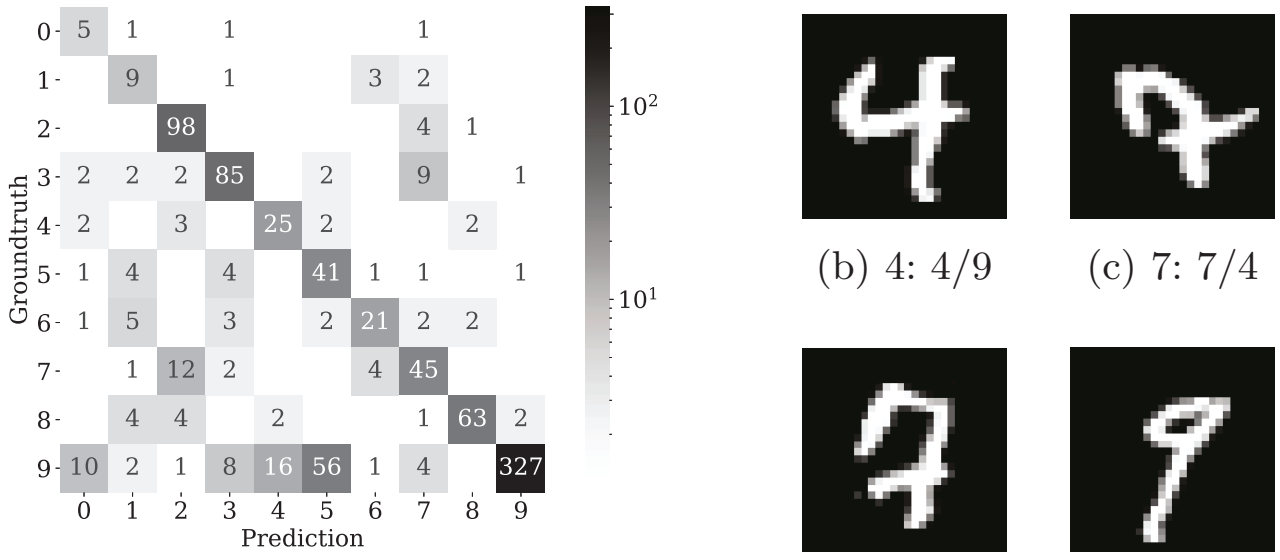

(b) 4: 4/9 (c) 7: 7/4

(a) Confusion matrix of final VGG8 classifier (d) 7: 7/9 (e) 9: 9/7

Fig. 5: Confusion matrix of MNIST images and sample images [2], their groundtruths and predictions made by one (a) or two (b – e) classificators. The label of images b – e show the ground truth followed by the different predictions of the two classificators.

For all following experiments, we will use both the traditional SoftMax loss [3, 181] as well as the ArcFace loss [18].

To visualize what the model has learned, the latent space is plotted in a 3D space. As expected, the different MNIST classes form clusters in this space. Datasets varying in difficulty will lead to differently well separated clusters. The better the clusters are separated, the more robust the model should be. This 3D visualization of the learned clusters is shown in Fig. 6. For both losses, different combinations of training and test datasets are shown. This overlapped view is useful to detect outliers arising from difficult data. From this visualization alone, we can see that when training with not only the “easy” images, but the complete train dataset (which also includes more difficult elements), a far better separability could be achieved. We can also notice that in our experiments, using the ArcFace loss [18] resulted in better separated clusters than using the SoftMax loss. This pattern is observed independently of the number of training epochs.

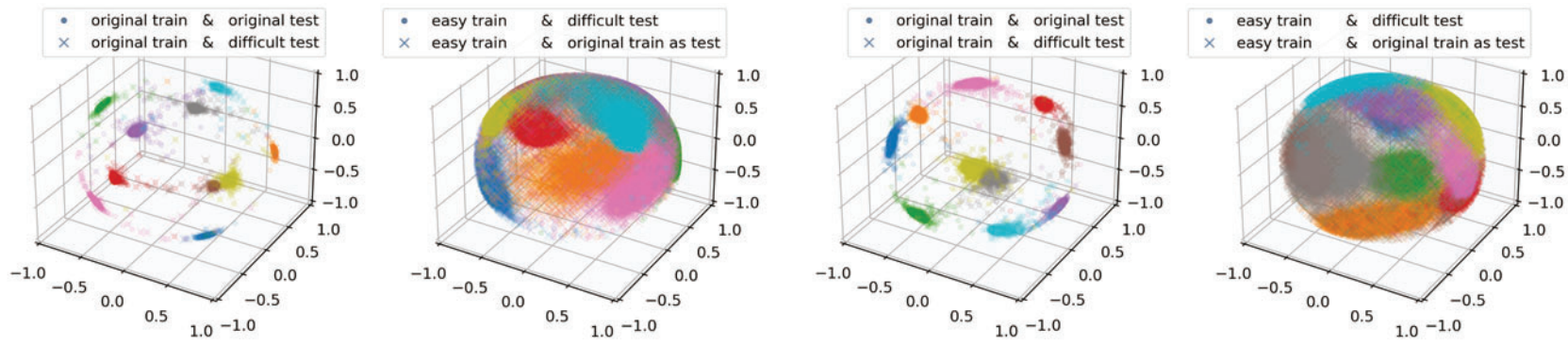

(a) ArcFace [18] loss function (b) SoftMax [3, 181] loss function

Fig. 6: Visualization of learned features in the latent space when trained for 100 epochs using different loss functions and different combinations of train and test datasets. Different colors represent different classes of the MNIST [2] dataset, different symbols are used to represent the different dataset combination.

Different Splits of difficulty lead to the distributions of silhouettes shown in Fig. 7. A high silhouette means an item is located near the center of the learned cluster.

The quantitative results of our experiments (shown in Table 2) and the visualized silhouette values (see Fig. 7) show the generally better performance of ArcFace over SoftMax and the benefits of training more epochs.

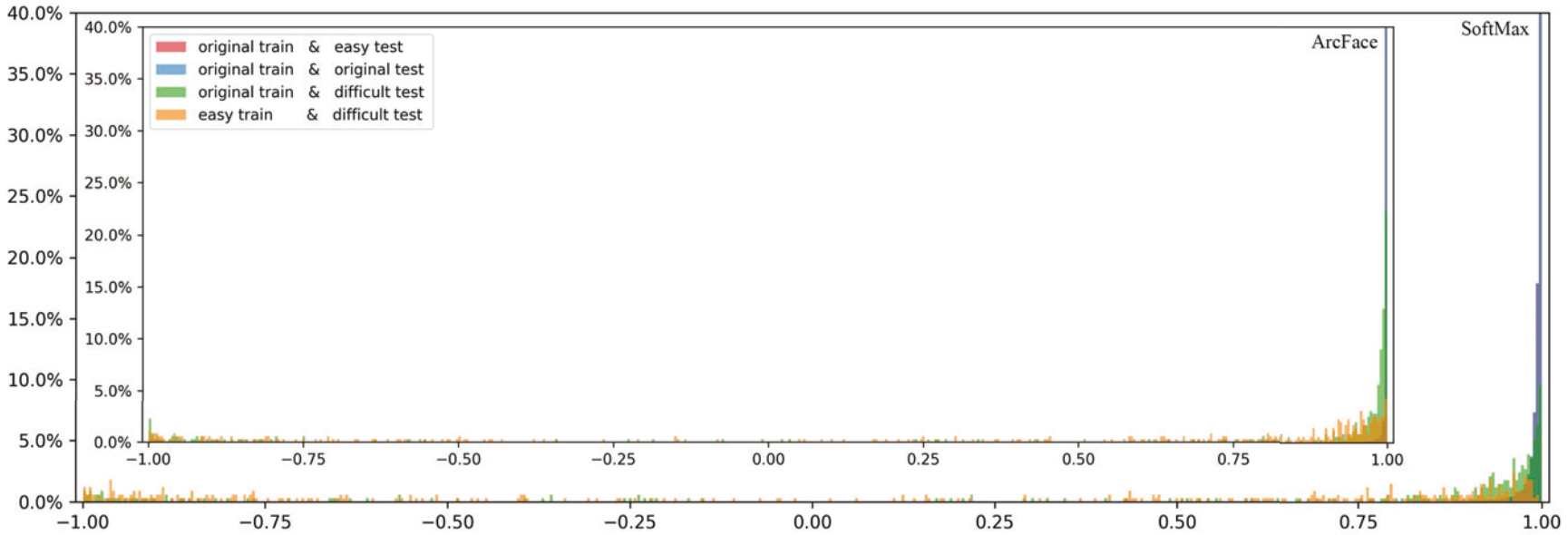

Fig. 7: Histograms of the silhouette values after training for 200 epochs with the SoftMax and ArcFace loss. Using ArcFace leads to more high silhouette values (especially in the rightmost bin) representing a better separation for all combinations of test and training datasets.

In general, if an image sample has a high silhouette value, it is well positioned in the clustered feature space. We can see that combinations that are deemed “difficult” lead to a wider distribution, e.g. training with easy images and testing difficult images, while “easy” combinations lead to a peak near the high end of the histogram. A wide distribution shows that there is no clear tendency to how good the elements are clustered, but a peak on the right side shows that most elements are clustered well. This is also represented in the entropy value of each graph, shown in Table 2. The wider the graph, the higher the entropy. The smallest value for the "easy" train and "difficult" test scenario is given when training with ArcFace for 200 epochs. Overall, we see that in most cases training using the SoftMax Loss is less effective than training with the ArcFace Loss. This is also shown in the histogram in Fig. 7, as with SoftMax the classic "blue" combination of training and testing using the official dataset splits shows a wider peak at the right side than with ArcFace, visualizing that more elements are worse positioned in the feature space.

Table 2: Entropy values measured in our experiments. A low entropy is an indicator of good cluster separation in the latent space.

| Dataset for training | testing | SoftMax Entropy after 7 | 100 | 200 Epochs | ArcFace Entropy after 7 | 100 | 200 Epochs |
|---|---|---|---|---|---|---|---|
| original train | easy test | 2.09 | 1.65 | 1.57 | 2.13 | 0.39 | 0.30 |
| original train | original test | 2.27 | 1.80 | 1.71 | 2.37 | 0.57 | 0.40 |
| original train | difficult test | 4.89 | 4.45 | 4.21 | 4.91 | 3.24 | 3.38 |
| easy train | difficult test | 5.39 | 4.96 | 5.09 | 5.52 | 4.99 | 4.81 |

## 6 Conclusion

We proposed a framework to separate a dataset into different levels of difficulty using a majority voting approach and evaluated how models which are trained on the "easy" data split only perform when the "difficult" data split is used for testing. We ensure a both qualitative and quantitative measurement by evaluating plots of the latent space and silhouette histograms as well as the entropy value of the silhouette histogram.

This theoretical approach has been applied to the MNIST dataset [2] to prove its efficiency. The results of the experiments are as expected: training with a dataset consisting of only "easy" data leads to less robust models than training with the full dataset that also contains "difficult" samples. We have also proven that the entropy of the silhouette measure histogram and both the visualizations are useful to determine the robustness of an AI-model. During these experiments, we also measured that using the ArcFace loss instead of SoftMax leads to a better clustering and therefore more robust models in most cases.

## 7 Future Work

In future work, this approach will be transferred to more complex applications using action recognition to support the use cases in the area of life sciences. For this we will use an activity recognition video dataset and matching machine learning models for video action classification.

Independently, in future projects, the majority voting approach can be replaced by using the silhouette coefficient to separate "easy" and "difficult" samples. The assumption that a high silhouette value implies that the data sample is easy to classify will result in shorter training time, as the number of required classificators is reduced.

## Acknowledgements

We thank J. Wetzel and T. Iraki for helpful comments and discussion. This work was funded by the Ministry of Science, Research and Arts of Baden-Württemberg (MWK) as part of the project Q-AMeLiA (Quality Assurance of Machine Learning Applications).

# Exercises in Human-Centered AI: On Shneiderman's Second Copernican Revolution

Dieter Wallach[1], Lukas Flohr[2], Annika Kaltenhauser[2], Sven Fackert[1]

[1] Hochschule Kaiserslautern
**dieter.wallach@hs-kl.de, sven.fackert@hs-kl.de**
[2] Ergosign GmbH
**annika.kaltenhauser@ergosign.de, lukas.flohr@ergosign.de**

**Abstract.** In his stimulating paper "Human-centered Artificial Intelligence" Shneiderman (2020) reframed AI research and application development by putting human users at the center of system design. Shneiderman requests a reunification of the view where humans are in the loop around algorithms and AI and suggests putting AI in the loop around humans, with a dedicated focus on the needs of users. From a UX designer's point of view, we discuss the idea of putting humans at the center and illuminate the implications of Shneiderman's arguments by referring to projects from our lab. By this, we emphasize the role of empirical research, (collaborative) UX Design, and evaluation in the development of human-centered AI systems. Our case studies exemplify a human-centered design approach of AI-injected systems in different domains and carve out the core learnings we gathered. The first case study, IMEDALytics, is taken from a project targeted at the development of a clinical decision-support system (CDSS) for individualized medical risk assessment, monitoring, and therapy management in intensive care medicine. We select this project to illustrate the indispensable nature of ethnographic user research to arrive at a holistic understanding of user needs. By visualizing the results of contextual observations and interviews in comprehensive user journeys, we shift the focus from problem solving through technology to the design of experience potentials (see Hassenzahl, 2010). We argue that it is paramount to present information to physicians in an unambiguous and understandable way, which classifies the task as an Explainable AI example in which answers to the following questions need to be derived:

- How can we combine human abilities of healthcare professionals – such as their general understanding, previous experiences, flexibility and creativity in the decision-making process – with the powerful possibilities of an AI-based system?
- How can we make diagnosis and therapy suggestions provided by the system accessible to healthcare professionals without depriving their self-efficacy?
- Which design processes are needed to design an interactive interface that leads to a long-term positive UX?
- Which influence has (the type of) presented information – e.g., in the form of information visualizations – on the perceived transparency or even trust in a CDSS?

Our second example focuses on the development of real-world autonomous mobility-on-demand (AMoD) public buses and required services for their operation. In autonomous mobility-on-demand systems, passengers are transported by self-driving cars, i.e., by vehicles with high or full driving automation capabilities. Comparable to taking a ride in a (shared) taxi, journeys in AMoD systems are temporally and spatially flexible. This means that there are neither fixed timetables nor fixed pick-up or drop-off locations required. Given that there is neither a driver nor an accompanying assistant available to answer traveler queries, AMoD rides vary greatly from current mobility-on-demand or taxi services. This new situation of riding in a driverless vehicle might feel awkward to passengers who are exposed to the decisions and actions of an autonomous system. Consequently, user interfaces capable of compensating the absence of a human driver are needed to establish a trustful AV-passenger communication. To counteract these challenges, we had to find answers to questions that include the following:

- How can and how should AI-infused AMoD systems communicate with passengers?
- How can we design and evaluate user interfaces while taking the complete user experience – before, during and after a ride – into account?

In our talk, we illustrate the applied formative design approach that is used for an iterative refinement of mobile and in-vehicle passenger UI prototypes (GUI- and Chatbot-based) and their subsequent empirical evaluation in simulators with increasing fidelity — ranging from video-based lab setups to real-word (Wizard-of-Oz) on-demand drives. In our third case study, AI science and AI engineering (Shneiderman, 2020) — i.e., emulating human behavior and developing useful applications — are combined in the creation of an AI-based (predictive) prototyping tool. The resulting tool allows the creation of complex interactive prototypes for which quantitative performance predictions are derived by running cognitive models. Such models are automatically generated by monitoring a designer's interaction while completing a task scenario using a prototype. The underlying models are based on the ACT-R cognitive architecture (Anderson & Lebiere, 2020) comprising hybrid (i.e., symbolic and sub-symbolic) structures and processes. ACT-R is a prominent example of a Unified Theory of Cognition integrating empirically supported assumptions about the interplay between human memory, learning, attention, perception and motor behavior which has been successfully applied to a broad range of tasks. The generated models can directly interact with a prototype, perceive its interface elements, learn task interactions and be able to manipulate the state of controls. In our talk we demonstrate that — by generating the behavior of synthetic participants — we can successfully predict human performance in real-world tasks ranging from mobile phone applications to the operation of commercial riveting machines in aviation industry. By using the very same tool, interface designers can create a prototype and receive almost instant evidence about its interactional efficiency by running artificial user models. Predictive prototyping opens the potential to significantly shorten iteration cycles by providing quantitative performance KPIs without the need to conduct effortful user studies. In the light of the case studies presented, we argue that the methodological apparatus of UX research and collaborative design practices contributes to the development of "human-centered" AI-based systems that result in positive user experiences — and thus increase the likelihood of adoption of AI-based systems in practice.

This work has been funded by the German Federal Ministry of Education and Research (BMBF) under the grant numbers 13GW0280B and 02L15A212 as well as by the German Federal Ministry of Transport and Digital Infrastructure (BMVI) under the grant number 16AVF2134G.

**Keywords:** User Experience Design, User Research, Prototyping, Autonomous driving, Clinic decision support systems, Evaluation, Human-Computer Interaction

# Interpretable Machine Learning for Quality Engineering in Manufacturing - Importance Measures that Reveal Insights on Errors

Holger Ziekow, Ulf Schreier, Alexander Gerling, Alaa Saleh
Business Information Systems, Furtwangen University of Applied Science,

78120 Furtwangen, Germany

{holger.ziekow,ulf.schreier,alexander.gerling,alaa.saleh}
@hs-furtwangen.de

**Abstract.** This paper addresses the use of machine learning and techniques of interpretable machine learning to improve quality in manufacturing processes. It proposes analysis methods on top of SHAP values to elicit useful insights from machine learning models. These methods constitute novel importance measures that support quality engineers in the analysis of production errors. We illustrate and test the proposed methods on synthetic as well as on real-world data from a German manufacturer.

## 1 Introduction

Recent advancements in machine learning have created increased interest in leveraging the potential of this technology in a range of application domains. Within this paper we address the application of machine learning to quality engineering in manufacturing. Here, the goal is to leverage machine learning for identifying and reducing causes of production errors. Like in other applications, the black-box nature of many machine learning approaches poses challenges to the applicability. This creates the need for methods to explain machine learning models. Specifically, there are two prevalent reasons for explaining machine learning models in quality management for manufacturing. One is to gain trust in the model decisions. The other is to leverage insights of the model to drive human data analysis. That is, quality engineers want to be pointed to factors or combination of factors that help to understand and reduce production errors. In this paper we explore existing and new methods of explainable machine learning to address this need.

Recently, a range of methods have been proposed that are intended to make black-box machine learning models explainable (see e.g. Gilpin et al. [1] or Molnar [2] for an overview). Among the existing methods are several means to assess feature importance. Feature importance can be used as guidance for humans about where they should focus their analysis of the data. However, the question is, what importance measures provide useful guidance. In this paper we argue that existing importance measures are not ideal for the application in manufacturing quality management. We also propose a range of new importance measures and test them on synthetic data as well as on real-world data from a German manufacturer.

This paper is a minor revision of our earlier technical report on the topic [10]. Section 2 presents the use case that motivates our work. Section 3 introduces our concept for identifying important features and defines corresponding measures. In Section 4 we evaluate the proposed importance measure on real-world data form a German manufacturer (SICK AG). In Section 5 we review related work. Section 6 summarize the content of this paper.

## 2 Motivating use case

The methods proposed in this paper may be of general applicability, but are mainly motivated by the use case of quality engineering in manufacturing. Specifically, we draw the requirements from the PREFERML research project [3] and the participating manufacturing company SICK AG. The task of quality engineers is ensuring high product quality while keeping the production cost low. Therefore, they strive to identify and eliminate root causes of production errors. All products undergo rigid tests, not only when they are finished. The final checks ensure that only high-quality products are shipped to the costumers. However, it is desirable for the manufacturer to sort out faulty products early in the production process and to avoid allocation of resources to later discarded products. Therefore, additional quality checks are conducted after each production step throughout the production line. Each checkpoint records a range of measurements and sorts out products that do not satisfy predefined criteria. The recorded data points are also the source for investigating error causes. For instance, a quality engineer might realize that products that show a value greater than X for property Y in production step A, have an increased chance of failing when checked in production step E. This insight might result in adjustments of check in step A to filter out such products early.
However, the data pool for investigation is very large. The manufacturer in the PEFERML research project takes several thousands of measurements for each product. For humans it is a very challenging task to find relevant relations in such a large pool of data. The methods that we propose in this paper provide means to guide quality engineers to relevant measurments.

## 3 Finding important Features for Quality Engineers with SHAP

The goal is to point quality engineers to interesting measurements (features) in the quality test data. We consider a feature interesting in the target context, if it provides actionable insights for quality engineers to adapt thresholds in the current quality checks. Hence a feature is interesting if it (a) enables a simple rule for predicting errors and (b) the prediction with that rule is of sufficient quality to reduce production cost.

### 3.1 Shapley Additive Explanations

In our work we leverage local explanations and specifically SHAP values [4] to reason about feature importance. Local explanations address feature importance for individual data points whereas global importance measures capture the general (e.g. average) importance of a feature. Local explanations have the advantage that they can reveal if features are sometimes (possibly only a few times) of great importance. The contribution of such features may be hidden in global importance measures. However, a feature that is very important a few times can be of major interest for quality management. This is because the interesting situations (predictable errors) are rare in a production process. With our methods we aim to identify such features. The proposed feature importance metrics are not global or local in the classical sense. The identified features are important in the sense that they allow for good and simple predictions for some regions.

### 3.2 Illustrative Example for using SHAP values for Quality Management

SHAP values provide insights into predictions for individual data points. For a given data set this results in a set of explanations. Analyzing this set of explanations can yield additional insights into the workings of a model and the modeled phenomenon. Lundberg et al. [5] proposed and implemented some analysis on top of SHAP values and SHAP interaction values. These include clustering of explanations and visualization of interactions.
This chapter introduces additional methods for analyzing SHAP values and SHAP interaction values. In contrast to existing methods for analysis, the proposed methods do not aim at providing a holistic understanding of the analyzed model or phenomenon. Instead, they aim at finding simple and good explanations for part of the analyzed phenomenon and thereby guide the analysis of quality engineers. We argue that a feature or relation between features is of potential interest, if it enables (a) prediction rules with high prediction quality and (b) the respective rules are simple enough for human comprehension as well as for taking corresponding action (e.g. adjusting a threshold in a quality check). In the motivating use case, prediction quality is high enough, if the ratio between true positives TP and false positives FP is high enough. That is, TP/FP must be bigger than the cost of a false positive divided by the savings for a true positive. (False and true negatives are irrelevant as they do not chance the as-is process.)
We further argue that the support for the rule is less important. Errors are rare in highly optimized production lines and any prediction rule will likely have low support. It is more important to identify strong effects because they enable clear actions. A weak but more common effect can yield a rule with good support but is more difficult to exploit in the production process. The core ideas behind the proposed analysis methods follow from the arguments above. That is, to sacrifice support in favor of confidence and simplicity. This is in contrast to established feature importance measures, which aim at finding features that are overall most important (e.g. important on average). We are looking for features and relations that are very important, but possibly only in a few cases. For the sake of illustration consider the subsequent simplified case: Assume a fictional manufacturer that has produced 10000 product items. An intermediate quality test measures the features A and B with uniformly distributed values (range 0 to 1) and have the following relation with production errors: (1) Products with feature A<0.8 have roughly a 20% chance of failing the final quality check. (2) Products with feature B>0.99 have roughly a 98% chance of failing. Overall, about 16% of the sample data reflect faulty products.
To support the quality engineer, we aim to identify the feature that yields the most useful insights. We therefore train a machine learning model on the quality test data and analyze the feature importance. The feature importance score should point the quality engineer to the most useful feature. For illustration, we build a tree model (see Figure 1) and compute the typical feature importance scores (see Table 1). We used the Python XGBoost library to build the tree, using a single tree for simplicity. Figure 1 illustrates the model[1].
Table 1 shows the values of established feature importance measures for the presented model. Note that the typical feature importance measures all consider feature A as most important. However, feature A is of little use in the targeted application domain. Through investigating feature A, the quality engineer can build a rule of the form “IF feature A < 0.8 THEN ERROR”. However, this rule is not of practical use. Sorting out products with a predicted error would remove 80% of the products and as many false positives. In contrast, an analysis of feature B would result in a prediction rule of the form “IF feature B > 0.99 THEN ERROR”. This rule affects a reasonable number of products and is almost always correct. It therefore enables an adjustment of the production process. Note that the typical feature importance measures fail to rank the features as desired. Table 2 shows the results from the measures that we introduce in this paper. They all correctly rank feature B first. The specifics of each importance measure follow in the corresponding subsections below.

[1] Positive values in the leave nodes correspond to the prediction of an error an negative values to the prediction of no error.

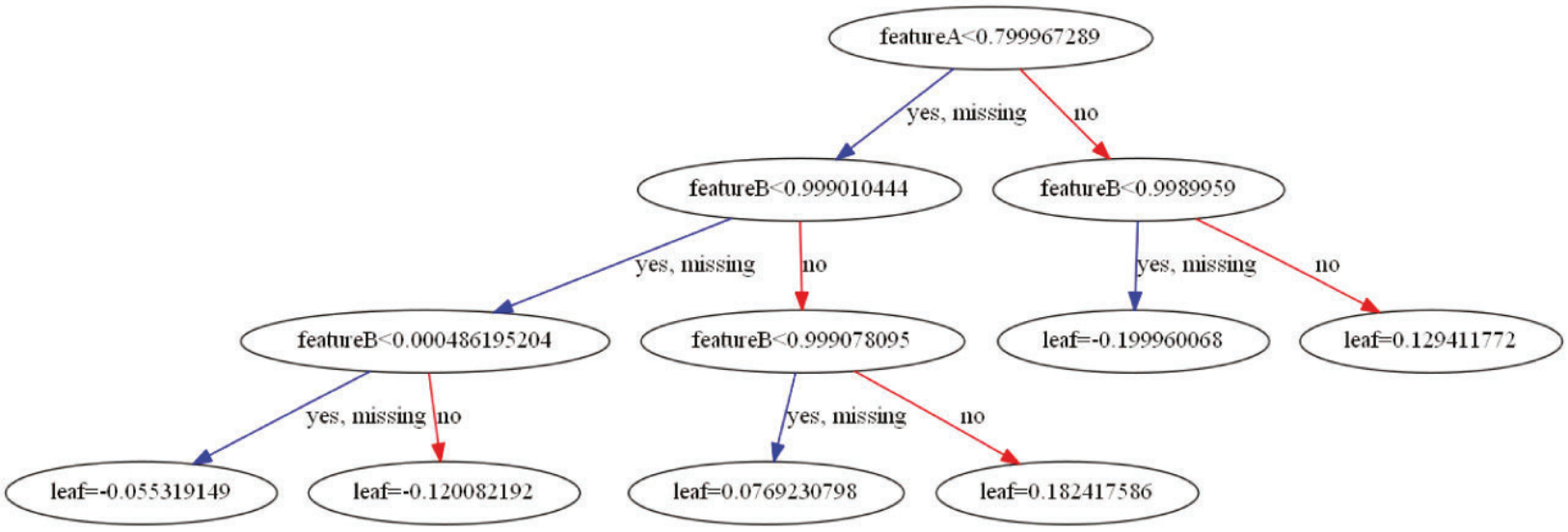

Figure 1. The first tree in the sample case model (using XGBoost)

Table 1. Established feature importance measures for the model in fig.1

| Importance Measure | Score for feature A | Score for feature B |
|---|---|---|
| Gain | 16.119608 | 7.564618 |
| Cover | 946.390676 | 676.451196 |
| Weight | 263 | 258 |
| Total Gain | 4239.456775 | 1951.671417 |
| Total Cover | 248900.747835 | 174524.408458 |
| Gini-importance | 0.680605 | 0.319395 |
| Average SHAP values | 0.866374 | 0.029806 |

Table 2. Proposed importance measures for the model in fig.1

| Importance Measure | Score for feature A | Score for feature B |
|---|---|---|
| Max SHAP | 0.705525 | 7.591009 |
| Range SHAP | 6.420452 | 8.724322 |
| Smoothed Range SHAP | 61.053000 | 65.920289 |
| Top-K SHAP (with k=10) | -19.984684 | 75.860939 |
| Max Main Effect | 0.667381 | 4.482486 |

## 3.3 Concept for Feature Importance measures on top of SHAP values

In this section we introduce a range of importance measure on top of SHAP values. All these importance measures leverage the local importance measures provided by SHAP values, to identify "locally interesting" features. By "locally interesting" we refer to features that yield good explanations for errors in at least some cases. This is in contrast to measures that identify features that are frequently or on average important.

**Top Importance for Top-K Predictions**

The idea behind the analysis of top-k predictions is to zoom in on the most relevant cases. We define top-k by ranking the predictions according to their confidence. In the target application we only care about predicting errors. Hence, we only look at error predictions with high confidence. Here we use a notion of confidence that does not imply quantifiable confidence intervals. That is, we use the SHAP values in our analysis to determine the top-k instances. We then average the local feature specific SHAP values for each of the top-k predictions. Leaning on the notation from Lundberg and Lee [4] and assuming that a high SHAP value refers to the error class, the importance measure is defined as follows[2]:

$$\text{Top-K}_i(f, D) = \frac{1}{k} \sum\nolimits_{x \in \{d \mid k > |\{d' \in D \mid f(d) < f(d')\}|\}} \phi_i(f, x)$$

[2] For the sake of simplicity we assume a total ordering of prediction scores.

Here, f is the model function and D the data sets for evaluating the feature importance, i the feature to score, and ϕ_i (f,x) the SHAP value of feature i for model f and data point x. The intuition behind this measure is the following: In the target application we look for predictions with high confidence only. Hence, we only care about features that play a role in the predictions with high confidence. We then aggregate the local SHAP values for the most relevant predictions. An assumption behind this approach is that the top-k predictions are similar from the perspective of the model. That is, they yield predictions with high confidence for similar reasons. This may not always hold true, especially for larger k. Hence, a natural extension of this measure is to consider clusters within the top predictions.

**Max SHAP**

The idea behind Max SHAP is to look for features that can have a high contribution to the outcome. It is simply defined as the maximum SHAP value for a given feature in data set:

$$\text{Max SHAP}_i(f, D) = \max\{\phi_i(f, x) | x \in D\}$$

Here, we again assume that the outcome of interest is associated with high SHAP values. One may change the sign of the measure or substitute the maximum with a minimum function, if the opposite is the case.

The intuition behind this measure is that it captures the highest effect that a feature can have. The rationale is that a feature with a high maximal contribution is in some cases of high interest. A weakness of this measure is that it is sensitive to outliners. We argue that confidence is much more important in the targeted application domain than support. However, isolated outliers can lead to undesirable results. One may therefore alter this importance measure by taking a certain quantile instead of the maximum value.

**Max Main Effect**

The idea behind Max Main Effect is to look for features that have the highest effects on their own. That is, we explicitly ignore the effect of feature interactions, which are otherwise included in the SHAP values. Leaning on Lundberg et al. [4], we define the measure as:

$$\text{Max Main Effect}_i(f, D) = \max\left\{\phi_i(f, x) - \sum_{j \neq i} \phi_{i,j}(f, x) \,\middle|\, x \in D\right\}$$

Here, ϕ_(i,j) is the SHAP feature interaction as defined in [4], and with ϕ_i (f,x) we denote the dependency on the model f and the data set D.

The intuition behind this measure is an enhancement of the Max SHAP measure. Again, we assume that the outcome of interest is associated with high SHAP values. One may change the foresign of the measure or substitute the maximum with a minimum function, if the opposite is the case. The maximal SHAP value of a feature may be heavily dependent on the interaction of features. In this case, the high contribution of the feature cannot be assessed in isolation. With the Max Main Effect measure we aim at identifying features, which have a high contribution regardless of the other features. This is appealing in the targeted use case, because such features support simple decision rules and visual analysis with only one dimension.

**Range SHAP**

The idea behind Range SHAP is to look for features that have a strong impact on the outcome over their value range. Unlike Max SHAP, it considers also negative SHAP values. That it takes into account, if certain value ranges of the feature are an indication for no error. (Again, we assume that the outcome of interest is associated with high SHAP values.) We define the measure as follows:

$$\text{Range SHAP}_i(f, D) = \max\{\phi_i(f, x) | x \in D\} - \min\{\phi_i(f, x) | x \in D\}$$

The intuition behind this measure is to look for features that can have a strong impact on the model output in either direction. By looking at the range, we capture the strongest local effects. That is, the score is high if the feature contribution varies strongly between some cases.

**Smoothed Range SHAP**

The idea behind Smoothed Range SHAP is to suppress variance for data points with the same feature value. Due to feature interaction, the same feature value may correspond to different SHAP values. Smoothed Range SHAP averages over a sliding window to suppress variation for the similar feature values. Using the formula for moving average the measure is defined as follows:

$$\begin{aligned}\text{Smoothed Range SHAP}_i(f, D) \\ = max\left(\left\{\frac{1}{W}\sum_{n=m-\frac{W-1}{2}}^{m+\frac{W-1}{2}} \phi_i(f, x_n) \,\middle|\, \frac{W-1}{2} \leq m \leq |D| - \frac{W-1}{2}\right\}\right) \\ - min\left(\left\{\frac{1}{W}\sum_{n=m-\frac{W-1}{2}}^{m+\frac{W-1}{2}} \phi_i(f, x_n) \,\middle|\, \frac{W-1}{2} \leq m \leq |D| - \frac{W-1}{2}\right\}\right)\end{aligned}$$

Here we assume that the data set D is sorted by x and xi is the i-th position.

The intuition behind this measure is an enhancement of Range SHAP. If a feature interacts strongly with other features, the SHAP values may have a high variance for the same or similar feature values. However, it is more interesting for the target application to find changes in the model output, which correspond to different feature values. Such changes are more helpful for identifying simple prediction rules and are therefore emphasized by this importance measure.

# 4 Evaluation

To evaluate the proposed importance measures with real-world data, we analyzed quality logs from the German manufacturer SICK AG. Note that we omit some details in the data description that are of minor importance for the evaluation. This is to protect internal information of the manufacturer. The analyzed data cover a time span of roughly one year and contain records about several ten thousand products of a specific type.

Specifically, we analyzed the test data from a production step A and the outcome of tests from the subsequent step B. We trained a model (i.e. XGBoost classifier) on 34% of the quality test data from step A, with the aim to predict errors in step B. For model training we used roughly 100 test parameters from step A as features and the test results from step B as label. For simplicity, we reduced different error labels to the binary outcome "passed" or "failed".

Within this setup we analyzed the trained model with regards to feature importance. We tested the established and new introduced measures listed in Section 3. The introduced measures require data instances and the model as input. Here we used the training data. We used the XGBoost library for the implementation of the model[3] and the established importance measures. To implement the new proposed measures, we set up on top of the SHAP library for computing SHAP values and SHAP interaction values. The parameters $K$ and $W$ for our measures were set to $K=10$ (i.e. the top-10 predictions) and $W=50$ (i.e. a smoothing window of 50 values).

[3] We used default parameters, with exception of scale_pos_weight. This parameter was adjusted to reflect the cost for false and true positives in the manufacturing line.

The analysis results in a ranked list of features for each tested importance measure. Table 3 shows the top 10 results for established measures and Table 4 depicts the top 10 results for the new introduced methods. Feature names are obfuscated to protect internal information of the manufacturer. In both tables we highlight feature Dfleft_col_1593 and Dfleft_col_379. We argue that - amongst the listed features - these two features are of most interest in the targeted application. Hence, we expect that an effective importance measure ranks these features high.
Figure 2 and Figure 4 provide the details for our argument about the importance of Dfleft_col_1593 and Dfleft_col_379. The figures show the distribution of feature values as histogram. The Y-axes displays the frequency of feature values and have logarithmic scale. The X-axis is scaled to cover the whole value range. We omitted axis labels to protect internal information of the manufacturer. The shading of the bars encode the percentage of errors (faulty product) in the respective bar (white 0% and black 100% errors). We show all features that are among the top three features of any tested importance measure.

Table 3: Features ranked by established importance measures

| Average SHAP | Cover | Gain | Total Gain | Weight |
|---|---|---|---|---|
| Dfleft_id | Dfleft_col_1125 | Dfleft_col_1125 | Dfleft_id | Dfleft_id |
| Dfleft_col_832 | Dfleft_col_832 | Dfleft_col_832 | **Dfleft_col_1593** | Dfleft_col_357 |
| Dfleft_col_705 | Dfleft_col_708 | Dfleft_col_711 | Dfleft_col_738 | **Dfleft_col_1593** |
| **Dfleft_col_1593** | Dfleft_col_705 | Dfleft_col_366 | Dfleft_col_1125 | **Dfleft_col_379** |
| Dfleft_col_1126 | Dfleft_col_711 | **Dfleft_col_1593** | Dfleft_col_832 | Dfleft_col_1322 |
| Dfleft_col_932 | Dfleft_col_366 | Dfleft_col_733 | Dfleft_col_357 | Dfleft_col_738 |
| Dfleft_col_738 | Dfleft_col_751 | Dfleft_col_745 | Dfleft_col_1214 | Dfleft_col_1214 |
| Dfleft_col_1566 | Dfleft_col_725 | Dfleft_id | Dfleft_col_711 | Dfleft_col_1267 |
| Dfleft_col_357 | Dfleft_col_738 | Dfleft_col_738 | **Dfleft_col_379** | Dfleft_col_1266 |
| Dfleft_col_711 | Dfleft_col_1126 | Dfleft_col_1126 | Dfleft_col_1126 | Dfleft_col_1566 |

Table 4: Features ranked by proposed importance measures

| Max Main Effect | Max SHAP | Range SHAP | Smoothed Range SHAP | Top-K SHAP |
|---|---|---|---|---|
| **Dfleft_col_1593** | **Dfleft_col_379** | Dfleft_id | Dfleft_id | **Dfleft_col_1593** |
| **Dfleft_col_379** | **Dfleft_col_1593** | **Dfleft_col_1593** | **Dfleft_col_1593** | Dfleft_col_738 |
| Dfleft_id | Dfleft_id | **Dfleft_col_379** | **Dfleft_col_379** | Dfleft_id |
| Dfleft_col_738 | Dfleft_col_738 | Dfleft_col_738 | Dfleft_col_738 | Dfleft_col_720 |
| Dfleft_col_703 | Dfleft_col_932 | Dfleft_col_832 | Dfleft_col_832 | Dfleft_col_1125 |
| Dfleft_col_708 | Dfleft_col_1322 | Dfleft_col_1322 | Dfleft_col_1125 | Dfleft_col_357 |
| Dfleft_col_1125 | Dfleft_col_357 | Dfleft_col_357 | Dfleft_col_720 | Dfleft_col_1214 |
| Dfleft_col_811 | Dfleft_col_1214 | Dfleft_col_703 | Dfleft_col_1210 | Dfleft_col_745 |
| Dfleft_col_1322 | Dfleft_col_720 | Dfleft_col_1214 | Dfleft_col_1266 | Dfleft_col_1126 |
| Dfleft_col_1214 | Dfleft_col_703 | Dfleft_col_932 | Dfleft_col_1267 | Dfleft_col_711 |

As Figure 2 shows, the features Dfleft_col_1593 and Dfleft_col_379 have desirable properties for the targeted use case (however the error numbers that support this observation for Dfleft_col_379 are rather small). For both features one can observe an interval with a low error rate. Outside this interval, the error rate is very high. This allows quality engineers to derive simple rules of the form "IF value > X or value < Y THEN ERROR". For Dfeft_col_1593 and Dfleft_col_379 this rule would apply in few cases and – according to the training data – has high prediction quality. Several of the other identified features share this property of feature Dfleft_col_1593 and Dfleft_col_379. However, the corresponding relations are weaker. That is, the error rates outside the respective intervals are lower and/or the rules would apply to fewer instances. We therefore argue that the feature Dfeft_col_1593 and Dfleft_col_379 are most important and should be ranked high by the importance measures. Note that this is the case for the new introduced measures and in particular for *Max Main Effect* and *Max SHAP*.

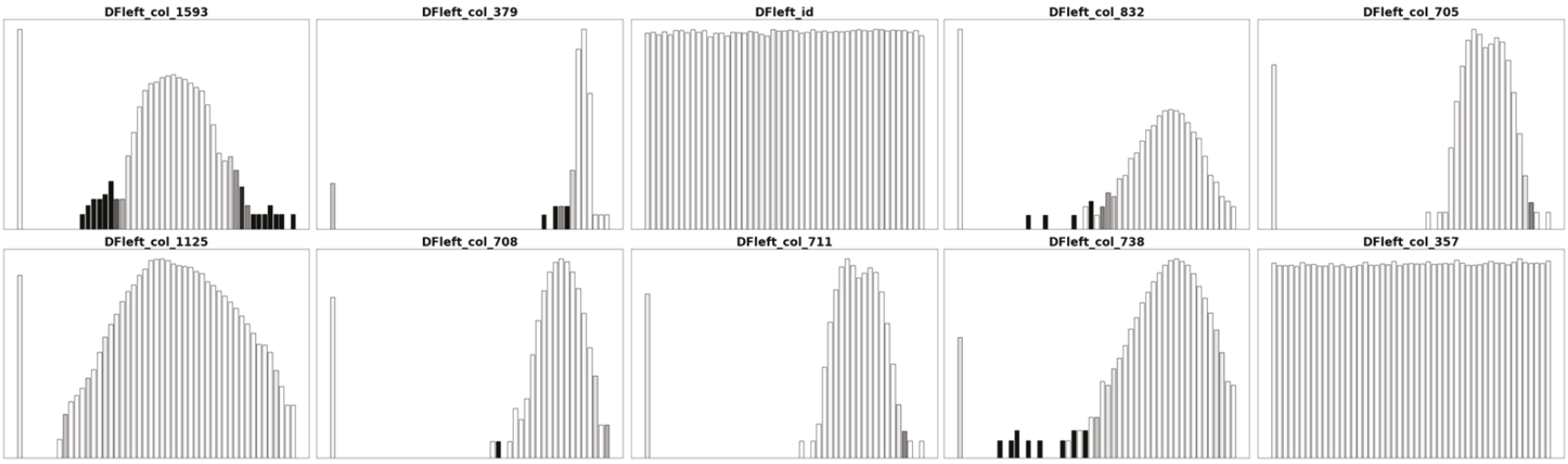

Figure 2: Distribution of feature values for training data. (Y-axes with logarithmic scale, color coded error percentage, white=0%, black=100%).

Furthermore, the features Dfleft_id and Dfleft_col_357 stick out. These features do not show a clear relation with errors in Figure 2. Figure 3 provides more details on these features. For Dfleft_id and Dfleft_col_357 it shows the SHAP values and feature values for each data point in the training data. The plots provide insights on why these features are considered important by some measures. For some data points the features have strong contributions to the model output. However, the variance of the contribution is high, in particular in regions with potential high impact. (This can be seen in the vertical arrangement of points below or above point with high contribution). The figure implies that these features have strong interactions with other features and do not enable good explanations on their own. Thus, the features are not useful for simple prediction rules that consider only one value.

As a side note, the feature Dfleft_id is an identifier value that roughly resembles a counter. This feature is not useful for predictions but can help analysis in retrospect.

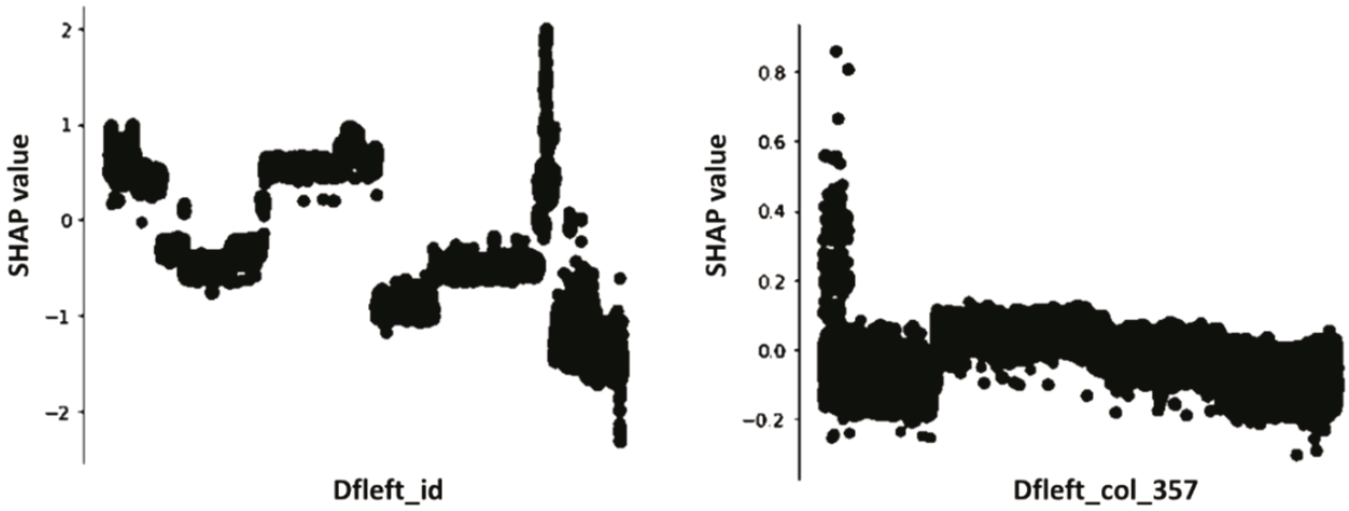

Figure 3: SHAP values for Dfleft_id and Dfleft_col_357

For our analysis we split the data into training and test data. However, the evaluation on the training data may be more meaningful in the targeted use case. This is because the main goal is not to build a general prediction model, but to point quality engineers to interesting phenomena in the data (possibly only in retrospect). In this case the human can judge the validity of the findings based on domain knowledge and validation through a separate test set may be of less importance. (See [2] for a more detailed discussion of using test sets or training sets for evaluating approaches of interpretable machine learning). In our test case, the insight from the model would have been available after 34% of the analyzed period and before the test data is available. However, it is still interesting to see how derived rules would have played out on the test data (see Figure 4). Overall, we find that insights from the training set continue to be valid in the test set. Yet, the total frequency of errors decreased. This is not surprising, since the observed process is subject to continuous improvements by the quality engineers.

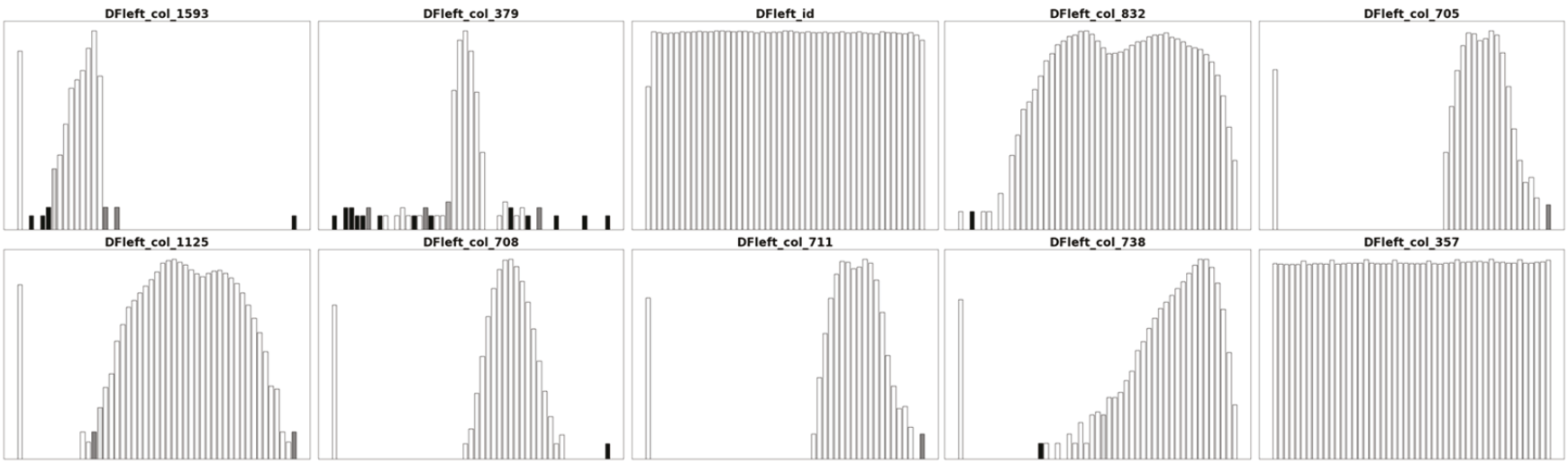

Figure 4: Distribution of feature values for test data. (Y-axes with logarithmic scale, color coded error percentage, white=0%, black=100%).

# 5 Related Work

Recently there has be a strong increase of interest in works on interpretable machine learning[4]. Such works aim to address the opaqueness of many machine learning models and to provide human understandable explanations. (See [1] or [2] for an overview.) Works from this category are generally related to our work. However, we are not aware of any approach that is tailored to the specific needs of quality engineering in manufacturing. The existing works typically aim at providing a holistic understanding of a model. In contrast, our work aims at electing specific insights from the model that are helpful for quality engineers.

More specifically related to our work are works on feature importance. We can distinguish between global importance measures and – more recently introduced – local importance measures. Global importance measures like gain [6], or and similar measures for tree models (e.g. as implemented in [7]) are well established. They are some form of aggregate that aim at capturing the typical (e.g. average) importance of a feature. This type of aggregate conceals features that are only important in some rare cases. However, such rare cases can be of most interest in quality management for manufacturing.

Local importance measures such as LIME [8] and SHAP [4] explain feature importance for individual data points. This is useful for understanding specific predictions. However, they require additional analysis to gain insights beyond the scope of single data points. Closely related to our work is the work of Lundberg et al. [5], who address visualizations on top of SHAP values. Such visualizations are helpful for a detailed analysis. Yet, they show individual data points and leave the interpretation to the user. In contrast, our work points quality engineers to features of interest. Lundberg et al. [5] also use mean SHAP values as global importance measure. This is related to our work in the sense that it is an importance measure on top of SHAP values. However, the measure is not designed for the specific needs of our use case and – as our experiments show – it is less effective than our measures in this context.

Cohen et al. leverage Shapley values in a feature selection mechanism [9]. Thereby they indirectly define feature importance on top of Shapley values, similar to [5]. However, they focus on maximizing the overall performance of classifiers. That is, like other global importance measures, they aim at capturing the typical importance of a feature. Therefore – unlike our measures– their analysis is not tailored to identifying useful features for quality engineers.

---

[4] We use the term to refer to approaches of explainable ML as well, while acknowledging an ongoing debate on how to specify the difference between the terms.

## 6 Conclusion and Future Work

In this paper we introduced feature importance measures that are tailored to the needs of quality engineers in manufacturing. They leverage SHAP values to identify locally important features. Along synthetic and real-world data we demonstrated the benefits of these measures. Our tests indicate that “Max Main Effect” and “Max SHAP” are most promising among the introduced measures. In future work we plan to further investigate the proposed measures to derive recommendations about their application. Other directions of future work are expansions to better leverage local information on feature interaction and to build regional models on top of the identified relevant features. We plan on expanding on the idea to build surrogate trees by (a) building trees only for regions and not as surrogate for the entire model, (b) building trees on a subset of features, and (c) training on the original data and not the model predictions. First steps in this direction have already been made in [11].

## Acknowledgements

This project was funded by the German Federal Ministry of Education and Research, funding line “Forschung an Fachhochschulen mit Unternehmen (FHProfUnt)“, contract number 13FH249PX6. We also like to thank the manufacturer SICK AG for the cooperation. The responsibility for the content of this publication lies with the authors.

# Application of Machine Learning Methods for the Development of Internal Combustion Engines – An Overview

Youssef Beltaifa, Shahida Faisal, Maurice Kettner

Karlsruhe University of Applied Sciences
Gas Engine Laboratory (GenLab)

Youssef.beltaifa@h-ka.de
shahida7085@gmail.com
Maurice.kettner@h-ka.de

**Abstract.** Machine Learning (ML) has a strong potential to improve the performance and effectiveness of several technologies and processes. In recent years, ML has gained in importance, primarily due to its matchless success in image recognition and computer games. These ML accomplishments have motivated to transfer and adapt its algorithms and modeling methods to most scientific disciplines. For instance, in mechanical engineering, ML is coming to hold a crucial position ranging from value chain optimization (production) to substitution of complex simulation models (research and development). In the case of traditional research and development approach, the analysis and optimization of a process are implemented according to the understanding of the governing mechanisms described by physical and mathematical rules. On the contrary, the intelligence of the ML method originates from the extraction of trends and laws based on data patterns, which produces surprisingly good results in many cases. However, it is not entirely evident why it performs so well. One of the most challenging mechanical engineering topics is the improvement of the Internal Combustion Engine (ICE) towards higher efficiency and lower negative impact on the environment. ICEs are very complex systems, which involve high-speed reciprocating motions, transient gas flow and combustion chemistry. Thus, the application of ML methods for ICEs opens new perspectives regarding the modelling, control and maintenance. These topics are addressed in detail in the course of this paper, based on the most relevant published results found in the literature, to provide an overview to the actual research and development of ICE using ML methods.

## 1 Introduction

Internal combustion engines will maintain their position as major power source during the coming decades, particularly for heavy-duty applications [1, 2]. Future internal combustion engines have to comply with tightening legislative emission-limits, high fuel-energy conversion-efficiency, affordable prices and customer requirements. To reach this target, engine researchers worldwide are working on innovative exhaust aftertreatment systems, alternative combustion processes, bio- and renewable fuels, lightweight materials, modern lubricants and advanced manufacturing processes. Within the development of innovative combustion processes (research focus of Gas Engine Laboratory at Karlsruhe University of Applied Sciences) mainly experimental (mostly at the engine test bench) and numerical investigations (0D, 1D and 3D-CFD) are performed. Engine tests are expensive (costly metrology, etc.) and very time-consuming. Moreover, numerical simulations are very dependent on the validation of the implemented physics-based models and necessitate in many cases a large computational capacity. Considering this facts, different alternative approaches that enable saving costs, time and computational power are required. One of the possible solutions that has been increasingly used in recent years is machine learning.
Machine learning is a branch of knowledge dealing with training computers to forecast output values or to classify things without having been explicitly programmed for such function. Machine learning success in many areas like image/speech recognition, effective internet search, self-driving cars is mainly lead by the availability of huge datasets. Machine learning methods can be categorized into two main groups: supervised and unsupervised algorithms, as shown in Figure 1, which depicts some of the most used machine learning algorithms.

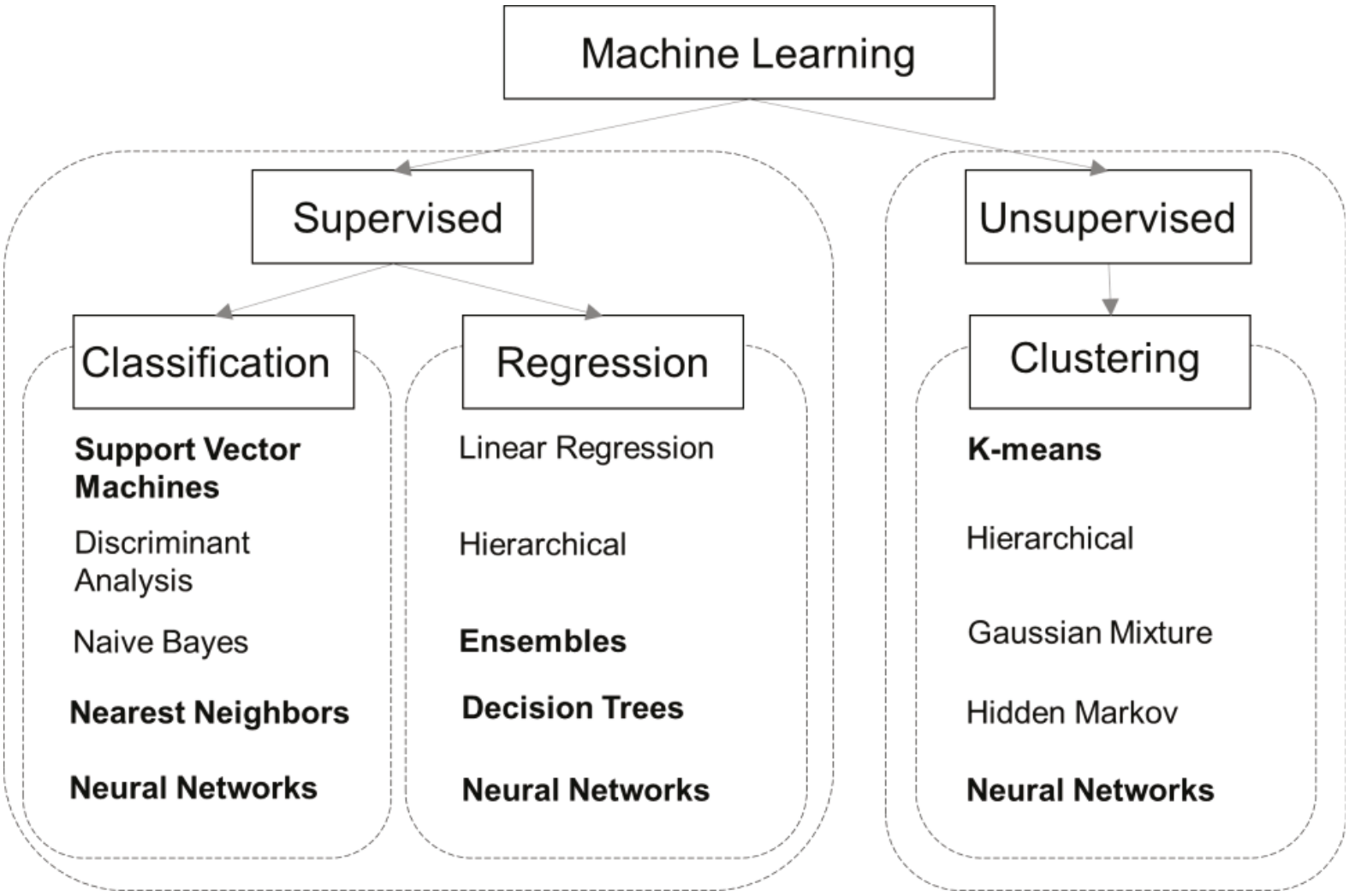

Figure 1. Classification of the most common machine learning algorithms

The methods marked in bold in Figure 1 are the methods that have been used the most in recent studies dealing with internal combustion engines, which are collected and analyzed within this paper. These works are classified within this paper depending on the intended use of machine learning into three categories: Prediction of engine operation parameters and emissions, anomaly detection and predictive maintenance, and real-time engine control. These three topics are covered throughout this paper in detail.

## 2 Prediction of Engine Operation Parameters and Emissions

Many studies [3-18] have demonstrated that engine combustion associated parameters and emissions can be predicted accurately using neural networks over a wide range of operating conditions, given that the training data provides good knowledge of the system's behavior. The combination of fast-computational time and the network's ability to analyze broad non-linear problems can potentially replace expensive exhaust gas sensors (FID, Gas Chromatograph, etc.) and physics-based, computationally intensive engine modeling approaches. Multi-Layer Perceptron (MLP) is a conventional artificial neutral network (ANN) structure that is commonly used for the prediction of engine operating parameters and exhaust gas components. MLP consists of input, hidden and output layers, as seen in Figure 2 on the left.

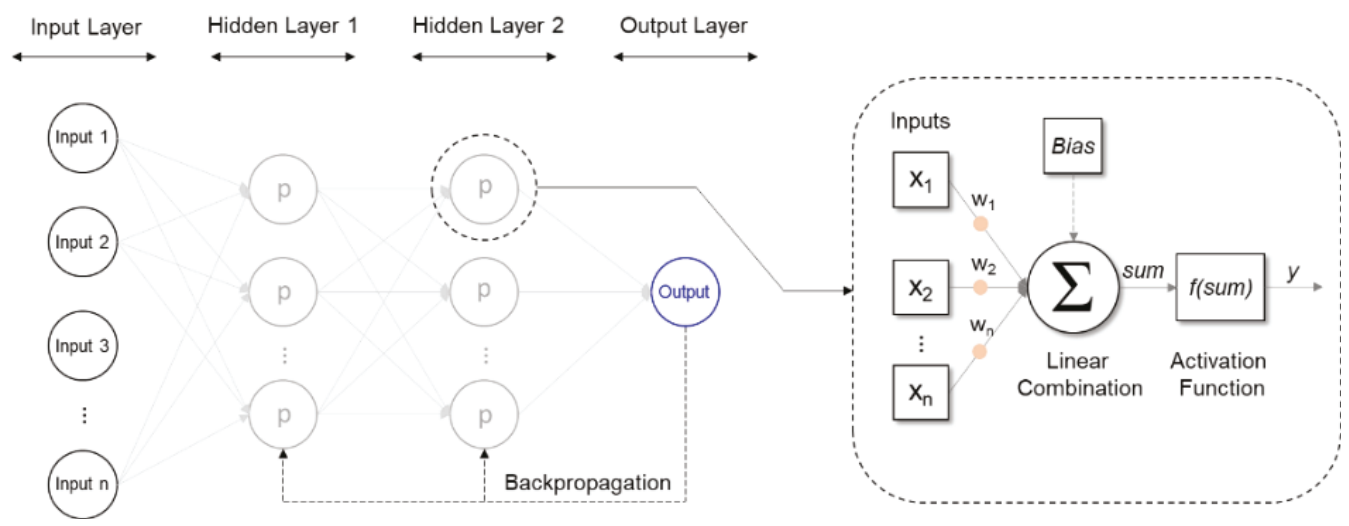

Figure 2. Structure of MLP with two (as example) hidden layers (left), Structure of Perceptron (right)

As shown in Figure 2 on the right, each input is assigned to a weighting factor, representing the importance of the input factor given by the model when predicting the output. Activation functions are employed in the hidden and output neurons, allowing mapping the non-linear relationships between outputs and inputs. For the model training, MLP uses among others gradient descent backpropagation algorithm, where the goal is to minimize the modeling error, meanwhile the weights between neurons are gradually adjusted. Usually, the network training takes place using the Levenberg-Marquardt (LM) backpropagation algorithm, known for high computational efficiency. LM algorithm is a curve-fitting method for solving nonlinear least-squares problems. LM combines the two minimization algorithms gradient descent and Gaussian-Newton to minimize the sum of the squared errors between the fitted model function and the experimental data [19]. MLP with the weighting approach can also be used to have an insight on the dependency of the model output on the input parameters. As an example, the authors in [20] analyzed the relative importance of the in-cylinder parameters affecting the $NO_x$ and HC model output by extracting the saved weights from the trained network. The results yielded that both HC and $NO_x$ were commonly dependent on the engine load and IMEP. The model showed a significant dependency of the $NO_x$ emissions on the peak pressure in the combustion chamber, which is physically reasonable. Higher peak pressures in the combustion chamber are associated with high charge temperatures, which result in turn in a high temperature oxidation of the diatomic nitrogen in the combustion air and the formation of "thermal" $NO_x$. Further studies demonstrating the success of artificial neural networks in predicting and modeling of engine-operation associated parameters and emissions are summarized in Table 1. For these studies, the statistical efficiency of the models lies between 94% and 99.9%. It is important to notice, that MLP with backpropagation is the most frequent encountered machine learning approach in the field of the research and development of internal combustion engines.

Table 1. Summary of MLP applications for the prediction of engine operation characteristics and emissions found in literature

| Author | Output Parameters | Inputs Parameters |
|---|---|---|
| Danaiah et al. [3] | Brake specific fuel consumption<br>$CO$, $CO_2$, $HC$, $NO_x$, $O_2$ | Tert butyl alcohol blend percentage<br>Load<br>Engine speed |
| Uslu et al. [4] | Brake specific fuel consumption<br>Brake mean effective pressure<br>HC, CO, $NO_x$ | Speed<br>Fuel blend (i-amyl alcohool in gasoline)<br>Compression ratio |
| Gürgen et al. [5] | Cyclic variability | Fuel mixtures<br>Engine speed |
| Togun et al. [6] | Specific fuel consumption<br>Torque | Iginition timing<br>Throttle angle<br>Engine speed |
| Tasdemir et al. [7] | Specific fuel consumption<br>Torque<br>Power<br>HC | Intake valve advancement speed |
| Roy et al. [8] | Specific fuel consumption<br>$NO_x$, $CO_2$ | Load<br>Diesel injected<br>Fuel injection pressure<br>EGR |
| Maurya et al. [9] | Ringing intensity at different conditions of a hydrogen HCCI engine | Combustion duration<br>Combustion phasing<br>Equivalence ratio<br>Engine speed<br>Inlet valve temperature |
| Martinez et al. [10] | $NO_x$ | Injection timing<br>Torque<br>Intake pressure<br>Engine speed<br>Ignition timing<br>Throttle angle |

| Author | Output parameters | Inputs parameters |
|---|---|---|
| Hariharan et al. [11] | Brake specific fuel consumption<br>$NO_x$, HC, CO | Load<br>$H_2$ massflow |
| Mehra et al. [12] | Torque<br>Brake specific fuel consumption<br>$NO_x$, CO, HC, $CH_4$ (methane slip) | Engine load<br>Excess air ratio<br>Spark timing |
| Ghobadian et al. [13] | Torque<br>Specific fuel consumption<br>CO, HC | Blend percentage (Waste cooking oil)<br>Engine speed |
| Kapusuz et al. [14] | Brake specific fuel consumption | Power<br>Torque<br>Fuel mass flow |
| Aydin et al. [15] | Brake specific fuel consumption<br>$NO_x$, HC, CO | Fuel injection pressure<br>Biodiesel blend<br>Load |
| Akkouche et al. [16] | Airflow<br>Pilot fuel flow<br>Exhaust temperature | Biogas mass flow (CNG engine)<br>Methane contents<br>Power |
| Oguz et al. [17] | Torque<br>Power<br>Fuel mass flow<br>Brake specific fuel consumption | Engine speed<br>Fuel type |
| Cay et al. [18] | Specific fuel consumption<br>Air-fuel ratio<br>CO, HC | Blend percentage (Methanol / gasoline)<br>Engine speed<br>Torque |

Further neural network concepts/architectures have been used in other studies. Taghavi et al. [21] considered in addition to the MLP network the non-linear autoregressive network with exogenous inputs (NARX) as well as the radial basis function (RBF) network for the prediction of start of combustion (SOC) of a HCCI engine. Input parameters were the intake mixture characteristics (Air-Fuel-Ratio, EGR, intake mixture temperature) as well as the engine speed. The NARX algorithm has, depending on the usage (training or prediction) two structures: The series-parallel (or open loop) and the parallel (or closed loop) architectures. The two network architectures are shown schematically in Figure 3.

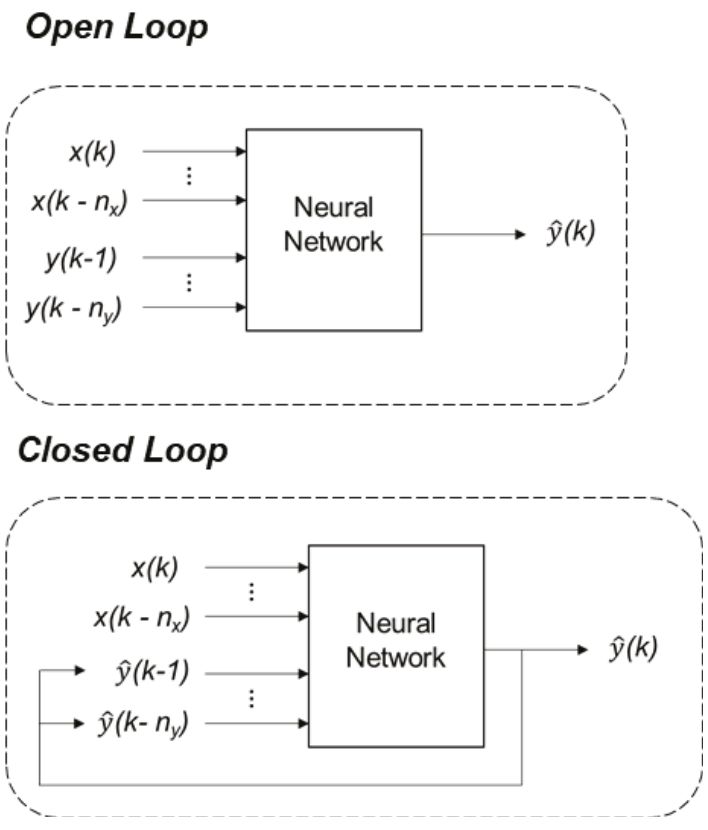

Figure 3. NARX networks architectures

The series-parallel architecture is used for training: the prediction at time-step $t + 1$ is provided based on real input and output values at the current time step $t$, as well as those from the previous $n$ time steps, as shown in Eq. 1. The pure feedforward architecture of the series-parallel (open-loop) structure is applied during training due to the fast static backpropagation [22]. By providing the real input-output pairs during training, the model is able to make future prediction with excellent accuracy. After training, the model produces a final set of adjusted weights, which minimizes the error between the predicted and the true output values. The adjusted weights together with the activation functions approximate the nonlinear mapping function F in Eq. 1. During the prediction stage, the open-loop structure is converted to a closed-loop architecture. Instead of using the real output when making future prediction (time step t+1), the trained model takes the output predicted by itself from the current time step as input, as well as those from the previously n time-steps, as shown in Eq. 2.

$$\hat{y}(t+1) = F\begin{pmatrix} \boldsymbol{y(t)}, y(t-1), \ldots, y(t-n_y), x(t+1), \\ x(t), x(t-1), \ldots, x(t-n_x) \end{pmatrix} \tag{1}$$

$$\hat{y}(t+1) = F\begin{pmatrix} \boldsymbol{\hat{y}(t)}, \hat{y}(t-1), \ldots, \hat{y}(t-n_y), x(t+1), \\ x(t), x(t-1), \ldots, x(t-n_x) \end{pmatrix} \tag{2}$$

Additionally, Taghavi et al. [21] applied the radial basis function (RBF) networks also for predicting the SOC using the same input parameters as in the case of the NARX network. The RBF network typically uses only an input layer, a single hidden layer and an output layer [23], as shown in Figure 4 on the left.

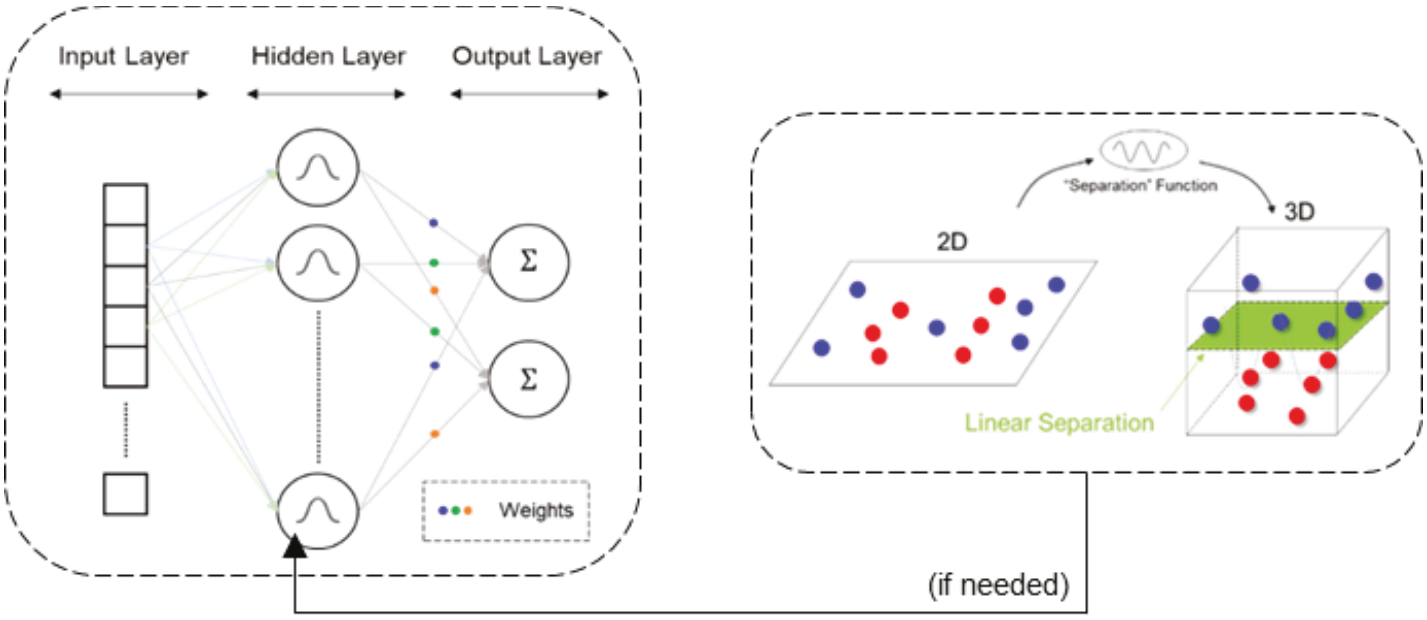

Figure 4. RBF network architecture (left), Schematic description of data dimensionality increase enabling linear separation (from 2D to 3D) (right)

The RBF network is shallow and its behavior is strongly influenced by the nature of the special hidden layer, which performs a computation based on a comparison with a prototype vector [23]. The structure and computations performed in the hidden layer are the key to the power of the RBF network. Here, a hybrid calculation involving two stages takes place. Within the first stage, the linear separability should be ensured: If needed, a projection of the original data points into a higher dimensionality, so that they become linearly separable, is performed. This is based on the Cover's theorem on separability of patterns [24]. For a simplified understanding, Figure 4 on the right shows this step schematically. The second stage is the RBF (Radial Basis Functions) computation, which is based on the comparison of the input units $\bar{X}$ with the prototype vectors $\bar{\mu}_i$ in the hidden layer units according to the equation (3) [23].

$$h_i = \varphi_i(\bar{X}) = \exp\left(-\frac{\|\bar{X} - \bar{\mu}_i\|^2}{2 \,.\, {\sigma_i}^2}\right) \tag{3}$$
$$i \in \{1, \ldots, \mathrm{m}\}$$

m is the total number of the hidden units. Each of these m units is created to have a high impact on a particular cluster of points, which is closest to its prototype vector $\bar{\mu}_i$ [23]. Therefore, m can be regarded as the number of clusters used for modeling, and it represents an important hyper-parameter available to the algorithm [23]. Each unit has a bandwidth $\sigma_i$, which is often the same for all units with the different prototype vectors [23]. After the RBF calculation in the hidden layer, the outputs from the RBFs are weighted and summed by a simple connection to the output layer. The values of the weights need to be learned in a supervised way, dealing with the specific studied case [23]. On the contrary, the hidden layer is trained in an unsupervised way [25]. This involves several parameters such as the prototype vectors, the bandwidths and the number of hidden neurons m. Elaborate description about the determination methods of these parameters can be found in [23]. In comparison to MLP and RBF, the NARX network featured a better prediction accuracy, reaching R = 0.99933 [21].

Another machine learning process used for the prediction of engine-operation related parameters is the Ensemble modeling. For the prediction of the performance as well as efficiency of an engine converted from the diesel CI to the natural gas SI combustion process, Liu et al. [26] applied ensemble methods (bagging and boosting) and compared their prediction performances. The model output was the indicated mean effective pressure (IMEP). Input parameters were spark timing, fuel/air-ratio and engine speed with overall 153 sets of data (122 for training and 31 for testing). "*Unity is strength*": This statement describes in three words the core idea behind the strength of ensemble methods in machine learning. Such methods improve the predictive performance of a single model by training multiple models and combining their predictions [27]. The base models building the ensemble model are "weak" learners, which feature either a high bias or much variance. These are combined within the ensemble method in such a way that they build a strong learner. The combination strategy of the base learners enables to group the ensemble methods in two main categories, depending on how the base learners are generated [28]. The first category is "bagging". Here the individual learners are created independently and their generation can be parallelized [28]. The second category, called "boosting", creates individual learners sequentially in a very adaptive way [28]. Both ensemble methods are shown schematically in Figure 5. For the first step of the bagging algorithm, multiple bootstrap samples (data subsets) are created. These subsets are almost independent datasets created from the original one using random selection [26]. It is important to notice, that the size of the original dataset should be large enough compared to the size of the bootstrap samples so that they are "sufficiently" independent. Subsequently, one "weak" learner (usually the same) is fitted for each bootstrap sample. The predictions of the base learners are then combined to the final prediction of the ensemble model in some kind of weighting process [27]. The combination of the base learners within the bagging method enables the reduction of the variance compared to the variance levels of the single base learners [28]. Therefore, base models with low bias but high variance are more suitable for bagging. Concerning the boosting algorithm, it is not suitable for parallelized computation. Boosting starts with training a first base "weak" learner and then adapt the distribution of the training data according to the output of the base learner such that incorrectly classified samples will have increased consideration from subsequent basic learners [28]. In other words, each new base learner focuses on the most difficult samples (wrongly predicted by the previous learner), so that we get a strong ensemble model with low bias. Hence, base learners with low variance but high bias are suitable to be combined within boosting ensemble methods. Liu et al. [26] found that boosting outperformed bagging, can deal with data set with uneven distributed conditions among the operating range, and provided a high accuracy prediction ($R^2 = 0.9623$) even for low frequency cases, which are poorly presented in the original data set.

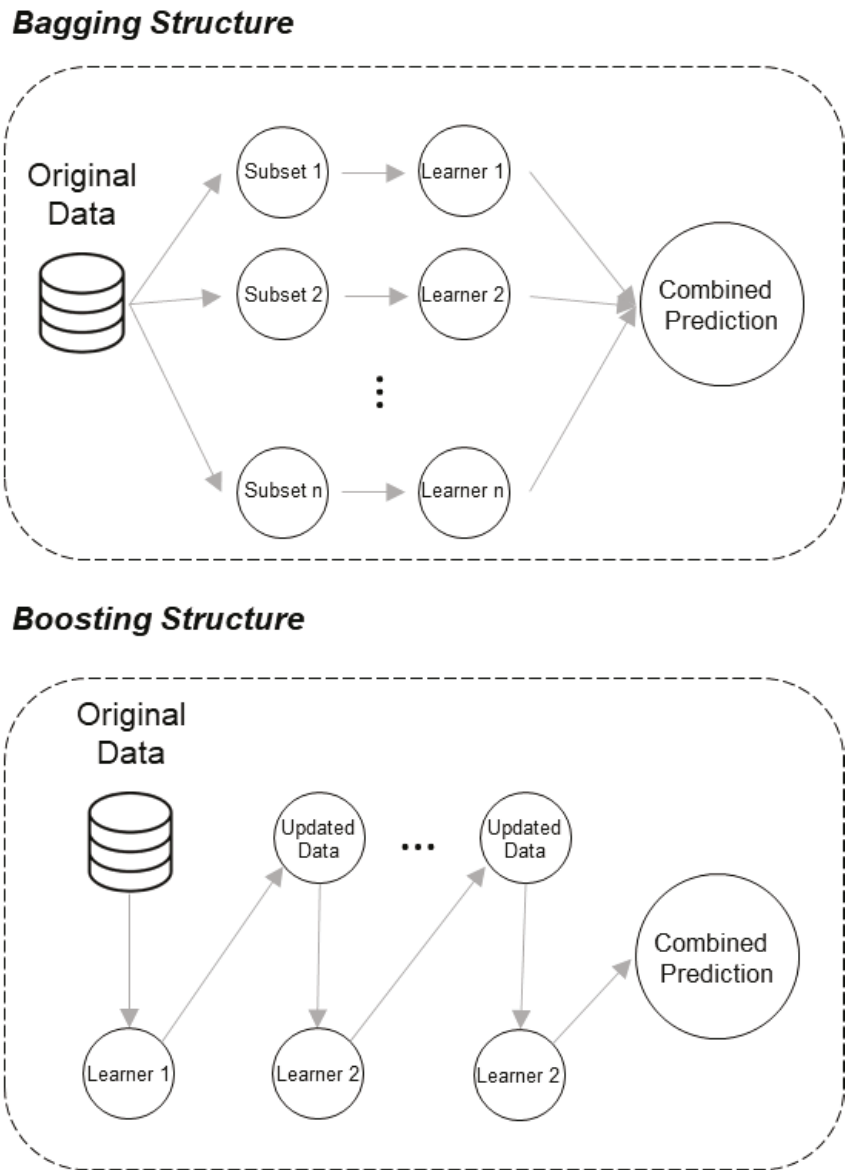

Figure 5. General structure of bagging (top) and boosting (bottom) ensemble algorithms

## 3 Anomaly Detection and Predictive Maintenance

With recent developments, powertrain systems are becoming more complex. Understanding this complexity and dealing with associated particular problems/failures requires evolved methods. New detection methodologies involving machine learning and predictive diagnostics have become the need of the hour [29]. In this frame, Farsodia et al. [30] proposed an approach combining unsupervised learning and clustering to detect anomalies, which may occur in engines or after-treatment-systems (ATS). To validate their strategy, Farsodia et al. [30] addressed the example of the backpressure problem occurring in the diesel particulate filter (DPF) of an automotive diesel engine. Figure 6 depicts schematically the approach proposed by Farsodia et al. [30].

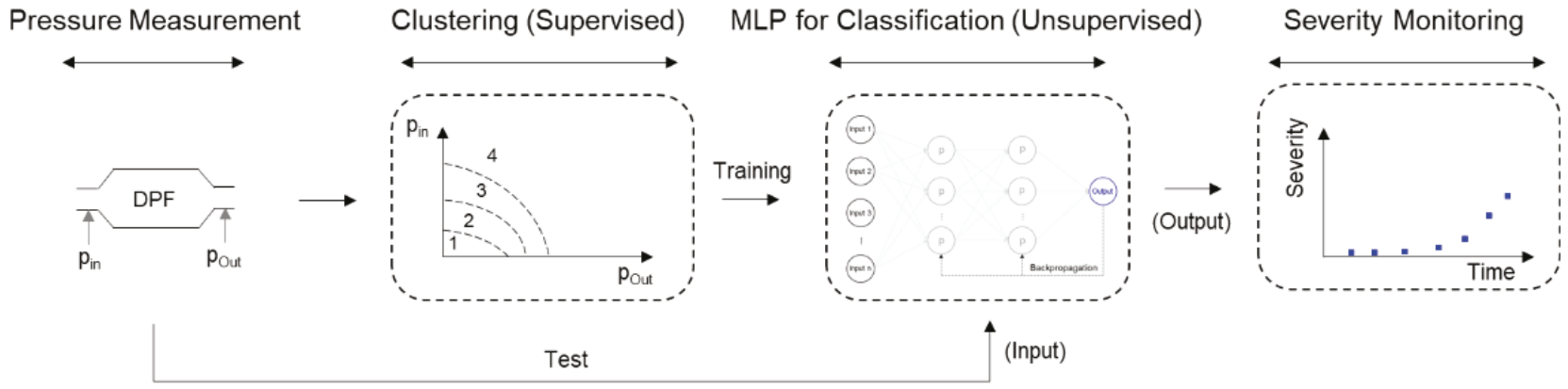

Figure 6. Schematic description of the approach proposed by Farsodia et al. [30] for anomaly detection

As shown in Figure 6, the pressure values before and after the DPF are measured and clustered in a supervised way using the k-means clustering algorithm. Here, the data set will be distributed into "k" clusters. Each cluster has a centroid, which is defined by averaging (taking mean) of the assigned data. First, centroids are determined randomly. Then, data points from the dataset are arranged to the nearest available centroid. The positions of the centroids change within an optimization process until further movement of centroid is not possible. The algorithm is one of the most commonly used techniques for clustering purposes, as it quickly finds the centers of the clusters. Detailed specifications to the k-means clustering algorithm can be found in [31]. In a further step, a classification MLP is trained with the defined classes (clusters) from the supervised clustering step and the associated data. The trained classification MLP is then used in a third step to predict in an unsupervised manner the operating mode of the DPF. For a better monitoring of malfunction cases, Farsodia et al. [28] defined a "severity factor", which enables a time dependent tracking of the DPF functionality degradation. The "severity factor" is derived based on the relative density of data (e.g. malfunction, backpressure, etc.) with respect to total available data points [30]. This "severity factor" gives a pre-warning about any component's malfunction, which will enable the end user to take necessary preventive measures [30].

In a further case, Farsodia et al. [30] presented a methodology involving the weighted k-nearest neighbor (w-kNN) algorithm to predict the temperature shoot-up events in a DPF, which are harmful for the ATS from thermal aging and safety perspectives. kNN is among the simplest statistical learning tools in density estimation, classification as well as regression and known to be trivial to train and easy to code [32]. The difference between the standard and the weighted kNN is that in the weighted approach the prediction of a test point is more depended on the nearest observations [30]. In other words, the k points within the neighborhood of the test point do not contribute equally to the final decision of the test point. Indeed, the closer an observation is from a test point, the more it contributes to its classification. For deeper insight into the w-kNN-methodology, please refer to [33]. After defining the most probably governing parameters on the temperature shoot-up event (engine speed, torque, airflow, HC injection quantity, etc.), Farsodia et al. [30] classified the training dataset, containing temperature shoot-up events, into three different risk categories: "high", "medium" and "low", using w-kNN within a supervised learning process. Category "high" risk implied that there are very high chances that there will be temperature shoot-up post DOC. When testing the trained model with test data from the same vehicle, the model released a warning signal about 60 seconds before the temperature shoot-up event occurred. The algorithm derived from the data is "smart" enough to detect the difference between the high-end temperatures and shoot-up events. However, the excellent beforehand prediction performance was not precisely explained by the authors. Especially, the relationship between the algorithm behind the occurrence of the warning signal and the previous classification step was not discussed.

For a 2.4L diesel excavator engine, Jang et al. [34] proposed also an anomaly detection model, which is depicted schematically in Figure 7. The main idea of the proposed approach is to extract abundant features from gathered data using an autoencoder and then to distinguish between normal and abnormal operating conditions with help of a one-class support vector machine (OCSVM). First, data was collected from 123 different sensors at high frequency (one value every 0.1 s) over 12 days. Due to the large learning dimension, raw collected data cannot be applied to the autoencoder. Therefore, the authors used statistical values instead (median, variance, deciles, etc.). This enabled the reduction of the data amount and the expression of data characteristics more prominently. In a second data-dimensionality-reduction step, the autoencoder is applied to the derived statistical indicators. Autoencoders are neural networks that can automatically (unsupervised) learn useful features from data [35]. Autoencoders work by compressing the data into a latent-space representation also known as bottleneck, and then reconstructing the output from this representation. Jang et al. [34] used compressed features from the latent space of the autoencoder network as input for the classification algorithm, which is the OCSVM, which is used in the context of pattern classification to discriminate between two classes [36]. More details to support vector machines can be found in [37]. Ten days of "healthy" measurement data were used to train the OCSVM model. The anomaly classification performance was evaluated using data from two days, where faulty events were present. The model accuracy reached up 73%. However, the model achieved an excellent recall score with 83%, indicating the model reliability to ignore false alarms.

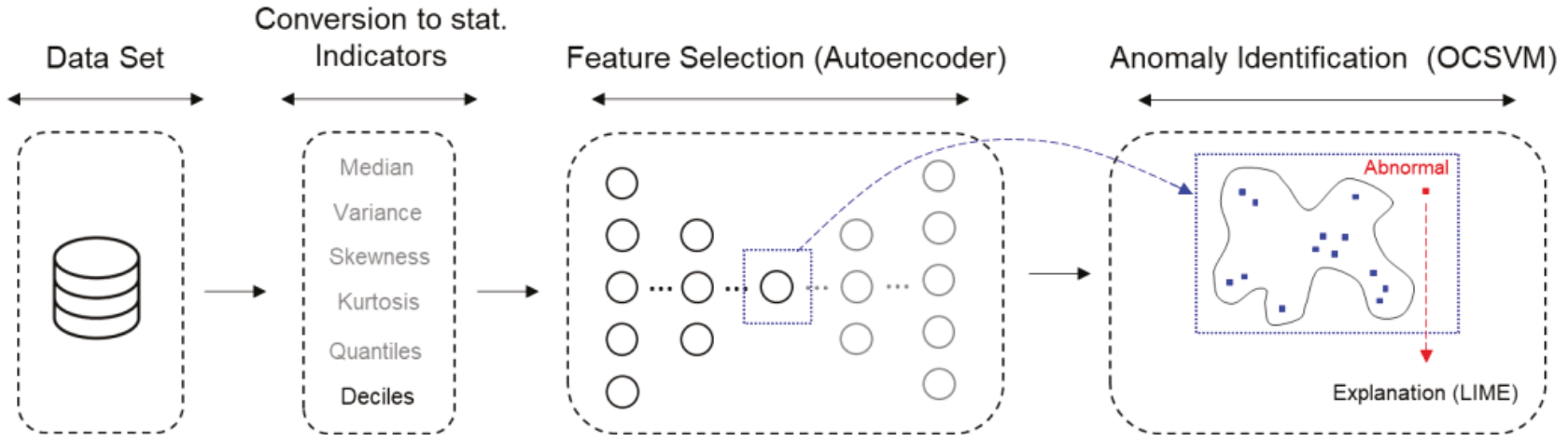

Figure 7. Schematic description of the approach proposed by Jang et al. [34] for anomaly detection

In a classification task, the positive or negative results are insufficient without explaining the classifier's decision-making. Therefore, Jang et al. [34] used Local Interpretable Model-agnostic Explanation (LIME) to get more in-depth interpretation regarding the most critical factors contributing to the classification results. LIME is an algorithm for providing interpretable explanations for the non-interpretable (black box) ML models such as neural networks. An in-depth explanation to the LIME approach can be found in [38].

## 4 Real-Time Engine Control

Conventionally, ICEs control is based on map-calibrations tuned by full factorial or design of experiments processes. To reach engine efficiency targets, manufacturers are increasing the number of actuators [39], leading to an increase in the calibration design space and thus affecting the real-time capability of the control unit, especially for transient operating conditions. Thus, new control techniques, which can better deal with increasing actuators number, are developed. In this context, Egan et al. [40] introduced a hybrid modelling approach involving the non-linear model predictive control (nMPC) in combination with static and dynamic (time-dependent behavior) artificial neural networks. nMPC is an advanced control strategy that has the greatest acceptance in the industry, because it provides an intuitive approach to the optimal control of systems subject to constraints [41]. Nevertheless, it has its drawbacks, mainly the large amount of calculation required, since an optimization problem is being solved at every sampling time [41]. Thus, non-linear MPC use for ICE is usually limited due to the short period available between engine cycles (~25ms at 5000 rpm) and the limited computational power of automotive control units [42]. Alone the evaluation of non-linear engine-models and its linearization take about 60%-75% of the total computational time per nMPC iteration [41]. Taking into consideration that neural networks can computationally efficient capture non-linear behavior and have the ability to be linearized in minimal time [40], Egan et al. [40] proposed to replace traditional engine modeling methods by artificial neural networks and use them within the nMPC framework, as shown schematically in Figure 8. Their aim was to accelerate the nMPC processing time and thus facilitating its integration into the engine control unit. Egan et al. [40] found that the proposed control system successfully controls the investigated engine with tractable computational load, opening doors for the application of their approach for future Engine Control Units.

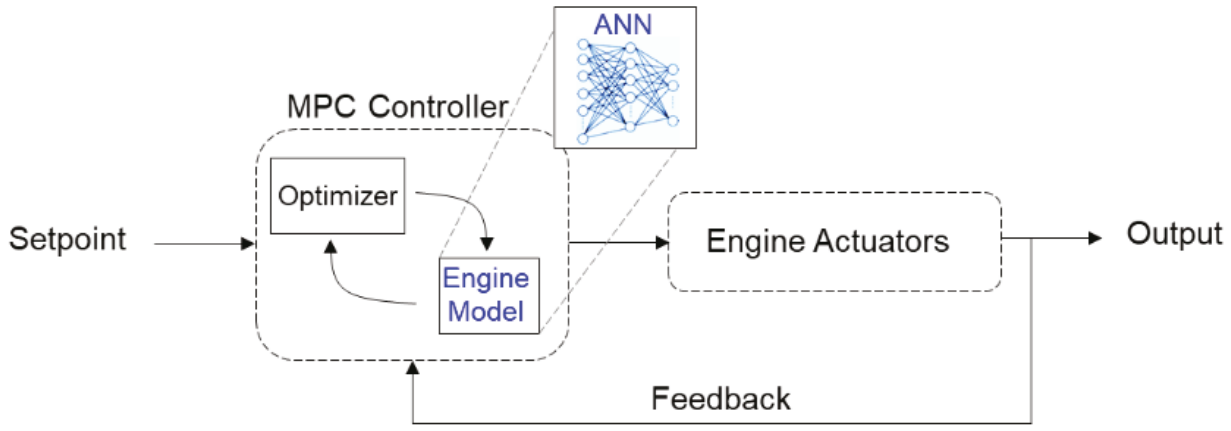

Figure 8. MPC architecture with ANN as engine modeling approach, approach proposed by Egan et al. [40]

## 5 Summary and Outlook

This paper deals with existing applications of machine learning in the field of internal combustion engine development. Most of the work found in literature handles empirical model building with the use of artificial neural networks, especially the MLP structure, which is quite suitable for mapping non-linear processes occurring in an IC engine. Other structures for modeling and predicting engine-related parameters have also been found in the literature, such as the NARX and the RBF networks. Further studies considered ensemble methods, which are well suitable for modeling engine parameters (especially the boosting algorithm).

In addition to model building, machine-learning methods in the field of combustion engines are also used for the predictive maintenance and anomaly detection. Especially, Clustering (k-means clustering, k-nearest neighbors, support vector machine, etc.) and “deep learning” (autoencoder, convolutional neural network, etc.) methods are used for these purposes. In this process, XAI (explainable artificial intelligence) methods (e.g. Local Interpretable Model-agnostic Explanation) were also employed to get more in-depth explanations and interpretations for the machine learning model decisions. These methods then ultimately allow a more advanced understanding of the engine's behavior.

One more field of machine learning application in the engine development segment is the real-time engine control. New engine Control systems face the challenge of dealing with growing engine complexity and thus increasing computational intensity. In this context, artificial neural networks offer the possibility to reduce the computational effort without affecting the number of actions to be managed with a given time slot. Indeed, the MLP structure have the advantage that it can simply map the engine operation (highly non-linear) and be easily linearized (from non-linear to linear), which motivates for its integration into existing control systems involving computing intensive optimizers.

However, machine learning also has its drawbacks. One cannot expect any magic from machine learning algorithms. Indeed, simple learning programs are unable to learn complex concepts from few input data. To deal with this fact, more data and “smarter” algorithms are needed. Therefore, researchers are increasing the application of deep learning (DL) methods such as convolutional neural networks (CNN) [43], generative adversarial networks (GAN), and autoencoders (AE), which has proven to enable the automatic detection of most significant features during the training phase and to exceed the prediction accuracy of the simpler ML models with conventional human-aided feature extraction. Therefore, we expect an increasing use of deep learning algorithms within the future research and development of internal combustion engines.

# Modeling for Explainability: Ethical Decision-Making in Automated Resource Allocation

Christina Cociancig[1], Christoph Lüth[2], Rolf Drechsler[3]

[1,2,3] University of Bremen; German Research Center for Artificial Intelligence (DFKI GmbH), Germany
[1] chrcoc@uni-bremen.de
[2] christoph.lueth@dfki.de
[3] drechsler@uni-bremen.de

**Abstract.** Decisions delegated to artificial intelligence face an alignment problem: humans expect the algorithm to make fast and well-informed decisions aligning with human morals. In the design and engineering process of algorithms, ethical principles enter the black box explicitly and implicitly as functional or non-functional properties, much to the detriment of explainability and transparency. Previous work has established surrogate modeling to promote explainability and transparency of the decision-making process. We extend on this, model in lower complexity decision trees and as labeled transition systems, which is a method inherent to bisimulation theory, as well as evaluate on synthetic data with a rule-based algorithm. As a case study, we analyze the triage processes in German and Austrian hospitals during the COVID-19 pandemic, based on official guidelines that regulate the allocation of intensive care unit beds. We discovered that the decision processes are similar, however, the systems do not behave in the same manner. The diverging behavior equates to a discrepant ratio of patients treated in intensive care in contrast to the general ward. Our insight leads us to the conclusion that our approach ensures ethical decision-making in healthcare and should be considered due to its explainability and transparency.

## 1 Introduction

Machine learning and artificial intelligence (AI) in general can support virtually any human decision. Whether we assume the decisions made or intervene to revise it, with data supply and automation, we entrust a mostly black box with coming to the best conclusion it possibly can. Often, these decisions we delegate require fast conclusions and a high degree of expert knowledge. While algorithms can fulfill these criteria of decision-making, a human decision is inherently one that draws upon human morals. That we demand of algorithms to emulate this, gave rise to a new field of research that is located at the intercept of computer science and philosophy: machine ethics [1].

Machine ethics, also referred to as AI ethics when defined in a narrower sense, became an increasingly vital factor in AI research, because the decisions we now automate, have direct implications on human lives. Machine ethics places an emphasis on the establishment of values that should steer the development and deployment of artificial intelligences in the form of guidelines for "ethical AI" [2]. Ethicists agree with the pressing issue of ethical algorithmic decision-making by advocating particularly for transparency [3] and explainability [4] of the decisions produced by the black box that a machine learning algorithm, or even more so a deep learning algorithm, can represent.

Related work in algorithmic explainability and transparency put forward various approaches, including but not limited to, surrogate modeling and formal verification. Previous research in the area of surrogate modeling advanced to complex cases of decision tree modeling of a neural net, with a well-founded result of reduced complexity, high fidelity, and comprehensibility [4]. Even though it has not yet been done in full terms, approximate-bisimulation has been employed to model (dynamic) neural networks and their behavior in terms of their input and output [5]. These approaches add to the extensive list of measures to analyze the decision-making process with the intention of optimizing for ethical decision-making of the system. However, they fail to consider that some explainability and transparency is better than none, especially for use cases that involve critical decisions in healthcare.

AI represents an evolution of informed decision-making in the medical field [6]. In the clinical environment, well informed decisions must be made fast. Not only in Germany this potential has been identified, and discussions to implement decision-making software are well under way or already implemented. SmED (short for “structured initial medical assessment in Germany”) is an algorithm that assists medical on-call services to decide where a patient’s healthcare needs can be addressed best: a general practitioner or an emergency clinic [7]. While both are not yet applied in the clinical context, OPTINOFA (short for “optimization of emergency care through a structured initial assessment using intelligent assistance services”) aims to provide an algorithmic assessment of the urgency of treatment in clinics [8]. The most striking difference between the softwares: SmED appears to be rule-based and is not openly accessible, OPTINOFA is composed of an AI and will be openly accessible.

In the interest of examining a contemporary decision process in healthcare, as a case study, we compare approaches to the decision-process of triage during the COVID-19 pandemic in two countries: Germany and Austria. The decision processes of triage are based on a practice of resource allocation historically attributed to military medicine, which categorizes patients and commonly prioritizes treatment of patients with a high chance of survival [9]. At the beginning of the pandemic, the allocation of resources, i.e., particularly intensive care beds that can accommodate a ventilator, has been regulated by strict guidelines in Germany [10] and Austria [11]. It is exactly this type of situation in which algorithms are capable to provide humans with relief to make well informed, fast decisions. However, it is of the utmost importance that the decisions provided by machines agree with human ethical values.

With this paper we provide a recommendation for transparent and explainable algorithmic decision-making in healthcare, that complies with the ethical principles of explainability and transparency in form of non-functional properties of the system. We build on related work in the area and propose formal surrogate modeling with decision trees, including associated entropy and information gain values indicating the informative strength of a node within the tree, as well as modeling as labeled transition systems, a method inherent to bisimulation, which provides a comparative analysis of the behavior of two systems [12]. With an algorithmic data evaluation, we support the findings of our analysis numerically, and demonstrate, that drawing on metrics inherent to the models provide reference for a comparative analysis of systems. With this recommendation, we hope to contribute an opportunity for ethical healthcare software to be transparent and explainable for medical professionals and patients alike.

## 2 Methods

This section outlines the methodology applied to construct decision trees and a bisimulation evaluation of triage processes, as well as a description of the algorithm we deployed to measure effects in ratios. Our hybrid approach of two comparative modeling systems and a test with synthetic data was chosen, because it gives a valuable insight into robust tools that can be accessed for the purpose of investigating strengths and weaknesses of systems such as the triage decision process. We compared and identified differences in the German and Austrian triage guidelines first by focusing on their underlying ethical principles and subsequently in terms of their functional properties.

Both guidelines are governed by implicitly or explicitly defined ethical principals. The Austrian guideline, “Allocation of intensive care resources due to the Covid-19 pandemic”, lists four ethical principles influencing every decision within the triage process: justice, non-maleficence, doing good, and the observation of autonomy of the patient [11]. The definition of each principle is linked to several more, some non-ethical, values that shall be upheld: using resources efficiently, allocating fairly, not endangering the supply system, serving the well-being of each individual patient, respecting guardians of patients, and respecting individual freedom [11]. Although the Austrian guideline does not connect these values and principles to individual decisions, each decision made within the process should be guided by them. The German guideline mentions ethical principles predominantly implicitly by connecting them to decisions in the assessment process, including the needs of the patient for intensive care unit (ICU) treatment and the patients will that are directly reflected in decisions within the process, whereas a prohibition of discrimination due to age, social characteristics, and disabilities, as well as fairness are implicit and not represented as decisions within the process per se [10].

## Decision Trees

To investigate the sequence of decisions and the associated informative value of decisions, we manually translated the triage guidelines into their respective decision trees and compared the metrics of entropy and information gain, both inherent to information theory. Each decision outlined in the triage guidelines translates into a decision node of a tree. The end nodes represent the decision for or against ICU treatment of an individual patient.

Entropy is measured in bits and represents the average level of information or uncertainty of possible outcomes of a variable. Given a variable $X$, with possible outcomes $x_1,\ldots, x_n$, with an associated probability of $P(x_1),\ldots, P(x_n)$, the entropy of $X$ is defined by Shannon [13] as:

$$H(X) = -\sum_{i=1}^{n} P(x_i) \log P(x_i) \tag{1}$$

Entropy can be calculated for each node in a decision tree. For a description of the strength of the node, the value of information gain expresses the change in information entropy from one node to the next:

$$IG(T, a) = H(T) - H(T|a), \tag{2}$$

where $H(T|a)$ is the conditional entropy of $T$ given the value of $a$ [14]. Information gain can therefore adopt values between zero and one, a higher information gain is associated with a strong decision node, at which an informative decision is made.

## Evaluation

To verify the differences in the performance of the two systems, we evaluated benchmark data flowing through the process modeled as decision trees. To this end, we implemented a rule-based algorithm that sorted and evaluated synthetic data of 100 patients based on the health data required for the German triage decision process.

Although the triage criteria to receive intensive care are different for German and Austrian patients, both guidelines have baselines in common, whose negation can in no circumstance lead to treatment in intensive care. For one, a patient must give consent to receive intensive care. Moreover, though the German guideline explicitly states that a necessity for intensive care must be assessed, the same can be assumed for the Austrian system. German medical personnel are furthermore urged to assess the *prospect of success* of intensive care for a patient as one of the first steps in the triage process, whereas Austria assesses *hopelessness* and *proportionality* of ICU treatment only as a criterion for the abortion of intensive care [11]. For our purpose, we equate the assessment of prospect of success in Germany with the assessment of hopelessness and proportionality in Austria and quantify this criterion with 96%, i.e., the average reported survival rate of COVID-19 in Germany and Austria [15].

Beyond the shared baseline assumptions, the triage systems additionally assess the patients on health criteria. These criteria have determined our algorithmic implementation and data development and have been summed to a health score. The criteria formulated for the health assessment of German patients consist of five points, which quantify scores or represent the presence or absence of a criterion: heightened severity of illness, e.g., acute pulmonary embolism, acute organ failure assessed on the sepsis-related organ failure assessment score (SOFA), a prognostic marker for COVID-19 patients (we assume this marker to be a positive COVID-19 test), comorbidity, e.g., neurological disease, and health status assessed on the clinical frailty scale (CFS) [10]. In Austria, health assessment is done in nine points: chance of survival via SOFA score, comorbidity, presence of cardiac insufficiency or failure, renal insufficiency or failure, presence of immunosuppression, dementia assessed on Activities of Daily Living score (ADL), pulmonary disease, other primary disease, and other relevant criteria [11].

As a first step to data creation, we affirmed the baseline assumptions for ICU treatment and created five health data points for each patient randomly, modeling the German health assessment. We randomly one-hot encoded for the presence or absence of a criterion and appoint scores where applicable, i.e., a SOFA score between 0-24, counts of comorbidities between 0-5, and overall health status of CFS score between 1-9. This encoding resulted in an overall health assessment score sum between 1-40, with a higher score being associated with a more critical condition. For scores under 20, we assumed care at the general ward as sufficient, patients with scores over 20 require intensive care. In the corresponding trees, this decision node is represented as a score of under 50% or over 50%.

As a next step we translated the patient data for the Austrian decision process of nine health assessment points, which involved the addition of more scores. As the survival chance in Austria is also indicated by the SOFA score, we adopted it from the German model patient. We transferred comorbidity counts over zero as the presence of a comorbidity and associate a heightened severity in Germany with a primary disease in Austria, as well as translate the CFS score of five and above as the presence of dementia. The remaining criteria, i.e., cardiac or renal insufficiency, immunosuppression and pulmonary disease were again one-hot encoded with a random distribution. The complete assessment results in an overall summed score between 1-31. We again divided the score in over 50% and under 50%, a score of 15 or lower does not receive intensive care.

## Bisimulation

To further investigate the difference in behavior of the two triage processes, we remodeled the decision trees as a bisimulation evaluation. Our notation for this evaluation was adopted from Davide Sangiorgi [12]. Specifically, we explore the states and transitions of the processes modeled as labelled transition systems (LTS), which are formally described with triples, i.e., $(Pr, Act, \rightarrow)$ where $Pr$ is a (non-empty) set, also referred to as the domain or the set of the processes of the LTS, $Act$ is the set of actions or transitions, and $\rightarrow$ denotes the transition relation between processes. Bisimulation is a binary relation on the states of two systems $P$ and $Q$, if for all $\mu$ we have:

$$\text{1) for all } P' \text{ with } P \xrightarrow{\mu} P', \text{ there is } Q' \text{ such that } Q \xrightarrow{\mu} Q \text{ and } P' R\, Q';$$

$$\text{2) for all } Q \text{ with } Q \xrightarrow{\mu} Q', \text{ there is } P' \text{ such that } P \xrightarrow{\mu} P' \text{ and } P' \, R \, Q'. \quad (3)$$

If the bisimulation is complete, meaning for each process in the system $P$ there is an equivalent process in system $Q$, the systems are bisimilar, i.e., they behave in the same way. If not all processes in system $P$ can be mapped to an equivalent process in system $Q$, the systems might have equal inputs and outputs, but internally do not behave the same way.

# 3 Results

## Decision Trees

Both triage decision processes were modeled as decision trees. Figure 1 is a comparison of the German triage system and the Austrian triage system. For reference, we added entropy ($H$) values for each decision node in the trees. Due to our assumption of the baseline criteria for intensive care treatment as being met, i.e., necessity of treatment (represented as the first decision node with $H = 1$ as is standard for decision trees) and consent of the patient, the respective entropy values do not amount to expressive decision nodes.

However, we were specifically interested in the decision node labeled *prospect of success*, as the location of this decision and the corresponding nodes in the trees is an identifiable difference between the two decision processes. Based on our assumption that the prospect of success of treatment, which is assessed early on in Germany, is equivalent to the criterion of hopelessness and proportionality, which is assessed after the ICU treatment has already commenced for the Austrian patient, the anticipated entropy values correspond to the same entropy of $H = .24$. Again, these values are identical, because we assume a survival chance of 96% for both countries, as it is the reported survival rate of COVID-19 in both countries [15]. The information gain, however, of the prospect of success node amounts to $IG = .76$ in the German system, compared to the Austrian system of $IG = .75$.

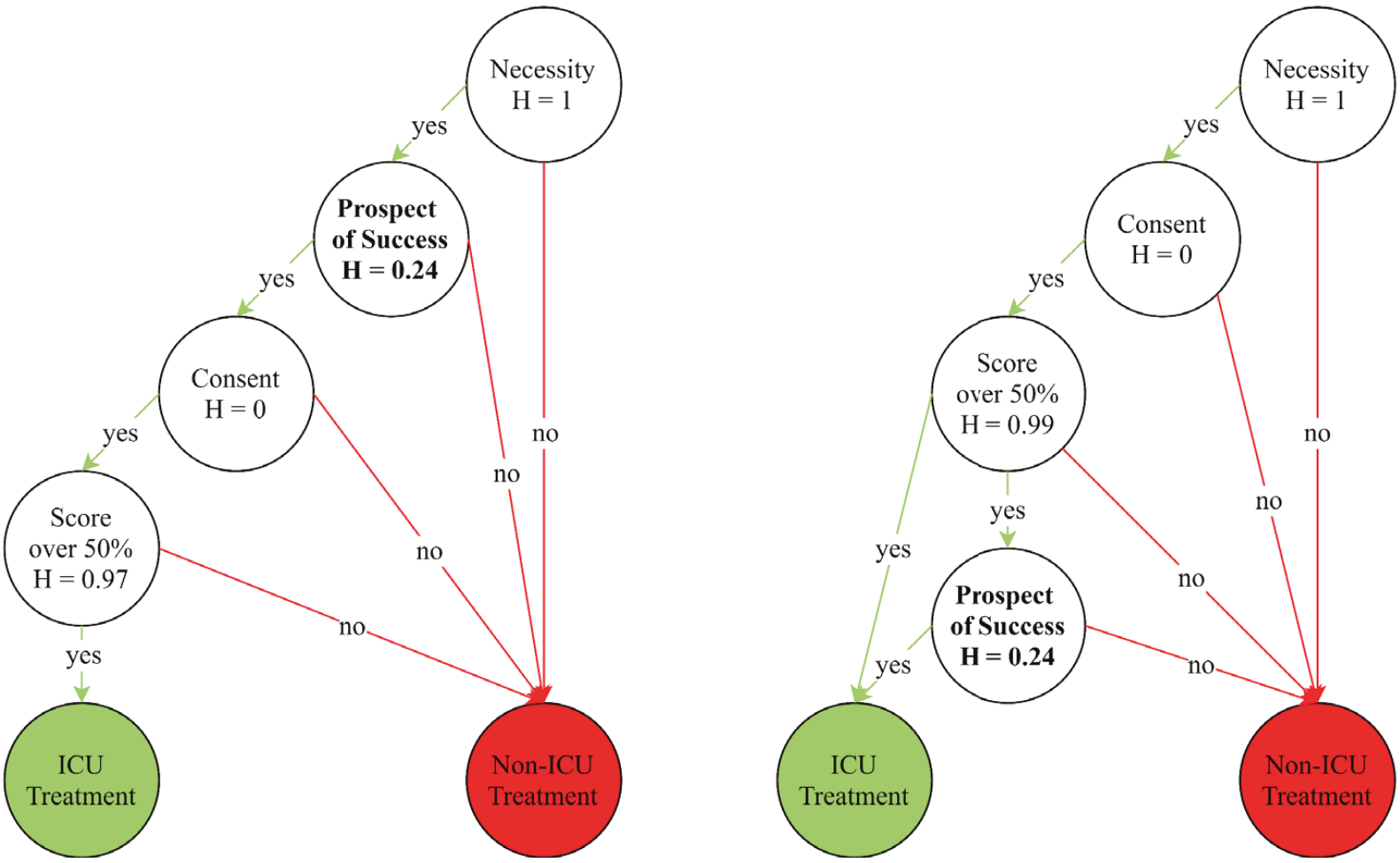

**Fig. 1.** German (left) and Austrian (right) triage decision tree, entropy value per decision node

## Evaluation

The rule-based algorithmic evaluation of our data of 100 patients revealed, that the Austrian system initially treats more patients in ICU, more specifically 56% in comparison to 39% in the German system. This ratio is not representative, however, of the mandatory re-evaluation which is featured in both triage processes and more visible in the labeled transition systems.

## Bisimulation

The LTS of triage in Germany and Austria offer valuable cues as to how the decision process is executed and can be described and analyzed formally. Figure 2 compares the LTS side by side and indicates their domain, actions, and transition relations. For the sake of clarity and brevity, we indicated transitions abbreviated, e.g., from $R_1$ to $R_6$, instead of "No Necessity assessed" as it would be formally described, we simply indicated "No Necessity".
Not only does the LTS comparison demonstrate that the Austrian triage ultimately has less states until it arrives at a final decision over ICU treatment or no ICU treatment, the formal description of the LTS is a further indicator for the similarity of the systems. As we examine the binary relation between the two systems, we pair processes with the same transition relations:

$$R = \{ (R_1, Q_1), (R_3, Q_2), (R_4, Q_3), (R_6, Q_5) \}. \quad (4)$$

Given that not all states in the German system have equals in the Austrian system, the bisimilarity of the systems cannot be proven and therefore indicate that the systems do not behave in the same way.

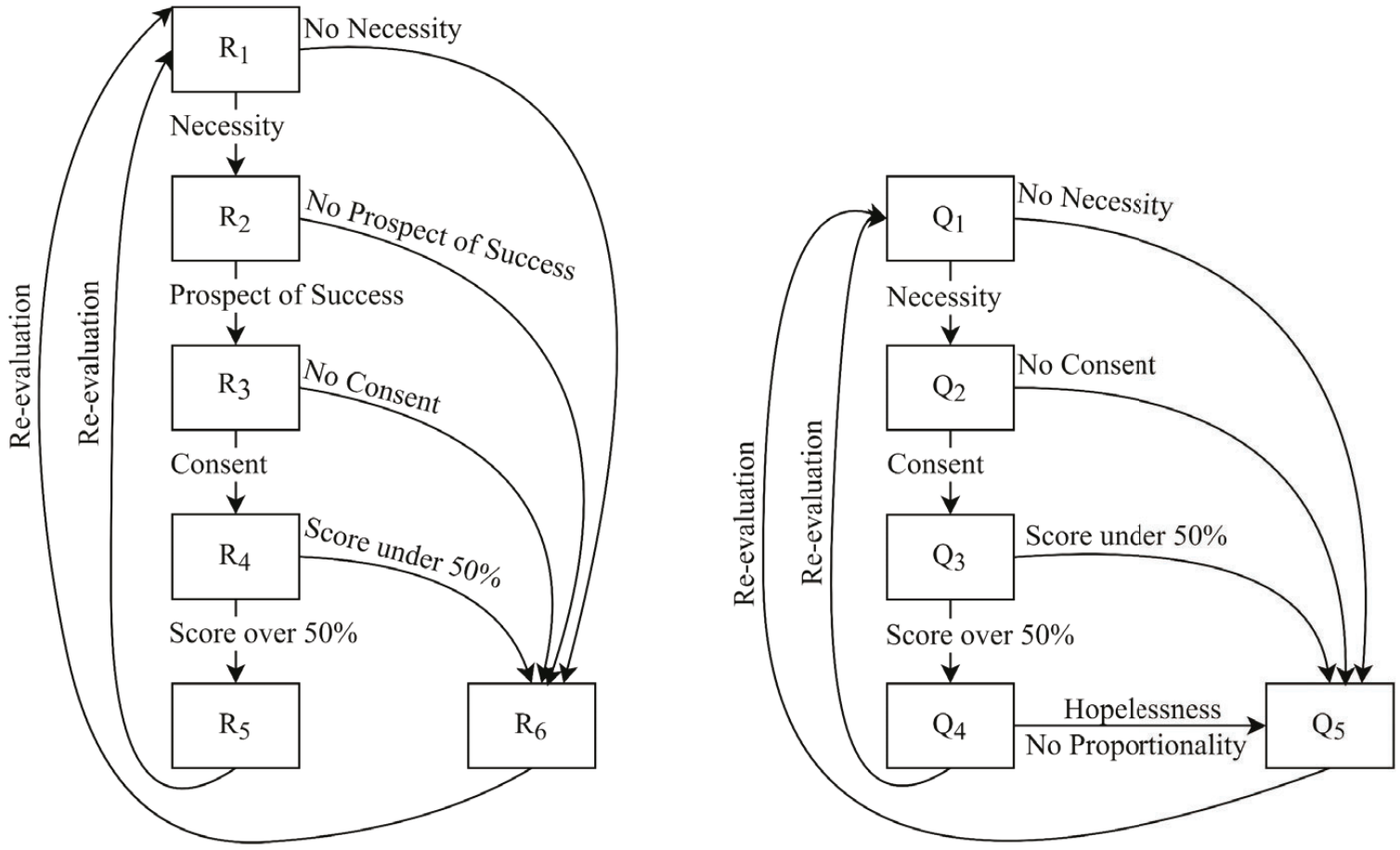

**Fig. 2.** LTS of triage in Germany (left) and Austria (right)

# 4 Discussion

The importance of interpretable models, that promote transparency and explainability cannot be emphasized too much. Our method of surrogate modeling as decision trees and labeled transition systems, as well as evaluating on synthetic data, has not only enabled us to identify the different approaches between the triage processes in Germany and Austria, but also given us the possibility to explain any given final decision by traversing through the models. Therefore, by modeling in comparison, we achieve transparent and explainable decision-making as it is promoted by machine ethics [1,3].

At first sight, the systems seem to align in their design, as they assess patients on similar criteria. With the data evaluated based on the decision trees, however, we see that the smallest differences in the processes have a substantial effect on the number of patients treated and thus possibly on the number of lives saved. The decision trees have provided a comprehensible model to enable further evaluation, as we expected and was touched on in [4]. The most striking observation to emerge from the evaluation of information gain, is the difference between the values for the *prospect of success* decision node. Even though the numeric difference between the German and Austrian node is marginally small, it indicates that the informational value of the node is in fact different, and this difference is due to its position in the decision process, i.e., the rank of the decision.

Even though a bisimulation of neural networks has not yet been successful [5], we were able to considerably benefit from modeling the rule-based triage systems as labeled transition systems to reproduce the finding of differences from the decision trees. As the binary relation is incomplete, the evaluation confirms that the decision processes are not equal, and furthermore, that this inequality can be attributed to an inequality in the patient's health assessment and the order of the decisions made in both countries. The model borrowed from bisimulation has provided us with the ability to formally describe with states and transition relations are represented in both decision processes.

Although we gained considerable insight from modeling the processes, the link to ethical principles, especially to the overarching principles of explainability and transparency of (algorithmic) decisions, were made by us. Neither the German nor Austrian guideline mentions these principles as essential to decisions in triage. Furthermore, despite the insight, we are not qualified to make statements regarding fairness of the decision process, even though the German triage guideline acknowledges this ethical value [10]. Synthetic data is simply not suitable to evaluate this metric on.

## 5 Conclusion

We have outlined a comparative method of evaluating decision-making by modeling the decision processes in form of labeled transition systems, as well as decision trees and the corresponding metrics of entropy and information gain, to promote the transparency and explainability of decisions within the triage process. For our case study of triage processes in Germany and Austria, we found that differences in the non-functional properties the decision process adheres to, i.e., in broader terms the ethical implications of the triage system, has consequences for the behavior of the system and these consequences can be measured. In the context of triage, the measurable difference of courses of action ultimately equates to lives saved or lost.

For decision-making in triage, we conclude that a formal modelling allows for the analysis and precise comparison of systems. This can be applied to systems of different countries, or two competing systems within one country. Our findings assert that a careful engineering of the decision process, whether with implicitly or explicitly translating ethical principles into decisions, can lead to more efficient decision-making. Besides efficiency, with modeling of the design we access insight about comparability and weaknesses of systems, and measurable metrics can lead to a better understanding of the outcome of specific decisions. In combination, formal modelling, i.e., decision trees and bisimulation, lead to transparency and explainability of the decisions made.

Our method could advance many other decision processes conducted by AI or machine learning in healthcare. Decision-making softwares, whether they include an AI or rule-based algorithm similar to what we employed and SmED appears to be, behave according to underlying ethical principles. As we have concluded from the triage guidelines in our use case, the link between these principles and the properties of the system they are embedded into, are not always functional, i.e., direct and apparent. Yet, the insight gained from the models and their corresponding metrics provides an excellent resource to ensure ethical decision-making. Future work evaluating its models algorithmically on benchmark data should, however, aim to collect or obtain organic data, which was beyond of the scope of this research.

# Condition Monitoring of Electric Motor with Convolutional Neural Network

Tanju Gofran [1], Maurice Kettner [1] and Dieter Schramm[2]

[1] IEEM-Institute of Energy Efficient Mobility, Karlsruhe University of Applied Sciences
gota0001@h-ka.de,maurice.kettner@h-ka.de
[2]University of Duisburg-Essen
dieter.schramm@uni-due.de

**Abstract.** Safe, efficient and uninterrupted operation of machine requires continuous monitoring of its health and modern autonomous smart factory demands a Condition Monitoring (CM) process without direct human involvement. Deep Learning (DL) algorithms have shown great success of learning directly from data in various real life applications and recently it become also popular in CM researches but still detail clarification of selecting the DL design and its relevance to learn the features from data are often missing. This paper shows a DL algorithm - Convolutional Neural Network (CNN) to CM of an Electric motor from its external vibration. The output of the deep layers of the learned model is analyzed to explain how the model extract features of raw vibration input and do the classification of different conditions.

## 1 Introduction

In the age of the fourth Industrial Revolution, application of Artificial Intelligence is not just a demand but a necessity. A smart factory involves numerous machines and sensors requiring machine to machine and machine to human communication without interruption. Condition monitoring and predictive maintenance of the machines to prevent failure in advance or detect any anomaly early enough before breakdown is one of the key trends of Industry 4.0.

Application of Machine Learning (ML) algorithms in the field of Condition Monitoring (CM) of Electric machines (EM) has been investigated and implemented in reality in various researches for the last several years, but this is still relatively new and has a lot of room for improvement. Vibration based CM of EM has been found very effective as the vibration frequency analysis can uncover several electrical, mechanical defaults and as well as running conditions of the machines. But for such analysis exact parameters of the machine and its drive is required and furthermore in real life impending fault signatures are not as ideal as theoretical fault signature. ML algorithms can learn from monitoring sensor data without prior knowledge of the EM and traditional ML based CM process involves extraction of useful information from raw data and use the extracted features as input of the ML and finally classify different faults. This feature extraction rules is often depend on the domain, so the same algorithm may not work for other domain or motor drive. Deep Learning (DL) algorithms which can directly learn the features from data have recently become very popular approach in many fields because of advancement of computation power, cloud computing, simpler tools or frameworks and also for easily accessible large database.

The presented paper is a continuation of previous work where novel convolutional neural network (CMCNN) architecture was shown to detect bearing faults using a public dataset [1]. In this work we used a newly generated vibration dataset for bearing faults to model the CMCNN architecture for multi-sensory input. Separately generated test data is used to evaluate the accuracy of the model and the learned model's deep layers are analyzed to understand the feature extraction process.

## 2 Related work:

The challenge of beginning researching ML and DL algorithms for CM or fault diagnosis is the access of dataset because creating a realistic mechanical fault dataset generating test-bench is complex and costly. For vibration based rolling bearing fault diagnosis the dataset produced by Case Western Reserve University (CWRU) is the most popular and easily accessible dataset that has been considered as standard reference in many publications [3]. Neupane and Seok reviewed a large number of publications regarding DL algorithms using CWRU dataset in their paper [3]. Smith and Randall have analyzed the entire dataset of CWRU to recommend benchmark for diagnostic technique [4]. CWRU dataset has mainly six classes of data: healthy, inner ring fault, rolling element fault and outer ring fault at three load zone [2]. The faults were implemented in sizes of 0.007 to 0.028 inch with Electric Discharge Machining and the monitoring bearings were either at Drive-side (DE) or Fan-side (FA) of the motor. All the vibrations are measured with three sensors located at DE, FE and at base plate and measurements were taken for four motor speeds.

Various DL algorithms like Deep Belief Networks (DBN), Autoencoder (AE), Generative Adversarial Networks (GAN), Recurrent Neural Networks (RNN), Convolutional Neural Networks (CNN) etc. are investigated to detect bearing faults using the CWRU dataset in the literatures. Stacks of AE based deep neural network (DNN) is applied to classify CWRU dataset among ten classes considering different fault sizes as different classes by Jia and et al, where they used the frequency spectrum of the raw data as the input [2]. Shao and et al. showed DBN based bearing fault classification using both simulated vibration data for inner and outer ring fault and the CWRU dataset dividing all the dataset into ten classes [6]. Jiang and et al proposed a deep recurrent network (DRNN) to automatically extract feature from input spectrum and diagnose rolling bearing fault in their work [7]. They consider frequency domain signal as input believing noisy vibration data may not be robust. The proposed DRNN has stack of recurrent hidden layers of long short-term memory (LSTM) units and classify the CWRU dataset into 12 conditions. GAN based fault diagnosis on CWRU dataset is studied by Jiang and et al [8]. Their idea of implementing GAN algorithm to differentiate faulty vibration from healthy vibration as anomaly detection, relating with real industrial scenario where faults appear in the bearings after millions of cycle hence data collection for faulty bearing is difficult. For Robust feature extraction and fault classification Shaheryar and et al proposed hybrid model (MCNN-SDAE) of two layers multi-channel CNN combined with three stacks of Denoising Autoencoder (DAE) using the CWRU dataset [9].

Gua and et al showed a hierarchical adaptive deep convolutional network (ADCNN) using CWRU data where the 1D vibration is converted to 2D matrix and they tested their model for both fault classification also fault size predictions [10].Wide first-layer kernels with deep CNN (WDCNN) model is proposed by Zhang and et al also using the CWRU data [11]. They used data argumentation technique which is basically dividing the long signal into segments to create bigger dataset in which the input width is 2048. Some sets of the training data were overlapped segments and some were not. Their five layer CNN was designed as the first layer has wide kernel size and following layers have very small kernel width and finally the model classified 10 labels. Other works of CNN based bearing fault diagnosis are presented in the literatures [12-15].

The investigated works in the literatures mostly used same dataset to test their model accuracy to test their domain adaptively for example the fan-end and drive-end vibration information should be clearly different and most cases it is not clear if they considered fault classes for both locations learn the domain robustly or not. Another most interesting note is many of the studies considered the faults sizes as separate classes, where the CWRU dataset fault sizes are clearly different (0.007 inch, 0.014 inch, 0.021 inch) which should be easily diagnosable. Among many DL based CM approaches, CNN has shown the most suitability of using raw data directly.

In our previous work we used the CWRU dataset to train CMCNN model and classified the classes considering both location of the bearing and fault sizes in same class. The aim of the current work is to introduce a new dataset to model the CNN model where same design approach is considered as CMCNN presented in previous work [1].

## 3 Dataset: IEEM - CMData

The dataset contains external vibrations of a motor having different types of faulty bearings at various speed and load combinations. External vibration means it should contain more noise or additional vibration from the rotating parts which is ideal for industrial applications. The bearing data generating test-bench is developed at the Institute of Energy Efficient Mobility (IEEM) of University of Applied Science and Technology Karlsruhe and supported by SEW-Eurodrive GmbH (SEW). In the Fig. 1 a view of the test-bench (left) and the CAD design (left) is shown.

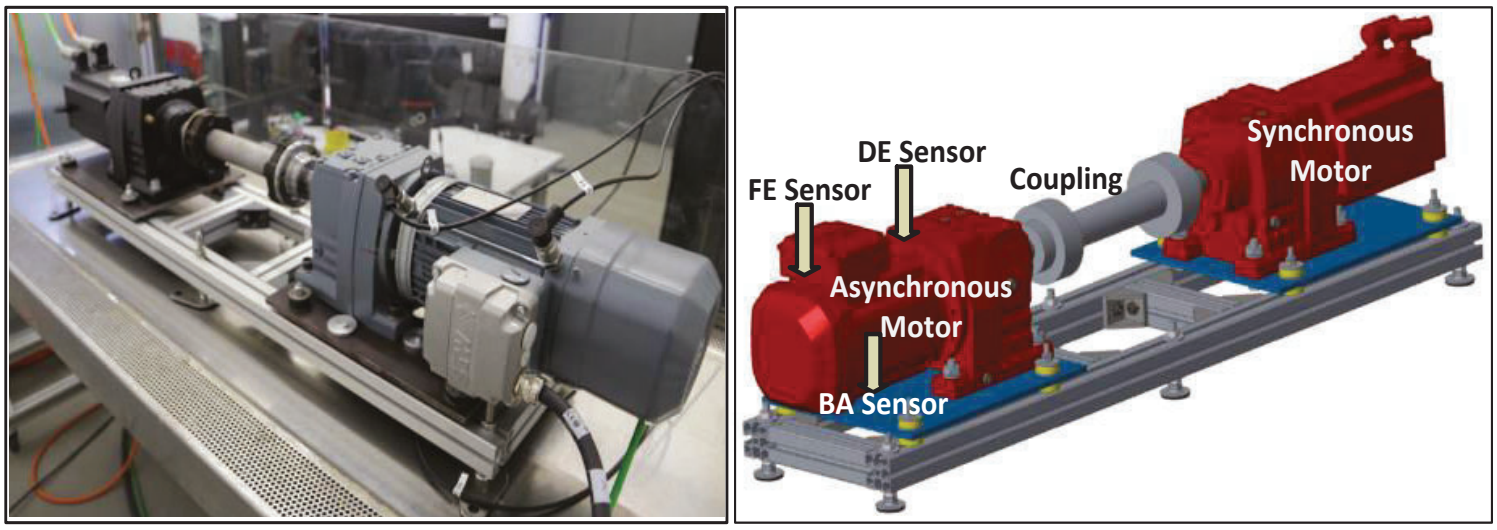

**Fig. 1.** IEEM-CMData test bench

The test-bench has test motor of power 0.75 kW and speed of 1440 RPM which is an asynchronous gear motor (R47 DRN80M4 by SEW) with 8.01 gear ratio connected thorough a highly flexible coupling with the load motor of output torque 144Nm, speed of 3000 RPM which is a synchronous gear motor (R47 CMP80M by SEW) with gear ratio 3.83. Artificial faults were implemented on different parts of the deep groove ball bearing (6304-2RSH by SKF). Three acceleration sensors (iCS80 by IDS Innomic GmbH) were installed near FE, DE and base plate (BA) to measure the vibrations. For training all types of data are generated for both bearings at Fan-End (FE) and Drive-end (DE) at two different sample rates. In this work the sample rate of all input data of the model is 12.8kS/s. National Instrument's cDAQ-9174 is used for data acquisition and data processing is done with MATLAB 2018a with additional package NI-DAQmx.

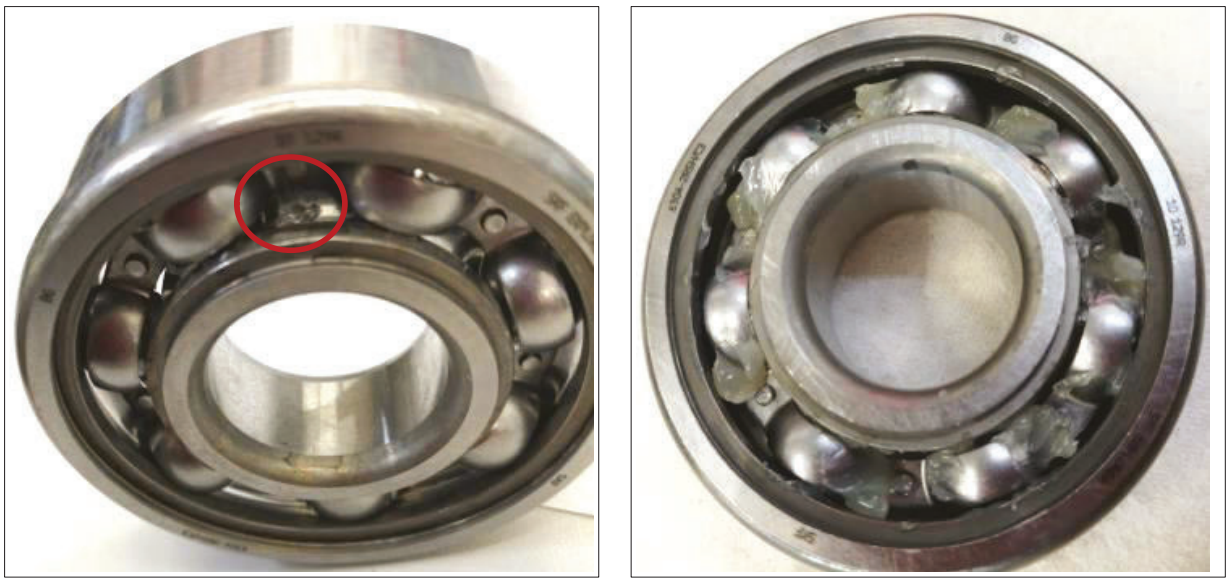

**Fig. 2.** Example of engraved spall in the inner-ring (left) and example of reduced amount of lubrication for measurement (right)

Artificial faults are implemented on different parts of the bearing to achieve different types of faults and one third of the recommended lubrication is used during measurements. Fig. 2 shows an example of a prepared bearing having a small inner-ring spall created by electric engraver (left) and the amount of lubrication used for the measurement (right). Table 1 contains the description of different types data created for the dataset with short names and labels of classes for the model. Among all prepared bearings, ten bearings (Training Bearings) are used to create training data for the CMCNN model and three bearings (Test Bearings) are kept for testing the model. Four speeds (Speed-1 to 4) and five loads (Load-0 to 4) were pre-selected for the measurements, which are called Known Speed-Load data used for training the model and some data are collected at randomly selected Speed and Load combinations which are called Unknown Speed-Load data used for testing the model accuracy.

**Table 1**. Fault description and short naming of the data types with labels for the model

| **Fault Description** | **Short Name** | | | **Class Labels** | | | |
|---|---|---|---|---|---|---|---|
| | **Fault** | **DE** | **FE** | **4 class** | | **8 class** | |
| Healthy(NO) | NoFault | DEOK | FEOK | 0 | 0 | 20 | 10 |
| Inner ring ((IR) spall of 2mm(S1) | IRSpall | DEIRS1 | FEIRS1 | 1 | 1 | 21 | 11 |
| Inner ring (IR) spall of 3.5mm(S2) | IRSpall | DEIRS2 | FEIRS2 | 1 | 1 | 21 | 11 |
| Outer ring (OR) spall of 2mm (S1) | ORSpall | DEORS1 | FEORS1 | 2 | 2 | 22 | 12 |
| Outer ring (OR) spall of 3.5mm (S2) | IRSpall | DEORS2 | FEORS2 | 2 | 2 | 22 | 12 |
| Rough rolling surface (RR) | RRSurface | DERR | FERR | 3 | 3 | 23 | 13 |

## 4 IEEM-CMCNN Architecture for Bearing Fault Classification

The model is named as IEEM-CMCNN; has input of three channels 1D data, six convolution layers, three Fully-connected layers and four or eight output classes. The detail architecture of the IEEM-CMCNN is described in the Fig. 3.

The input is a three-channel 1D vibration data considering three sensors at three positions. The first channel contains the main-sensor data, second channel belongs to the opposite-sensor data and third channel for the base-sensor data. Main-sensor for the FE bearing is the sensor at FE and sensor at DE is the opposite-sensor; for DE bearing this is reversed accordingly. During training the input of IEEM-CMCNN is a fixed-size: 1 x 1000 x 3 vibration data. The one dimensional vibration input length is considered as approximately one revolution of the motor shaft as described in previous paper [1]. No pre-processing is done on the training dataset.

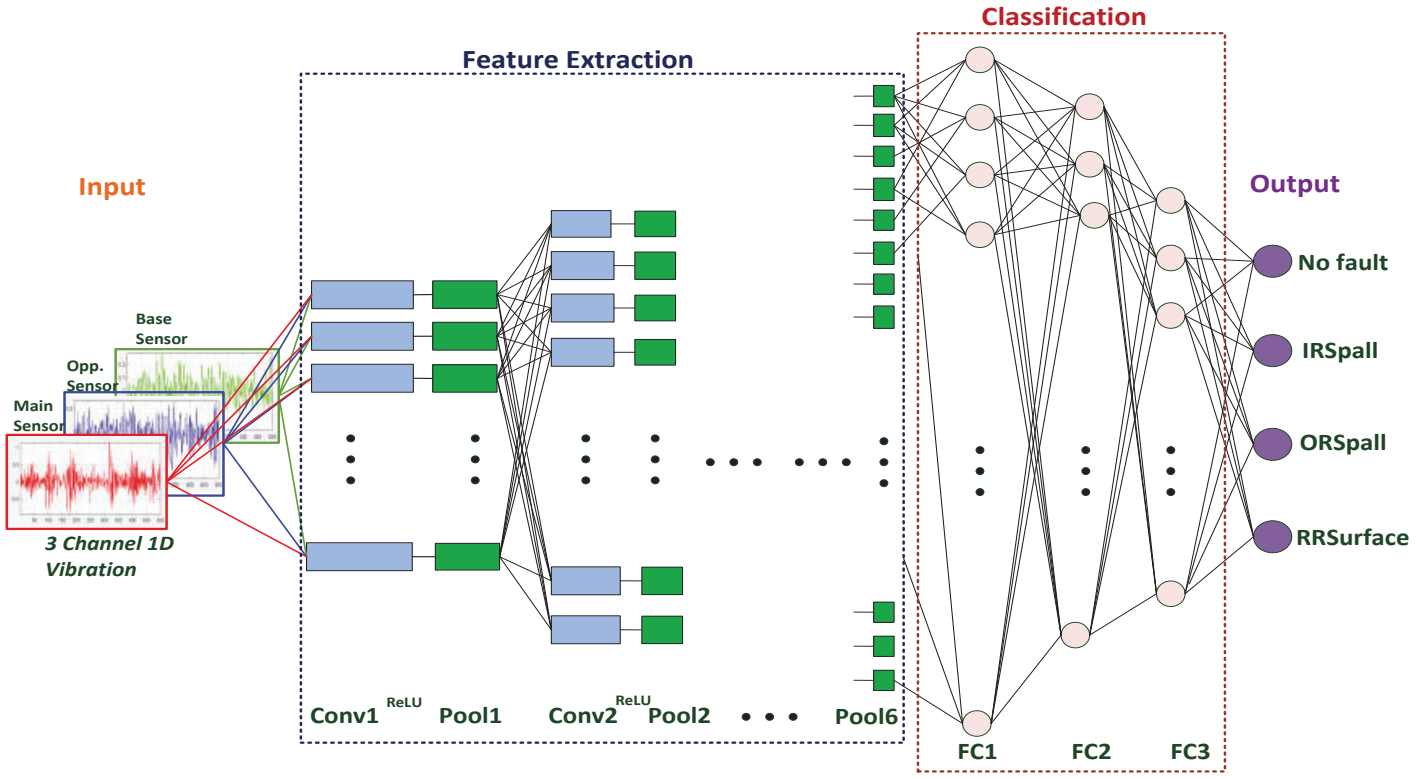

**Fig. 3.** IEEM-CMCNN architecture for four fault classes

The three sensor vibration input is than passed through stack of six convolution (Conv) layers. At the first layer the filters have very large receptive field i.e. $1 \times 50$ and gradually reduced sized filters towards higher level thus at final layer filter size become $1 \times 2$. The convolution stride is fixed to $1 \times 1$ and padding varies from lower layers i.e. $24 \times 24$ to higher layers i.e. $2 \times 2$ (where the stride and padding size is to capture left/right centre). The last Conv Layer padding is $0 \times 0$. The convolution stride and padding in layers are calculated in way to preserve the most of length of the input of each layer. After first Conv layer one batch normalization layer is kept. Each convolution layers are followed Rectified-Linear unit Layer (ReLu) to remove the negative value, those followed by Max-Pooling layers (Pool) of window size 1x2 with stride 2 and zero-padding.

The stack of Conv layers is then followed by three fully connected (FC) layers: first FC layer has 1024 channels with a ReLu layer, second FC layer has 1000 channels also with one ReLu layer and third has same number of channel as number of class. The final layer is soft-max layer. We compared different architecture of different number of filters after analysing the filter activities at each Conv layer: in this work the developed architecture has similar number of filters as VGG16 [16].

The training was stopped when accuracy is not improving after 3 epochs.

## 5 Model Accuracy Analysis

As discussed in Section-2, most of the literatures considered to classify all data types where FE and DE data should be easily detectable. In this work we compare two models: 1) training the model for four classes (Model: 4-Class) where location of bearing (DE and FE) is not known to the model and 2) training the model for eight classes (Model: 8-Class) where two bearing locations belonged to different classes.

The accuracy of the models are also evaluated by testing Unknown speed-load data from training bearings and test bearings as well as Known Speed-Load data from test bearing. This way, the test data can be divided into three groups: 1) Unknown Speed-Load data from Training Bearings (UnSpLd_TrBr), 2) Known Speed-Load data from Test Bearings (KnSpLd_TsBr) and 3) Unknown Speed-Load data from Test Bearings (UnSpLd_TsBr).

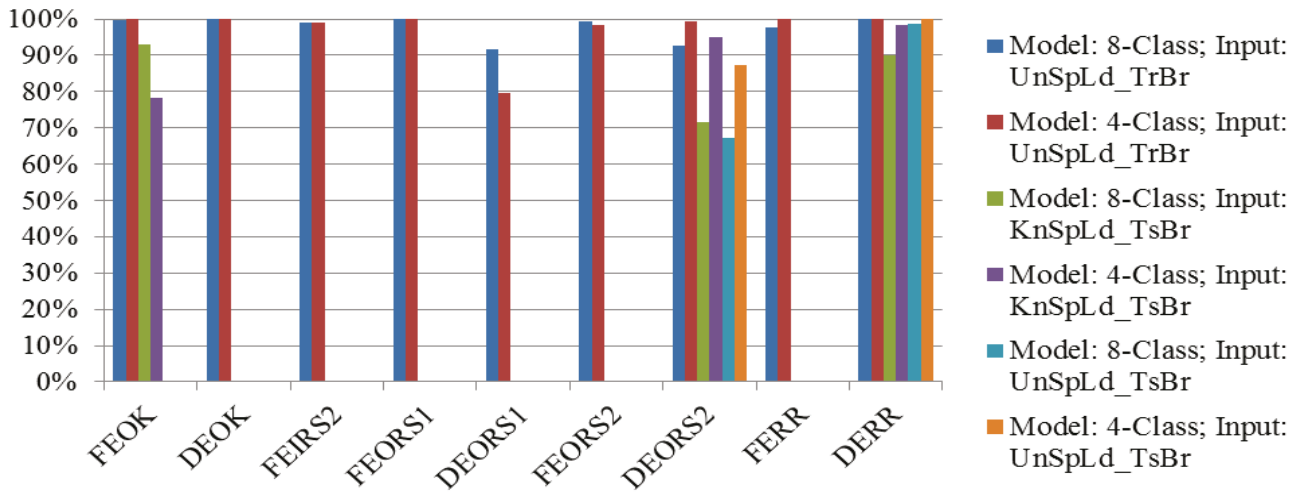

**Fig. 4.** Test accuracy comparison for 4 Class Model Vs 8 Class Model

Model: 4-Class and Model: 8-Class both has average training accuracy above 99% and to evaluate the performance accuracies are checked per class labels. In Fig. 4 the performance of two models are compared for fault sizes and location of the bearings. The labeling of the classes is given in Table 1.

## 6 Feature Map Analysis

DL based CM of electric machine has been successfully applied in many researches but in general it is still not clear why the fault detections were made with high accuracy and how the network is learning the features from vibration. In a previous work [1], we analyse the first Conv layer output by converting them to frequency domain and showed that a significant range of frequencies were learned by each filters for each classes. In the paper [11] the authors also focused feature visualization with FFT and showed feature distribution for each layer and each 10 classes using Stochastic Neighbour Embedding (t-SNE). In this work, we focused on understanding the how in all convolution layers features are learned and thus the classes are separated.

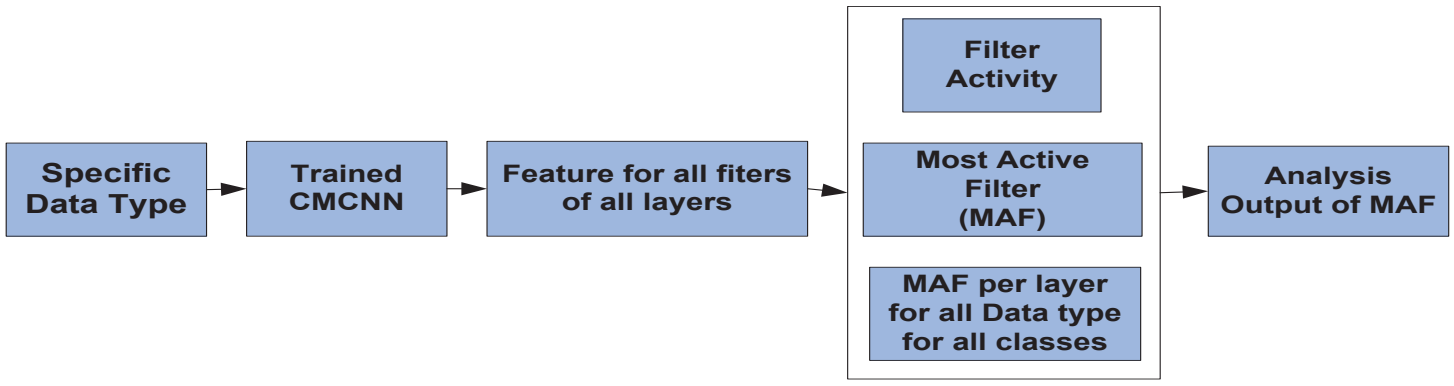

**Fig. 5.** Tasks involed for analysis the feature map of trained IEEM-CMCNN

One feature visualizing technique in computer vision is to feed the network with large amount of dataset and keep track of which images highly activate some neurons, which is shown by Girscick et al [17]. We followed similar approach, aiming to understand among number of filters in the convolution layers if certain filters are contributing to learn the features comparing other filters of the same layers.

The tasks involved to analyse the feature map of IEEM-CMCNN can be described by Fig 5. As the trained dataset has different motor operating conditions and two bearing locations in first step we define the Data type, so one specific class (i.e. IRSpall) has 4 Speeds times 5 Loads in total 20 types of data type for both bearing location. Each type of data is feed through the trained model and all the feature maps of all the filters of all layers are saved for later analysis. Filter activity is measured by calculating the area under the curve of the Pool output. In Fig. 6 the Conv output or the feature map of all six layers (left) for one input for DE bearing at Speed-4, Load-4 and the filter activity pointing the most active filter in red star at all layers for the same input (right) is shown. It is seen that at each layer one filter is highest active than others; for example at layer-1, filter-3 is most active among the 16 filters and at layer-6 filter-338 is the most active among 512 filters.

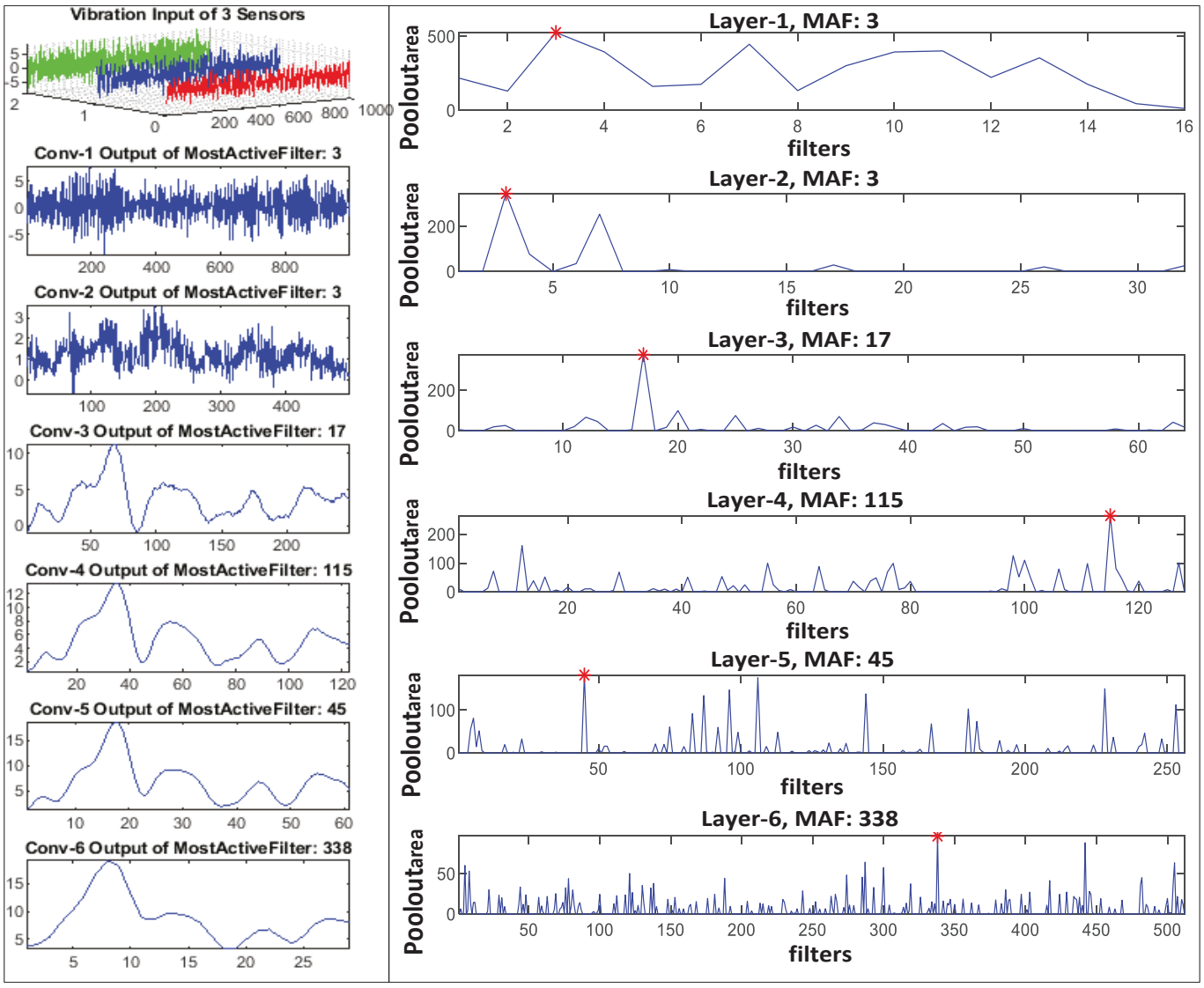

**Fig. 6.** Convolution output or Feature Map of all layers (left) and filter activity for the same input at all layers (right)

In the next step for all data type Filter Activity is checked thus the most active filter (MAF) of all data type is known. MAF means for all 128 inputs of a specific data type most of the time (i.e. 90% times) one particular filter is always highest active. In Fig.7 MAF for all 20 data of for both bearings are shown for four layers. In the similar figure MAF for one test data (TsKnLd) is also plotted and it is shown that almost all time MAF for test data and training data are same. In this way it can be concluded that for one trained model some certain filters are contributing to learn the class features and now these features of MAFs can be examined to know if the features are more differentiable for classes and thus fault classification is highly accurate. In Fig. 8 extracted feature or Conv output of MAFs for four classes is plotted over each other over for 1$^{st}$, 2$^{nd}$ 4$^{th}$ and 6$^{th}$ layers and it is observed that from to higher layers the classes are becoming more distinguishable and thus easily diagnosable as different class.

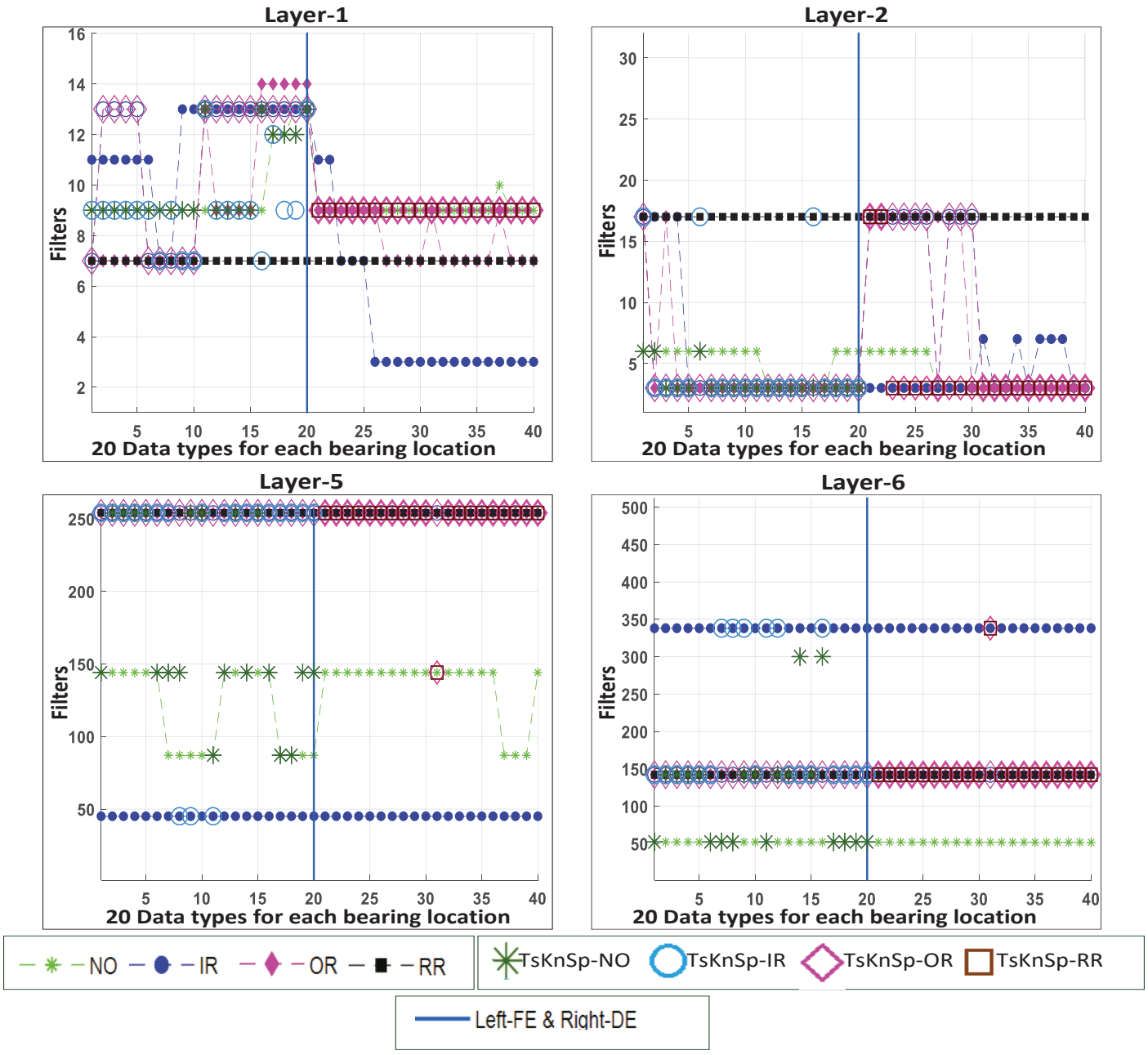

**Fig. 7.** MAF for all data type for four layers ($1^{st}$, $2^{nd}$, $5^{th}$ and $6^{th}$) of trained IEEM-CMCNN for 4 classes

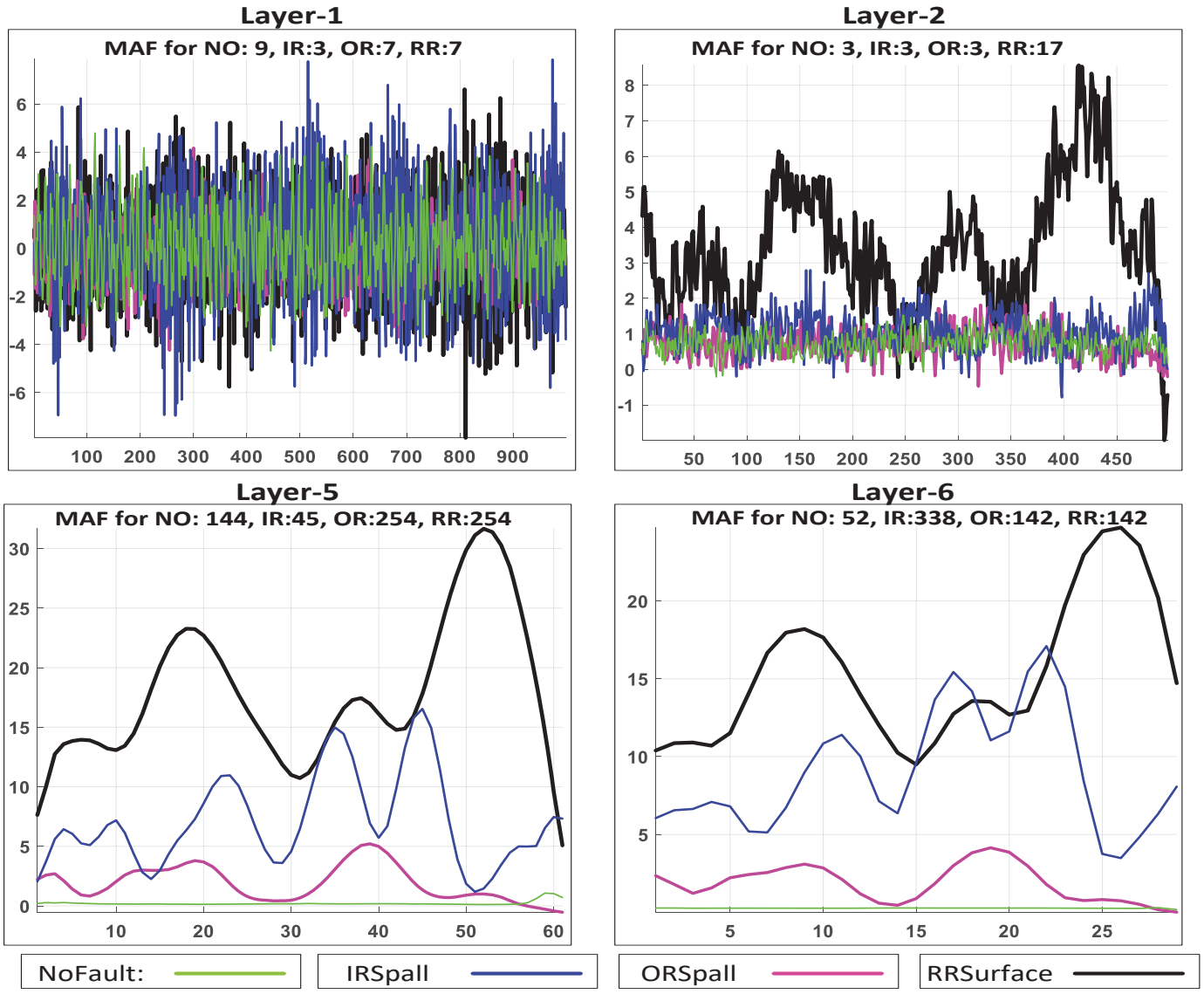

**Fig. 8.** Extracted features or Conv output by the MAF relevant to the classes at four layers ($1^{st}$, $2^{nd}$, $5^{th}$ and $6^{th}$) of trained IEEM-CMCNN for 4 classes

The plotted Conv outputs of all classes are of same Speed-Load, that means all the inputs has same noise or vibration infused in their data and the different fault classes should have different significant pattern after de-noising. In Fig. 8 we can see that the infused noise in this case the healthy or NOFault (in bright green colour) is becoming less and less visible in higher layers and the patterns of the fault classes are becoming more visible in higher layers.

## 7 Conclusion

This work shows that the design criteria for the CNN architecture in previous work can be adapted for different test-bench data. The accuracy comparison in section-5 (Fig. 4) reveals that both models can detect fault classes with high accuracy for both Training Bearings and Test Bearings. The feature analysis explains how in deep layers the model learns the features for different classes.

This work shows the approach of designing the input size for vibration based bearing fault detection applied for CWRU dataset also adaptable for IEEM-CMData. The multi-channel input design can be considered in other applications where multiple sensors are involved. Feature analysis shows that how features are learned from noisy data similar like computer vision where it is known that lower layers detect the low-level features like edges, dark spots and high-level features like shape, object are learnt in the higher layers. It shows in vibration based detection the lower layers are de-noising the data and in higher layers the pattern of the vibration curve which is becoming trainable for the classifier or final layer with high accuracy. This analysis approach can be implemented in different vibration based problems and number and thus sizes or numbers of filters in each layer can be analyzed to optimize the model architecture.

# Classification and Prediction of Bicycle-Road-Quality using IMU Data

Johannes Heidt and Klaus Dorer

Institute of Machine Learning and Analytics, Offenburg University
johannes.heidt@hs-offenburg.de
klaus.dorer@hs-offenburg.de

**Abstract.** The present work ties in with the problem of bicycle road assessment that is currently done using expensive special measuring vehicles. Our alternative approach for road condition assessment is to mount a sensor device on a bicycle which sends accelerometer and gyroscope data via WiFi to a classification server. There, a prediction model determines road type and condition based on the sensor data. For the classification task, we compare different machine learning methods with each other, whereby validation accuracies of 99% can be achieved with deep residual networks such as InceptionTime. The main contribution of this work with respect to comparable work is that we achieve excellent accuracies on a realistic dataset classifying road conditions into nine distinct classes that are highly relevant for practice.

## 1 Introduction

In November 2019 the Bundestag passed the climate package [1]. There it was decided that at least 900 million Euros should be invested in the expansion and renewal of the German cycling infrastructure. This raises the question of how to obtain an overview of the current state of cycle tracks.

While the Federal Ministry of Transport and Digital Infrastructure (BMVI) is responsible for recording and assessing the condition of the major highways, responsibility for inner-city roads lies with the communes. The BMVI carries out these assessments at fixed intervals of four years, using special measuring vehicles that use laser technology and digital cameras[1]. At the communal level, it depends on individual requirements [2]: Expensive measuring vehicles are used as well, and often random surveys of the residents are conducted on the condition of the roads. With passing the climate package in November 2019, this topic is likely to take on a greater role for local authorities in Germany in the near future [1]. The government decided to invest at least 900 million euros in the expansion and renewal of the cycling infrastructure.

The aim of this work is to develop a new approach for the condition monitoring of cycling tracks. For this purpose, we attach an acceleration, gyroscope and GPS sensor to a bicycle. With the help of machine learning methods, the recorded acceleration values and angular velocities are used to predict the road type and condition. Here, the classifier differentiates between the road types asphalt, cobblestone and gravel as well as the conditions smooth, rough and bumpy. Finally, the system sends the recorded sensor data to a server, which is responsible for the classification and visualization of the results. In this way, our setup determines the condition of the roads concurrently, so that it requires no further monitoring tasks by the cyclist.

[1] https://www.bmvi.de/SharedDocs/DE/Artikel/StB/zustand-netzqualitaet-der-fahrbahnen.html

For the training and testing of the machine learning algorithms, we collected the required sensor data manually. The setup consists of a smartphone and a Bosch XDK which is equipped with an accelerometer and a gyroscope. Furthermore, we developed an app which communicates with the XDK via Bluetooth – this enables us to annotate the data. Alternatively, the XDK can persist the sensor data on its memory card. In this case it can be labeled manually afterwards. In this way, we created two data sets with different sampling rates and trained, evaluated and compared different classifiers. The models use common algorithms such as tree-based procedures or state-of-the-art residual networks. To test the transferability of the models, we collected a data set with another bicycle and used it for testing the previously trained models. Finally, we created a setup which allows the wireless transmission of the sensor data to a classification server.

## 2 Related Work

[3] dealt with both the classification of road quality and the detection of bumps. In order to accomplish this, the authors attached a smartphone to the steering wheel of a bicycle and covered a distance of almost 14 km for a total of 16 times [3][p. 40 f.]. During the rides, they used the smartphone to record GPS coordinates at a rate of 1 Hz and accelerometer data at 37 Hz. From this data, the authors extracted five attributes - inclination, speed, as well as mean, variance and standard deviation of acceleration. Finally, they combined the features into segments based on the GPS coordinates and annotated them manually.

In this case, Hoffmann et al. distinguished three classes: smooth, rough and bumpy. For prediction they chose the classifiers k-Nearest Neighbors (kNN) and Naive Bayes. Furthermore, Hoffmann et al. tested different segment lengths (1 m, 2 m, 5 m, 10 m, 15 m, 20 m) and feature combinations. In the end, they achieved the best results by choosing a segment length of 20 m and relying only on the three acceleration features. Moreover, the other two features proved to be unhelpful, because the speed could not contribute to the prediction and the slope even confused the classifiers. The class distribution was not very balanced with 6772 smooth, 2044 rough and 646 bumpy segments. For evaluation, the authors performed a tenfold cross-validation, with the kNN classifier achieving a slightly better result than Naive Bayes with a mean accuracy of 77.457% [3][p. 41 f.]. In our work, we achieve significantly higher accuracies, mainly due to the higher sampling frequency.

Litzenberger et al. also deal with the classification of bicycle tracks using accelerometers [4]. They compared different experimental setups [4][p. 1]. They recorded all data on three flat and straight sections, each 100 m long. Furthermore, the authors chose three different section types: cobblestone, gravel and asphalt. These also represent the classes to be predicted. For the data set generation, they drove each route at three different speeds (10/20/30 km/h) and three different tire pressures (3/4/5 bar). Consequently, each route was covered nine times.

Furthermore, the authors used two different devices for data recording [4][p. 2]: First, Litzenberger et al. attached an accelerometer sensor with a sampling rate of 500 Hz to the fork of a bicycle. On the other hand, the rider had placed a smartphone in his back pocket, which recorded accelerometer data at a frequency of 100 Hz. Subsequently, they formed samples that encompassed time windows of different sizes. Values of one and two seconds were tested here. In the next pre-processing step, they calculated 16 summary statistics for each of the four channels of the accelerometer (X, Y, and Z direction and total acceleration). These statistics are: average, maximum, and minimum values, average peak distance and amplitude, median FFT (Fast Fourier Transform) signal frequency, maximum and minimum FFT signal frequency, maximum and minimum FFT signal amplitude, average positive and negative slope, maximum and minimum positive slope, and maximum and minimum negative slope.

For the classification task the authors trained and tested different tree-based methods and SVMs in a fivefold cross-validation. In the end, an SVM with a polynomial kernel was able to achieve the highest accuracy in the detection of three classes - both when using accelerometer data (99.2%) and smartphone data (97.7%). In another test, Litzenberger et al. tried to additionally predict speed and tire pressure, resulting in 27 classes. Here an ensemble of boosting trees scored best. With the data from the accelerometer, the classifier scored 97.9%, whereas the smartphone data lead to an accuracy of 48.4%. In all trials, the time window of two seconds produced slightly better results.

Hoffman et al. only relied on one track that is run a total of 16 times [3]. This is not optimal, because it means that similar data is used in the process of training as in the validation. Furthermore, the validation accuracy can possibly be improved. Litzenberger et al. drove three selected routes several times as well [4]. Although they used different tire pressures and speeds, the selected courses are only 100 m long. Consequently, the question arises to what extent the models are suitable for application in practice. In this work, we achieve a similarly high or even higher validation accuracy compared to Litzenberger et al. At the same time, a larger and more varied route selection was performed while attempting not to run the same tracks more than once.

## 3 Approach

The goal of this work is to find classifiers, that predict road types and their condition with the help of a wireless sensor network. The sensor attached to the bicycle is supposed to transmit the data to a classification server. Ideally, the model should achieve the highest possible prediction accuracy for unknown roads. The pursued approach is described in this chapter.

### 3.1 Hardware

With regard to the classification task, we chose a Bosch XDK 110 which has a BMA 280 accelerometer and a BMG 160 gyroscope. Both sensors offer a sampling rate of up to 2000 Hz. The measuring range of the accelerometer is ±16 g, that of the gyroscope ±2000 ∘ /s. The other sensors of the XDK are not needed in this case. There is also a slot for micro SD cards. Three programmable LEDs and two programmable buttons are available for user interaction. Finally, the XDK offers two options for wireless data transmission: Bluetooth 4.0 Low Energy IEEE 802.15.1 and Wireless LAN IEEE 802.11b/g/n.

### 3.2 Measurement

For data set generation, we mounted a plastic box modified for this application to the basket of a 28-inch trekking bike (see figure 1 bottom left). Before that, we attached the bicycle basket to a carrier, which is located above the rear wheel (top left). The Bosch XDK can be hooked into the mounted device (middle picture). The lockable plastic box has proven to be a reliable rain protection. Both the bicycle basket and the box device are securely fixed with several cable ties so that no additional bouncing occurs during the ride. The two components only "vibrate" together with the carrier.

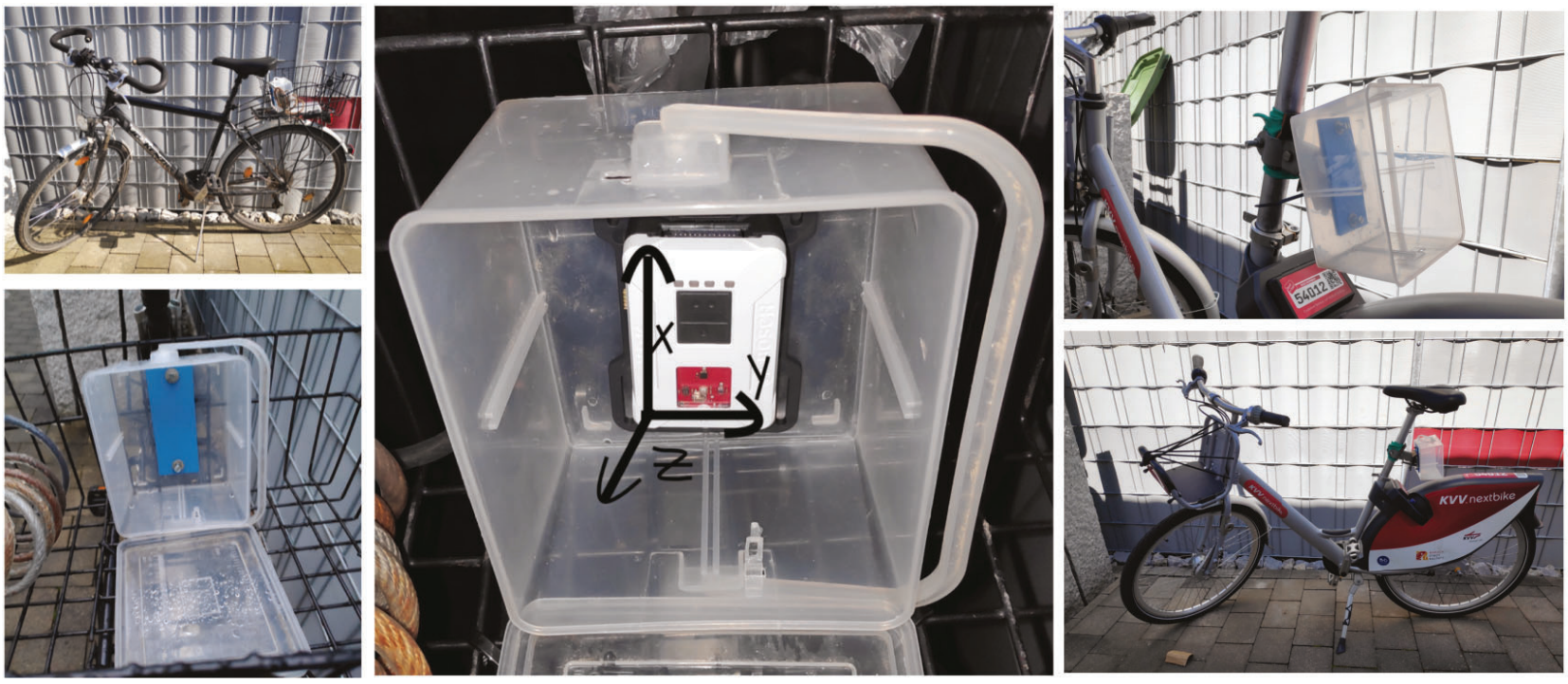

Fig. 1: Setup for data recording.

We created a total of three data sets: Two training sets at different sampling rates and a final test set. The data collection took place with as many different routes as possible. At the same time, we ensured that the data set had a relatively balanced class ratio, with each of the nine classes accounting for about one-ninth of the entire dataset. As a result, routes of rarely occurring classes were sometimes used several times. When creating the data sets, the cyclist tried to vary the driving speed. With regard to the amount of data to be collected.

Moreover, we developed a labelling app to create the first training set. It connects via Bluetooth to the Bosch XDK, which in turn sends the recorded sensor data to the app. Via the app the cyclist can assign the data to one of the nine classes. Afterwards the app persists the sensor values on the smartphone. However, since the Bluetooth protocol limits the length of a message to 20 bytes, it is not possible to achieve a very high sampling rate. In this way, the system recorded 734,441 values per sensor axis (x-, y- and z-values of the accelerometer and the gyroscope) in a period of 376 minutes, which results in a sampling rate of 32.5 Hz. We drove 91.34 km with an average speed of about 14.6 km/h. In general, there were no multiple runs on the same route - except for the rarely occurring classes of smooth and bumpy cobblestone. Furthermore, we achieved a relatively balanced class ratio (the distances covered range from 9.2 to 11.3 km per class).

With regard to the second data set, we aimed to maximize the sampling rate. Therefore, we omitted the Bluetooth app and stored the data directly on the SD card of the XDK. For the labelling of the collected sensor data we programmed a simple user interface using the buttons and LEDs of the XDK. With this approach, the XDK was able to record 5,632,000 values per sensor axis over a period of roughly 314 minutes. Compared to the first data set, the total recording time is about 62 minutes shorter, but we collected significantly more data (factor approx. 7.7). This results in an approximate sampling rate of 294.1 Hz - more than nine times higher than in the first approach. We sampled 76.85 km, with the routes largely identical to those of the first data set. Moreover we omitted some shorter tracks for practical reasons, so that we drove in total about 15 km less. The average speed was about 14,5 km/h and is very close to that of the first data set (14.6 km/h). Lastly, the class ratio is relatively balanced, similar to the first case.

Finally, we created a last data set that is only used for testing purposes and not for training models. By using another bike to collect the sensor data, it should be checked to what extent the trained models are transferable to another bike. Therefore, we chose a bike from the local rental system. As shown in figure 1 (pictures on the right), we placed the plastic box device in a similar way above the rear wheel. For data acquisition again the variant with the memory card was our choice because it offers the higher sampling rate. Apart from the new bike, the procedure is the same as for the creation of the second data set. In about 34 minutes, the XDK recorded 514,500 measurements per sensor axis, resulting in a sampling rate of approx. 252 Hz. Thus the size of the test data set corresponds to roughly 9.1% of the second data set. For each of the nine classes, we recorded a distance of around one kilometer, totalling 9.32 km. Compared to the previous data sets, the average speed of 16.4 km/h is slightly higher. However, the class ratios are not quite as balanced as with the previous data sets: The number of measured values per axis is between 40,000 and 76,000 for the individual classes.

### 3.3 Data Preprocessing

Before the collected data is used, several preprocessing steps are performed. To reduce noise in the data, idle times are removed from the data set. These can occur when stopping the bike to annotate the data with the smartphone app. Since the timestamps of all sensor data are known, including the GPS coordinates, the idle times can be removed in an automated manner.

Before the data is used as model input, it is normalized. The individual measurements of the six sensor channels are transformed so that they have a mean value of 0 and a standard deviation of 1. In addition, the transformation is performed independently for training and test data. Finally, this is the model input for the deep learning-based methods. For all other methods, the data are statistically summarized before normalization, whereby the following metrics are calculated for the individual features of a sample: arithmetic mean, standard deviation, maximum and minimum value, mean width and amplitude of signal peaks, mean peak distance, mean positive and negative slope as well as maximum positive and minimum negative slope. A total of 14 new features are calculated for each of the six features (accelerometer and gyroscope each with x-, y-, and z-values). Consequently, the original sample matrix with dimensions $s_i \times 6$ becomes a sample vector of length $s_j = 84$. Here, $s_i$ is the sample size respectively the number of time steps and thus a configurable hyperparameter.

## 4 Results

This chapter presents the results of the classification using several machine learning models that were trained and evaluated using cross-validation.

### 4.1 Hyperparameter Selection

We used the scikit-learn library to implement the machine learning methods [5]. Only an implementation for XGBoost is missing, so here the XGBoost library is used [6]. Lastly, we implemented the deep-learning based methods using keras [7] and TensorFlow [8]. When implementing our algorithms, we had to choose the parameters carefully. Eventually, we tested different parameter combinations for each classifier in a grid search, whereby each parameter combination of this grid was tested using a tenfold stratified cross-validation. The final parameter selection is presented in the following chapters. It should be noted beforehand that all the other standard parameters, which are not explicitly mentioned in the following parameter grids, remain unchanged and can be looked up in the according documentation [5–7].

**Tree Based Methods** In general, tree based methods provide good results for different types of classification problems [6][p. 785 f.]. Furthermore, their decision-making process is easy to interpret. That is why this evaluation process includes random forest and Extreme Gradient Boosting (XGBoost) [9, 10]. In the case of the first data set, we used the following parameter grid for the grid search: $t \in \{5, 10, 25, 50, 100, 200\}$, $max_depth \in \{10, 25, 50\}$, $min_samples_leaf \in \{1, 2, 5\}$, $min_samples_split \in \{2, 5, 10\}$ and $n_estimators \in \{2, 5, 10\}$. Please note that $t$ is not a parameter of the scikit-learn or XGBoost models which we used. It describes the number of time steps for which we calculated summary statistics such as mean values and standard deviations. With the following parameters the random forest has achieved the best result: $t = 200$, $max_depth = 50$, $min_samples_leaf = 1$, $min_samples_split = 5$ and $n_estimators = 100$, resulting in a mean accuracy of 68.546% with a standard deviation of 3.271%. In contrast, the highest mean validation accuracy of the XGBoost classifier is slightly better with 68.963%. At the same time, the standard deviation of 1.646% is slightly lower than with the Random Forest. The parameter selection $t = 200$, $max_depth = 50$, $min_samples_leaf = 1$, $min_samples_split = 2$ and $n_estimators = 300$ leads to this result.

We adjusted the first grid slightly for the second data set, regarding its higher sampling rate: $t \in \{50, 100, 250, 500, 1000, 2000\}$ and $n_estimators \in \{100, 300, 1000\}$, while the possible values for the other three variables remain the same. The best result of the random forest is achieved with $t = 2000$, $max_depth = 25$, $min_samples_leaf = 2$, $min_samples_split = 5$ and $n_estimators = 300$ - this leads to a validation accuracy of 83.333%. With XGBoost the parameter values $t = 2000$, $max_depth = 25$, $min_samples_leaf = 1$, $min_samples_split = 2$ and $n_estimators = 1000$ result in the highest validation accuracy (86,17%).

**Support Vector Machine** The support vector machine (SVM) also proves to be a reliable model for a wide range of classification tasks [11]. Here we use the following parameter grid in connection with the first data set: $t \in \{5, 10, 25, 50, 100, 200\}$, $kernel \in \{linear, poly, rbf, sigmoid\}$, $C \in \{2^{-5}, 2, 2^{10}\}$, $gamma \in \{scale, auto\}$ and $degree \in \{2, 3\}$. Regarding $gamma$, a value of $auto$ means $1/n_{features}$ and $scale$ means $1/(n_{features} \times \sigma_x^2)$. With the parameter combination $t = 200$, $kernel = linear$ and $C = 2$, we achieved a mean accuracy of 66.539% with a standard deviation of 2.377%, which is the best result.

For training with the second data set, we adjusted only the time steps $t \in \{50, 100, 250, 500, 1000, 2000\}$. With the parameter selection $t = 2000$, $kernel = linear$ and $C = 2$ we achieved the highest validation accuracy of 86.702%.

**Deep Learning Based Methods** As deep learning approaches we tested long short-term memory (LSTM) and convolutional neural networks (CNNs) with different architectures [12, 13]. Hence, we tested a single LSTM layer, which is fully connected to the output layer. When working with CNNs, we also adjusted the *sample_size* and *dropout*. The $sample_size$ specifies the number of instances which we combine into one input. It can be considered as the number of time steps, but in comparison to $t$ we calculated no summary statistics. Additionally, we analyzed different values for the number of $filters$, $kernel_size$ and $pooling_size$. Here, we tested different architectures: a simple CNN consisting of two convolutional layers with dropout and max pooling followed by a fully connected hidden layer which is again fully connected to the output layer. Next, we evaluated a CNN-LSTM which has a similar architecture like the simple CNN, we only replaced the hidden layer with an LSTM layer. Lastly, we tested two residual networks from Fawaz et al.: ResNet [14] and InceptionTime [15]. The main part of the architecture are the residual blocks, which contain three convolutional layers and use batch normalization in between. The blocks themselves are also linked via residual connections. ResNet consists of three and InceptionTime of two blocks. Moreover, the residual blocks for InceptionTime include an inception module, which enables the use of multiple, concatenated filter types. In this paper both model architectures remain unchanged.

For the plain LSTM architecture the parameter grid for the first data set looked like this: $sample_size \in \{20, 50, 100, 150, 200\}$, $recurrent_dropout \in \{0, 1/4, 1/2\}$ and $units \in \{32, 64, 128\}$. With the parameters $sample_size = 50$, $units = 128$ and $recurrent_dropout = 1/4$, we achieved the highest mean validation accuracy - 65.329% with a standard deviation of 1.03%. The tested CNN architecture uses the following grid: $sample_size \in \{20, 50, 100, 150, 200\}$, $filters \in \{32, 64, 128\}$, $dropout \in \{0, 1/4, 1/2\}$, $kernel_size \in \{3, 5, 7\}$, $pool_size \in \{2, 5\}$. With the parameters $sample_size = 100$, $dropout = 1/2$, $filters = 32$, $kernel_size = 5$ and $pool_size = 5$, we achieved a mean validation accuracy of 62.912% with a standard deviation of 1.792%. The CNN-LSTM architecture uses the same grid as the CNN and achieved a slightly better result with a validation accuracy of 70.222% at a standard deviation of 1.699%. The model uses the parameters $sample_size = 200$, $dropout = 0$, $filters = 64$, $kernel_size = 3$ and $pool_size = 5$. In the case of residual nets, we examined only different values for $sample_size \in \{20, 50, 100, 150, 200\}$. With a value of $sample_size = 200$, the residual nets achieved the highest mean validation accuracy: InceptionTime scores 77.645% with a standard deviation of 1.925%, ResNet is just below that with 77.484%. But the standard deviation of 1.382% is slightly lower than with InceptionTime.

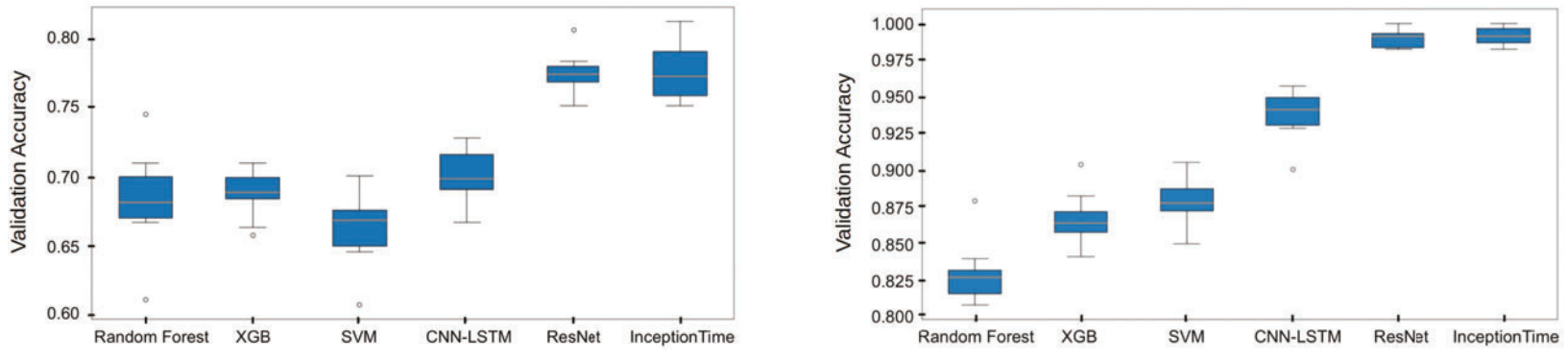

(a) First dataset (b) Second dataset

Fig. 2: Comparison of classification results.

In the case of the second data set, the parameter grid for all the deep learning based models remained almost the same. We only adjusted the $sample_size$ for the higher sampling rate: $sample_size \in \{50, 100, 250, 500, 1000, 2000\}$. The best result of the LSTM is 71.129% which uses the parameters $sample_size = 100$, $units = 32$ and $recurrent_dropout = 0$. When using a CNN, we achieved the highest accuracy of 77.852% with the parameters $sample_size = 250$, $filters = 128$, $dropout = 1/2$, $kernel_size = 7$ and $pool_size = 2$. The CNN-LSTM scored a significantly higher validation accuracy of 91.312%. Already during the tests with the first data set we noticed that the accuracy of LSTMs drops rapidly with $sample_size >= 150$, which seems to be a common issue [16]. To prevent these problems, we adjusted the pooling size ($pool_size \in \{5, 15, 50\}$) compared to the first parameter grid. The parameter selection $sample_size = 2000$, $filters = 32$, $dropout = 0$, $kernel_size = 7$ and $pool_size = 50$ finally leads to the best result for this model. All in all, the residual networks provide the best results. With a sequence length of $sample_size = 2000$, InceptionTime achieves 98.404%, the ResNet scores with 98.227% marginally below.

We trained all models over 200 epochs, only the residual nets over 1000 epochs as it took significantly longer for overfitting to occur. In all cases, we stopped the training process as soon as the training accuracy did not improve over 30 consecutive epochs. Furthermore, we used a local computer with 64 GB of RAM and an nVidia GeForce GTX 1080 graphics card.

### 4.2 Classification

Next, we compare the presented models with each other using the hyperparameter selection from the previous chapter.

**First Data Set** As seen in figure 2 (a), the deep-learning-based methods deliver the best results - especially the two deep residiual networks ResNet and InceptionTime. While InceptionTime has the highest mean validation accuracy of the examined models with 77.65%, ResNet with 77.42% has a slightly higher median than InceptionTime (77.26%). Furthermore, the variance in the validation accuracy of the ResNet is lower than that of InceptionTime because the ResNet has a lower standard deviation (1.382% vs. 1.925%) and a smaller interquartile range (1.129% vs. 3.226%). The top two models are followed by the CNN-LSTM, which has an average validation accuracy of 70.22% with a standard deviation of 1.699%. In addition, the median lies at 69.84% with an interquartile range of 2.484%. After that follow the models XGBoost, Random Forest and SVM with median values of 68.90%, 68.23% and 66.88% respectively. The interquartile ranges are 1.553%, 2.958% and 2.605% respectively.

During the tests in chapter 4.1 we observed that almost all classifiers achieved their best results with longer sequences. Consequently, the question arises why no values $t > 200$ or $sample_size > 200$ were examined. In the end, we set this value as the upper limit for practical reasons: At an average speed of about 14.6 km/h and a sampling rate of 32.5 Hz, 200 time steps result in a time period of about 6.15 seconds and thus an approximate distance of 25 m. In terms of application of the model in practice, a classification of shorter sequences seems preferable.

To gain a better understanding of the strengths and weaknesses of the single classifiers, we examined the precision and recall rates for the individual classes. It became apparent that all models have problems in recognizing the classes rough asphalt, rough cobblestone and bumpy gravel. When also looking at the confusion matrices of these models, it is noticeable that the models often confuse the problem classes with each other. The reason for this could be the sampling rate: At an average speed of about 14.6 km/h, roughly 4.05 m are covered in one second, so that the smartphone app records a sensor measurement approximately every 12.5 cm. However, there are cobblestones that have a shorter length than this. Also in the case of asphalt and gravel, bumps can occur at shorter intervals. Consequently, the tests with the next data set should show what influence the sampling rate can have.

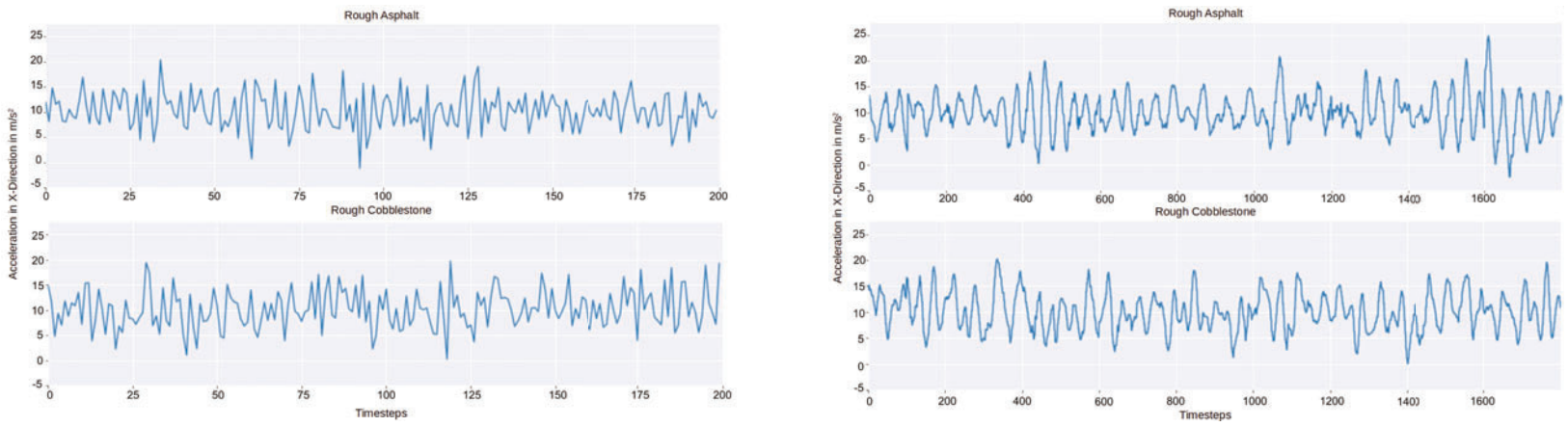

(a) Rough asphalt and rough cobblestone (32.5 Hz)

(b) Rough asphalt and rough cobblestone (294.1 Hz)

Fig. 3: Comparison of selected sequences at different sampling rates.

**Second Data Set** As in the test with the first data set, the two deep residual networks clearly provide the best results (figure 2 (b)). Between the two, InceptionTime achieves slightly better results: The mean validation accuracy is 99.18% and the median is 99.11%. For the ResNet, it is 98.97% and 99.11%. Again, the variance metrics are nearly equal: While the validation accuracy of InceptionTime has a standard deviation of 0.60% and an interquartile range of 0.98%, the corresponding values of the ResNet are 0.60% and 0.98%. The CNN-LSTM also achieves very good results with a median of 94.14% and an interquartile range of 1.87%. The SVM follows with a median of 87.77% at an interquartile range of 1.54%. Finally, the tree-based methods XGBoost and Random Forest achieve a median of 86.40% and 82.74%, with an interquartile range of 1.42% and 1.60%.

Already during the tests with the first data set we observed that most models benefit from longer sequence lengths. This is also the case with the second data set, but again we limit the maximum sequence length for practical reasons. In case of the first data set we selected 200 time steps as maximum. Since the sampling rate is almost ten times higher for the second data set, we set a corresponding value of $t = 2000$. At a sampling rate of 294.1 Hz, an average sequence of length $t = 2000$ records a period of about 6.8 seconds. Thus, at an average speed of 14.45 km/h, the cyclist covers a distance of roughly 27.3 meters.

Figure 3 (a) shows some selected sequences that indicate why a correct classification can be difficult with a lower sampling rate. Rough asphalt and rough cobblestone are two classes that were very frequently confused with each other in the case of the first data set. The measured value is the acceleration in the x-direction of the XDK (perpendicular to gravity) and it is plotted over 200 time steps. It is noticeable that both time series have very similar characteristics: The measurements vary quite regularly between -5 and 15 m/s$^2$ with just a few outliers. In addition, the number of peaks and the average period lengths appear to be similar. Consequently, a simple distinction of the classes based on the graphs is not possible. For comparison with sequences at 294.1 Hz, we slightly adjusted the x-axis in 3 (b): 1800 time steps are mapped here, which corresponds to a duration of about 6.12 seconds. This is about the same time period mapped in the left graphs (6.15 seconds). In addition, the measured values of the right-hand figures fluctuate within approximately the same range as in figure 3 (b). Furthermore, the number of peaks is comparable to the left-hand time series. Nevertheless, the higher frequency sequences seem to be more easily distinguishable. In the case of the first dataset, our setup recorded a sensor measurement on average about every 12.5 cm. With the second dataset, the average speed is 14.45 km/h, so that the cyclist covers roughly 4.014 m in one second.

Consequently, our second setup performs a sensor data recording approximately every 1.36 cm. This allows a much more precise ”scanning” of the road surface.

**Third Data Set** As already described in chapter 3.2, we chose the same procedure for data acquisition as for the second data set to record a third data set. The setup only included a different bicycle. The purpose is to check whether the previously trained models are transferable to a new bicycle.

When looking closer at the recorded test data, in particular the standard deviations in relation to the measured values of the individual sensor axes, it is noticeable that these in some cases differ significantly from the training data. Compared to the second data set, almost all standard deviations are about 20-40% lower. Even more significant differences can be found with the gyroscope values, which sometimes deviate by a factor of 3 to 4. The reason for this is most likely the choice of a different bicycle and the slightly different mounting of the XDK support (see figure 1).

Using the third data set, we tested the models from chapter 4.2, which we trained using the second data set. The differences between the training and test data are reflected in the classification results of the models: ResNet achieves the highest accuracy with 26.459%, InceptionTime reaches 26.07%, and all other models fall below the 20% mark. This results conclude that the data from chapter 4.2 is not sufficient to easily transfer the models to any bicycle. Besides the choice of the bicycle, the mounting of the XDK holder and the tire pressure might also influence the recorded sensor data. To train a robust model that delivers solid results regardless of all these factors, the training data must represent these elements.

### 4.3 Comparison to Related Work

Chapter 2 presented the work of Hoffmann et al. [3], which distinguishes three classes of bicycle tracks: smooth, rough and bumpy. They covered one track with a distance of 14 km 16 times in total and recorded accelerometer data at 37 Hz. The best classification result is a mean validation accuracy of 77.457%, determined by tenfold cross-validation. With respect to the present work this is most comparable to the results of the first data set, since we used a similar sampling rate with 32.5 Hz. However, in addition to the accelerometer data, we also used gyroscope data, which were not available in the case of Hoffmann et al. On the other hand, this work distinguishes nine classes, whereas Hoffmann et al. distinguish three. The data set of this work is smaller by a factor of about 2.5 and we attempted to avoid multiple runs on the same track. Finally, the highest mean validation accuracy with respect to the first data set is 77.645% (InceptionTime) which is slighlty higher than the best result of Hoffmann et al. (77.457%). However, when using our second data set (294 Hz), the validation accuracy of InceptionTime is significantly higher: 99.183%. Itt should also be noted that our training and test data include a greater variety of tracks compared to Hoffmann et al., while avoiding multiple runs of the same routes as far as possible. This makes our models particularly suitable for real-world applications.

The second work discussed is a paper by Litzenberger et al. [4]. They selected three routes of 100 m length each and have run each of the three routes nine times. They combined three different tire pressures (3, 4 and 5 bar) with three different speeds (10, 20 and 30 km/h). They also recorded accelerometer data, firstly with a smartphone (100 Hz) and secondly with a sensor device (500 Hz). Here the authors distinguish the classes of asphalt, cobblestone and gravel, with the highest mean validation accuracy at 99.2% (500 Hz) and 97.9% (100 Hz) respectively, determined with a five-fold cross-validation.

This is comparable to the results of the second data set of this study, since the sampling rate of 294 Hz is almost in the middle of the sampling rates used by Litzenberger et al. There are also differences between the two works: The data set used in this paper is significantly larger and the tracks to be predicted are much more diverse. In contrast to Litzenberger et al., we additionally use gyroscope values. When predicting nine classes, InceptionTime achieves the highest average validation accuracy of 99.18% within the tenfold cross-validation, which is minimally lower than the best result of Litzenbeger et al. (99.2%), but on nine classes making the results more practically relevant.

In order to make our results more comparable to the work of Litzenberger et al., we also tested the classification for three classes - asphalt, cobblestone and gravel. For this purpose, we combined the road types of the second data set so that there is no distinction between the road conditions smooth, rough and bumpy. Consequently, the three combined classes of asphalt, cobblestone and gravel now comprise 1,735,500, 1,887,500 and 2,009,000 measurements per sensor channel. Using this data, we trained the two residual networks InceptionTime and ResNet using the previously determined parameter settings from chapter 4.1. We performed a tenfold stratified cross-validation, whereby the models trained in each run for 500 epochs and the highest validation accuracy is used for evaluation. In this way, the residual nets slightly exceeded the results previously achieved with nine classes: The mean validation accuracy of the ResNet is 99.54% with a standard deviation of 0.163%. InceptionTime performs slightly better with 99.72%, only the standard deviation is a bit higher with 0.213%. To sum up, our models (99.72%) have a slighty higher prediction accuracy compared to Litzenberger et al. (99.2%) At the same time, our route selection for training and testing is significantly larger and more varied, so that our models should be way more suited for use in real-world applications.

## 5 Conclusion and Future Work

This paper describes the development of an approach for condition assessment of cycling tracks. In total, we collected three different data sets: Two training sets with different sampling rates (32.5 Hz and 294 Hz) as well as a test set using a different bicycle. For the first data set, the two residual networks achieved the highest validation accuracy of about 77%. However, all models show difficulties in recognizing certain classes, and mix-ups often occur. A closer look at these classes led to the assumption that a higher sampling rate could solve these problems. With the second data set, all models were able to achieve a significantly higher validation accuracy with the residual nets achieving 99% accuracy.

The evaluation of the transferability of the models to a different bicycle did not lead to good results. The recorded data of the third set differs too much from the training values of the second data set, so that none of the models can make reliable predictions. Once again, the two residual networks record the highest test accuracy, but they reach only about 26%. Despite the poor performance of the third data set, the initial results suggest that this is a very promising approach. The models developed provide higher validation accuracy than existing approaches. At the same time, we use a much larger and more varied selection of cycling tracks, which results in robust models that should be more suitable for use in real-world applications.

In future work, a variety of routes, bicycles, riders, speeds and tire pressures should be taken into account when creating the training data. Furthermore, the final setup for the transmission of sensor data could be adapted so that it can be used for the annotation of training data. For this the already developed bluetooth labelling app would have to be adjusted only slightly. In this way, the user experience could be made much more convenient compared to the previous recording with the XDK's memory card. Lastly, the final test setup could be made more user-friendly by removing the smartphone dependencies from the current architecture. At the moment, the GPS sensor of a smartphone is still required. To resolve this dependency, the Bosch XDK could be extended with a corresponding sensor via its GPIO ports. Finally, the XDK needs a Wifi-hotspot - for this purpose a mobile router could be mounted on the bicycle. With these changes, the recording and classification of the road condition could be completely automated, so that the cyclist can fully concentrate on the riding itself.

# Modeling natural convection in porous media using convolutional neural networks

Mohammad Reza Hajizadeh Javaran[1], Amadou-oury Bah[2], Mohammad Mahdi Rajabi[1], Gabriel Frey[3], Florence Le Ber[3], Marwan Fahs[2]
[1]Civil and Environmental Engineering Faculty, Tarbiat Modares University, PO Box 14115-397, Tehran, Iran.
[2]Université de Strasbourg, CNRS, ENGEES, ITES UMR 7063, 67000 Strasbourg, France.
[3]Université de Strasbourg, CNRS, ENGEES, ICube UMR 7357, 67000, Strasbourg, France
**mrhj75@gmail.com, amadou-oury.bah@etu.unistra.fr, mmrajabi@modares.ac.ir, g.frey@unistra.fr, florence.leber@engees.unistra.fr, fahs@unistra.fr**

**Abstract**. Deep learning has become increasingly prevalent in a wide range of engineering contexts. In this work, we tried to make a connection between the groundwater engineering community and the field of deep learning. Natural convection in porous media is usually simulated using common numerical modeling tools with high computational costs. In this work, we aim to use supervised learning in input-output pairs (porous media characteristics-heat map distribution) in an image regression task, employing an encoder-decoder convolutional neural network (ED-CNN) to develop a meta-model that is able to predict the distribution of heat map resulting from a natural convection process in porous media or to estimate the characteristics of the porous domain when the heat map distribution is given. In order to achieve this objective, a training data set of  samples is prepared using Comsol Multiphysics numerical modeling and is trained with the proposed encoder-decoder CNN. We also employed several evaluation metrics such as root mean squared error (RMSE), coefficient of determination ($R^2$-score) to assess the robustness of the developed network. We observed promising results in both approaches, as well as accuracy and speed, indicating the network's relevance in a variety of groundwater engineering applications to come in the future.

## 1 Introduction

Natural convection is an important concept in porous media problems [1]. It is encountered in several applications such as in heat storage in aquifers, $CO_2$ sequestration in geological formations, geothermal energy extraction, and geological deposition of nuclear waste. Physics-based numerical models are commonly used for simulating natural convection in porous media. Despite the effectiveness of these models in most cases, they encounter some critical challenges. One key challenge is the computational time cost, which is more noticeable at large time and space scales, especially in repetitive runs. In recent years, several meta-models, such as polynomial chaos expansions and feed-forward neural networks, have been proposed in order to reduce the simulation time of natural convection models. These meta-models have demonstrated acceptable performance in low-dimensional domains, but they do not scale well to high-dimensional problems [2]. To overcome this challenge, we propose the use of a convolutional neural network (CNN) architecture [3]. We apply the proposed ED–CNN in the context of '*image-to-image* regression to (a) estimate the entire heat distribution resulting from a specified permeability or (b) estimate the permeability from a heat map.

# 2 Methodology

We first develop a numerical model based upon a hypothetical square porous media example, generating heat map distribution images as training data. Each image references a unique value of a porous domain characteristic, known as the Rayleigh number. The generated data are then trained and validated using an encoder-decoder CNN, and results are analyzed using various methods.

## 2.1 Example description and governing equations

A hypothetical, two-dimensional saturated square porous media is considered. As demonstrated in **Fig. 1**, Dirichlet temperature boundary conditions are assigned to the side walls. $TL$ and $TR$ have constant values and $TL > TR$. We also consider Neumann boundary conditions for the bottom and the upper boundaries, which emphasizes impermeable and thermally adiabatic conditions. The flow is assumed to be steady-state with a Newtonian and incompressible fluid following Darcy's law. The test case is a homogeneous media, with equal hydraulic and thermal properties considered as Rayleigh number (Ra). The natural convection in porous media is explained by the heat transfer equation showing the energy balance, the continuity equation for mass balance, coupled with a variable fluid density function. The governing equations are [4]:

$$\frac{\partial u}{\partial x} + \frac{\partial v}{\partial y} = 0 \tag{1}$$

$$u = -\frac{\partial_p}{\partial_x} \tag{2}$$

$$v = -\frac{\partial p}{\partial y} + Ra.T \tag{3}$$

$$Ra = \frac{k.\rho_c.\beta.g.\Delta T.H}{\mu.\alpha} \tag{4}$$

$$u\frac{\partial T}{\partial x} + v\frac{\partial T}{\partial y} = \frac{\partial^2 T}{\partial^2 x} + \frac{\partial^2 T}{\partial^2 y} \tag{5}$$

Where $u(\frac{m}{s})$ and $v(\frac{m}{s})$ are velocity in the $x$ and $y$ directions, respectively, $p$ is the pressure, and $T$ is temperature. The dimensionless Rayleigh value is defined while $k$ $(m^2)$ is hydraulic conductivity, $\rho_c$ $(kg/m^3)$ is the fluid density, $\beta(1/K)$ is the fluid thermal expansion, $g(m/s^2)$ is the acceleration due to gravity, $\Delta T$ (K) is the temperature gradient between the left and right walls (i.e., $TL - TR$), $H$ $(m)$ is the size of the domain, $\mu$ $(kg/m.s)$ is the fluid viscosity, and $\alpha$ $(m^2/s)$ is the medium equivalent thermal diffusivity coefficient.

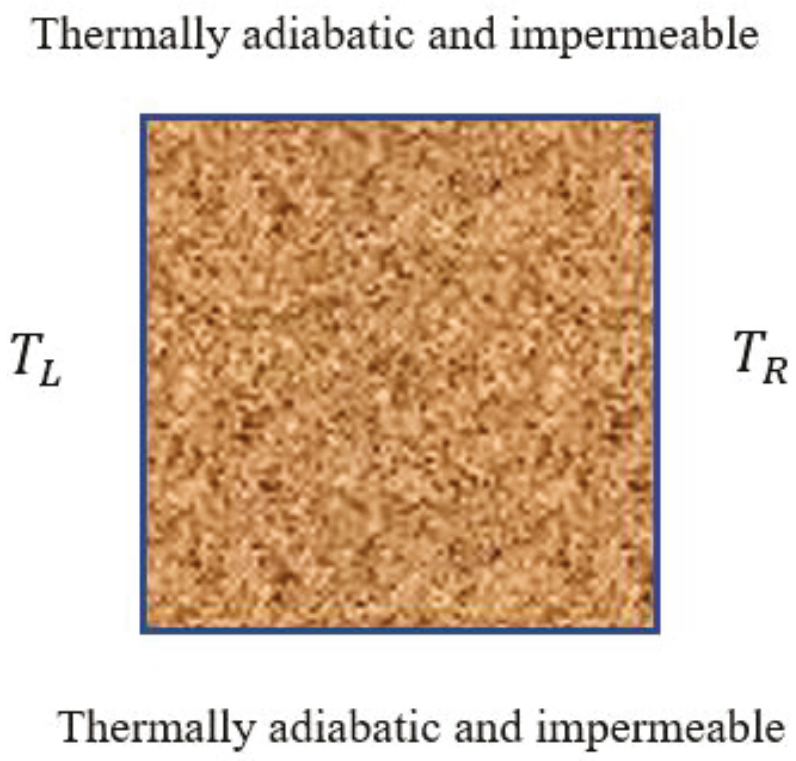

**Fig**.1. Schematic of the problem domain

## 2.2 Training Data preparation

In order to train the CNN models, we generated data using a COMSOL Multiphysics model, solving the above-mentioned equations, which takes about 600 minutes to reach a steady state. Using uniform probability distribution, sampling is done using Latin hypercube sampling, and independent Rayleigh numbers are chosen on the interval [10, 210] to generate 2000 heat map images. To train an image-to-image regression model, we converted the Rayleigh value numbers to 32×32 images using Numpy and Matplotlib packages, each representing a specific Rayleigh value pertaining to a heat map image and pixel values of images are normalized between 0 and 1 in the preprocessing step of neural network training. All pixels of Rayleigh images have the same values for each image due to homogeneity; this is because we are developing a methodology, and though it might seem counterintuitive, we are using a homogeneous case as a first step. The input-output pairs are used to train an encoder-decoder CNN.

## 2.3 Encoder-Decoder CNN

We employ an encoder-decoder architecture for this problem, consisting of two separate subnetworks; encoder is a subnetwork that extracts features through a contracting process, followed by a decoder, which reconstructs the image [5],[6]. Decoder CNNs usually have the same network architecture as encoders, except that they are oriented in the opposite direction [7]. They recover the spatial resolution lost at the encoder by deconvolution and up-sampling and construct output maps based on the feature maps from the encoder [8],[9]. After data preparation, we trained the model with a maximum number of 2,000 samples, where 50% are used for training, 30% for validation, and 20% for testing. We developed two ED-CNNs, one as a meta-model and the other as an optimizer. The meta-model is trying to estimate the heatmap distribution as an output while the input Rayleigh parameter images are fed to the model. Furthermore, a similar methodology has also been employed to develop a model that acts as an *optimizer* to estimate the Rayleigh number from the heat distribution. The ED-CNN models have been developed using Keras and Tensorflow python machine learning libraries. **Fig.2** shows our proposed ED-CNN [2], which is constructed using convolutional layers, each of which is followed by a batch normalization layer, which regularizes the network while enhancing the accuracy [10] Two times, down-sampling and rebuilding is done using two pooling layers and two upsampling layers, respectively in the middle of the network. Furthermore, the activation function is rectified linear unit (Relu), but the sigmoid function is also used in some layers, and the loss function is mean squared error. The model is trained with 300 epochs using batch size 24 and the learning rate of 0.0001 with Adam optimizer.

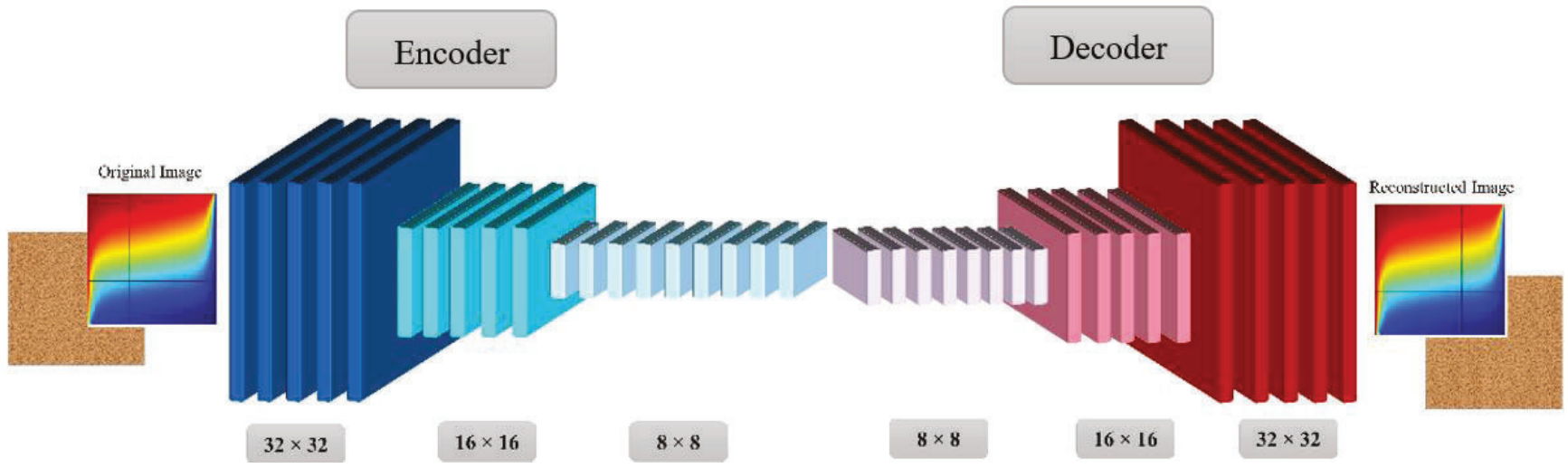

**Fig 2**. The architecture of the proposed encoder-decoder CNN

# 3 Result and discussion

## 3.1 CNN as meta-model

Different numbers of training input-output images (including 100, 500, 1000, and 2000) generated from the numerical model are employed to train the proposed networks. Two evaluation criteria are used to assess the performance of the developed ED-CNN modes: (1) the root mean squared error (RMSE) [11], and (2) the coefficient of determination ($R^2$-score), a number that shows a good prediction as it gets closer to 1 [12].

$$RMSE = \sqrt{\frac{1}{N}\sum_{i=1}^{N}\left(y_{pred} - y_{\,true}\right)^2} \tag{6}$$

$$R^2 = 1 - \frac{\sum_{i=1}^{N}\left(y_{pred} - y_{\,true}\right)^2}{\sum_{i=1}^{N}\left(y_{pred} - y_{mean\ true\ values}\right)^2} \tag{7}$$

**Fig3.a** illustrates RMSE decay with different numbers of sample sets. It is apparent from the plot that training the network with about 60 epochs could be enough to reach a stable value of errors. Increasing the number of samples to 2000, the RMSE converges to an acceptable value of 0.0186. **Fig3.b** shows $R^2$-score changes with the number of samples. As it is apparent from the plot, increasing the number of samples from 100 to 2000 samples slightly improves the accuracy, which is more than 0.97, conforming the RMSE plot results. As an example of the results, the performance of ED-CNN used as the meta-model is visualized for a specific value of the Rayleigh number in **Fig4** using different numbers of sample sets to assess the effect of the number of samples. In this figure, we compare the CNN's predicted heat map with the numerical modeling result, which shows a prediction with a decreasing error as we increase the samples from 100 to 2000. Using only 100 images shows a noticeable spatial error with a total error of 0.05, but increasing samples to 2000 decreases the total error to about 0.01. The spatial distribution of the error, that is, the absolute value of the difference between CNN and numerical model predictions of temperature, is calculated pixel by pixel. In the meta-model case, we can see that in the middle of the domain, errors are more prominent.

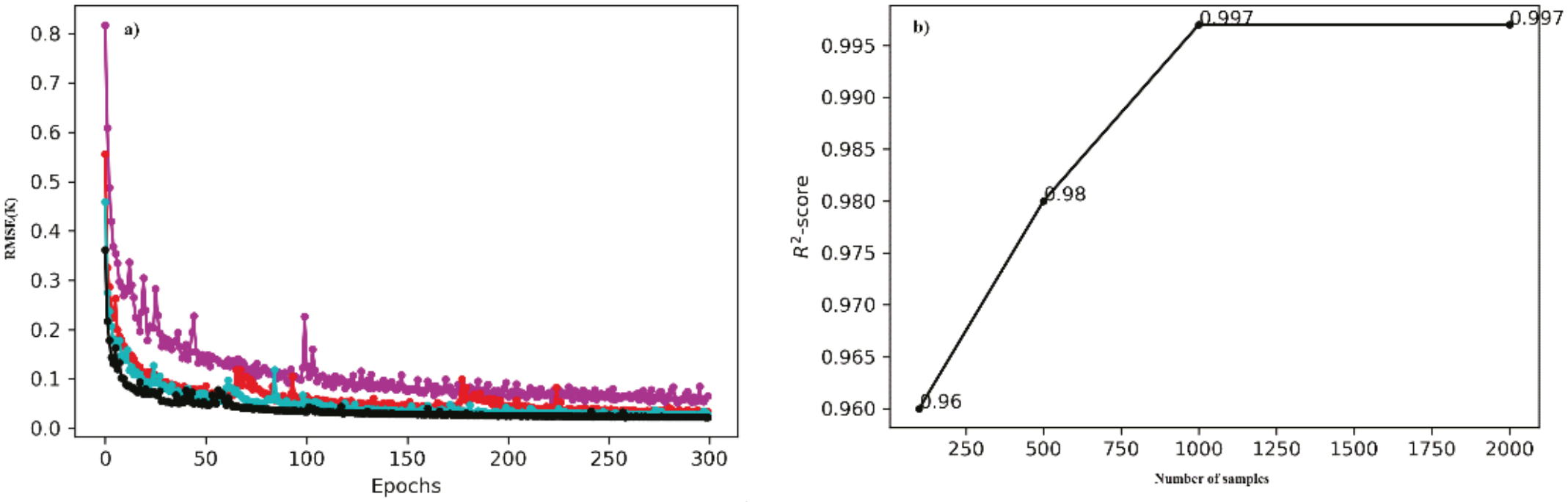

**Fig 3**. a) RMSE(K) and b) $R^2$-score ED-for CNN as meta model

| Number of samples | Real HeatMap | Predicted HeatMap | Spatial distribution of the error |
|---|---|---|---|
| 100 | 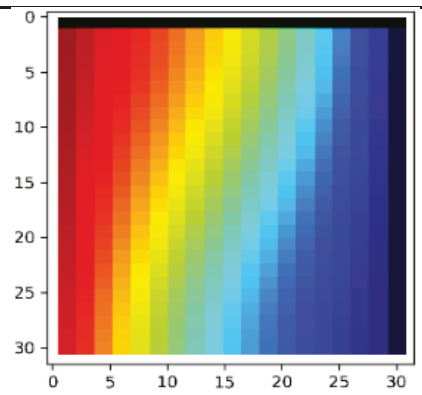 | 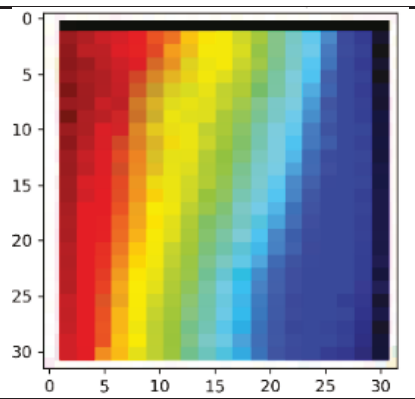 | 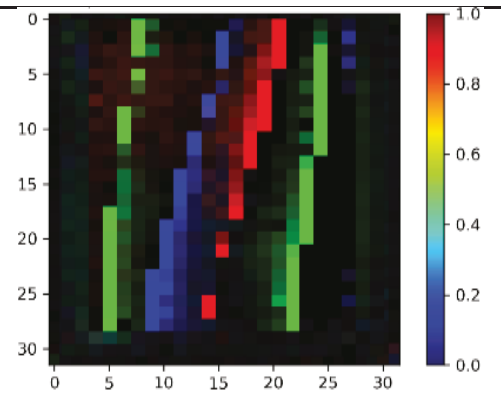 |
| 500 | 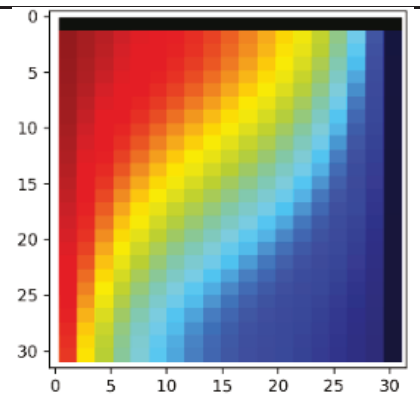 | 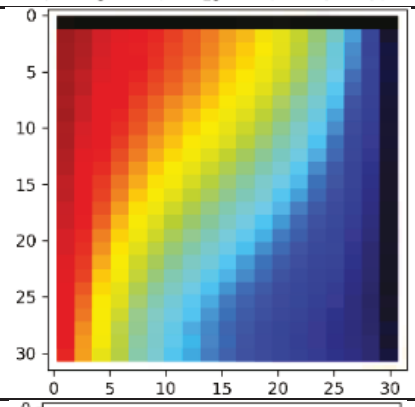 | 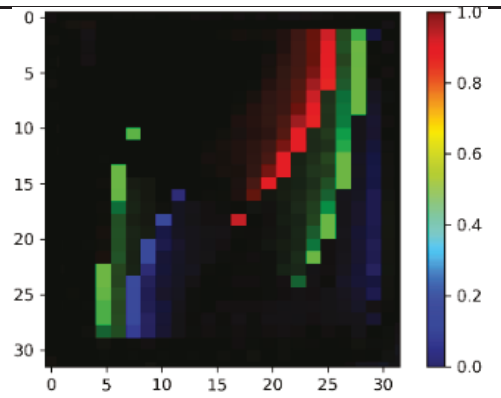 |
| 1000 | 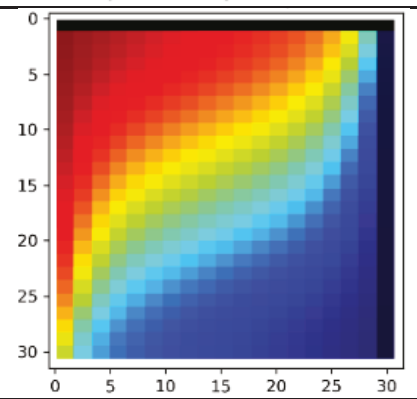 | 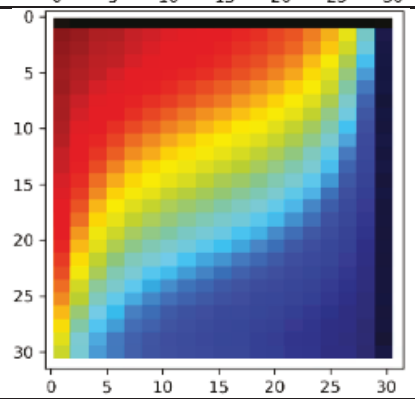 | 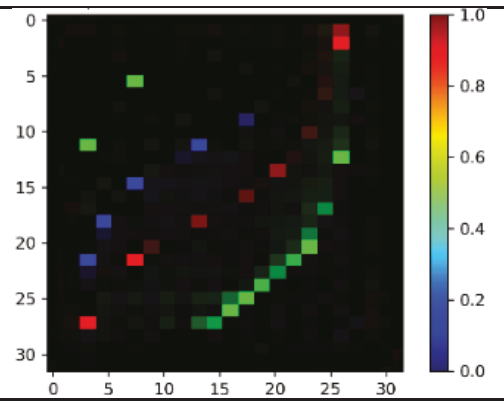 |
| 2000 | 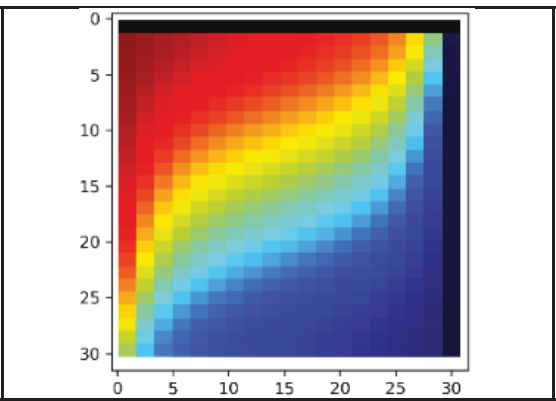 | 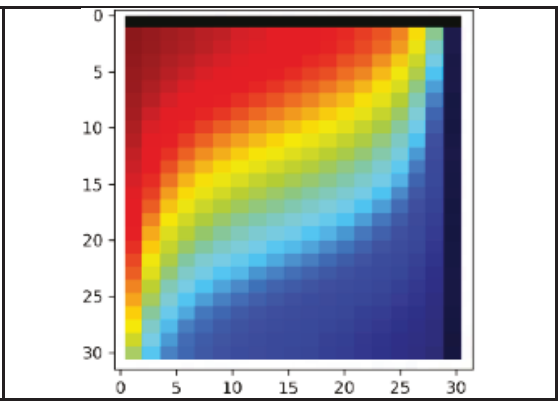 | 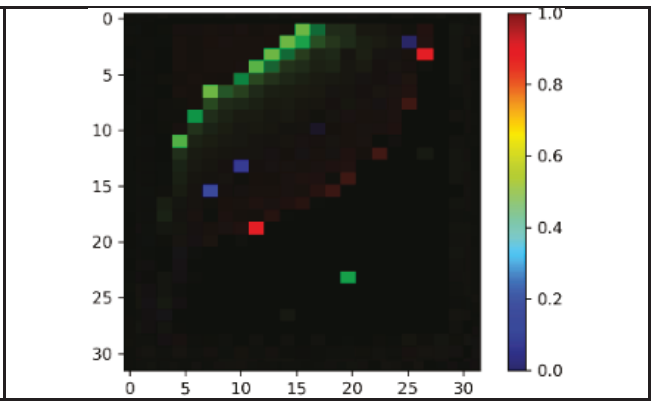 |

**Fig 4**. Real and predicted heat map comparison for different number of samples for CNN as meta-model

## 3.2 CNN as an optimizer

To evaluate the proposed network for parameter estimation, the same performance metric, RMSE and $R^2$-score are employed. **Fig5 a.** illustrates the RMSE decay with the number of epochs. It can be inferred from the plot that increasing the number of samples from 100 to 500 significantly decreases the error while using more than 500 does not affect the RMSE noticeably; this fact is also approved by **Fig5. b**, while using 500 samples instead of 100, enhances the $R^2$-score from 0.59 to a value of more than 0.98. The other assessment method is shown in **Fig. 6** as a scatter plot, comparing the predicted and actual values using a maximum training data of 2000. As it is apparent from the plot, the majority of predicted and real cases cluster around the $45^{\circ}$ line, approving the network's effectiveness. Furthermore, an exemplary table of random predicted and test values also confirms the robustness of the network showing a deficient relative error.

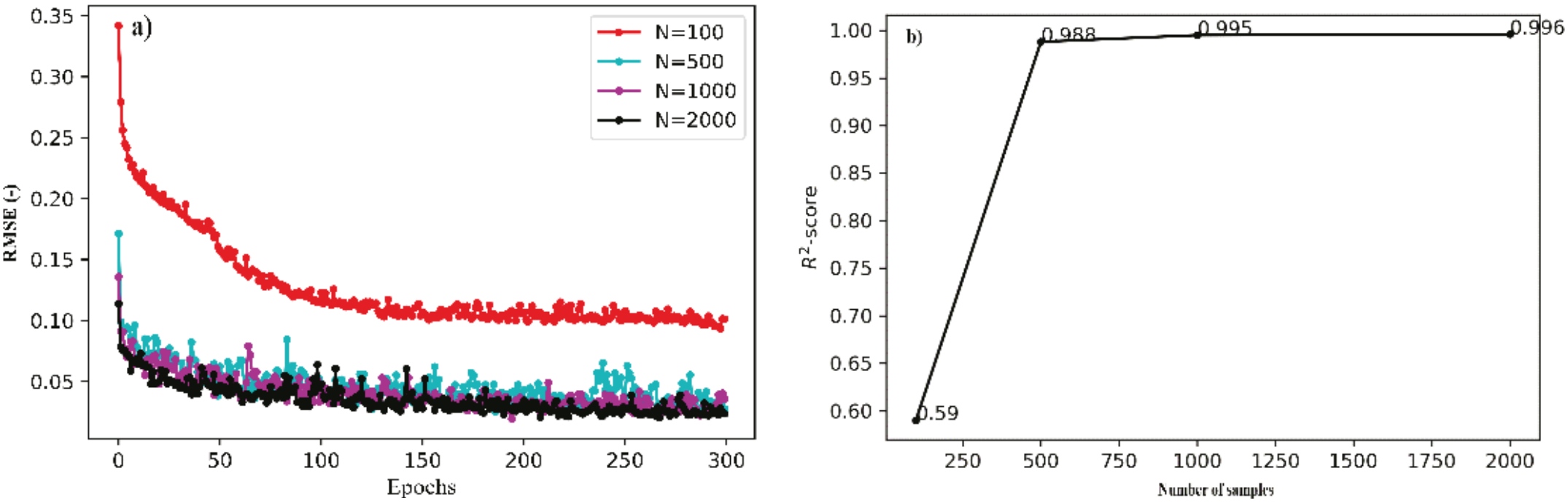

**Fig5**. a) RMSE(-) and b) $R^2$-score for ED-CNN as optimizer

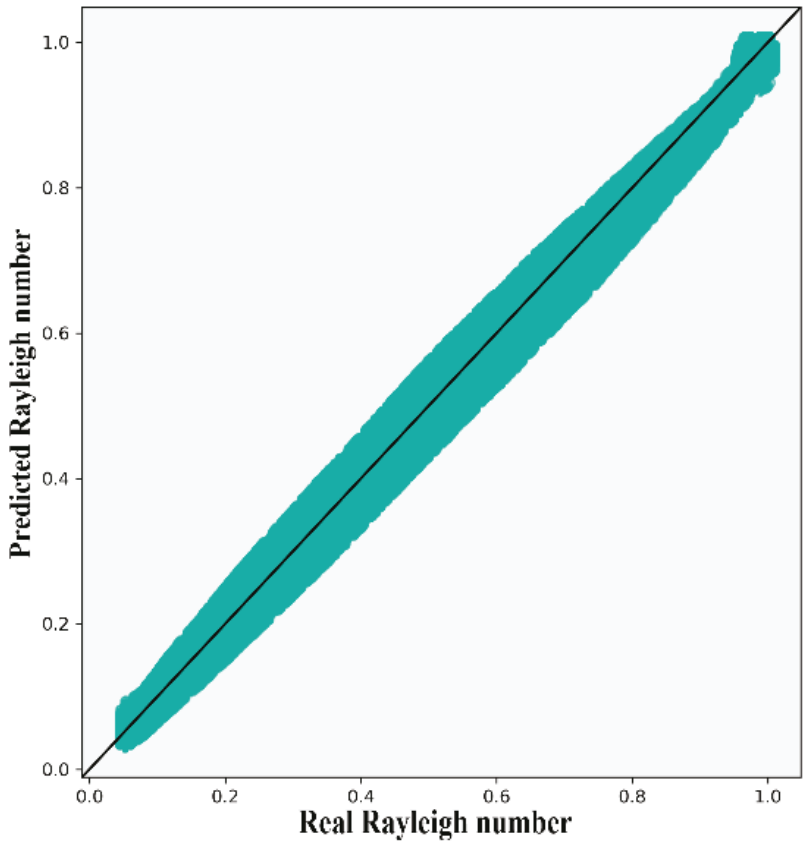

**Fig 6.** Real and predicted Rayleigh value scatter plot

**Table 2**. Relative error of real and predicted Rayleigh value for random test cases

| Real Rayleigh number | Predicted Rayleigh number | Relative error (%) |
|---|---|---|
| 49.11 | 50.16 | 2.13 |
| 56.25 | 57.30 | 1.86 |
| 60.0 | 60.8 | 1.33 |
| 89.4 | 89.8 | 0.44 |
| 117.5 | 117.7 | 0.17 |
| 192.68 | 190.16 | 1.30 |

## 4. Conclusion

In this paper, we have developed an encoder-decoder CNN to achieve a meta-model and optimizer for natural convection problems in porous media, considering the time cost and accuracy of the model. We initially generated 2000 heat map images; The data is then trained using similar architectures as meta-model and optimizers. It is apparent from the accurate results that the proposed methodology can be employed as a tool for estimating natural convection heat distribution as a meta-model and estimating the properties of the porous cavity in inverse modeling. It is also observed that this network is trained in about 40 minutes while the numerical modeling process takes more than 600 minutes, which means it saves time, more than 93% compared to numerical modeling tools, showing its robustness in solving the time cost problem. In summary, this network can be used as a meta-model and optimizer and should be also useful for uncertainty analysis.

# Point Cloud Capturing and AI-based Classification for as-built BIM using Augmented Reality

Thomas Klauer and Bastian Plaß

i3mainz, Institute for Spatial Information and Surveying Technology,
Mainz University of Applied Sciences
**(thomas.klauer, bastian.plass)@hs-mainz.de**

**Abstract.** The benefits of using Building Information Modeling (BIM) have been proven in architecture, engineering and construction industry. However, implementing BIM in in facility management has not been achieved yet due to missing complete and accurate as-built BIM. Modeling comprehensive information for as-built documentation from 3D point cloud data is referred as Scan-to-BIM but lacks automation caused by unstructured data and high user input. We tackle the main issue of structuring the 3D point cloud data by using artificial intelligence while capture. With both, a highly reliable and low-cost technology we achieve less time-consuming point cloud capturing and segmentation contributing to a novel Scan-to-BIM approach with promising initial results.

## 1 Introduction

Similar to many other sectors, digitalisation is advancing rapidly in the architecture, engineering and construction (AEC) industry. An important component here is the concept of the "digital twin" of a new building to be constructed or an existing building to be operated on. Building Information Modelling (BIM) has been established as the method for this, in which a 3D building model represents the core element. There are various ways to create such a model: in the case of new buildings to be planned, the model is generated "form the scratch" by the planners with specialised design software, while in the case of existing buildings it is necessary to capture the real building geometry with its relevant component information in reality. In addition to planning and construction, the management of existing buildings also benefits from digital BIM solutions, so that facility management will be able to exploit the advantages of BIM in the future with the ongoing development of efficient solutions for digitising existing buildings.

There are various approaches to 3D as-built modelling, such as deriving from 2D CAD plans (as-planned BIM) or capturing the up-to-date building representation by metrological methods (as-built BIM). In the latter, state-of-the-art 3D point cloud data are obtained from laser scanning or structure from motion (SfM) methods and serve both registered and manually pre-processed as the information basis for modelling a semantically enriched as-built BIM (Scan-to-BIM). The as-built modelling process lacks automation yet due to missing, sparse, outdated and complex information about the captured objects, relationships and attributes as well as customisable uses of the BIM [1,3-4]. In addition, the use of professional and therefore expensive metrology hardware has been required to perform such scans and specialised experts are needed to carry out both the scanning and modelling processes. This paper will show how these processes can be improved in terms of automation and simplification.

For an initial simplification of the scanning process, preliminary work [2] has shown that sufficient point cloud quality can be achieved with inexpensive consumer products such as the Apple iPad Pro or the Intel RealSense L515 for as-built modelling of indoor scenes. This hardware can be handled by non-experts after a brief instruction, resulting in a significant cost reduction for capturing the 3D building geometry. In order to automate and thus simplify the modelling process, methods are needed that can provide semantically structured information of the scanned building geometry to identify relevant, constituent objects, such as building components, furnishings or building services elements.
One way to achieve this is to divide the raw and unstructured 3D point cloud into semantic regions by means of segmentation, in which objects are then automatically recognised [3] and finally geometrically approximated by standard geometries. Structuring the 3D data into semantic regions for further understanding thus represents the initial technical step in the BIM modelling process (c.f. Fig. 1). Artificial Intelligence (AI) methods such as machine learning (ML) can be used for segmentation and classification of 3D point cloud data showing various characteristics and dealing with different conditions usually. Published research [1,4-5] confirms the successful use of automated approaches for highly simplified building representations, but less so for the representation of complex reality.

This paper presents the development of an intelligent 3D data acquisition and processing method using LiDAR-based consumer hardware, designed for Apple's Pro Series such as the iPhone 12 Pro and the iPad Pro. The prototyped 3D data application called "Semantic Data Capture" provides a detailed and semantically structured point cloud using AI based on high-resolution depth data acquired by Apple's vision technology without prior calibration and less technical knowledge. This is done by capturing a depth map with the mobile device's built-in sensors. In combination with MLCore, extended by a third party ML model, captured geometries can be classified simultaneously into pre-defined building component categories. The prototype also uses Apple's augmented reality framework (ARKit), which allows users to visualise the results of the captured and classified data in the overlaid AR-image of the integrated RGB-camera synchronously. An application prototype has been developed that is usable also for non-experts, able (a) to generate detailed geometric representations of interior scenes, (b) to combine them with semantic attributes in real time and (c) to deliver a structured 3D point cloud for the further Scan-to-BIM process. As an application example, an interior analysis in the care sector was chosen here, which, for example, checks the suitability of living sites for certain diseases or care situations. A key feature of this novel application in academia is the simultaneous geometric capture and AI-based 3D data classification through a low-cost optical technology with direct visualisation for non-expert users, which has the potential to establish a new state-of-the-art Scan-to-BIM method.

## 2 Requirements for Scan-to-BIM and why it lacks automation

Referred as Scan-to-BIM, the automated reconstruction of existing buildings for BIM modelling is based on 3D point clouds, acquired by consolidated techniques such as laser scanning or SfM. Both techniques allow a rapid and up-to-date acquisition of as-built components with high spatial resolution but produce a huge amount of data as a consequence, that needs to be processed in a time-consuming and almost entirely manual process chain as presented in Fig. 1. The BIM model generation from acquired point cloud data can be roughly organised into the four main categories data preprocessing, segmentation, classification and BIM modelling.
Representing indoor scenes full covered by point clouds requires a variety of scan stations and viewpoints that need to be registered immediately after acquisition. The scope of the preprocessed step is to remove outliers, smooth noise effects and for example downsampling or transforming the point cloud data into a usable format for subsequent processes [6]. Following that, the segmentation serves as the first technical step to transform the unstructured point cloud into several subsets according to the semantic property of points with respect to the scene characteristic. In line with the subsequent demand, the subsets can address rooms, unique constituent elements or specific regions of interest such as interiors or furnishing elements. Aside the segmentation, classification becomes relevant for mapping the segmented but disordered point cloud data by feature extraction and component identification into regular forms that are capable for the final BIM modelling.

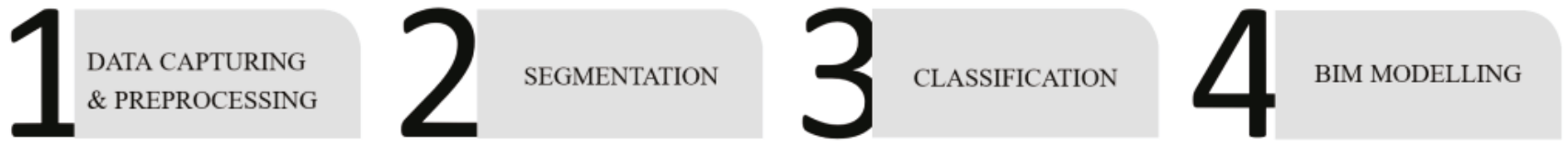

**Fig. 1.** Scan-to-BIM stages according to [6]. The presented approach summarises the first three categories in order to achieve automation and simplification. Consequently, it reduces the manifold process chain for as-built BIM reconstruction.

Despite the increasing popularity of BIM for years and a quantity of technical solutions for automated Scan-to-BIM, the overall process remains costly and manually. Recent research efforts have been made in the development of automated point cloud segmentation and classification methods, mostly using ML approaches based on artificial neural networks and deep learning [7-12]. However, point cloud processing is still in its infancy because of several challenges described below.

As a consequence of evolving 3D data acquisition technologies, point clouds represent the state-of-the-art in surface reconstruction for engineering and 3D modeling tasks in the last decades [13]. Given a number of discrete points that sample a surface, point clouds are used prevalently but their processing still faces many challenges due to data imperfections and a variety of special data structure characteristics. The most challenging properties dealing with point clouds are the arbitrary sampling density that results in high redundancy on the one hand but still missing, sparse or obsolete data on the other hand due to occlusions, clutter and noise [1,4-5]. Apart from that, 3D point clouds have no inherent order such as pixel neighbours. That is why permutation invariance is another main property to take care of. Furthermore, point clouds are often dense with millions of single points that will provoke big data and large computational times raising cubicly on the number of point instances. All these properties affect the automatic processing of point clouds. Nevertheless, many efforts have been made to improve the quality of automated point cloud processing and to detach manual work. Even if the methods differ in approaches according to [14,15], they all used to apply after acquisition in postprocessing. Considering the large amount of data, postprocessing methods require the point cloud to be patched or significantly downsampled resulting in a loss of information and additional cost. While performing segmentation tasks on 2D image data using DL techniques represents state-of-the-art, a transfer to 3D point cloud data lacks due to above-mentioned reasons. Therefore, this paper proposes a new approach for point cloud segmentation in real-time during the acquisition process using deep learning techniques for both data paradigms, 3D mesh classification and reprojected 2D object classification result by a pre-trained model. Further details about our approach that is also using Augmented Reality (AR) for result visualisation are given in the following chapter.

## 3 Mobile AI-based Augmented Reality Approach

Mobile devices, such as tablets and especially smartphones with integrated LiDAR sensors, are in principle well suited for the 3D acquisition of buildings and building components, as they are widely used as personal devices and thus the users are usually familiar with their functionality and usability. Compared to professional acquisition hardware, they are also comparatively cheap to purchase and operate with and can be used also by non-experts. Just a few apps like[1,2,3,4] on the consumer market enable 3D scanning of indoor spaces. However, these apps are not able to carry out the necessary process steps such as segmentation and classification after the pure geometric acquisition. In order to avoid processing steps with a separate software after or detached from the capture, the main scope of the concept presented here was to integrate this directly into the mobile application, hereinafter app. The "Semantic Data Capture" app introduced here integrates the functionalities pointed out in Tab. 1.

**Tab. 1**. Functionality of the presented approach based on a mobile application.

| |
|---|
| Capturing the 3D geometries of inner building structures and interiors. |
| Segmentation, classification and recognition of components and furnishings. |
| Immediate visualisation of the capture and recognition with corresponding usability. |
| Saving of the results. |
| Exporting the results in relevant and open 3D formats. |

### 3.1 Geometry Capturing

On a current mobile device from Apple, geometry capturing and also parts of object recognition are performed with the ARKit framework [16] using the built-in LiDAR sensor. In an AR session, ARKit stores all information that belongs to the captured environment. The core functionalities are tracking, i.e. the ability to follow objects relative to the position of the device, and scene understanding, i.e. the ability to collect information about the detected objects. In addition, ARKit offers simple integration into existing visualisation libraries (e.g. SceneKit and SpriteKit) as well as into individual visualisation solutions with Metal, Apple's computer graphics API. ARKit uses the built-in sensors of the mobile device such as the namely HD camera, 6D sensor for rotation, position and accelerometer as the basis for so-called Visual Inertial Odometry (VIO), for e.g. the determination of position and orientation supported by the RGB camera feed. The captured optical data is superimposed with the other device sensors (e.g. gyroscope and accelerometer) and then position and movement of the device in 3D space are calculated.

A method called raycasting is used to capture the spatial geometry, which allows 3D scenes to be displayed quickly. This is done by using a virtual ray that is projected from a point on the screen into the real world and allows the calculation of the intersection point with real objects (c.f. Fig. 2b). During the ongoing capture, ARKit then creates a virtual world of the captured geometries. To ensure that the computer-generated elements remain in their real positions, so-called virtual anchors are generated. The anchors used in the app presented here contain 3D geometry data that describe the objects from the environment in the form of nodes, polygons and normals, which in turn form a polygonal mesh. In addition, semantic information can be predicted and assigned to each polygon from the mesh. Eight classes are currently supported with ARMeshClassification [17], namely ceiling, door, floor, seat, table, wall, window and *none,* when ARKit cannot predict the class of the polygon. Since these eight classes are not sufficient to recognise interiors with various other components and furnishings, an AI-based extension has been developed.

[1] 3d Scanner App: https://apps.apple.com/de/app/3d-scanner-app/id1419913995
[2] Canvas: Pocket 3D Room Scanner: https://apps.apple.com/us/app/canvas-pocket-3d-room-scanner/id1514382369
[3] Trnio 3D Scanner: https://apps.apple.com/de/app/trnio-3d-scanner/id683053382
[4] Capture: 3D Scan Anything: https://apps.apple.com/de/app/capture-3d-scan-anything/id1444183458

### 3.2 AI-based Mesh Classification and Object Detection

In order to be able to recognise objects in the virtual representation of a captured interior that are not part of the ARMeshClassification classes, i.e. they have been assigned to the class *none*, an AI-based extension was designed and prototyped. Various libraries or models are available here that can detect objects from (moving) images in real time. The YOLO (You Only Look Once) model was chosen in this work [18]. The approach of YOLO, in contrast to many other systems that work based on Convolutional Neutral Networks (CNN), is to perform object detection in a single pass – hence "you only look once". To make this possible, the CNN YOLO was trained with data from Microsoft's Common Objects in Context (COCO) database [19]. This trained network can now be applied to images or videos to perform multiple object detection in a fast manner. The model recognises features across the entire image and creates individual bounding boxes that assign a class to recognised objects according to the highest probability. Images are divided into a symmetric grid, where frames are suggested from each cell. Class probabilities are also calculated per cell, corresponding to the number of known classes in the training dataset. The class probabilities depend on the probability that an object is present in the cell.

The captured images are first preprocessed with Apple's vision framework. Afterwards a collection of the objects found is returned in the form of observations or an empty array if no objects were found by ARMeshClassification. An observation contains the class of the object and the normalised coordinates of the origin as well as the width and length of a frame within which the object should be located. This data is then passed to CoreML, Apple's ML framework, where it is classified in the app using YOLO, which is one of several models that can be integrated into CoreML. As an example for indoor scenes, five new classes were implemented, namely tvmonitor, laptop, bed, sink and toilet. The integration of YOLO was a particular challenge in terms of software technology, as mesh classification with ARKit and object detection with YOLO could not be processed in the same procedure. Instead, the processes were split, first the classification with ARKit and then detection with YOLO. The processes could also not be parallelised in the prototype because of the permanently regeneration of the mesh geometry by ARKit. That is why YOLO processes static areas after they are already classified as *none*.

### 3.3 AR-based Interactive Visualisation

To visualise the detection process on the mobile device, the classified mesh is overlayed with the real camera image, i.e. AR is created. In case of successful object detection with both the eight ARKit standard classes and the extended five YOLO classes, the affected area of the mesh is coloured associated to the object class. The SceneKit framework [20] (among others) was used for visualisation on the mobile device screen in real-time. SceneKit displays a 3D scene, such as the virtual image created while capturing the interior spaces, on the screen. It calculates which elements of the generated mesh are visible from the current camera angle and displays them on the screen. Since the colouring of the mesh runs parallel to the classification, the mesh generated by ARKit is coloured first and then, after the scan has been completed, all grey (i.e. ARKit class *none*) mesh parts are processed with YOLO and coloured in terms of highest class probability accordingly to the object detected. For this purpose, the objects detected in the camera frames by YOLO,as explained in the previous section are then spatially assigned and reprojected from the 2D screen to all anchors in the 3D mesh by raycasting operations. The result can be seen in the following Fig. 2, where an object of the class *laptop* is classified first as *none* (c.f. Fig. 2a). After that, the region is detected by YOLO, thus framed in yellow and reprojected to the generated anchors of the 3D mesh by ARKit (c.f. Fig. 2b).

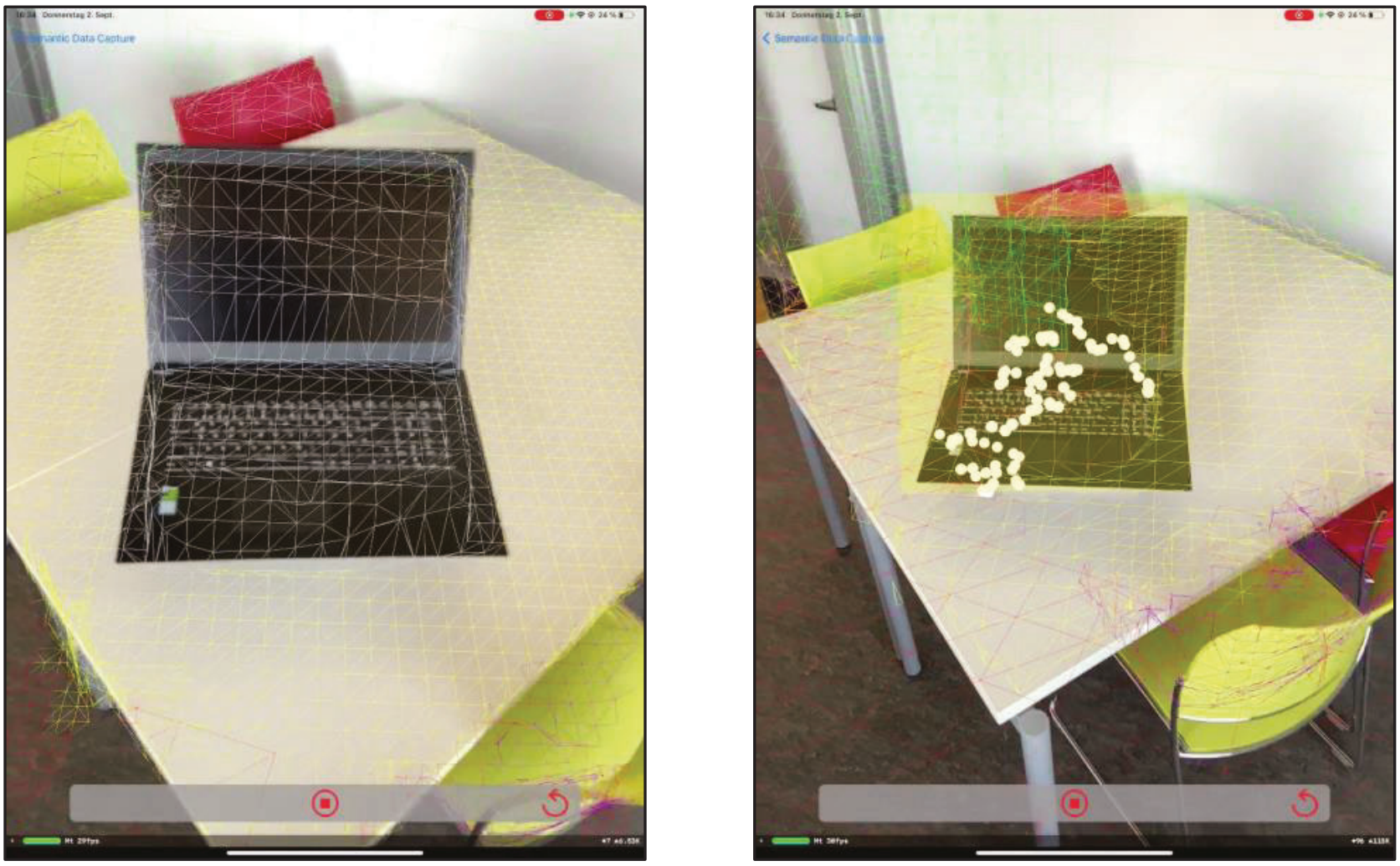

(a) (b)

**Fig. 2.** Acquisition process of a specific object that is classified first as *none* with ARMeshClassification (a) and later finalised with YOLO as *laptop* (b). The yellow bounding box represents the region of interest for YOLO that is projected to the captured mesh surface by ray traces. The video sequence[5] according to that figure was captured by an Apple iPad Pro in debug mode with 30 fps.

## 4 Result and Outlook

This paper presented the mobile app “Semantic Data Capture” based on iOS for (a) capturing 3D building sites with consumer products and (b) recognising interior structures and furnishings using efficiently augmented reality (AR) and machine learning (ML) methods. This serves to support the AEC industry in general for planning purposes and facility management for e.g. automated space analysis with respect to BIM. It was shown how user-centred working can also be made possible for non-expert users with the help of an integrated LiDAR sensor and augmented reality. For this, consumer hardware from Apple was used, which is currently the only manufacturer offering the functionality of active low-cost and real-time 3D scanning. With a combination of both, in-built sensors and close software libraries using AR and ML techniques, a new method for Scan-to-BIM was suggested and successfully prototyped. Here, Apple's own recognition methodology for furnishings was extended with classes from the freely available YOLO model. The results after the data export are shown in Fig. 3.

[5] Video sequence of “Semantic Data Capture”: https://video.hs-mainz.de/Panopto/Pages/Viewer.aspx?id=6838959a-83af-4e36-918d-ad970123048d

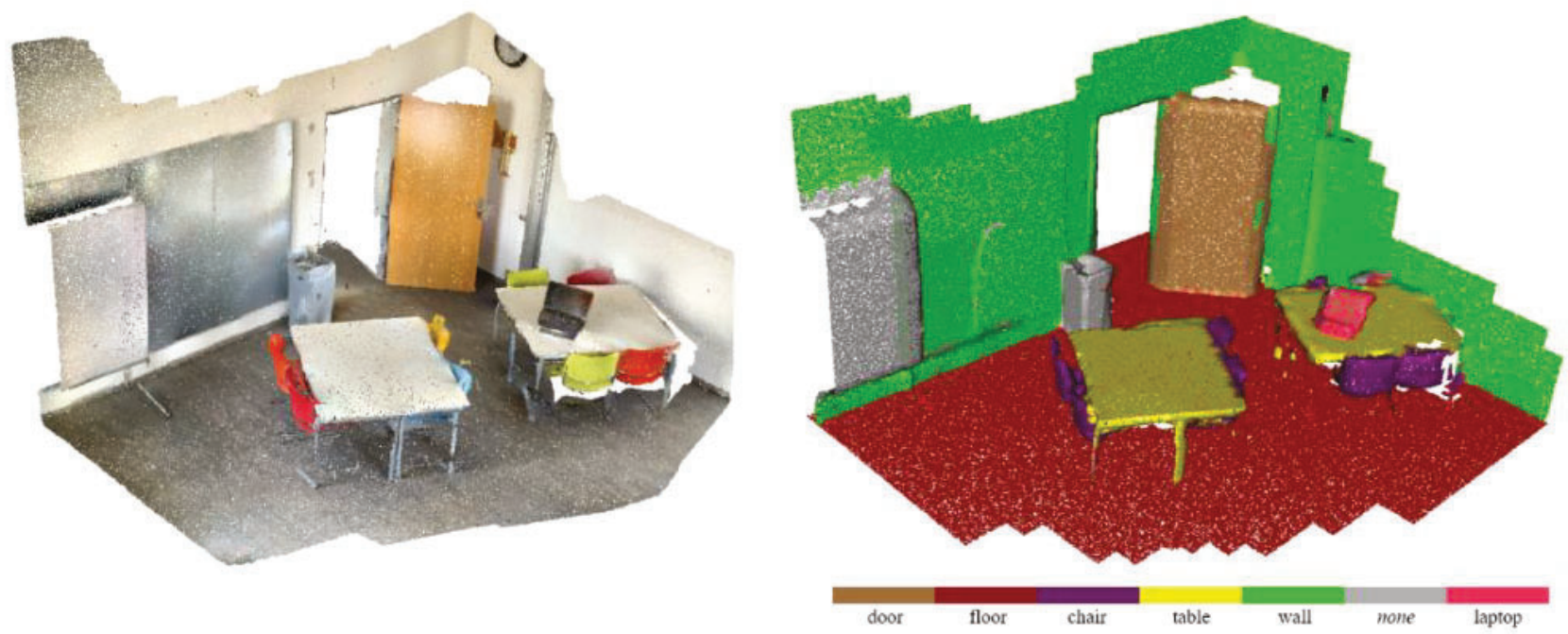

(a) (b)

**Fig. 3.** Results of the app "Semantic Data Capture" (b) illustrated by CloudCompare in comparison to an RGB coloured point cloud captured by 3D Scanner App[1] (a). The first six classes are from the ARKit mesh classification while the class *laptop* originated from the extended YOLO model.

Despite the successful implementation of the prototype, there are several challenges that need to be considered in future. First, the simultaneous mesh classification by ARKit and object detection by YOLO could not yet be implemented. This is one of the improvement possibilities envisaged for the immediate future. Improving this aspect, the dual data acquisition could be avoided and working in real-time will be possible. Further additions can be made in the area of object detection with ML models. For example, YOLO or other models can be used to add further classes for additional components or furnishings at free will. As another main improvement, the acquired data could be used in combination with existing models to generate more precise ones individually prepared for the situation through training. One of the core disadvantages of the solution presented here is the utilisation of libraries provided by Apple that cannot be viewed and changed by their black box character. Considering the geometrical accuracy of the captured and classified objects, LiDAR could not keep up terrestrial laserscanning or SfM techniques in terms of reliability and precision. As a consequence, LiDAR results serves currently just for coarse BIM modelling.

Nevertheless, in future, the ability of LiDAR will increase with respect to acquisition and registration accuracy. By means of that, LiDAR will be among the most important acquisition methods for 3D point cloud capturing even for fine geometry and BIM valid model fitting. In collaboration with powerful and smart mobile devices, the possibilities to process 3D data are so far not limited as the prototype shows. "Semantic Data Capture" provides a new paradigm of building indoor acquisition and processing simultaneously in order to achieve the final scope of Scan-to-BIM: the automated 3D modelling of buildings with appropriate accuracy by using mobile devices.

## 5 Acknowledgements

The authors would like to thank D. Iordanov for the contribution to the development of the application within his studies at Mainz University of Applied Sciences.

# A Reference Model for Dialog Management in Conversational Agents in High-Engagement Use Cases

Nima Samsami, Stephan Kurpjuweit

Hochschule Worms – University of Applied Sciences
samsami@hs-worms.de
kurpjuweit@hs-worms.de

**Abstract.** The objective of our research is to systematically derive a refined reference model for the dialog management component of a conversational agent. Firstly, we characterize high-engagement conversational agents and derive solution strategies to address this class of agents. Secondly, we propose a set of conceptual components that refines the dialog management component and addresses the solution strategies. Thirdly, we survey implementation approaches for the individual components of the reference model.

## 1 Introduction

With the general availability of smart speakers since 2017, conversational agents have gained increased popularity among consumers [1]. Based on our experience, the use cases of conversational agents in the consumer domain can be characterized as either "task-oriented use cases" or "high-engagement use cases".
The objective of our research is to systematically derive a reference model for the dialog management component of a conversational agent from the requirements of high-engagement use cases. Our research is based on the following approaches: We characterize high-engagement conversational agents (section 2) and derive solution strategies to address this class of agents (section 3). Then we propose a set of conceptual components that refines the dialog management component and addresses the solution strategies and survey implementation approaches for the individual components of the reference model (section 5).

## 2 Quality characteristics of high-engagement conversational agents

Based on our experience, the use cases of conversational agents in the consumer domain can be characterized as either "task-oriented" or "high-engagement": For task-oriented use cases the objective is to answer the users‘ information needs or to complete a task in as few conversational turns as possible. Example domains include banking, customer service or directory services. As the user wants to 'get a job done', satisfying the following two quality characteristics is essential:

(1) Relevance and (2) Focus: By nature, the bandwidth of conversational agents (i.e. the amount of information that can be communicated to the user per time) is small compared to other - esp. screen-based - digital channels like web or mobile apps. Thus, conversational agents must deliver relevant and focused responses and reduce the number of conversational turns users have to take (‘get to the point‘).

For high-engagement use cases the objective is to keep the user engaged in the conversation for as long as possible. Example domains include media, news, entertainment or conversational commerce. In the context of our research, the level of user engagement is characterized by (a) how often the user starts a session with the agent per time [2], (b) how much time the user spends per session [3], and (c) for how long the user is active overall (customer lifetime).

While focus and relevance is central to all conversational agents, it may not be enough to ensure a high level of user engagement. Depending on the concrete nature of the agent, other quality characteristics should be taken into consideration, including:
(3) Variety: Agent should provide a natural, varied language and avoid repetitive phrases (’don’t bore me).
(4) Topicality: Agents should provide pieces of information that are current and new to a user, so that users frequently feel the need to engage with the agent (’satisfy my curiosity’).
(5) Discoverability: Agents should suggest follow-up actions that may be of interest to the user (’show me what else you can do for me’).
(6) Adaptability: Agents should be personalized and adapt to the user’s needs over time (’become my companion’).

## 3 Solution strategies for high-engagement conversational agents

To address the quality characteristics of the high-engagement conversational agents outlined above, concrete solution strategies have to be implemented. The following list describes generic solution strategies which we expect to be beneficial for most high-engagement conversational agents:
(1) Text variation generation: To avoid repetitive phrases, text variations should be generated (ideally automatically).
(2) Personalized content: Personalized and current content should be selected and delivered to the user.
(3) Education: Short messages that explain additional features and follow-up actions should be delivered to the user.
(4) Graceful error handling / disambiguation: Instead of entering error flows, the agent should try to understand the user’s intent, e.g. via disambiguation.
(5) Context-awareness: The agent should adapt to the usage context. For example, the agent may decide to deliver a longer response of the user is driving in a car.
(6) Modular responses: To deliver varied responses, personalized content, educational messages, etc. the response should be composed of text fragments in a flexible way.

## 4 Conversational agent reference architecture

Figure 1 shows a well-adapted reference architecture for conversational agents, which decomposes the dialog management component into four sub-components. This architecture serves as a basis for the refined reference model in section 5:
(1) Natural Language Understanding (NLU): Identifies and parses a user’s text input to obtain semantic tags that can be understood by computers, such as entites and intents [4].
(2) Conversational State (CS): Maintains the current conversation state based on the

conversation history. The conversation state is the cumulative meaning of the conversation history, which is generally expressed as slot-value pairs.
(3) Conversational Flow (CF): Outputs the next system action based on the current conversation state.
(4) Natural Language Generation (NLG): Converts system actions to natural language output [5].

We select this modular architecture over an end-to-end architecture (see [6]). End-to-end architectures are based on successes of deep learning approaches in recent years. The system consists of a large neural network that handles all tasks such as NLU, NLG, CF, etc. This model is still being explored and is as yet rarely applied in the industry [6]. Although the trend is toward end-to-end systems, these approaches are still limited and cannot clearly outperform the traditional methods [7]. In practice, it may not be feasible to implement a specific agent solely based on an end-to-end architecture due to a lack of training data.

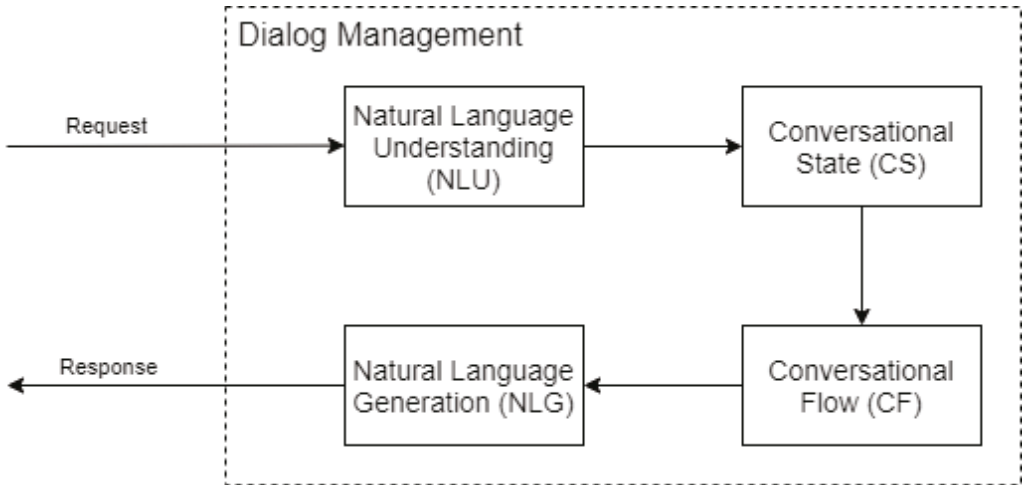

**Fig. 1.** Modular structure of dialog system

## 5 Reference model for high-engagement agents

Based on the strategies outlined in section 3, we derive a reference model for high-engagement agents. The reference model is a based of the reference architecture described in section 4. The reference model consists of a set of conceptual software components (see figure 2). Some components are optional and not required for all agents.

For the decomposition of the dialog management component we apply the following criteria: (1) Each component has a manageable set of responsibilities. (2) The implementation approaches for the components can be chosen - to a large degree - independently from each other. (3) The model can be implemented on the basis of existing conversational AI platforms and frameworks like RASA [8] or IBM Watson [9]. Available implementation approaches for the components typically range from traditional (e.g., rule-based) approaches to more sophisticated machine learning (ML)-based approaches. In practice, it may not be feasible to implement a specific agent solely based on ML approaches due to a lack of training data or development budget. Thus, in our opinion the latter criterion is important to allow for a selective component-by-component migration of a traditional implementation towards a ML-based implementation.

For each component we describe its responsibilities and survey possible implementation approaches. Figure 2 lists the conceptual software components. Flows between the components are omitted for clarity.

## 5.1 Component: Conversational Memory

**Responsibilities:** One limitation of conversational agents is that they cannot go back and forth in a conversation. This makes natural and dynamic communication between humans and computers difficult. A conversation can be carried across multiple topics. To do this, the agent must store what it has already talked about.
(1) Usage History: The entire usage history of a user stored for analytics purposes.
(2) Session State: The usage history of the current conversational session is stored with the goal to determine the current conversational context, i.e. which pieces of information a user request may refer to.
**Approaches:** The literature contains descriptions of many models of conversational memory. These models mainly seek to reflect how the human brain implements memory. Elvir et al. also describe an Episodic Memory Architecture to address this problem. [10] Vinkler et al. present an architecture consisting of two memory types. A short-term memory to understand the context and a long-term memory to allow the conversational agent to refer to previous information in the conversation [11].

## 5.2 Component: Personalization & Context-Awareness

**Responsibilities:** The personalization and context-awareness component accesses the usage history and calculates context information that may be required to interpret a new user request and determine the response. There are multiple flavours of conversational context which can be addressed by individual sub components:
(1) Personal Preferences: The personal preferences capture which intents and entities the user is especially interested in. These may be set explicitly or derived from the usage history. Personal preferences can be used to prioritize the text fragments selected for a user.
(2) Usage Context: The usage context captures aspects like the time of day, the usage environment (at home, in car, etc.), the device type (smartphone, smart speaker, etc.), and the interface type (chat, voice, multi model, etc.), which all may impact the response delivered to the user.
(3) Emotion Detection: The human being is an emotional being. Each person conditioned by his emotions and type expresses himself differently. To carry out a pleasant communication, it is therefore important to address the emotional intelligence aspect of communication.
**Approaches:** Hao et al. present a method for using content-consistent conversation to also engage in emotion-consistent communication. Emotional Chatting Machine (ECM) addresses this factor with three new mechanisms that respectively (1) model high-level abstraction of emotion expressions by embedding emotion categories, (2) capture the change of implicit internal emotion states, and (3) use explicit emotion expressions with an external emotion vocabulary. [12]

## 5.3 Component: Conversational Flow

**Responsibilities:** Check if all pieces of information to answer the user request are available (intent, slot values, context information) with sufficiently high confident values and decide whether to (1) answer the user request (standard path), (2) ask a disambiguation / clarification question (esp. if indicated by the entity disambiguation detection component), (3) enter an error handling flow or (4) hand over the conversational flow to a human (optional). Especially check if the input makes sense in the context of the current conversational state (e.g., if the agent is waiting for a response to a specific question) (5) Conversational State: The conversational state captures "what the conversation has been about" so far, so that the user can refer to entities mentioned in previous conversational turns and ask follow-up questions. The conversational state also determines if a specific type of input is expected in the upcoming conversational turn.

(1) Intent Ambiguity: aims to clarify the semantics of an Intent in context by finding the most appropriate meaning from a predefined Intent.
(2) Entity Ambiguity: Beyond word sense disambiguation, a word can mean something different in different contexts. E.g. Mars, Galaxy and Bounty are all delicious. It is difficult for an algorithm to figure out if it is talking about an astronomical structure or chocolate tokens.
(3) Conversational State: Maintains the current Conversational state based on the conversation history. The conversation state is the cumulative meaning of the conversation history, which is generally expressed as slot-value pairs.

**Approaches:** Decisions can be made rule-based. Decision criteria are the request data, the conversational state, and the confidence levels. The rules can be specified as part of the language model. Jan-Gerrit Harms et al. define Dialog Management as a component of Conversational Agents that processes the dialog context and determines the agent's next action [13]. Yinpei Dai et al. kategorisieren Dialog Management in three Generations. a) The first-generation dialog systems were mainly rule-based. b) Second-generation dialog systems driven by statistical data (hereinafter referred to as the statistical dialog systems) emerged with the rise of big data technology. At that time, reinforcement learning was widely studied and applied in dialog systems. A representative example is the statistical dialog system based on the Partially Observable Markov Decision Process (POMDP) proposed by Professor Steve Young of Cambridge University in 2005 c) third-generation dialog systems built around deep learning have emerged. These systems still adopt the framework of the statistical dialog systems, but apply a neural network model in each module [6]. In general, third-generation dialog systems are better than second-generation dialog systems, but a large amount of tagged data is required for effective training. Therefore, improving the cross-domain migration and scalability of the model has become an important area of research [6]. To solve the problems of domain dependency in end-to-end systems, Lu Chen et al. propose to use a multi-agent system, where the tasks are passed from a domain specialized agent to an agent trained on another domain [14] . Jan-Gerrit Harms et al. show a taxonomy of the approaches for managing dialogs and a classification of a selection of tools. [13]
Ambiguities can be determined by analyzing the entity and synonym lists of the language model. The problem can also be addressed using entity linking (EL). EL aims to resolve such ambiguities by establishing an automatic reference between an ambiguous entity mention/span in a context and an entity (persons, locations, organization, etc.) in a knowledge base. [15] Neural networks are used for this purpose as end2end systems [16] or in conjunction with ontologies [17] [18]. Sevgili et al. use graph embeddings as an efficient method [15]. María G Buey et al. present a method that work even if the ontology is not known at training time [19].

### 5.4 Component: Response Generation

**Responsibilities:** Decide which types of text fragments to include in the response and in which order, request the individual text fragments from the text generation components, build the response to the user by concatenating the text fragments.

(1) Response Assembly: Decide which types of text fragments to include in the response and in which order, request the individual text fragments from the text generation components, build the response to the user by concatenating the text fragments

(2) Text Fragment Generation: Generate the natural language response of a specific type. The response types are usually specific for the intent at hand. However, there are response types that can be used across intents. Examples include: Disambiguation / clarification questions, error messages and educational messages (which suggest additional features to the user).

(3) Text Variation Generation: This module ensures that the texts vary based on the situation and the course of the conversation to enable a dynamic conversation. It avoids that always the same answers follows to the same questions.

(4) Education: Helps the user to learn how to use the agent from agent itself and improve his experience with the agent.

(5) Personal Recommendations: Through entertainment history and usage of the agent, the agent learns more about the user and can include this information in the answer. E.g. in the form of interesting facts.

**Approaches:** Decision which types of text fragments to include can be made rule-based. Decision criteria are the intent and the context information (esp. the conversational state). The component may query the text generation components upfront to figure out, if a new text fragment of a specific type is available.
Text generation can be done rule-based by filling in data from a structured data source into text templates for individual sentences and concatenating the sentences. Traditional language generation methods are based on pipelines, such as the well-known standard Architecture six Component Pipeline, which was originally proposed by Reiter [20] and has been further developed by others. This includes the following stand-alone components: (1) Content Determination [21] (2) Document Structuring [22] (3) Lexicalization [23] (4) Referring expression generation [24] (5) Sentence aggregation [25] (6) Linguistic realization [20] for this module exists different flavors: Hand-coded grammar-based systems, Templates and Statistical Approaches [5] New approaches are based on deep learning. Santhanam et al. divide these into four categories [5] (1) Language Models [26] (2) Encoder-Decoder Archiecture [27] (3) Memory Networks [28] (4) Transformer Models [29]. Lowe et al. present a system for high engaging dialog generation [30].

## 6 Conclusion

In this contribution we propose a reference model for the dialog management component of a conversational agents which addresses high-engagement use cases.
The reference model may serve as a basis for multiple tasks, especially: system design (as a starting point to design both individual agents and agent creation platforms), system evaluation (as a structure to evaluate and compare agent creation platforms), and research (as a framework to structure future research projects and to put individual research contributions in context).

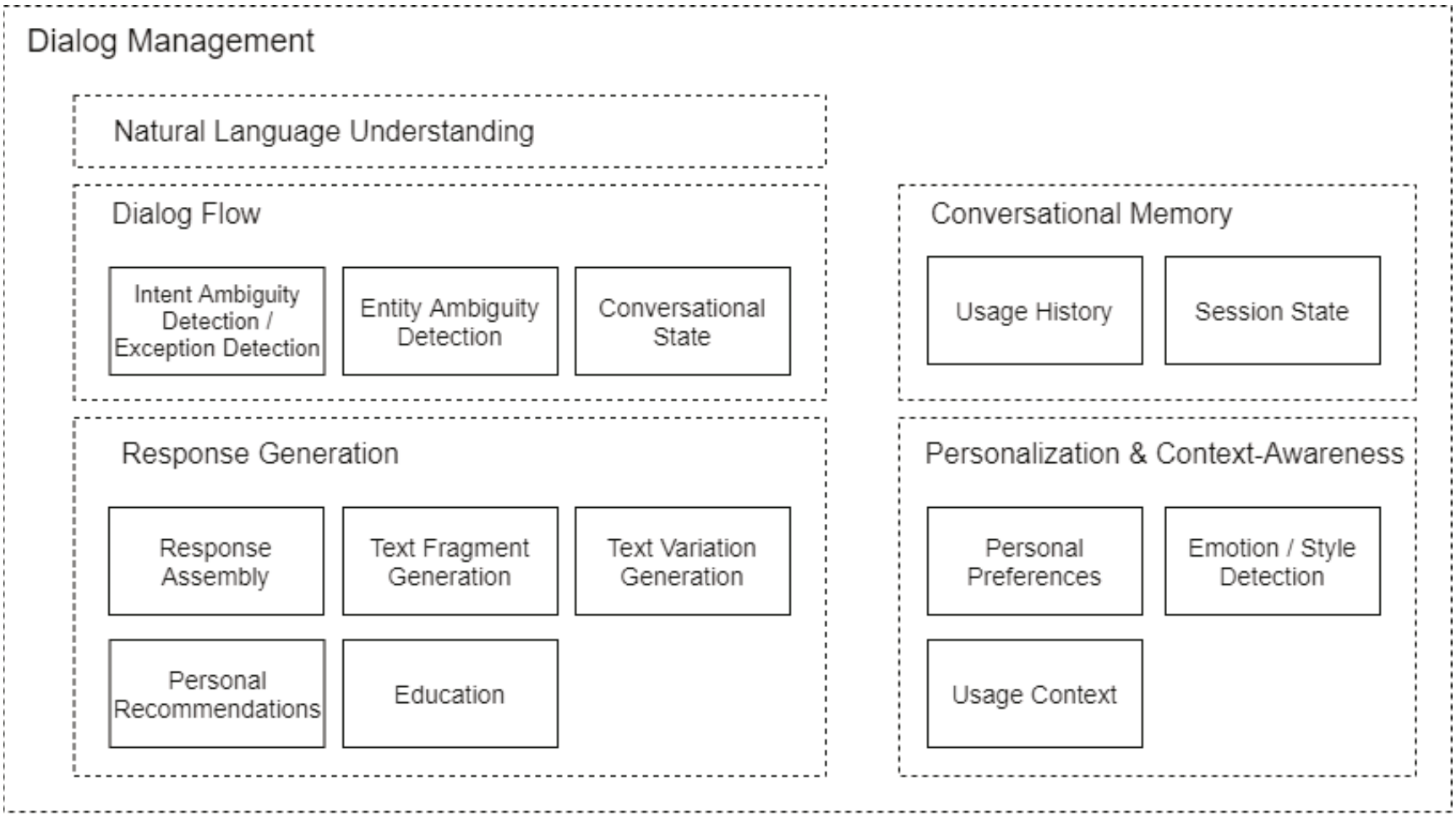

**Fig. 2.** Dialog Management Refernce Model (Conceptual Sub-components)

# Verify Embedded Systems Faster and more Efficiently with Artificial Intelligence

Alexander Schwarz[1], Björn Morgenthaler[2], Victor Vaquero Martinez[3], Miguel Garrido García[4], Manuel Duque-Antón[5]

[1] comlet Verteilte Systeme GmbH
**alexander.schwarz@comlet.de**
[2] comlet Verteilte Systeme GmbH
**bjoern.morgenthaler@comlet.de**
[3] comlet Verteilte Systeme GmbH
**victorvaquero.etereot@gmail.com**
[4] comlet Verteilte Systeme GmbH
**miguel.garrido@comlet.de**
[5] comlet Verteilte Systeme GmbH
**manuel.duque-anton@comlet.de**

**Abstract.** Embedded systems are the basis for many electronic devices. As a combination of hardware and software designed for specific purposes, Embedded Systems ensure the functionality of Connected Cars, Autonomous Driving, Smart Farming, Industrial Internet of Things and Smart Homes. The enormous competitive pressure forces manufacturers to significantly shorten their time to market and thus reduces the corresponding production cycles. This challenge is directed to the same extent to quality assurance. Due to the constantly growing number of (regression) tests, it is no longer practicable to perform all verifications in all development phases up to the finished product: each quality feature is planned and configured individually. But this approach is usually carried out manually with a lot of effort and is rather rarely adapted over time. On the other hand, software changes very quickly: new features are added, new dependencies arise or are resolved. Communication between individual components change. The probability that errors are found by tests (too) late is substantially increased with each change.

This paper presents an approach that successfully mitigates this challenge with the help of suitable Artificial Intelligence methods. To reduce lead time, a mechanism is developed that reduces the number of required tests. Based on the data from previous verifications, a (significantly) smaller subset of tests, which is sufficient to verify the correctness of the change, is selected. The remaining probability that tests necessary for negative verification of the software are not considered, is thereby accepted. Initial results, using data from several open-source projects as well as the use of a prototype machine learning pipeline, show promising results with respect to their predictive capabilities.

## 1 Introduction

As a technology partner and solution provider, comlet connects embedded systems of both global manufacturers (OEMs) and medium-sized companies into an overall system and offers intuitive usability by Artificial Intelligence (AI) while generating increased value. For example, predictive maintenance can be offered in the Industrial Internet of Things and new security risks can be detected in digital networking with the help of intelligent anomaly detection. To assure quality, comlet can also reduce the complexity of necessary tests with the help of AI while maintaining reliability.

The rise of monolithic databases, agile methodologies doing fast integration and multimillion source code line projects have created a practical problem in terms of computing power. What previous methodologies could do with branch manual testing and nightly runs is impossible or too time consuming to do for Continuous Integration (CI) teams.

## 2 Problem

Approaches have been established that take the introduced challenge into account: one of which being Continuous Testing (CT) as the integration of automated tests into a CI pipeline (see *Figure 1*), including the requirement of detecting errors as early as possible ("fail fast").

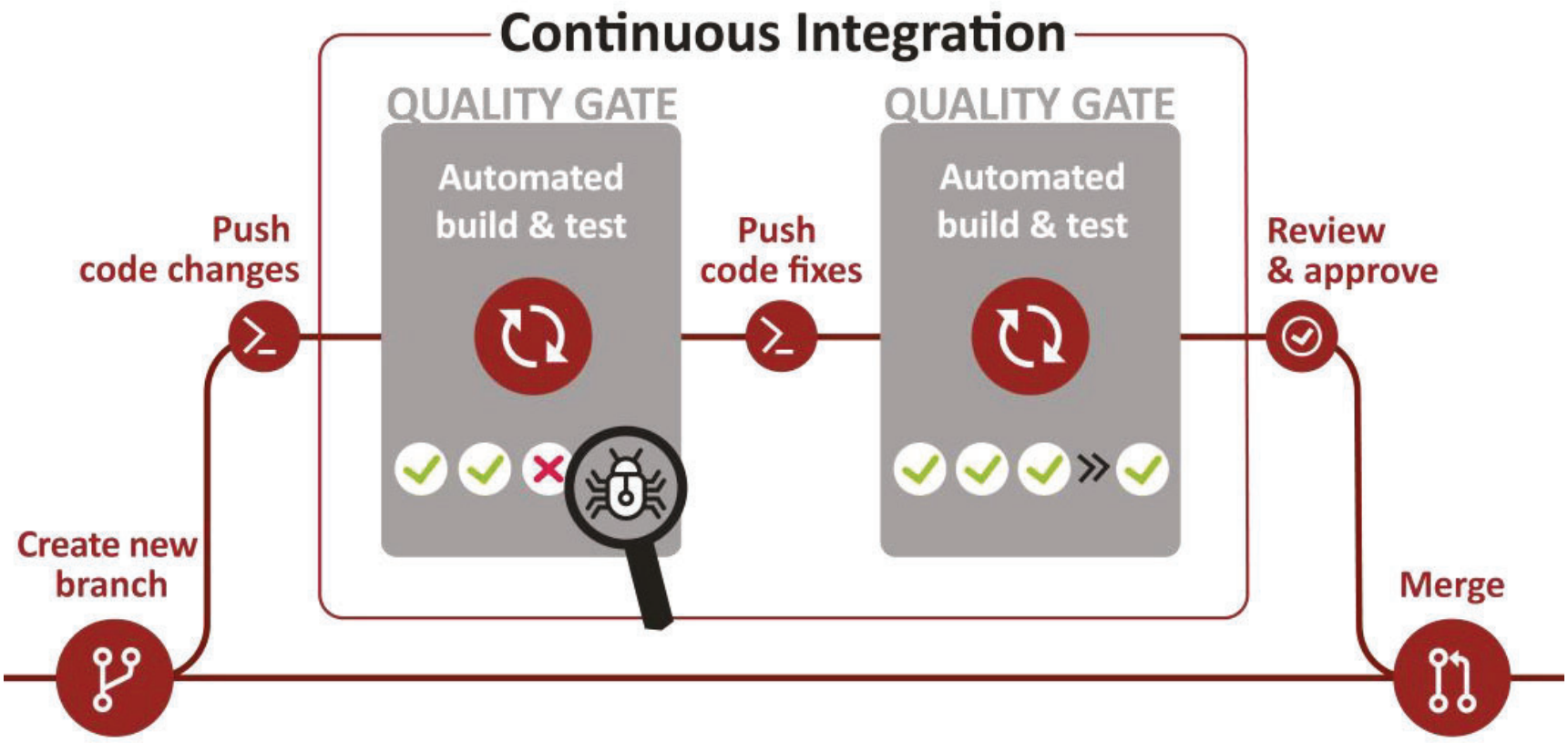

**Fig. 1.** Test automation as quality gate for a CI pipeline

In a CI pipeline, software is continuously developed in separate branches and eventually merged with the main development branch (trunk) once it has been completed. To assess the quality level before the individual branches are merged, various checks are performed at the so-called quality gates (QG).

Each QG is adequately planned at each integration step and configured with respect to the trade-off between duration, the risk of not detecting errors and feedback (see *Figure 2*).

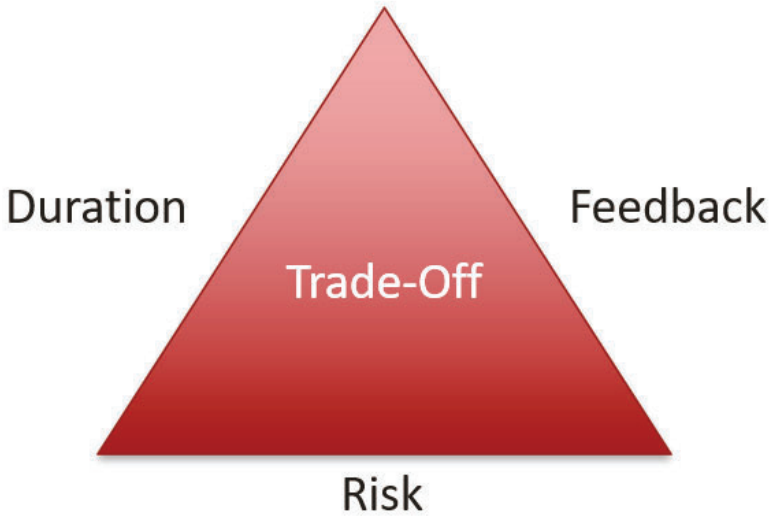

**Fig. 2.** Trade-Off while planning QGs

This process is usually carried out manually and must be adapted according to the changing requirements during development.

The developer also receives immediate feedback by the QG whether the changes contain errors and is obliged to eliminate these accordingly before integration.

As every step of development gets moved to the trunk, there is a higher need for quality control before and after the integration. The problem resides here, as projects with hundreds or thousands of commits and thousands or more of test suites cannot bear the computing power needed for running every test on every new code change – even with CT.

## 3 State of the art

To manage the previously mentioned problems, there have been multiple solutions: Google [1], for example, uses a simplification by accumulation. First, they use a bigger aggregation of tests named test targets. Secondly, they batch multiple code changes until they reach a specific threshold, e.g., 100 commits. At that moment the new code version, with the complete recollection of code changes, is run as a single unit on every test target (needed) that has not been run since the latest quality control.

This solution is based on a couple of heuristics and approximations but at its core it runs on something named change-based testing. The main property of this scheme is to only run the tests that somehow (the means change between methods) depend on the code changes, and as such are likely to fail, to reduce the resources needed. For this, useful features like historical failure data or code dependencies could be used. It can be done by any manual or automatic heuristic, but any non-automatic means bear a problem of scalability.

As of now there have been quite a few ad-hoc attempts at resolving the change-based testing task, and specifically, selecting an optimal subset of tests. Other companies commonly do it with specialized automatic solutions for some business or project - sometimes in conjunction with manual work [2] [3] [4]. There have also been some attempts at solving this problem [5] [6] by using simple heuristics and rules like recent failures, similarity between test and code; and specially code coverage. Some try to develop a set of diversified tests, with coverage measures or input analysis to cover the whole code change through search methods like greedy or local beam search [7]. There's also some more general development in risk prediction through specific features like historic failure data, similarity measures, code changes [2], etc.

To the author's knowledge there are only a couple of public papers, from Google [1] and Facebook [8], trying a similar approach of using machine learning to resolve the problem of selecting an optimal test subset but, as mentioned before, only for their own companies in an ad-hoc fashion. Another paper [9] used a reinforcement learning method only with historical failure input data.

The tool "Sherlock" by OMICRON [10] [11] tries to blend automatic and manual test selection by means of code-test dependency by dynamic analysis but without taking care of the remaining information.

To the author's knowledge up until today there is no homogenized approach that analyzes the general problem for every possible available input.

## 4 Solution

To solve the above stated problem more universally, the development of a pipeline capable of predicting a minimal test subset through training on historical test failure data is approached (see *Figure 3*).

Main objective then is to create a data model to input this learning algorithm and identify the minimum features required, the cost-effectiveness of the features, the most suitable algorithms and finally the actual predictive capability of the learning model.

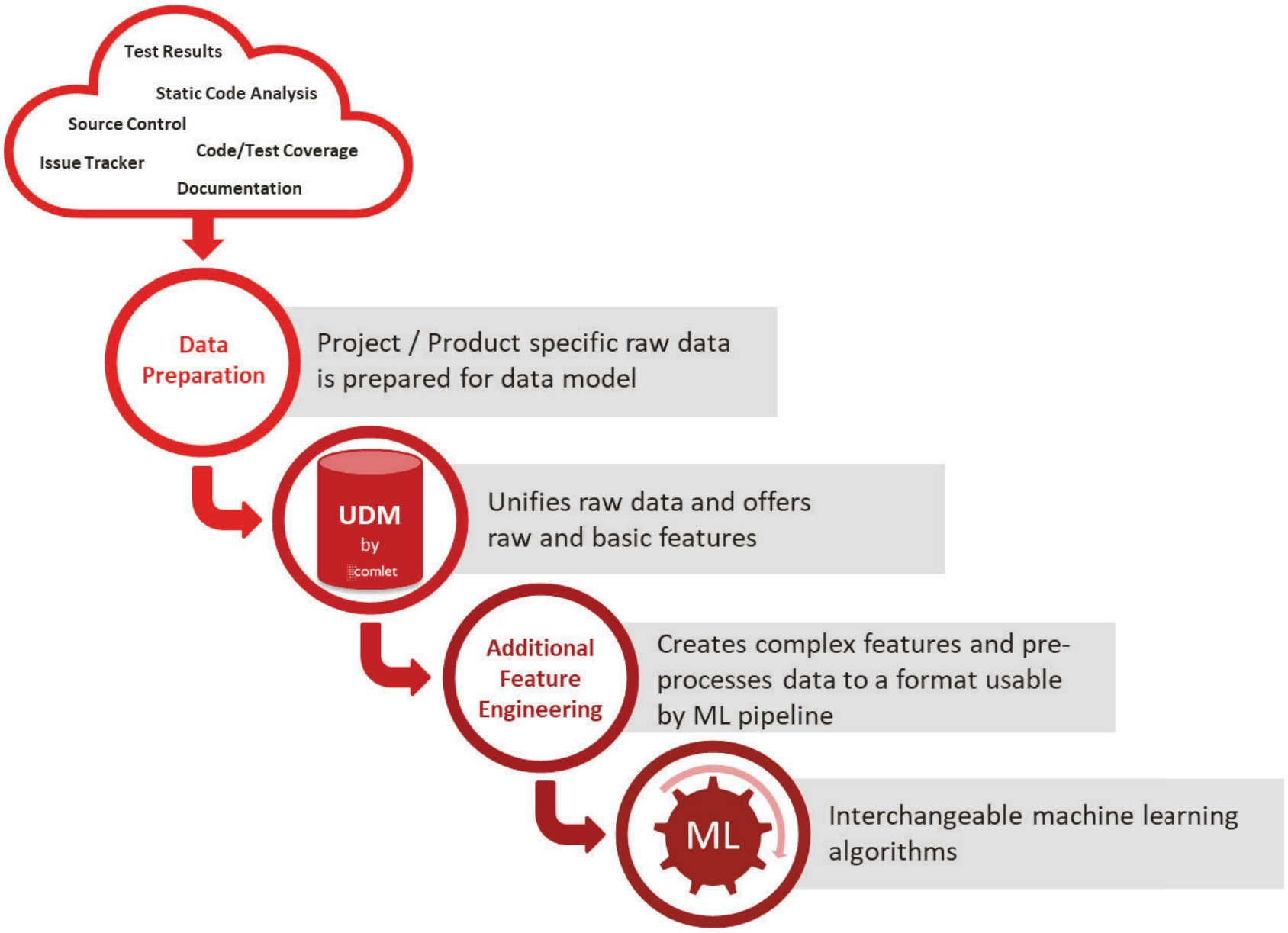

**Fig. 3.** Overview of proposed pipeline

The pipeline transfers all available data into a common, unified and abstract data model, the Unified Data Model (UDM). This data can come from various, very different sources, such as issue or bug trackers, source control management, documentation, (static) code analysis and automation servers.

Prior to transferring the data from all possible sources into the UDM, preprocessing and data preparation is required. The raw data presents itself in very diverse formats, for example all sort of databases such as SQL or No-SQL or file formats like JSON or CSV. Therefore, those (raw) data points are converted into a generalized data structure which is easy to work with. While this structure is created, it is also assured that the different values like dates or time are converted to common formats as well. Eventually, all data points can be classified into the following five categories:

- *Info Point*: Abstraction of any documentation about architecture, tasks, bugs, issues, pull requests

- *Code Block*: Abstraction of any possible aggregation of code as it is in the current moment in time. There are multiple possible types, but some would be modules, files, classes, functions, etc.
- *Code Change*: A basic data point that records every historic modification to the code base and serves as a version reference.
- *Test*: One of the basic data points comprised of the most updated information about the tests that are currently in use.
- *Execution*: Record of every time a *Test* has been run, keeping historical data about the successes or failures of every *Test* and *Code Change* to which every *Execution* is related to.

These five categories define the basis for the Unified Data Model which serves as an interface between source specific raw data and the unified feature engineering and machine learning algorithms, resp. models. Thus, it completely separates feature engineering and training of machine learning models from the specific (raw) data. The UDM itself provides raw and basic features out of the box that do not require further processing. Of course, a more thorough data analysis including data preprocessing and complex feature engineering based on the unified data is also performed to optimize the machine learning results.

Based on the data provided by the UDM, an attempt is now being made to reduce the time required to test software together with the time a developer waits for feedback whether the changes are causing test failures. This results in two different goals:

- *Subset selection*: Given a set of *Tests* and *Code Changes*, output the subset that finds all the failures. The preferred method to achieve this is to analyze every test individually and determine a value between 0 and 1 that reflects the probability of failing. With these probability values for all tests, it is possible to create an optimal subset.
- *Test prioritization*: Given a set of *Tests* and *Code Changes*, output the optimal order of the set. This optimality can be measured in several different ways, such as mean time or maximum wait time. The objective is to give feedback to the developer as soon as possible. It is also possible to use the same setup as in the subset selection algorithm simply by choosing those tests with a greater risk of failure first.

The specific approach as an initial step towards the general solution is to identify a minimal subset of tests that are very likely to fail, i.e. to fulfill the subset selection goal. No further prioritization of these tests will be done for the time being. For this purpose, data is collected from two large open-source projects, Pytorch [12] and Chromium [13]. This data is then merged into the UDM.

Since the UDM can manage different projects, it also ensures that non-existing features and missing values in the raw data are filled with meaningful values. This is necessary so that all further steps such as feature engineering or machine learning can be performed independently of the source project.

The UDM and the general approach of selecting a set of tests that are likely to fail are validated with a simple decision tree (DCT) algorithm. The DCT was chosen because it is easy to understand and a look at the internal structure is possible at any time. This gives more insight into the underlying data and features.

## 5 Results

The first step is defining performance metrics to compare the quality of different experiments objectively. After that, multiple experiments are conducted with different sets of features. From these experiments the first ones use only raw features which are directly available from the data and the next experiments use more complex engineered features.

### 5.1 Performance / Quality metrics

Predicting the outcome of a test is mainly a classification problem. Therefore, the data is labeled with two classes: the successful tests are labeled with "success" and the failing tests with "failed". Because the goal is to predict if a test fails, the failing tests are considered as positives. The following performance metrics are used to evaluate the quality of the trained models and the overall approach.

*Precision* is a measure for false positives. This means that a successful test is predicted as failed test. A low precision leads to a subset of tests with many tests that are successful.

$$Precision = \frac{True\ Positive}{True\ Positive + False\ Positive}$$

On the other hand, there is the *Recall* which is a measure for the false negatives. False negatives are failing tests that were predicted as successful, so a low recall will lead to the problem that many failing tests are not predicted as those.

$$Recall = \frac{True\ Positive}{True\ Positive + False\ Negative}$$

The overall A*ccuracy* is simply a measure for how many tests were correctly predicted.

$$Accuracy = \frac{True\ Positive + True\ Negative}{True\ Positive + True\ Negative + False\ Positive + False\ Negative}$$

The focus lies on improving the recall as much as possible without decreasing the overall accuracy and precision too much. This is due to the higher-level goal in predicting failing tests as early as possible. With a high recall fewer failing tests will be wrongly predicted.

### 5.2 Raw features

For the first prototype, it was decided not to work with individual test cases, but on the next higher hierarchical level: test suites. Test suites combine several tests that belong together thematically. This makes it much easier to deal with the huge amount of data in the first step. Another advantage is that it is easier to create a measure of complexity. In the case of test suites, complexity is the number of associated test cases. If individual test cases were considered, such a complexity measure could be, for example, the number of lines of code or even the number of function calls. This information is not directly available in the open-source projects considered and is very complicated to generate from the existing data. Another limitation chosen for the first prototype is to set the focus on whole commits and not on single files that changed. A commit can consist of changes in multiple files.

On this basis, initial experiments were conducted with a small subset of the collected data to better assess the underlying data. *Figure 4* shows the results of a decision tree when all directly usable features from the raw data are included. These are for example added, removed, changed lines of code, number of changed files, number of comments in an issue, number of reviews, number of tests inside a test suite, etc.

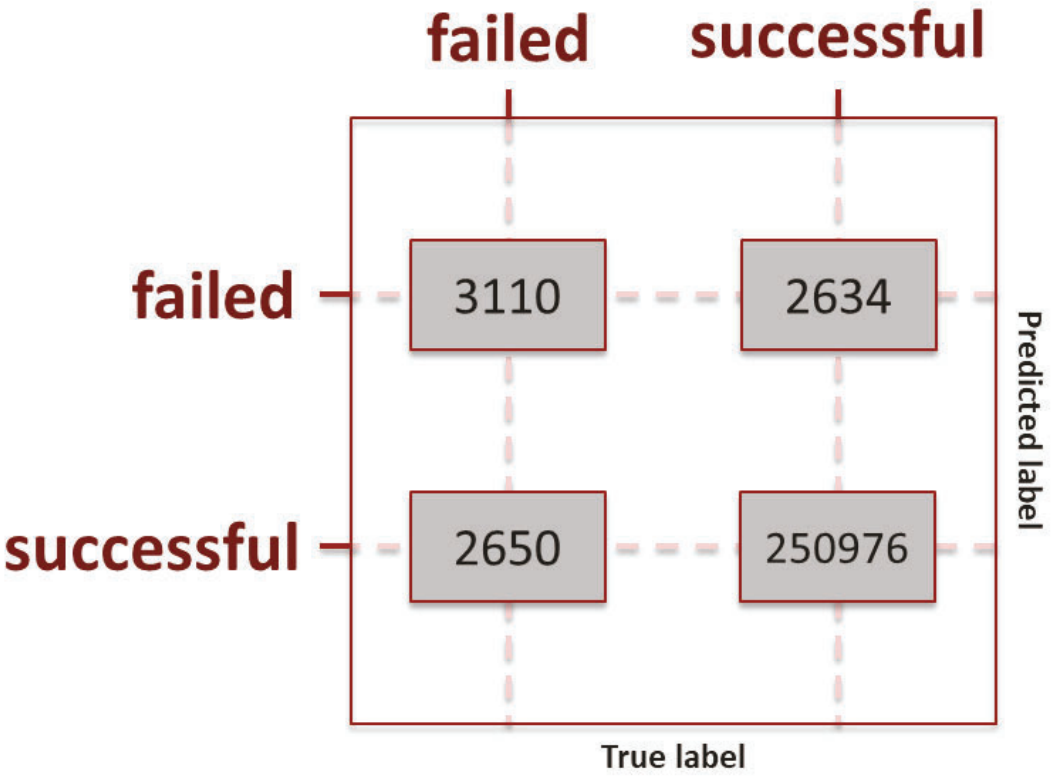

| Metric | Value |
|---|---|
| Precision | 0,5414 |
| Recall | 0,5399 |
| Accuracy | 0,9796 |

**Fig. 4.** Confusion matrix and metrics using only raw features

These results are from a 10-fold cross validation. The overall accuracy of the decision tree is close to 100%, as expected, because there is a strong imbalance in the data: there is much more data from successful tests than from failing ones. But despite this imbalance, already more than 50% of the failed tests are correctly identified as such, which is reflected by a recall of 53.99%. In principle, this shows that it is possible to predict the probability that a test will fail.

### 5.3 Additional and more complex features

Since it is not sufficient to predict only half of the failed tests, the recall must be increased even further. This is achieved by encoding the unused information in the raw data so that it can be used to train machine learning models and by engineering entirely new features from the existing data.

An example of such raw data that can still be usefully encoded is the build target, i.e. the underlying architecture or the operating system. In the raw data this is only present as a simple string containing all this information. A feature that was created completely new is the number of historical failures of a test; more precisely: how often a test failed in the last n runs.

With these new features, a recall of 66.4% is already achieved, as can be seen in *Figure 5*. Compared to the previous results, this value is already an improvement of 22.98%.

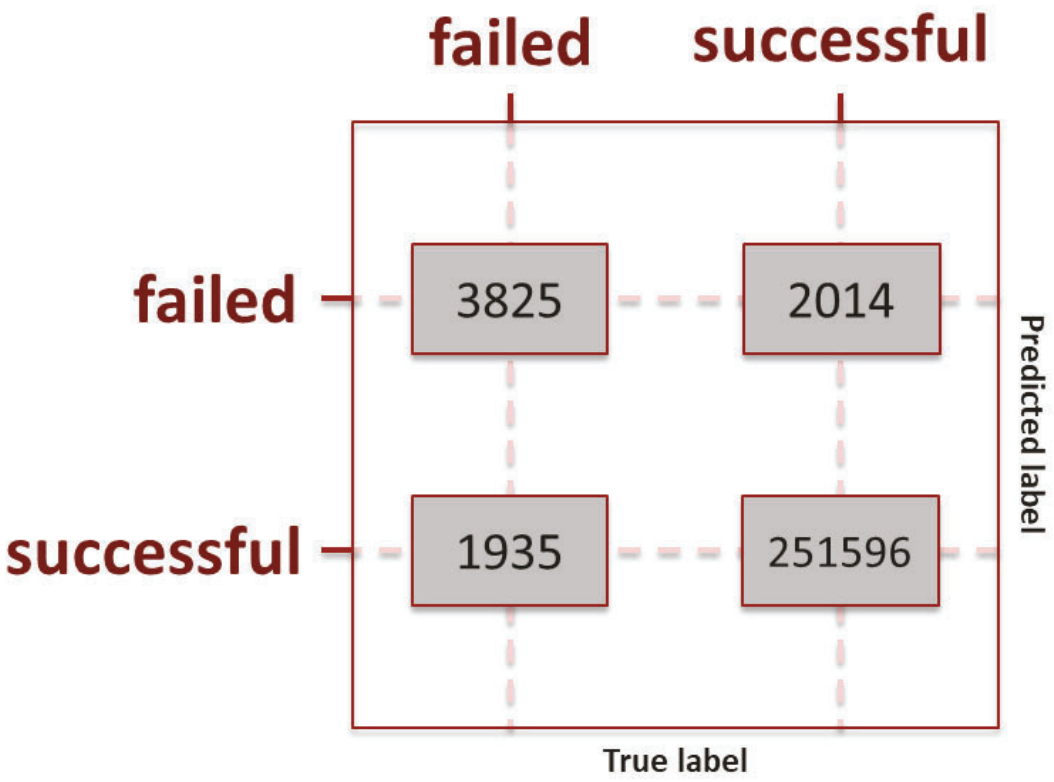

| Metric | Value |
|---|---|
| Precision | 0,6550 |
| Recall | 0,6640 |
| Accuracy | 0,9847 |

**Fig. 5.** Confusion matrix and metrics with additional features

It is expected that with further features the recall can be increased significantly without degrading the accuracy or precision. In the raw data, there is still a lot of information available, from which suitable features are yet to be generated through appropriate encoding and feature engineering.

## 6 Conclusion

The results from the experiments show that the introduced Unified Data Model works. The data from different software projects can be unified with it, which was examined at the example of two open-source projects. This makes it possible to carry out further data processing and feature engineering independently of the specific project. Also, the used machine learning algorithm can be exchanged now without adjustments. This makes it much easier to compare the performance of different algorithms and to select the best suitable one. The results also show that, despite the strong imbalance in the data in favor of successful tests, a recall of over 50% can be achieved even with a relatively small set of features. That is, over half of the tests that fail are correctly predicted. This recall could even be improved significantly with the first additionally engineered features.

The next steps are now to validate the Unified Data Model using a third open-source project as well as the first results of the machine learning model with significantly larger data sets. Also, the recall must be improved by further feature engineering and appropriate encoding of the still unused information in the raw data. Furthermore, it should also be investigated what results can be achieved if the granularity is changed away from test suites back to individual test cases. Likewise, it could improve the recall again clearly, if the focus lies no longer on the individual commits, but directly on the changed files. This would also give the opportunity to use some sort of code/test dependency if available as feature. It should also be possible to learn this dependency from historical test executions if it shouldn't be available.

# Suitability analysis of machine learning algorithms: Processing three-dimensional spatial data for automated robot control

Benjamin Peric, Michael Engler

Hochschule Furtwangen University, Faculty of Business Administration and Engineering, Robert-Gerwig-Platz 1, D-78120 Furtwangen, Germany

**benjamin.peric@hs-furtwangen.de**

**Abstract.** Global competition, rapidly rearranging market requirements and shorter product life cycles are expressed in constantly changing environmental conditions, which further complicate the demands on the production process. Given smaller batch sizes in small to medium-sized companies, the importance of flexibly varying handling tasks, which must be implemented through a robot gripping system, increases. Standardized workflows are difficult to establish in undefined environments since the products to be handled vary strongly in orientation and position.

The work aims to determine whether artificial intelligence can be developed through the combination of a color camera including an infrared depth measurement, which enables industrial robots to interact with the environment. The following two research questions arise: 1. to what extent can the potentials of artificial intelligence and its success of the recent period be adapted for the application of a robot gripping process and 2. how this symbiosis effects the use of industrial applications. The combination of intelligently controlled robotics using artificial intelligence and the processing of data without server-driven computing power at the end device form the basis of the investigation. The behavior of neural networks in scenarios with a small amount of data is the focus of the question. The realization of artificial intelligence is carried out in an iterative approach and the development process is available in written form.

The overall context of the approach is questioned via a suitability analysis to gain an understanding of possible applications and to name the limits of the system in the given scenario. With this approach, it can be examined which factors support the use of neural networks in the outlined context and whether they can be used successfully, despite of additional aggravating environmental influences.

## 1 Introduction

Self-learning computer programs are conquering economic structures as a sustainable branch of industry. The efficiency increase in all areas of a company indicates the potential of digital data processing. Machine learning algorithms can be identified as an essential driver for monitoring, regulation and control of industrial processes [1]. The degree of automatization is to be recognized as a fundamental prerequisite for the long-term safeguarding of competitiveness in manufacturing industries along with the key technology of artificial intelligence. The increase in flexibility, humanity, quality and productivity lies at the core of digitalization and Industry 4.0 in this branch of industry [2]. Autonomously controlled robot processes, digital quality assurance requirements, operational resource planning and preventive process analyses are mostly based on intelligent sensor technology [3].

Robotic applications are becoming increasingly challenging in the field of automation due to growing demands for flexibility. These requirements increase exponentially because of smaller batch sizes which creates an even more difficult starting situation in small to medium sizes companies. Common computer vision approaches could enable applications to deal with a dynamic scenery characterized by the opposing trend currents of automation and flexibility. Automated object manipulation, motion planning and even navigation through a dynamic scenery are the focus of these problems. Application areas within logistics, assembly and production include these problem issues [4].

In addition, the initial situation inhibits access to data collection and data processing. Large information structures often enable machine learning and are widely accepted as a prerequisite. Therefore, the impact of small data is the core of the following investigation.

Low computing power at the end device is another supplemental condition that is closely rooted in the initial scenario. Digital infrastructures cannot be established as a requirement in the manufacturing industry of small to medium-sized enterprises. Server-driven calculations are therefore not available as a solution option for answering the research questions posed [5].

## 2 Trend Research

Trend research is a methodological tool that attempts to identify the development process based on observations of technical and social changes. The newly acquired knowledge serves as an aid for the user to name individual trends, avoid surprises, assess interactions and enable interpretations through the collection of observations [6]. The scope is limited here to the analysis of already recognized trends and their origins. This approach aims to narrow down the relevant content of the suitability analysis by identifying the potentials and accurately define the core fields for the use case of the work. The high complexity inevitably requires the reduction of manageable factors.

### 2.1 Drivers of Artificial Intelligence

The applications of artificial intelligence have gained momentum in the last decade. The rapid increase in development can be attributed to the interaction of several factors, which can be seen in the following figure.

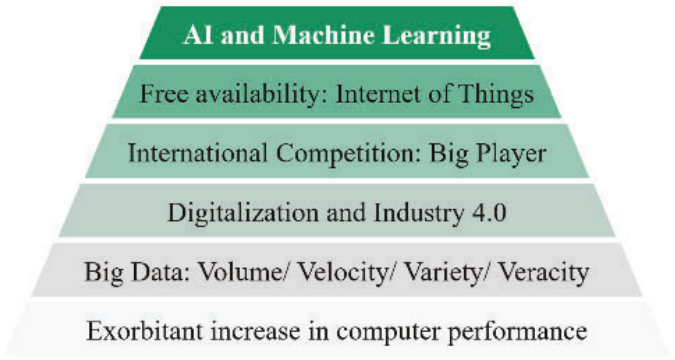

**Fig. 1.** Factors influencing the development of artificial intelligence

One of the strongest influencing factors is the massive increase in the computing power of intelligent circuits. Already in 1958, a regularity was derived by G. Moore through observations, in which the doubling of the performance can be continuously determined within a period of two years. The so-called Moore's law comprises more than 32 doubling cycles in the present period [7].

The applications of machine learning algorithms benefit especially when processing large amounts of data due to the exorbitant increase in computing power. The handling of large data structures can be described with the term Big Data. At the same time, it implies the possibility of being able to gain target-oriented insights by skillfully processing and analyzing mass data [8]. Symbiosis can be derived in which both the processing task and the medium to be processed will benefit from the constant growth [9].

In a business context, providing useful information, at the right time and in a usable form is critical to success. In addition to real physical data sets, so-called digital twins also serve to calculate and assess real scenarios. These furthermore support the creation process of well-founded data structures [10].

This core value creation process can be derived from the investments within the technology, which have increased fiftyfold to over 15 billion in the last ten years [11]. The leadership position is being fought by big players such as Google, Amazon, and Facebook on an international level. However, access to data sets is not only reserved for the technology giants. The German Federal Statistical Office estimates a tripling of the sensor market by 2025 [12]. The digitalization of industrial production and the high degree of connectivity favor the evaluation of the collected information. In the literature, terms such as "Internet of Things" or "Internet of Everything" are therefore already chosen to describe the development [13].

### 2.2 Alignment: Research and real-life applications

The following chapter offers a more specific insight into concrete fields of robot applications, which are optimized and implemented with the help of neural networks.

The subject area of robotics mostly covers technical apparatuses that usually interact with the physical world employing mechanical movements. Actuators are operated via kinematic variables by control systems, whereby a specific task can be performed automatically due to its structure [14].

One first milestone was published in October 2020 by Knapp AG in an interview. Through cooperation with Covariant Embodied Intelligence Inc., the transfer of research results to a real-life scenario has taken place. As an operational packaging robot, the system supports an electronics distributor for wholesale within logistics. Smaller goods are automatically transported to the robot arm via a conveyor belt. The gripper system recognizes the type, the lay, and the position of the objects. Transported via a pneumatic suction cup, each of these arrives at the desired packaging location. According to the company's information, this categorizes and recognizes almost 78,000 different small goods, whereby the products are initially unordered in a carton. A short-term maximum of approximately 600 objects per hour is stated. The system detects the products with an accuracy of more than 99% during operation. The gripping process appears very precise considering the error rate of less than one percent. Nevertheless, the manufacturers state a daily working time of just 14 hours [15].

The selected example impressively shows the different demands between research work and real-life solutions. An error rate of less than one percent is to be recognized as a groundbreaking and remarkable performance in the context of new research results. Within one of the most well-known international competitions, such as the ImageNet Challenge, a similar network could replace the previous titleholders [16]. However, these results represent the minimum level of acceptance in real-world use cases. The following short thought experiment illustrates the reason for this fact through an overall system effectiveness calculation.

With an average operating speed of approximately 400 objects per hour, with the assumed error rate of one percent, 96 objects are incorrectly detected during the day and therefore not processed. If the cause of this problem can be solved within five minutes, a machine downtime of approximately eight hours per day can be expected, assuming shift operation as the work design in this example. The downtime under consideration is only caused by a purely technical system error. Other recovery times, such as maintenance and servicing, occupancy, or manning times are disregarded in this consideration. Therefore, the specified working time of the "Pick-it-Easy" robot seems realistically evaluated.

## 2.3 Challenges of Artificial Intelligence in the manufacturing industry

In the following chapter, some decisive influencing factors are named which limit the use of a machine learning solution in a corporate context. Based on this very compact presentation, the limits of this technology will be critically examined and considered.

**Failure of drivers as amplifiers.** One of the obvious challenges in implementing meaningful AI-based robotics solutions is the absence of the drivers in business applications.

**Hardware requirements.** High computing capacities and memory requirements are a condition of practical applications. The necessity has an additional effect on the acquisition and operating costs of the system, which are particularly significant when a battery is used.

**Performance standards.** Whereas momentary successes of classification in worldwide competitions reach unprecedented accuracies, these results represent only the minimum of acceptance for real requirements.

**Specialist personnel and interface disciplines.** The pure basic knowledge of machine learning processes can only be implemented successfully in combination with specialist know-how. The generated results can only be validated based on a critical examination. The application of AI-based systems is to be defined as a highly interdisciplinary work process, which requires the combination of trained professional competencies [17].

**Flexible in use - rigid in applications.** Artificial intelligence can be embedded in almost any core activity of a company across industries. The integrity of the technology is highly dependent on each individual dataset used. The quality and quantity of training data ultimately determine the validity of the entire system. Changing environmental conditions limit the use of neural networks enormously. As soon as these changes are not reflected in the dataset used, this harms the capabilities of the whole system [18].

## 2.4 Experimental approach

The simulation of a simple gripping and joining process serves as the essence of the investigation. The analysis is based on a wooden game with small geometric wooden figures. The handling of unknown figures in a partly undefined environment is to be tested. The following investigation addresses common computer vision tasks, which allow object recognition. Besides the classification and detection, the localization and positioning of objects is the focus of this work. Six degrees of freedom are available to the target objects, which must be defined before grasping. In addition to three translational movements, the geometric bodies can rotate around all three spatial axes [19].

Applications within the sub-discipline of "Transfer-Learning" represent a variant to deal with very small amounts of data. This application area uses pre-trained networks, which are connected beforehand of the algorithms. The stored knowledge of the exorbitantly large networks transfers to the use case by interconnection and the probability of successful generalization increases in the ideal case [20].

The number of parameters in structures such as AlexNet, VGGNet, or ResNet reaches into the high millions. Due to the predefined application criteria, this discipline is left out in the following analysis. Consequently, approaches are chosen which hardly require any major computing power in operation. In addition to the redefinition of learning processes via a so-called Siamese network, compression via knowledge distillation, synthetic data generation, and data augmentation is the focus of the experiments.

A lightweight and very compact time of flight design with stereovision is offered by the infrared depth camera "Intel Realsense D435". The manufacturer of the product offers a comprehensive development platform compatible with free programming libraries and includes the most widely used programming languages [21]. Therefore, this depth sensor technology is the basis of the following work to create three-dimensional datasets and perception of the environment.

## 3 One-Shot-Learning: Rotation determination of unknown objects

Grasping objects are often located in a two-dimensional disordered initial situation. The localization of the target objects is only one part of the necessary scene determination. The rotation of an object cannot be determined with the help of the image segmentation illustrated above. If the components lie on a surface, it is necessary to extract the position and rotation of the objects, based on which the gripping and joining process can be derived.

It is assumed that the position of the target objects is already clearly determined. Consequently, the grasping objects are perceived in the zenith from the top view. Now the rotation of the object figures on the image plane must be determined. The rotation symmetry properties of the objects specify the maximum rotation on the image plane.

**Synthetic data preparation.** A single image is captured of each target object. Using image processing techniques, the subsequently visualized test figures are aligned and rotated with a rotation step size of half a degree. Consequently, each class receives a single original image to train. As a test set, five new images are taken of each object, which is augmented using the same methodology. The trainingset of the objects can be seen in the figure below.

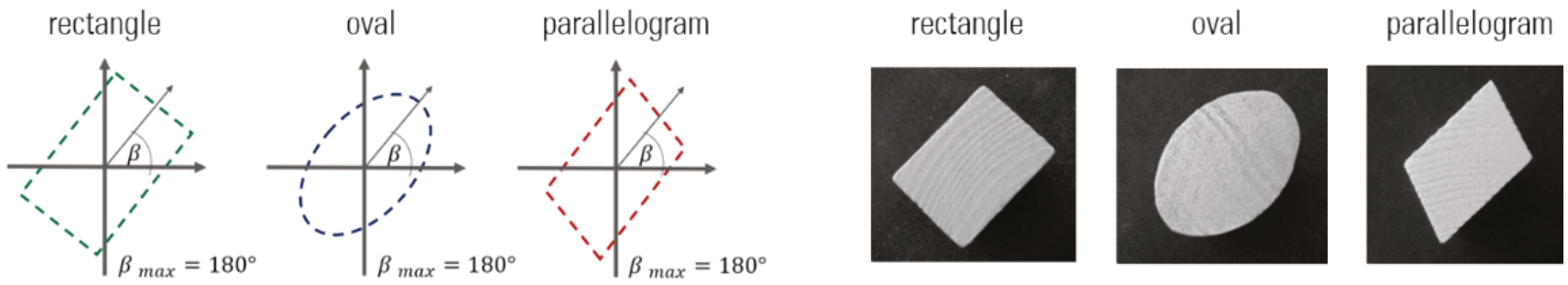

**Fig. 2.** Rotation determination of three geometric figures

**Problem definition.** The rotation is to be determined via a regression. Trigonometric relations represent periodic processes as mathematical elementary functions. Therefore, the use of a direction vector allows a regression with the support of oscillation functions. All rotationally symmetrical properties can be mapped via compression and extension of the oscillation functions. Here, the determination of the direction vectors is the focus of the neural regression. The neural network outputs the real vector elements in this approach.

**CNN for rotation determination.** Two convolutional layers with a kernel of size 3 x 3 scan the characteristics of the figures in 6 and 12 feature maps using relu activation functions. Via a dense layer, the data reach the output, which contains two neurons. The tangent hyperbolic function has the same range of values as the sine and cosine, so the network can determine the real vector elements using the mean squared error.

**Table 1**. Comparison: Vector and unit vector regression for rotation determination

| **Approach** | **Δ Mean** | **Δ Standard deviation** | **Δ Maximum** | **Intolerance < 2.5°** |
|---|---|---|---|---|
| Vector output | 4.28° | 2.49° | 21.46° | 54.91% |
| Unit vector output | 1.25° | 0.95° | 5.63° | 90.65% |

The validation shows that the regressive determination of the output over a simple vector fails. Only 55% of the test data fall within the set tolerance. The regression is therefore extended by the determination of the unit vector. The unit vector amount must always have the length of one. As soon as this condition is included in the output layer, it improves the rotation determination of the three target objects enormously, as shown in the table above.

## 4 Handling small data sets: Classification of unknown objects

The small geometric figures are to be classified and recognized based on their shapes. This enables the robot system to deal with unknown scenery.

**Data acquisition.** Four small target figures are measured as a test by the depth camera and stored in a data set. In total, this comprises 10,000 images per class. The captured images have a size of 100 x 100 pixels. The subsequent neural networks each receive a greatly reduced data number of 500 images per class to approximate the problem. The remaining data points are available to the test as validation. The data set represents the point clouds of the target objects from all viewing directions and varies strongly in the distance to the target objects.

**The problem of small data.** The phenomenon of overfitting occurs especially with small data sets, where the networks over-specify on the existing data points. Overfitting massively counteracts the primary goal of generalization, which is why neural networks can respond poorly to new input in this case [22].

**Augmentation.** Synthetic augmentation provides an effective method for extracting meaningful information structures despite having few representative data points [23].

The operations to augment the set of data points can be implemented using an integrated development tool called "Keras-Experimental" within the TensorFlow programming library and prepended to the architecture of the network.

The following methods are randomly selected and executed for each input.

- Zooming in and out of the original image up to a maximum of 20% of the existing image dimension
- Mirroring around the horizontal and vertical axis
- Average intensity contrast change by a maximum of 10% of the original image
- Shifting the image by a maximum of 10% in the horizontal or vertical direction

**Structure of Siamese Neural Network.** The Siamese Neural Network (SNN) is an architecture that enables the handling of very small amounts of data. It constantly receives temporally staggered impulses, which are to be linked together. The two input images are processed in parallel by the same convolutional neural network (CNN) architecture [24]. The SNN receives input pairs of data images with a size of 100 x 100 pixels. Two convolutional layers including max-pooling, alongside 24 and 48 feature maps respectively, process the data each with a kernel of 3 x 3 weights. Using a dropout of 50%, the data is transformed into the dimension of a vector with 50 elements as a result of a dense layer. The parallel processing of the input images enables the SNN to compare the vectors over the Euclidean distance of the intermediate output $\vec{v}_1$ and $\vec{v}_2$ , which represent the similarity of the input pair in an abstract form. This scalar value is calculated using a sigmoid activation function.

$$\phi(\vec{v}_1, \vec{v}_2) = \phi(|\vec{v}_1 - \vec{v}_2|) = 1/(1 + e^{-|\vec{v}_1 - \vec{v}_2|}) \quad (1)$$

The more similar the input images are, the smaller is the Euclidean distance of the vectors and the output approaches zero. Hereby the similarity of the three-dimensional shapes of the objects can be measured. A simple comparison image set of the figures to be compared is sufficient for classification since the closest match of the unknown input to the comparison set can be used as a discriminator.

To validate this architecture, almost identical CNNs are used, which map the number of classes to be distinguished in the output layer. With the help of a softmax activation function, the class probability distributions of the target object can be directly specified. The SNN uses the binary cross-entropy due to the binary similarity output structure, whereas the simple CNNs use the pure cross-entropy loss.

**Table 2**. Comparison: CNN with augmentation vs. Siamese Neural Network

| Network | Epochs | Train: 500 images | Test: 9500 images |
|---|---|---|---|
| CNN without augmentation | 100 | 99.17 % | 66.24 % |
| CNN with augmentation | 1000 | 91.17 % | 81.87 % |
| SNN with augmentation | 500 | 99.01 % | 90.21 % |

The very small datasets cannot be successfully distinguished without augmentation. The classification task of only four targets cannot be performed by simple CNNs for this small dataset. The augmentation methods allow the approximation of the problem, which can counteract the phenomenon of overfitting. The SNN outperforms the results by more than 10%, despite a nearly equal architecture of the model. However, a tenfold calculation time of an epoch must be accepted by doubling the input data. The validation of the SNNs is repeatedly done by pairing the test data. The achieved performance of more than 90% is to be considered as very good in this context since the use of several comparison images per class can additionally raise the result.

## 5 U-Net Compression with Knowledge Distillation: Image Segmentation

The three-dimensional scenery only acquires a complete meaning through the understanding within the pixel level. In addition to the classification, the object recognition and extrapolation of the object surfaces is therefore the focus of the next task. The recognition of a reference surface enables the robot system to form a normal, which can be used to determine gripping points. Object recognition is one of the central issues of a gripping process, in addition to grip evaluation, behavior coordination, and the determination of minimum holding forces [25].

**Data acquisition.** The question reflects a binary image segmentation of the point clouds. The generation of the target masks is typically done by hand, which is why the data preparation itself takes a very time-consuming process. However, the Intel-Realsense sensor technology offers some synergy effects. The recorded point clouds can be overlaid with a color image, whereby both data streams reproduce the same scenery. With the help of a simple color filter, the color channels can be used to distinguish the body surfaces. The top of the searched target figure is cut out from the rest of the environment in the color stream, allowing the corresponding mask to be formed to the depth data. The captured scene contains eight different geometric figures. The objects are set in motion on a tabletop, creating a dynamic environment. Different viewing directions and distances are again included in the dataset. The point clouds and the mask images assume a dimension of 200 x 200 pixels and contain 7,500 data points each. The test-training split assumes the ratio of 30 - 70%.

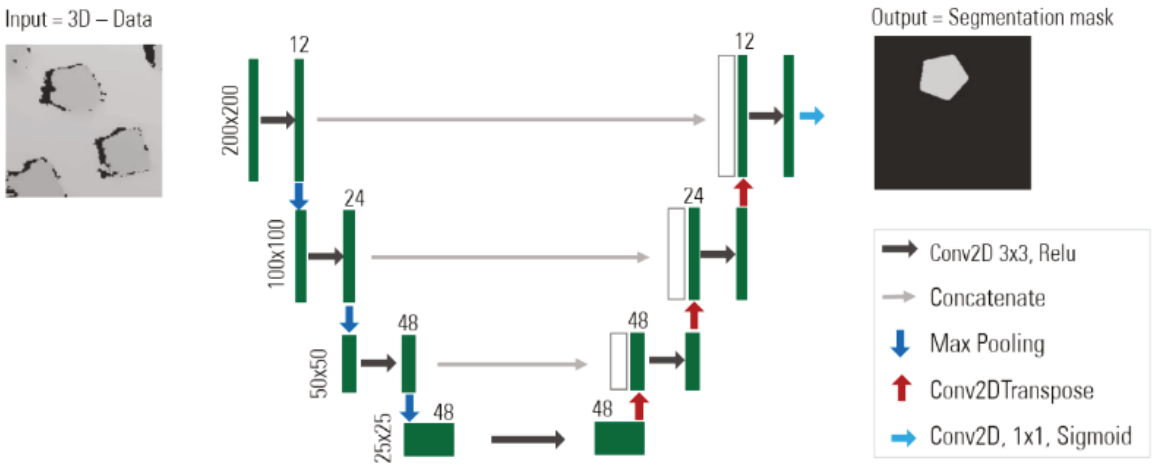

**Fig. 3.** U-NET Architecture for binary image segmentation

**U-Net architecture.** The original U-Net structure is used for binary image segmentation and reduced in size for the present application as shown in the figure above [26]. Via convolutional layers, the input data is compressed and brought back to the original size. During deconvolution, an additional dropout of 10% is integrated after each max-pooling layer, which is not visible in the figure above.

**Knowledge Distillation.** Considering a binary pixel classification, a sigmoid activation function is present in the output layer. The pure U-Net architecture outputs the pixel-probability mask $q$, which is compared to the target masks $y$ via the binary cross-entropy.

$$D_{BCE}(q, y) = -\sum_{(i,j)} [\, y_{ij} \log(q_{ij}) + (1 - q_{ij}) * \log(1 - q_{ij}) \,] \tag{2}$$

The architecture is trained using the existing data sets and their target masks. This Teacher Network is then used to develop a compressed architecture using Knowledge Distillation. The compressed U-Net structure receives as knowledge distillation loss the pure binary cross-entropy in combination with the Kullback-Leibler divergence [27].

$$D_{KD}(q_s, q_t, y) = \alpha D_{BCE}(y_t, q) + (1-\alpha) * D_{KL}(q_t^* \| q_s^*) \quad (3)$$

The products of the teacher and student architecture are aligned via the Kullback-Leibler divergence. However, even softer probability distributions $q^*$ of the student and teacher outputs are used. The divergence considers the temperature $\tau$ during the final computation with the activation function $q^* = \phi(x/\tau)$. The binary cross-entropy falls with the value $\alpha = 0.1$ to 10% in the combined loss of the student. The temperature $\tau$ receives the value 3. Hereby, the student not only receives the hard mask targets but also the corresponding soft probability distributions of the larger teacher network to approximate the problem.

**Table 3**. Comparison: Compression of U-Net Architectures with Knowledge Distillation

| U-Net Network | Parameter | Accuracy | Loss | IoU of the target class |
|---|---|---|---|---|
| Teacher | 80,485 | 99.02 % | 0.0257 | 85.72 % |
| Student from scratch | 7,753 | 97.63 % | 0.0592 | 67.54 % |
| Student with KD | 7,753 | 98.74 % | 0.0323 | 81.27 % |

Given the number of parameters that can be trained, the student U-Net architecture represents a tenfold reduction of the teacher, while the basic architecture remains the same. Accuracy does not correctly represent the binary image segmentation problem. Therefore, the Intersection over Union (IoU) is used for validation. The direct comparison of the student architectures with and without knowledge distillation shows optimization of the IoU of 14 percentage points. The tenfold compression therefore takes place with a drop of less than 4 percentage points.

# 6 Conclusion

The development of neural networks on the end device extremely complicates the development of neural networks. The experiments address extreme cases of an application under the chosen conditions. It was shown that some approaches facilitate the use of small data. Considering real application scenarios, however, the gap between the research results and real requirements becomes strongly apparent.

# VizNN: Visual Data Augmentation with Convolutional Neural Networks for Cybersecurity Investigation

Amélie Raymond[1], Baptiste Brument[1], and Pierre Parrend[2,3]

[1] Télécom-Physique – University of Strasbourg, France
amelie.raymond@etu.unistra.fr, baptiste.brument@etu.unistra.fr
[2] EPITA
pierre.parrend@epita.fr
[3] ICube laboratory
University of Strasbourg

**Abstract.** One of the key challenges of Security Operating Centers (SOCs) is to provide rich information to the security analyst to ease the investigation phase in front of a cyberattack. This requires the combination of supervision with detection capabilities. Supervision enables the security analysts to gain an overview on the security state of the information system under protection. Detection uses advanced algorithms to extract suspicious events from the huge amount of traces produced by the system. To enable coupling an efficient supervision with performance detection, the use of visualisation-based analysis is a appealing approach, which into the bargain provides an elegant solution for data augmentation and thus improved detection performance. We propose VizNN, a Convolutional Neural Networks for analysing trace features through their graphical representation. VizNN enables to gain a visual overview of the traces of interests, and Convolutional Neural Networks leverage a scalability capability. An evaluation of the proposed scheme is performed against reference classifiers for detecting attacks, XGBoost and Random Forests.

## 1 Introduction

Security Operating Centre (SOCs) continuously experience increasing amounts of supervision data, that require scalable processing capability coupled with fine-grained detection of known and unknown (zero-days) attacks. New solutions are thus required to back the investigation efforts of security analyst teams. The core requirements of these solutions are: explicability and traceability of alert to the original trace; investigation support through visualisation; scalability of detection.

The availability of scalable detection algorithms is thus key for building efficient SOCs capable of handling the data deluge. To this aim, neural networks such as Convolutional Neural Networks [1] are very competitive candidates, but still lack of maturity for rapid operational deployment. One of the key challenges for using neural networks in operation environment is the capability of performing suitable data augmentation in the analysis flow [2, 3]. Data augmentation can serve several goals: completing the data representation is areas where available information do not support satisfactory learning (scarce zones of valid data, or imbalanced dataset); altering the information to avoid overfitting and anticipate random modifications of the observed behaviours; changing the data format, for instance to apply algorithms with specific data input such as images to other data types such as text, sequences or qualitative features.

In this paper, we propose VizNN, a visual data augmentation model with Convolutional Neural Networks (CNN) for cybersecurity investigation. Its goal is to leverage scalability and detection capability of CNNs to the analysis of quantitative logs for detection of known attacks. Alternative detection solutions such as Random Forests [4] and XGBoost [5] also provide excellent results for medium size dataset. They are used as reference for benchmarking.

To evaluate the relevance and efficiency of the proposed approach, we apply it to the analysis of security properties of the DoH protocol [6]. DoH (DNS-Over-HTTPS) aims at encrypting DNS requests by encapsulating them in a classic HTTPS stream. Its goal is to solve the vulnerabilities of DNS (Domain Name Systems) which is used for each and every request on the Internet to bind human-readable domain names with machine-readable IP addresses. DNS experiences several majors weaknesses in its default configuration [7, 8], which urge the community to propose adequate countermeasures.

The evaluation of VizNN scheme is performed on CIRA-CIC-DoHBrw-2020[4] dataset [9], created by the Canadian Institute of Cybersecurity (CIC) and funded by the Canadian Internet Registration Authority (CIRA). Data is collected from the top 10,000 visited websites according to Alexa rankings. The raw data, in PCAP format, was processed and converted into CSV files using the DOHMeter tool.

The paper is organised as follows. Section 2 introduces the related works. Section 3 defines the data processing methodology, section 4 details the data preparation and exploration phase, whereas section 5 defines the propose data anaysis scheme. Section 6 evaluates the results. Section 7 concludes this work.

## 2 Related Work

### 2.1 Neural networks for cybersecurity

The advent of Deep Learning (DL) for the analysis of cybersecurity events takes its roots in the limitation of pre-existing Machine Learning (ML) algorithms. Machine Learning is used mainly for classification or regression, and keeps relying on feature engineering [10]. The expected advantages of Deep Learning is its capacity to perform well when data amount increases, and to provide very efficient test operation even though training phase is usually significantly longer. One of the key factors for the performance of Deep Learning is the numerous matrix operations it relies on, which let it be preferably executed over GPU.

Deep Learning comes in numerous flavours when applied to cybersecurity: Deep Neural Networks, which extracts hidden patterns from in the internal layers of the network, Recurrent Neural Networks such as LSTM (Long Short Term Memory) [11] which retain the memory of previous states, Convolutional Neural Networks [1, 12] to process data having a high degree of similarity, Restricted Boltzmann Machines for data generation or classification, Deep Belief Networks as brick of larger neural nets, Deep auto-encoders for trace de-noising or classification [13, 14].

[4] https://www.unb.ca/cic/datasets/dohbrw-2020.html

### 2.2 Data augmentation for neural network learning

Data augmentation is the process of artificially enriching real data with synthetic mock-ups to improve learning, in particular to remove learning bias [2]. One typical application of data augmentation is the densification of data zones where events are legit but scarce to avoid the generation of false negatives. The principle of data augmentation originally referred to statistical data enhancement like SMOTE [15], which is meant for handling imbalanced data. It increasingly refers now to the generation and enrichment of images to be analysed, in particular in the context of Convolutional Neural Networks. The objective is to leverage the capability of data volume scaling of deep learning approaches. It proves to be efficient for the detection of object landmarks such as invariance in shape, pose and illumination [16].

Image augmentation can be performed through basic image manipulations such as geometric transformations, flipping, colouring, cropping, noise injection, through geometric and photometric transformations such as kernel filters, mixing images, random erasing, or through deep-learning such as feature space augmentation, adversarial training, Generative Adversarial Networks or neural style transfer [2]. The combination of these operations, especially the simple ones, can be performed automatically to improve the validation accuracy like with AutoAugment tool [3]. The first image augmentation tool can be considered to be the auto-encoder of Hinton, which is applied for dimension reduction, noise reduction, data and image generation bases on multi-layer architecture with internal small-dimension layers [17]. More recent solutions focus on Generative Adversarial Networks (GAN) for data and image generation, which base on a pair of Deconvolutional/Convolutional Neural Networks [18]. This approach has made radical qualitative progress since its inception in 2014 [19].

### 2.3 Data augmentation for cybersecurity

Data augmentation for cybersecurity is used both for supervised and for unsupervised detection. It is used for supervised detection of malwares to address the variability of malicious code by adding noise in the training mode under Gaussian, Poisson or Laplace model. Then, a Convolutional Neural Networks performs the learning operation [20].

For unsupervised anomaly detection, the strategy is to oversample normal, rare data which usually causes most false positives. This approach is only applicable to well-defined data distribution if one does not want to inject excessive bias in the dataset. Lim proposes to augment not the input data, but a representative latent vector at the core of the neural network, through multivariate Gaussian sample generation [21]. It integrates an adversarial auto-encoder (AAE), which is an extension of GAN [22].

## 3 Data processing

Performed data analysis is performed in two steps: first, a generic DAP (Data Analysis Process), which is meant for reuse independently of the analysis methodology; then, VizNN, the visual data augmentation model we propose for Convolutional Neural Networks.

### 3.1 DAP - the Data Analysis Process

The Data Analysis Process entails following steps:

- Capture: the system behaviour is collected and gathered in a dedicated datalake for immediate analysis or later reference. The behaviour can be represented as actions (logs), system measures (scalar probes) or action measures (scalar measures derived from logs).
- Cleanup: incomplete or inconsistent data is removed, such as action or probe occurrences (data lines) with missing values or features (data columns) with non discriminating values.
- Standardisation: data is normalised or enhanced with metadata, in particular through FAIRification[5] (*i.e* to make it Findable; Accessible; Interoperable; Reusable).
- DE – Data Exploration: a manual scrutiny of the data is performed to highlight its particularities.
- DA – Data Analysis: automated statistical or Machine Learning algorithms are applied to the data in order to perform anomaly detection (in unknown rare events such as an abnormal access to a resource) or classification (discriminate known events such as a known attack).
- Visualisation: the data or extractions thereof is presented to the user.
- Investigation: the expert, here the security analyst, performs computer-assisted complementary inspection of the data to understand the output of the automated analysis and to identify behaviours not characterised by automation.

### 3.2 VizNN - Visual data augmentation for Convolutional Neural Network

The VizNN pipeline, shown in Figure 1, consists of five steps: data import and clean-up; selection of the features you want to keep; creation of images; preparation of images; training of the CNN model.

*Data import and cleanup* The import and cleaning steps are similar to those of a Machine Learning project.
After importing the data, missing values are completed or deleted. Categorical features are then converted to numerical ones. A correlation analysis may be relevant to remove unnecessary features. Finally, a "label" column is added to each row for classification.

*Selection of the features you want to keep* Rather than converting the entire dataset into images, a feature selection step is performed to only keep the columns that maximize performance.

Therefore, a basic XGBoost model is trained in order to retrieve the list of most important features on the model's gain.
Our proposal is to create images with the N most important features. To create images, this number cannot be a prime number since the width and the height have to be integers. The data is then filtered so that only the N most important features are kept.

*Creation of images* The filtered DataFrame is converted to a list of grayscale Image objects using PIL library[6].
Images then undergo data augmentation through resizing with bicubic interpolation to generate new pixels and therefore enlarge them.

[5] https://www.go-fair.org/fair-principles/fairification-process/
[6] https://pillow.readthedocs.io/en/stable/reference/Image.html

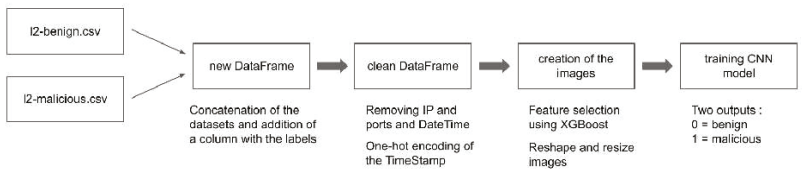

**Fig. 1.** Full pipeline process

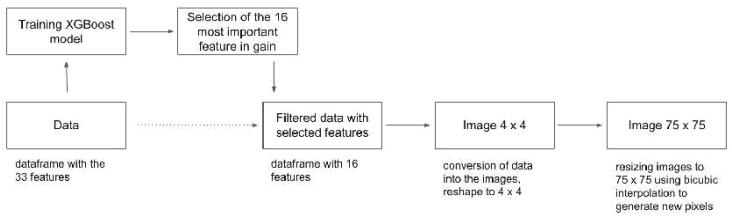

**Fig. 2.** Image creation process

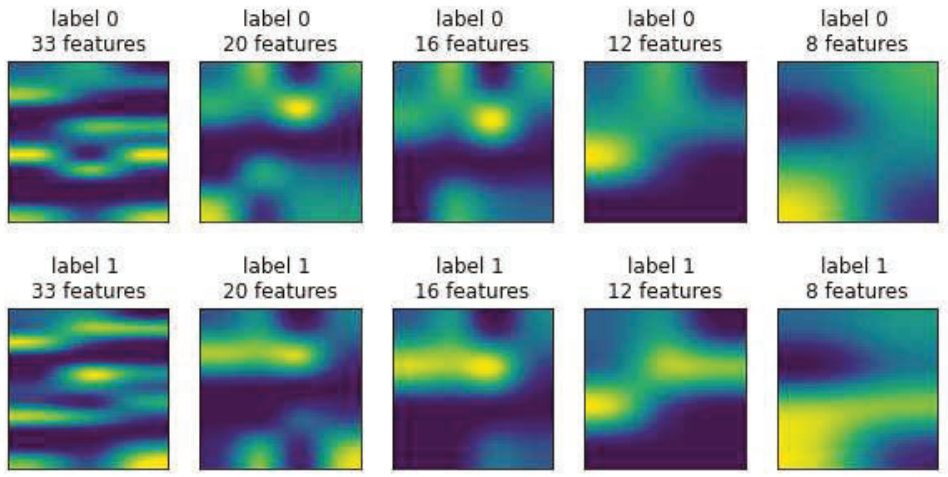

**Fig. 3.** Examples of augmentated feature images for VizNN

*Preparation of images* Several operations are performed on the images to transform them into CNN inputs.
First, the list of Image objects is converted to a NumPy array. Second, the data is split into a training and a testing dataset with the "StratifiedShuffleSplit" method. It shuffles and separates the data while respecting the proportion of each class. Thus, it ensures that even with unbalanced data, models are trained and tested on all classes with representative proportions.
Third, a dimension is added to the image so that it has three dimensions: width, height and number of channels(1 for gray-scale images and 3 for RGB images). Finally, images are normalized to have, for each pixel, values between 0 and 1.

*Training of the CNN model* The Convolutional Neural Networks is trained using the images generated by the reference dataset.

## 4 Data Preparation and Exploration

Data exploration is performed by data cleaning and subsequent scrutiny through highly efficient, through very widespread, learning algorithms: Random Forests and XGBoost.

### 4.1 The dataset

The CIRA-CIC-DoHBrw dataset entails two kind of traffic: HTTPS traffic without DOH and DOH traffic. In case of DOH traffic, attacks and normal traffic were distinguished. Thus, 4 subdatasets are considered: “non DOH”, “DOH”, “benign DOH” and “malicious DOH”. These datasets include 33 features that define the traffic.

This study focuses on the classification of DOH traffic (benign or malicious). The two corresponding CSV files are imported and combined into one DataFrame with an added column entitled "label". Label 0 corresponds to benign DOH traffic and label 1 to malicious DOH traffic.
The dataset is unbalanced: there are 67% of attacks and 23% of benign traffic. Several columns such as "sourceIP", "DestinationPort" were removed since they are artifically crafted for analysis stakes. Afterwards, the "TimeStamps" column is transformed into categorical data. In this way, four categories are used to describe the period of the day: "morning", "day", "evening" and "night".

### 4.2 Random Forests

Random Forests is a classification algorithm based on ensemble learning, which consists of using so-called "weak" models to make predictions and then to combine them into a larger model. It operates by creating a given number of decision trees. More specifically, an improved version of bagging is used in order to reduce the correlation between each tree. The samples of the training data are provided as an input for each tree. The latter then make their prediction. The results are aggregated using a majority rule which means that the final predicted class is the class that was predicted the most by the individual decision tree.

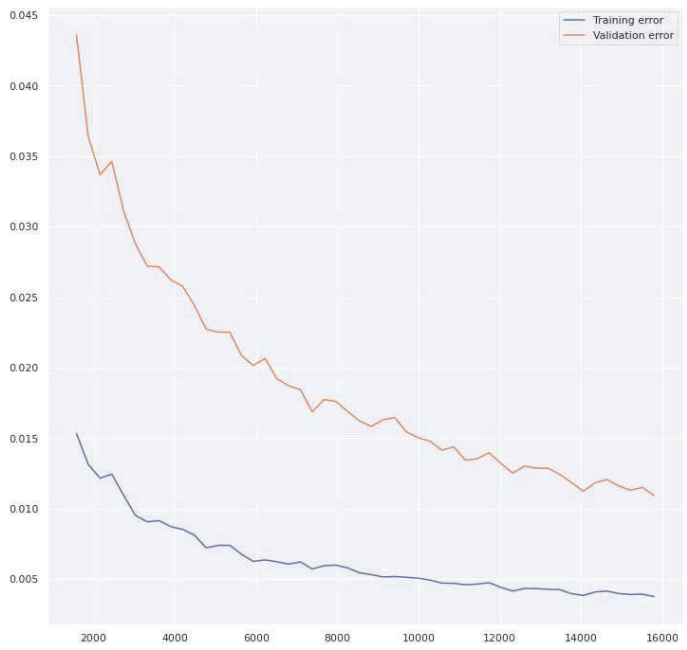

**Fig. 4.** The learning curves for Random Forests

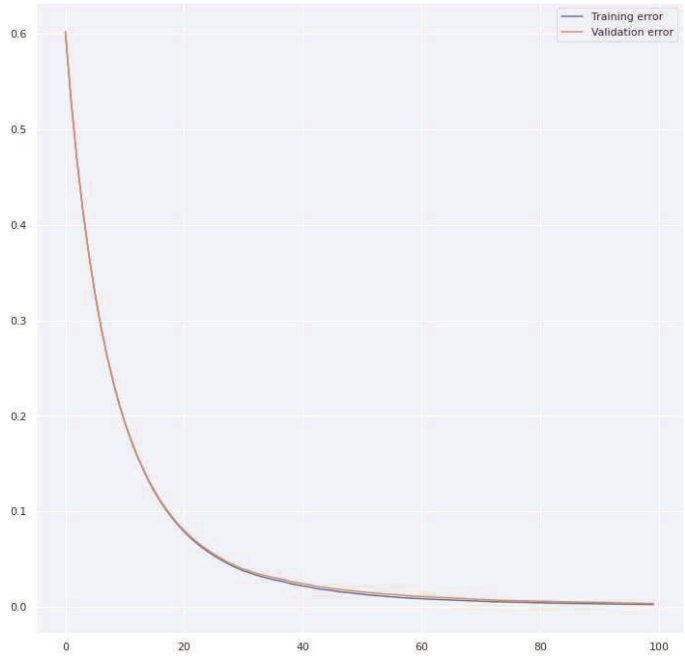

**Fig. 5.** The learning curves for XGBoost

Figure 4 shows the loss curve for Random Forests learning. Note that the ordinate spans from 0 to 0.045, which makes it a very close zoom. Training curve do not stabilise before 14,000 iterations.

The extraction of feature importance by gain, for the Random Forests algorithm, shows that the main features are statistics derived from the packet length: mode, mean, median and standard deviation. Note that the original dataset entails such aggregated data, and not directly raw communication traces.

### 4.3 XGBoost

XGBoost stands for Extreme Gradient Boosting [5]. This algorithm is also based on ensemble learning strategy. Gradient boosting is a subclass of boosting algorithms. In boosting, each sub-model is weighted. This allows greater flexibility as it gives the ability to give more importance to certain models in certain learning cases. Thus, this type of models is particularly effective when the training data is not balanced. The specificity of gradient boosting is that the contributions of the decision tree models to the final model are calculated from the optimization of the gradient descent. This technique gives very accurate results. XGBoost is considered one of the best algorithms in the current state of the art classification algorithms.

Figure 5 shows the loss curve for XGBoost learning.

The extraction of feature importance by gain, for the XGBoost algorithm, shows that the main features there are packet length mode, number of flow bytes received, duration, as well as packet length statistics.

## 5 DA – Data Analysis with VizNN

This section presents the implementation of VizNN and the search for optimal model parameters for the data preparation flow. The Convolutional Neural Networks takes a series of images representing the features to be analysed as input. It is built by a convolutional layer, maxpooling layer, renewed convolutional and maxpooling layers, two flatten and dense neuron layers, and a two-value output layer. RELU activation is used. Each experiment is performed with 50-folds cross-validation, a validation set consisting of 20% of the training data and over 20 epochs. Figure 6 shows this architecture.

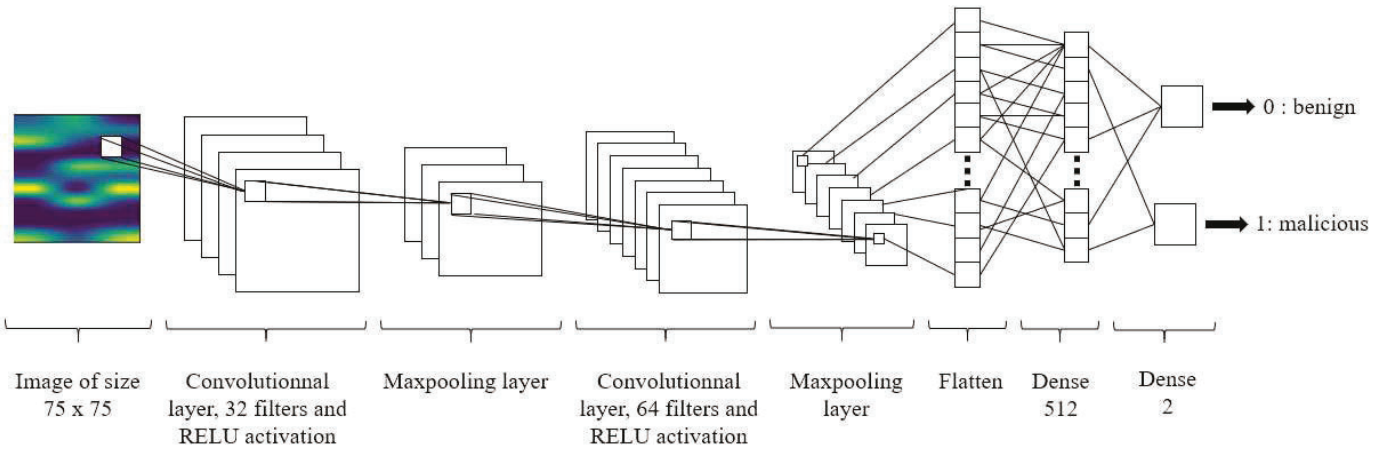

**Fig. 6.** VizNN architecture

*Influence of image size* To evaluate the impact of image size on the quality of the detection, we first evaluate the detection performance for square images with side ranging from 50 to 95 pixels. Images are created using all dataset features. Several models with the same architecture are then trained on these images, each model testing a different image size. The results indicate that image size has a significant influence on performance, as shown in Figure 7. According to the scores, $75 \times 75$ is the size that maximises the performance. Hence, the size is set as $75 \times 75$ for the rest of the study.

*Influence of the number of features* Another image-specific parameter that can influence the results is the number of features used to create images. Following the same procedure as for the image size search, several models are trained with images created from different number of features. For the dataset used for the experiment, keeping the 16 (out of 33) most important features from XGBoost's gain provides the best scores, as shown in Figure 8. Also, the model is under-trained when there are not enough features because the training images are not representative enough. On the contrary, if too many features are kept, the model is over-trained.
For the rest of the study, the number of features is thus set to 16.

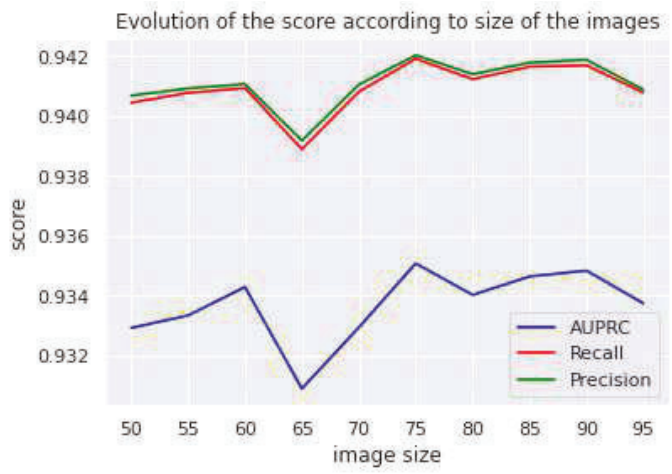

Fig. 7. Performance of VizNN according to the size of the images

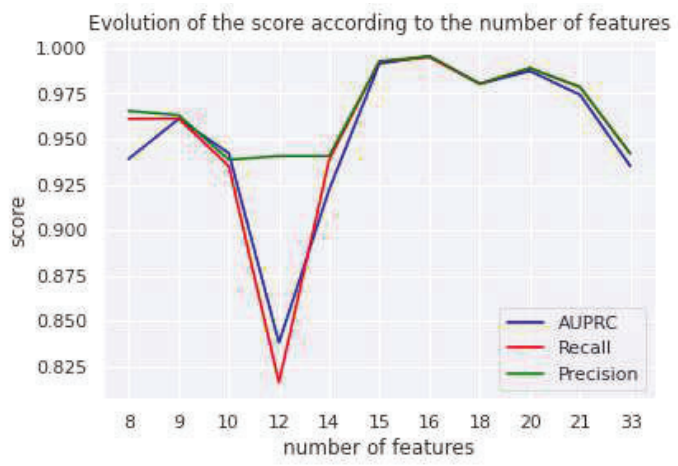

Fig. 8. Performance of VizNN according to the number of features

*Influence of the architecture* Once the image-related parameters are set, several CNN architectures are tested by changing the number of layers and filters. As shown on Figure 9, an architecture with 32 and 48 for the first and the second convolutional layer respectively provides the best detection capability for the dataset under consideration.

Figure 10 shows the loss for resulting architecture. Learning stabilised after 12 epochs for both training and validation.

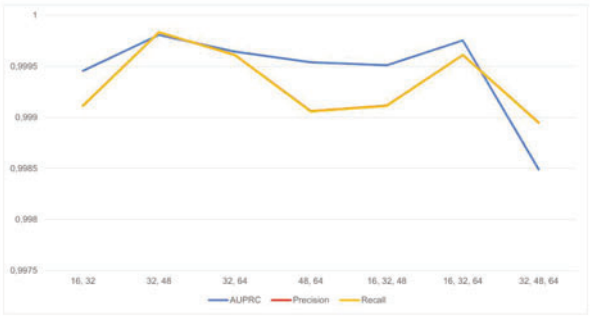

Fig. 9. Impact of CNN architecture on performance

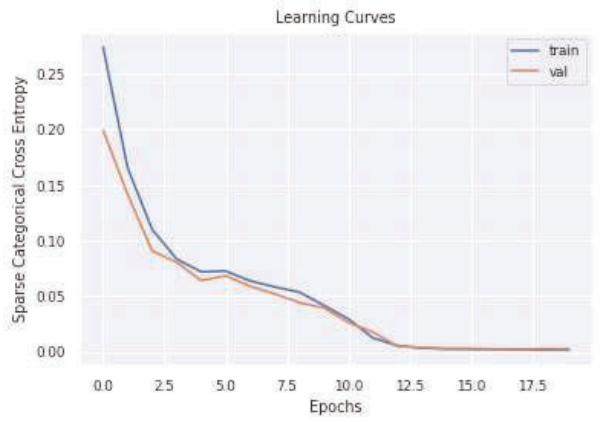

Fig. 10. The learning curves for VizNN Model

These evaluations for parameter settings show that each parameter has a specific optimum for optimising learning performance on the current dataset. It is highly likely that these values are dependant on the data, and can not be considered as general recommendations. Rather, the described process should be perform over again when the security context changes.

## 6 Evaluation

This section presents the overall results of the experiments on CIRA-CIC-DoHBrw-2020 dataset. Learning performance is evaluated using the AUPRC (Area under the Precision Recall Curve, suitable for imbalanced data), precision and recall metrics. Training and prediction time is evaluated for the VizNN and the two reference algorithms,

Random Forests and XGBoost. Each score corresponds to an average over a 50-folds cross-validation process.

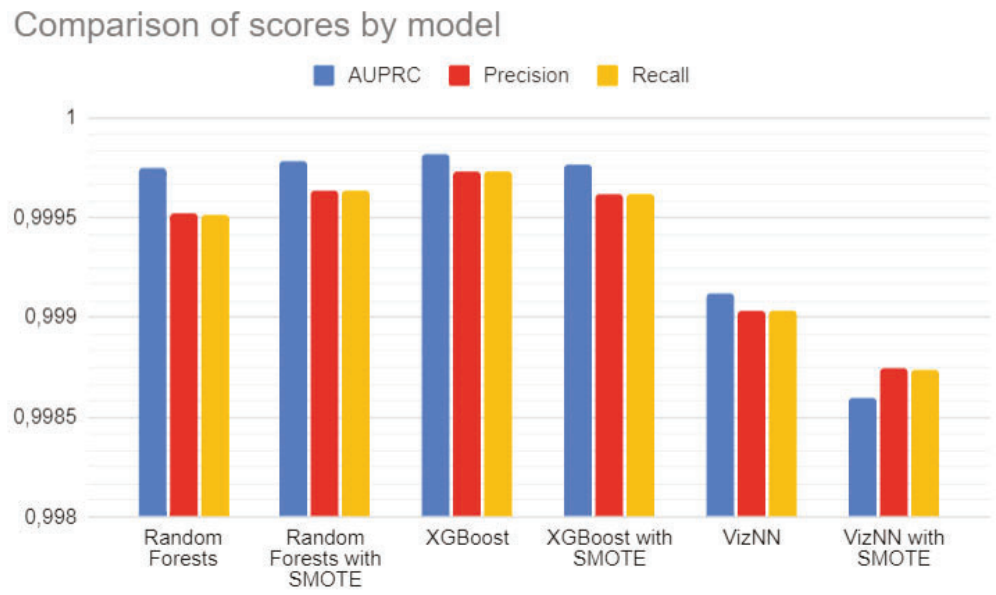

**Fig. 11.** Comparison of scores by model with and without SMOTE

Figure 11 shows the comparison of the different models: Random Forests, XGBoost and VizNN, without and with SMOTE (Synthetic Minority Over-Sampling TEchnique) oversampling strategy.

The first observation is that XGBoost outperforms Random Forests and VizNN for the three metrics AUPRC, precision and recall. Indeed, for the tests without SMOTE XGBoost reaches 0.9998 in AUPRC and 0.9997 for the precision and the recall. Random Forests comes in second place with an AUPRC of 0.9997, a precision of 0.9995 and a recall of 0.9995 VizNN gets an AUPRC of 0.9991, a recall of 0.9990 and a precision of 0.9990. Concerning the experiments on the dataset increased by SMOTE, the results are slightly different but the order remains unchanged. XGBoost provides an AUPRC of 0.9998, a precision of 0.9996 and a recall of 0.9996. Random Forests gets better results and reaches the same scores than XGBoost, namely 0.9998 for its AUPRC, 0.9996 of precision and 0.9996 for the recall. Finally, VizNN yields to 0.9986 in AUPRC, 0.9987 in precision and 0.9987 in recall.

For all these evaluation rounds, VizNN proves to be very competitive though slightly behind the reference algorithms – less that 0,08%.

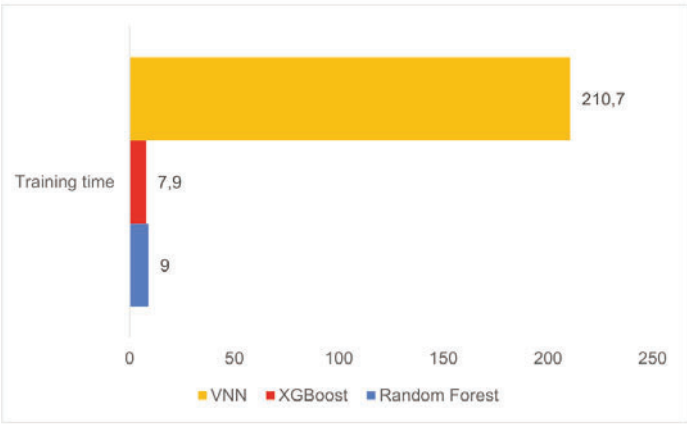

**Fig. 12.** Comparison of training time

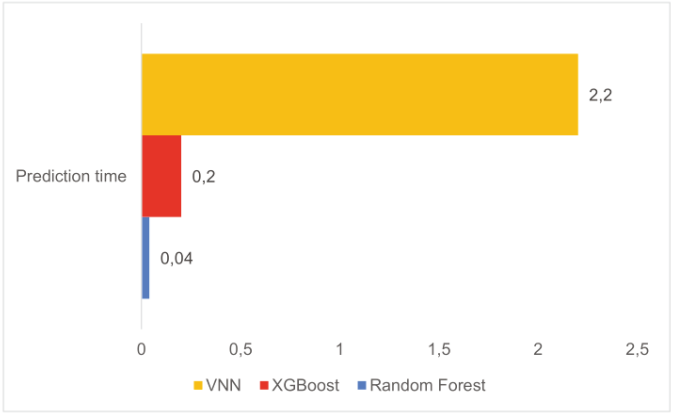

**Fig. 13.** Comparison of prediction time

Figure 12 shows the respective training times of Random Forests, XGBoost and VizNN. Training requires 9 seconds for Random Forests, 7.9 seconds for XGBoost, and 210.7 seconds for VizNN.

Figure 13 shows the respective prediction times of Random Forests, XGBoost and VizNN. Prediction requires 0.04 seconds for Random Forests, 0.2 seconds for XGBoost, and 2.2 seconds for VizNN.

In VizNN, the detection process is performed through a CNN which architecture is the following one: a convolutional layer with 32 filters and RELU activation; a maxpooling layer; a convolutional layer with 64 filters and RELU activation; a second maxpooling layer; and three final filtering layers. The configuration of our experimental setup has optimal results for 16 features, image size of 75x75 pixels and an architecture of 32 and 48 layers.

The evaluation of the proposed scheme shows that it is highly competitive with the reference algorithms XGBoost and Random Forest. It is performed by detecting attacks present in CIRA-CIC-DoHBrw-2020 dataset, which entails 60178 rows and 35 columns. The AUPRC metric achieves 0,9991 for VizNN, against 0,99982 for XGBoost and 0,99975 for Random Forest. The use of SMOTE has a slightly positive impact on detection capability in the Random Forest case only ($AUPRC = 0,99970$), and slightly negative impact for VizNN (0,9985) and for XGBoost (0,99976).

## 7 Conclusions

In this work, we propose VizNN, a new scheme for applying Convolutional Neural Networks to the detection of cybersecurity attacks through data augmentation. VizNN works by generating images from the system logs. It deliberately ignore non-representative information such as IP addresses and ports, and uses one-hot encoding of categorical information and timestamps. VizNN selects the most importance features as ranked by the gain they bring to learning, converts this features to image and resize them for optimizing the detection.

VizNN succeeds at integrating data augmentation in the analysis flow and at applying a highly scalable detection approach through Convolutional Neural Networks. It enhances the visibility of the analysis process by providing intermediate graphical representations of the system states to the security analyst. Nonetheless, VizNN still experience several limitations that need to be solved in future works. First, the graphical representations need important experience to be readable by the security analyst, like medical X-Rays require experienced practitioners. Second, the explicability and traceability of alerts to original traces are still better performed by alternative solutions like Random Forests or XGBoost.

The next step of this study is the evaluate to proposed scheme for massive datasets ($> 10^6$ data entries) to challenge the scalability capability of the various schemes under evaluation. Such an evaluation would be more representative of the amount of data that a Security Operating Centre (SOC) handles daily.

# Towards generating complex programs represented as node-trees with reinforcement learning

Andreas Reich and Ruxandra Lasowski

Hochschule Furtwangen University
reich@digitalmedia-design.com
ruxandra.lasowski@hs-furtwangen.de

**Abstract.** In this work we propose to use humanly pre-built functions which we refer to as nodes, to synthesize complex programs. As e.g. in the node-tree programming style of Houdini (sidefx.com), we propose to generate programs by concatenating nodes non-linearly to node-trees which allows for nesting functions inside functions. We implemented a reinforcement learning environment and performed tests with state-of-the-art reinforcement learning algorithms. We conclude, that automatically generating complex programs by generating node-trees is possible and present a new approach of injecting training samples into the reinforcement learning process.

## 1 Introduction

Many modern computer programs, especially in the 3D sector, offer users to interact with their software via so-called nodes, constructing so-called node-trees. Nodes are atomic parts of node-trees. Every node represents a function, a pre-programmed sequence of computer code visually. Node-trees are a visual high-level representation of non-linear sequences of nodes and are simpler to understand then regular code [1]. Due to the increasing prevalence of nodes in computer software, the idea arose to utilize machine learning to automatically generate node-trees with neural networks. Because nodes represent computer code, automatically generating node-trees with AI means automatically generating computer code with AI. Therefore, our approach is classified in the field of neural program synthesis.

The task of program synthesis is to automatically find programs for a given programming language that satisfy the intent of users within constraints [2]. Researchers like Bunel et. al. use supervised and reinforcement learning techniques to generate programs by concatenating low-level nodes linearly [3]. In contrast to their approach we propose to to generate programs concatenating high-level nodes non-linearly because this allows for more complex programs when using the same count of nodes. Our approach aims for automating the manual node-tree generation pipeline with AI that uses non-linear and high-level nodes. Using high-level nodes in the generation process is beneficial:

When compiled, all non-linear node-trees are transformed into linear sequences of code. The use of high-level nodes that consist of many low-level nodes is beneficial for certain use cases since more complex tasks can be solved with high-level nodes than low-level nodes (if the correct nodes are available) because more code is executed.

Most existing node-driven software solutions have a highly optimized set of high-level nodes that are software-specific. Consequently, users can perform a great variety of tasks solely using few domain-specific nodes. Automatically creating and combining these specialized nodes with the help of AI would save users a lot of time and resources, while keeping full editability of generated node-tress.

Developing algorithms that search for valid node-trees in a search space (space of all possible connections) is challenging because the search space grows exponentially when adding nodes or depth and best regular program synthesis algorithms like Chlorophyll++ are currently able to find programs in a search space of $10^{79}$ [4]. This means if the aforementioned algorithm would work with nodes it could reliably find programs out of a pool of 100 individual nodes and a depth of 39 nodes. According to Bodík finding programs in a search space of $10^{79}$ is not sufficient to solve complex problems, e.g. like implementing the MD5 hashsum algorithm, located in a search space of $10^{5943}$ [4], since the algorithm works with low-level functions.

However, the generation of programs becomes easier if few high-level functions, that achieve the same result as many low-level functions, are combined. For instance, a program could just consist of 5 pre-assembled, high-level functions that themselves consist of hundreds of lines of code. Most algorithms could find such a program, just consisting of 5 nodes, with ease. The complexity of finding valid programs is constrained to the size of the search space. This means the complexity of finding node-trees built from high-level and low-level nodes are equal since the complexity is constrained to valid connections. Nonetheless, training an AI with high-level nodes is more costly computationally since more code needs to be executed.

This paper strives to point out the benefits of working with high-level over low-level functions. Performing some tests with the 3D Engine Blender and utilizing a node-based modeling tool showed that a specific node-tree in Blender with as little as 4 nodes already represents more than 1000 lines of computer code, whereas approaches from the Google Brain team are able to reliably generate programs with up to 25 lines of code with AI [5]. Figure 1 visualizes the difference between the count of nodes with the complexity of the resulting programs.

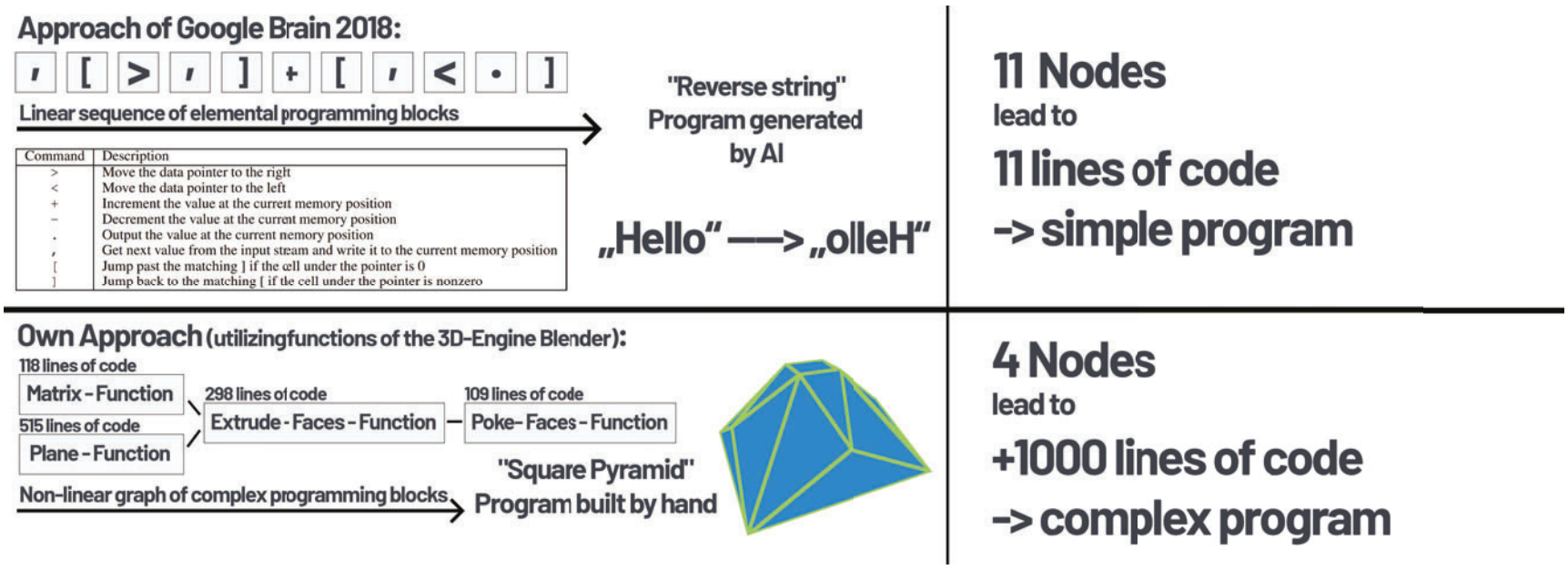

figure 1: complexity of high-level and low-level nodes

A possible real-world use case for using automatically synthesized node-trees could be the software Substance Designer with a completely node-driven workflow for texture and material creation. The Substance Designer documentation states that there are about 300 individual (high-level) nodes, yet also that most of these nodes only find application in rare, specific use cases [6]. Assuming one could build basic substance materials with 50 different nodes and a node-tree length of 40, the complexity of the search space would be $50^{40} \approx 10^{68}$ – less than what is currently possible in regular program synthesis. Nevertheless, a lot of the Substance Designer nodes have internal states that need to be adjusted, so the number of nodes necessary increases since changing an internal state can also be represented by the creation of a specific node.

Assuming one could create nearly all common Substance Designer materials by using all 300 nodes and 300 extra nodes that represent internal state changes, the creation of node-trees with lengths of 100 nodes leads to a search space of $600^{100} \approx 10^{278}$. Compared to solving MD5 with low-level functions and a search space complexity of $10^{5943}$ [4], a search space with the complexity of $10^{278}$ would still be way out of scope, yet more realistic to be achievable in the upcoming decades.

Besides Substance Designer, there are many node-based software solutions that offer a set of highly specialized nodes for different use cases. Making use of a predefined set of nodes is beneficial since most applications with node collections are capable of performing almost all tasks in a specific domain. We therefore identified certain areas where generating nodes with AI could be useful:

- Parameterization of 3D meshes (input: 3D mesh, output: node-tree approximating the modelling steps necessary to generate the mesh)
- Parameterization of photoscanned point clouds (input: point cloud, output: node-tree approximating the modelling steps necessary to generate the point cloud)
- Creation of 2D materials from images (input: 2D image, output: node-tree approximating a PBR-material)
- Reverse-engineering of functions and programs for optimization (input: program/function, output: optimized node-tree that approximates the program/function)

## 2 Purpose and Methods

We start our research for node-tree synthesis with the task of calculating a number with combinations of mathematical low-level nodes, e.g. plus and multiplication nodes, that are organized in a tree. In this way, we first heavily abstract more complex tasks to show feasibility. Figure 3 shows a node-tree reliably generated by our AI. We do not directly try to synthesize node-trees for complex graphic industry software because they require pipeline API's that could be implemented once the feasibility is shown. The purpose of our research is to show that we are able to automatically create node-trees from nodes with AI. Furthermore we want to point out the benefits of creating more complex programs by using high-level nodes instead of low-level nodes. Positive results could enable node-driven software solutions to utilize AI to automate processes that currently are solely performed by human users.

We use the qualitative method to investigate the state-of-the-art through literature research, especially concerning the comparison of publications and results of other researchers. Furthermore, we use the quantitative method to build experiments to prove that creating node-trees with supervised and reinforcement learning is possible.

To investigate the feasibility of generating node-trees with AI we implement an OpenAI Gym [7] reinforcement learning environment for automated node-tree generation and perform about 50 training sessions over the course of nine months. We use the reinforcement learning neural network architecture DDPG [8] and adjust its parameters.

To detect valid programs our actor explores a space of $8^{25}$ ($\approx 3.7 * 10^{22}$) in which it can perform 25 actions out of a pool of 8 individual actions. The following types of actions can be performed by the network:

- creating and automatically switching to new nodes
- changing internal states of nodes
- switching back to the previous node

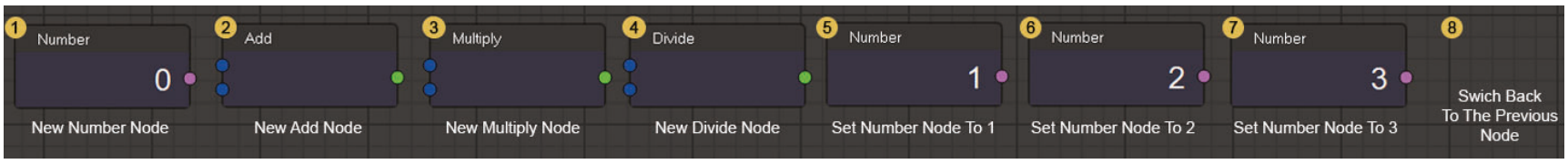

figure 2: All actions that can be performed by the network

The following states are observable by the network:

- specification/goal number
- relative distance to target
- number of nodes in total
- count of open connections
- type of the active node
- id of the active node
- internal state of the active node

Since our actor is not capable of finding valid samples without guidance we use an approach which we refer to as sample injection (figure 4). During exploration we randomly inject a valid action sequence (complete episode), which fulfils a given specification, into the current batch instead of the actors own chosen action sequence. This yields good results, since the actor learns from these optimal action sequences and is still capable of exploring the environment. This method is a combination of supervised- and reinforcement learning. The supervised samples are random sequences of valid connections between nodes that lead to a result (node-tree) which can be used as a specification for training. Therefore the actor can learn how to use which node in which context in order to fulfill a specification. Due to this on-the-fly generation process no data collection is necessary. Moreover the actor can learn how to use all available nodes in different contexts.

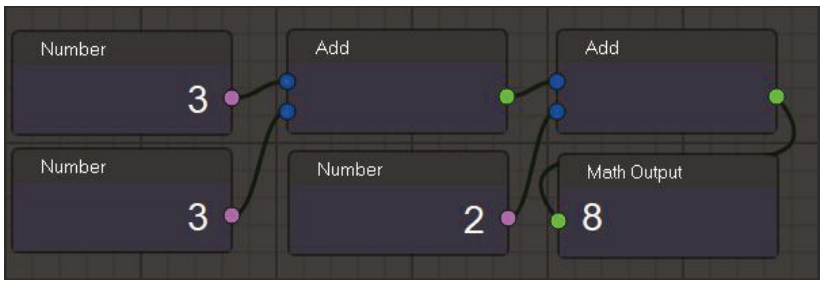

figure 3: node-tree found reliably in a search space of the size of $8^{25}$
(8 individual actions and max. 25 actions)

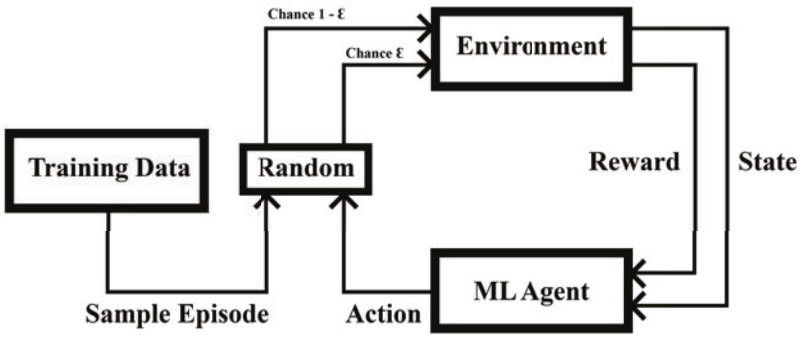

figure 4: sample injection

Because the search space in which valid node-trees can be found is very large our machine learning agent only receives sparse rewards and needs to learn from many samples. Therefore we decrease the learning rate to $10^{-5}$ and increase the replay buffer size to $10^7$. For the most successful training sessions we train the network for $\approx 20$ million steps. We use the following reward function, which rewards the actor for approximating a specification. Since we use mathematical nodes the goal number serves as specification and the current result of a node tree can be used to calculate the current distance which

is minimized:

$$\frac{((goalNumber - currentDistance) - (goalNumber - previousDistance))}{goalNumber} * incentive$$

DDPG uses the random Ornstein-Uhlenbeck process [9] which results in few valid programs to be found. We therefore complement the process with a supervised process to inject samples to enhance results. For most training sessions we inject samples 10% of the time instead of the action of the neural network. To reduce the impact of outliers we use the Huber loss [10] to enhance regression quality.

Figure 5 shows the network architecture of DDPG for the actor (left side) and the critic (right side). Figure 6 shows the parameters used for training. Figure 7 shows the pseudocode of DDPG with sample injection.

```
actor = Sequential()
actor.add(Flatten(input_shape=(1,) + env.observation_space.shape))
actor.add(Dense(16))
actor.add(Activation('relu'))
actor.add(Dense(16))
actor.add(Activation('relu'))
actor.add(Dense(16))
actor.add(Activation('relu'))
actor.add(Dense(nb_actions))
actor.add(Activation('linear'))
```

```
x = Concatenate()([action_input, flattened_observation])
x = Dense(32)(x)
x = Activation('relu')(x)
x = Dense(32)(x)
x = Activation('relu')(x)
x = Dense(32)(x)
x = Activation('relu')(x)
x = Dense(1)(x)
x = Activation('linear')(x)
```

figure 5: DDPG network architecture: left: actor, right: critic

```
memory = SequentialMemory(limit=10000000, window_length=1)
random_process = OrnsteinUhlenbeckProcess(size=nb_actions, theta=.15, mu=0., sigma=.3)
agent = DDPGAgent(nb_actions=nb_actions, actor=actor, critic=critic, critic_action_input=action_input,
                  memory=memory, nb_steps_warmup_critic=1000, nb_steps_warmup_actor=1000,
                  random_process=random_process, gamma=.99, target_model_update=1e-3,
                  supervised_learning=True, action_process=env.Supervised_Action_Process(env))
agent.compile(Adam(lr=.000001, clipnorm=1.), metrics=['mae'])
```

figure 6: parameters used for training DDPG

## 3 Results

Our experiments show that our trained AI agent is capable of generating certain node-trees reliably from low-level nodes when given a specification. Some generated node-trees are found with 100% accuracy in regard to a given specification whereas other node-trees approximate a given specification to some degree. Our results therefore show the feasibility of generating node-trees from low-level nodes with AI. Hence, We observe that more complex specifications, that need more nodes to be solved, lead to decreasing accuracy of our agent (table 1). This is due to the fact that the search space grows exponentially.

**Algorithm 3** DDPG algorithm with sample injection

Provide valid action trajectories for injection learning
Randomly initialize critic network $Q(s,a|\theta^Q)$ and actor $\mu(s|\theta^\mu)$ with weights $\theta^Q$ and $\theta^\mu$.
Initialize target network $Q'$ and $\mu'$ with weights $\theta^{Q'} \leftarrow \theta^Q$, $\theta^{\mu'} \leftarrow \theta^\mu$
Initialize replay buffer $R$
**for** episode = 1, M **do**
  Initialize a random process $\mathcal{N}$ for action exploration
  Receive initial observation state $s_1$
  With the probabilty of 1 - Ɛ select an episode for injection learning
  **for** t = 1, T **do**
    Select action $a_t = \mu(s_t|\theta^\mu) + \mathcal{N}_t$ according to the current policy and exploration noise
    If episode is selected for injection learning, sample all episode actions from valid trajectory
    Execute action $a_t$ and observe reward $r_t$ and observe new state $s_{t+1}$
    Store transition $(s_t, a_t, r_t, s_{t+1})$ in $R$
    Sample a random minibatch of $N$ transitions $(s_i, a_i, r_i, s_{i+1})$ from $R$
    Set $y_i = r_i + \gamma Q'(s_{i+1}, \mu'(s_{i+1}|\theta^{\mu'})|\theta^{Q'})$
    Update critic by minimizing the loss: $L = \frac{1}{N}\sum_i (y_i - Q(s_i, a_i|\theta^Q))^2$
    Update the actor policy using the sampled policy gradient:

$$\nabla_{\theta^\mu} J \approx \frac{1}{N} \sum_i \nabla_a Q(s, a|\theta^Q)|_{s=s_i, a=\mu(s_i)} \nabla_{\theta^\mu} \mu(s|\theta^\mu)|_{s_i}$$

    Update the target networks:

$$\theta^{Q'} \leftarrow \tau\theta^Q + (1-\tau)\theta^{Q'}$$
$$\theta^{\mu'} \leftarrow \tau\theta^\mu + (1-\tau)\theta^{\mu'}$$

  **end for**
**end for**

figure 7: DDPG with sample injection pseudocode

# 4 Limitations

Due to time and hardware limitations we are not able to find programs in a search space of $10^{79}$ and therefore work with a smaller search space. Furthermore, we are only able to set up our experiments using low-level nodes. However, the size of the search space of high- and low-level nodes are equal when the same count of actions and nodes are available. Therefore, the results of our experiments can be transferred and therefore also apply to high-level nodes.

**Table 1.** Node-trees reliably generated by our algorithm in a search space with the size of $8^{25}$ and their accuracy in relation to a given specification (sorted by accuracy)

| Specification | Exemplary action trajectory | Accuracy |
|---|---|---|
| 1,2,3,4,5,6,9 | add, num, set(2), back, num, set(2) | 100% |
| 10 | add, multiply, num, set(3), back, num, set(3) | 90% |
| 7,8 | add, add, num, set(3), back, num, set(3), back, num, set(2) | 86% |
| 11 | add, multiply, num, set(3), back, num, set(3) | 81% |
| 12 | add, multiply, num, set(3), back, num, set(3) | 75% |

## 5 Conclusions

In this work we use reinforcement learning in combination with supervised learning and the technique of sample injection to tackle the problem of solving the combinatorial search over node-trees that lead to the user specified result. We think that our technique could be used to ease many tasks in computer graphics like 3D mesh generation, VFX compositing, material generation and node-tree based scripting. We show that it is possible to automatically create node-trees from nodes with AI. Furthermore we point out that the size of the search space when using low-level nodes and the size of the search space when using high-level nodes are equal. This means that one can create more complex programs with the same amount of high-level nodes since high-level nodes consist of more lines of code. We therefore conclude that our technique will find application in node-driven software solutions and will leverage academia's interest in node-based neural program synthesis.

# Use of Artifical Intelligence and Image Segmentation for 3-Dimensional Modeling

Michael Weber[1,2], Tobias Weiß[1,2], Franck Gechter[2], and Reiner Kriesten[1]

[1] Institute of Energy Efficient Mobility
Hochschule Karlsruhe - University of Applied Sciences, HKA
Karlsruhe, Germany
{michael.weber, tobias.weiss, reiner.kriesten}@h-ka.de
[2] CIAD (UMR 7533)
Univ. Bourgogne Franche-Comte, UTBM
Belfort, France
LORIA-MOSEL (UMR 7503)
Université de Lorraine
Nancy, France
franck.gechter@utbm.fr

**Abstract.** To use Augmented Reality in an automotive vehicle for testing Advanced Driver Assistance Systems a new development approach with high computing power is needed. Reasons for this are a high vehicle speed as well as fewer possible orientation points on an urban test track compared to using AR applications inside a building. With the help of Image Segmentation, Artificial Intelligence for Object Detection, and Visual Simultaneous Localization and Mapping a 3-Dimensional Model with precise information of the urban test site is to be generated. Through the use of AI and Image Segmentation, it is expected to significantly improve performance like computing speed and accuracy for AR applications in automotive vehicles.

## 1 Introduction

Camera-based Advanced Driver Assistance Systems (ADAS) such as the active lane departure warning system and traffic sign recognition support the driver, offer comfort, and take responsibility for increasing road safety. These complex systems go through an extensive testing phase, which results in optimization potential regarding quality, reproducibility, and costs. ADAS in the future will support ever-larger proportions of driving situations in increasingly complex scenarios. Due to the increasing complexity of vehicle communication and the rising demands on these systems in terms of reliability to function safely even in a complex environment and to support the driver and increase safety, the test scenarios for ADAS are constantly further developed and adapted to higher requirements. European New Car Assessment Programme (Euro NCAP) has introduced a series of new safety tests for ADAS into its program and created a road map until the year 2025 [1] [2].

Today's test methods can be separated into two categories. On the one hand, the testing of the ADAS with the help of virtual worlds and on the other hand, the testing in reality on the test track using objects in real life. The central idea of the virtual test procedure is to transfer vehicle behavior to virtual test drives as realistically as possible. The approach for virtual tests is aimed at benefit from the advantages of simulation in terms of reproducibility, flexibility, and reduction of effort. In this way, specifications and solutions derived from them should be able to be tested and evaluated at an early stage of the development process. The use of suitable simulation methods enables the efficient design, development, and application of vehicles and vehicle components. However, virtual development methods cannot yet replace real-life driving tests in all respects. Due to the complex physical conditions in which a vehicle is transferred when testing ADAS, real-life driving tests are still necessary to the current status. For example, the weather, the surface texture of the road, and other influencing parameters take a decisive role in the evaluation process of ADAS test drives [3] [4].

The presented research background of this paper combines the advantages of ADAS-tests in a virtual simulation and these of ADAS-tests in a real environment. The camera images of the vehicle are augmented with additional virtual information. The augmentation of virtual road lanes allows, for example, the testing of a lane departure warning system independent of the test track. Scenarios such as the appearance of temporary lane markings or the absence of sections can be tested on the same test area. Narrowing and widening of lane markings can be represented as well as international differences between road markings. For testing traffic jam assistance systems, vehicles driving ahead can be augmented with camera images. In the first phase of testing, second vehicles including drivers can thus be dispensed with, reducing the costs of the tests and increasing the safety of the test engineers. Furthermore, ADAS-test cases with traffic signs as well as pedestrians and cyclists can be augmented situationally and quickly.

Furthermore, by using Augmented Reality (AR) for testing camera-based ADAS, new possibilities for testing complex, critical, and even forbidden test cases arise. For example, for testing the lane departure warning system, the traffic lane can be inserted into the image in any given width, regarding the lane and the white stripe itself. Therefore it is possible to test the system to its limits, a feature not possible by testing in reality on the test track.

For the use of AR, the system must be located (position and orientation) in its environment. This technology requires precise 3-dimensional (3D) modeling based on existing sensors. Usually, AR-applications are designed for human users and are mostly used inside buildings. Through a variety of orientation points inside a building and the movement speed of the user at walking speed, this technology is already quite advanced. The approach of this research project, by contrast, is being developed for a Electronic Control Unit (ECU), which requires a novel development approach with high computing power due to the high vehicle speed. Furthermore, compared to using AR-applications inside a building, fewer orientation points are available on a test site, so a new concept has to be developed here as well. The target of research described in this paper is to use Image Segmentation to analyze the environment of an mostly urban test site. Based on these results, a 3D model of the environment is to be created. In a further step, the 3D model is to be added by objects such as traffic signs, road markings, pedestrians, cyclists, etc. The use of Artificial Intelligence (AI) should provide precise information on the depth of the environment using 2-dimensional (2D) image sequences. Through the use of AI and Image Segmentation, it is expected to significantly improve the performance like computing speed and accuracy of the environment model. Moreover, conventional algorithms such as Simultaneous Localization and Mapping (SLAM) will be used for comparison within the research project.

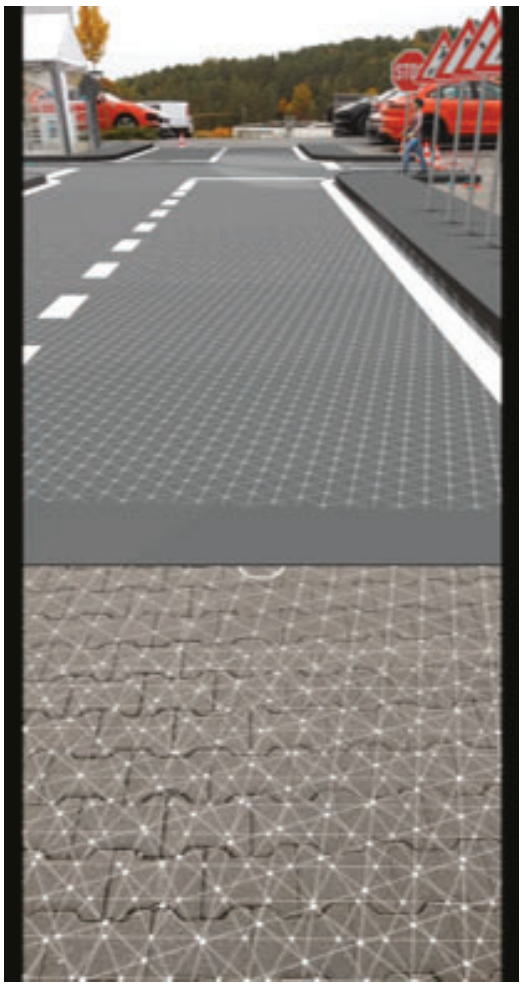

**Fig. 1.** Augmented Reality application showing a possible scenery

The main contributions of this paper include:
1. An overview of challenges for the use of AR in the automotive vehicles with regard to camera-based ADAS.
2. An introduction of a novel approach based on visual SLAM (vSLAM) and using of AI for object identification and thus increasing the accuracy and reproducibility of ADAS in automotive vehicles.

## 2 Necessary Criteria for Augmented Reality

To use AR in ADAS of automotive vehicles different criteria are necessary compared to conventional AR-applications like on a smartphone. This section will describe the contrasting criteria for this approach.

### 2.1 Augmented Reality for Conventional Applications

According to a proposal by Azuma, Augmented Reality can be defined as a combination of three fundamental features: the combination of real and virtual worlds and precise three-dimensional registration of the real and virtual objects, both in an interactive real-time environment [5].The basic principle of AR is best known by the mobile phone game Pokémon Go, published in 2016 by Niantic [6]. Within this game, the users can interact with digital creatures through their smartphones. These creatures are placed virtually in the environment of the user. Such an AR application can be seen in Figure 1 [6]. Figure 2 shows the three parts of the algorithms behind augmented reality: image analysis, 3D modelling, and augmentation.

The image analysis serves to detect points or regions of interest within the given image. Feature detections like corner detection or edge detection are often used for this step [7]. With the results of the image analysis, a three-dimensional model of the environment is created. The kinds of algorithms used for this step vary depending on the type of AR application. For AR in unknown locations, simultaneous localization and mapping (SLAM) or structure from motion (SfM) algorithms are widely spread [8]. The augmentation is based on the results of the 3D modelling. The scene model is usually provided as a positional description of a plane or a coordinate system representing the real world [9]. With this information, a virtual object can be placed upon the plane or in the coordinate system with adequate characteristics such as size and orientation. After the object placement, virtual content is combined with the real-world image [10].

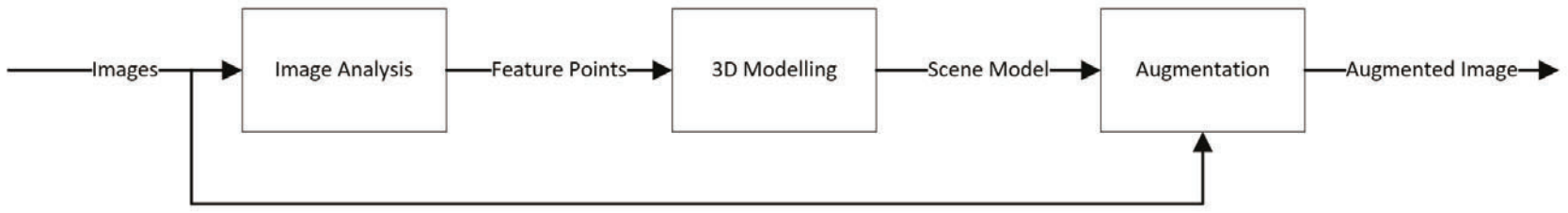

**Fig. 2.** Augmented Reality steps

There are several publications of applications for AR. These applications vary heavily in their fields, from the usage of AR in psychology [11] to the use in operating rooms in hospitals [12] to mobile games [6] to military applications [13]. What all these applications have in common is that the reality of a human is augmented. With the human as the user of AR, there are some implicit consequences for the application. One of them is that the human user is, in most cases, lenient towards virtual objects not placed precisely within a small range of error. Furthermore, the velocity of human movement and therefore the distance travelled by any given time is limited. By these restrictions, the requirements for localization, mapping, object placement, and runtime are not as high as in an automotive environment, as is discussed in the next chapter.

### 2.2 Augmented Reality for Advanced Driver Assistance Systems (ADAS) in Automotive Vehicles

For the use of AR in automotive vehicles and the associated specific use of ADAS-sensor technology on conventional test tracks, special criteria are to consider. For instance, the test sites are located outside of buildings and are usually therefore low textured [14]. In addition, there are quick scene changes due to the speed of the automotive vehicle. These points lead to the fact that conventional AR-SLAM approaches cannot perform the necessary localization- and mapping-process for SLAM-Algorithm (a reference to section 3) with the desired accuracy and resolution. Due to the desired integration of the presented approach into a serial automotive vehicle without any additional sensor technology, the aim is to generate information about the depth and texturing of the environment based solely on the installed camera. This camera is usually a mono-view-camera system that is often integrated into the rear view mirror of the automotive vehicle. Mono-view-camera systems are established vehicle hardware, which is mostly used in low-priced series models due to their compact design, high resolution, robustness, long-range and low cost. On the other hand, high-priced vehicles use stereo-view-camera systems, which enable spatial vision like a human [15].

In addition to the low textured environment of the test track and the fast change of scenery, other aspects such as weather influences like rain and the sun position, soiling of the windshield, bumps of the road surface and the lack of road markings have to be considered [16]. Furthermore, when augmented reality is used in automotive vehicles, the end-user is not the human driver, but an ECU. This implies that very high accuracy and a high realism, e.g. correct shadowing and occlusion of the augmented objects are required in the overall process [17]. In comparison to a human driver, the ECU must not detect any difference between reality and the augmented reality, otherwise, the ECU will be transferred to an error state. It is also highly relevant to consider the constant further development of ADAS, which persistently demands increased requirements for realistic test scenarios. This approach aims to achieve the same driving behavior as in reality.

In addition to accuracy, the runtime of the overall algorithm is also of great importance. Nowadays camera systems work with a frame rate of 30 to 60 Frames per Seconds [fps]. The resulting maximum overall runtime for handling one frame can be found in Table 1.

| **Framerate** | **Maximum runtime** |
|---|---|
| 10 fps | $\frac{1}{10}\,s = 0.1000\,s$ |
| 30 fps | $\frac{1}{30}\,s = 0.0333\,s$ |
| 40 fps | $\frac{1}{40}\,s = 0.0250\,s$ |
| 45 fps | $\frac{1}{45}\,s = 0.0222\,s$ |
| 50 fps | $\frac{1}{50}\,s = 0.0200\,s$ |
| 60 fps | $\frac{1}{60}\,s = 0.0167\,s$ |

**Table 1.** Several Framerate and the according maximum runtime.

For a successful evaluation of ADAS-test scenarios, the AR system must be able to orient itself in the environment very accurately [18]. One cause is the missing feedback about the impact intensity of test dummies when crashing them. For this reason, it is necessary to know the exact position of the car on the test track to calculate the intensity of the impact based on the braking distance. When using Euro NCAP test scenarios, velocities up to

$$130\,\frac{km}{h} \mathrel{\widehat{=}} 36.111\,\frac{m}{s} \tag{1}$$

are tested. The AR algorithm must have a faster runtime compared to the speed of the camera system. The distance $d$ the vehicle covers within a frame at any given velocity and framerate can be calculated by:

$$d = \frac{v_{Vehicle}\,[\frac{m}{s}]}{Framerate\,[\frac{frames}{s}]} \tag{2}$$

At a speed of $130\,\frac{km}{h}$ and a camera framerate of 30 fps, the vehicle travels

$$d = \frac{36.111\,[\frac{m}{s}]}{30\,[\frac{frames}{s}]} = 1.204\,\frac{m}{frame}. \tag{3}$$

Accordingly, for a framerate of 60 fps at the same speed, a distance of

$$d = \frac{36.111\,[\frac{m}{s}]}{60\,[\frac{frames}{s}]} = 0.602\,\frac{m}{frame} \tag{4}$$

is covered. A deceleration of one frame means a deviation of the test results of 0.602 to 1.204 meters.

Based on the high speed of the car and the camera, and the high need for precision in object placement, it is clear that the requirements for this application of Augmented Reality are far more strict than for the usual application for human users.

# 3 Development Approach

Simultaneous Localization and Mapping (SLAM) is a method for obtaining the 3D structure of an unknown environment and sensor motion in the environment. This system was initially intended to achieve autonomous control of robots [19]. Due to continuous development, SLAM-based applications have also found their way into mobile device applications and self-driving cars. To increase the accuracy of SLAM algorithms, various approaches allow the integration of different sensors, such as laser range sensors, rotary encoders, inertial sensors, Global Position Systems (GPS), and cameras. These algorithms are summarized in the following papers [20] [21] [22] [23]. Since cameras primarily are used for the most part in automotive vehicles, the approach presented in this paper is based on a subcategory of SLAM algorithms - visual Simultaneous Localization and Mapping (vSLAM). In the following section, the State of the Art for vSLAM-techniques are described. Based on these methods a new approach for AR using SLAM in automotive vehicles is presented.

## 3.1 State of the Art - Visual Simultaneous Localization and Mapping (vSLAM)

The approach of vSLAM uses only visual inputs to perform localisation and mapping. This means that no vehicle sensors other than the vehicles camera system are needed to create a 3D model of the environment thus making this approach more flexible than LIDARS, Radars, and Ultrasonics. The framework of vSLAM-algorithm is mainly composed of three basic modules: Initialization, Tracking, Mapping, and two additional modules: Relocalization and Global Map Optimization (including Loop Closing) [24].

Basic modules:

1. Initialization: To use vSLAM, the fundamental step is to define a specific coordinate system for camera position estimation and 3D reconstruction in an unknown environment. Therefore, the global coordinate system should be defined first during initialization. A part of the environment is therefore reconstructed as an initial map in the global coordinate system [24].
2. Tracking: After the initialization process, tracking and mapping are performed. Tracking involves following the reconstructed map in the image to continuously estimate the camera position of the image to the map. For this purpose, distinctive matches between the captured image and the created map are first determined by feature matching or feature tracking in the image [24].

3. Mapping: The mapping process expands the map by understanding and calculating the 3D structure of an environment when the camera detects unknown regions where mapping has not been done before [24].

Additional modules:

4. Relocalization: When tracking has failed, Relocalization is required. Reasons for this can be, among others, fast camera movements. In this case, relocalization makes it possible to recompute the current camera position about the reconstructed map [24].

5. Global Map Optimization (including Loop Closing): The map usually contains a cumulative estimation error corresponding to the distance of the camera movement. To eliminate this error, Global Map Optimization is usually performed. In this method, the map is refined considering the consistency of the whole map information. If previously recorded map elements are recognized, loops are closed and the cumulative estimation error can be corrected from the beginning to the present. Loop Closing is a method for obtaining reference information. While closing loops, a closed loop is first searched by comparing a current image with previously acquired images. Generally, relocalization is used to recover the camera position and loop detection is used to obtain a geometrically consistent map. Pose Graph Optimization is widely used to suppress the cumulative error by optimizing the camera positions. Bundle Adjustment (BA) is also used to minimize the map reprojection error by optimizing the map and the camera positions. In large environments, this optimization method is used to efficiently minimize estimation errors. In small environments, BA can be performed without loop closure as the cumulative error is small [24].

For the use of SLAM in automotive vehicles and the associated properties such as fast scene changes and low texturing of the environment, various approaches are available using vSLAM-Algorithm, which can be found in [14]. In this paper, different SLAM approaches are compared based on accuracy and robustness, among others. Some other approaches, which are not compared in [14] but seem promising for the presented approach in this paper, are briefly described in the following:

ORB-SLAM:

The ORB-SLAM algorithm was first presented in 2015 and seems to be the current state of the art as it has higher accuracy than comparable SLAM algorithms [25]. Here, ORB-SLAM represents a complete SLAM system for monocular, stereo, and RGB-D cameras. The system operates in real-time and achieves remarkable results in terms of accuracy and robustness in a variety of different environments. ORB-SLAM is used for indoor sequences, drones, and cars driving through a city. The ORB-SLAM consists of three main parallel threads: Tracking, Local Mapping, and Loop Closing. A fourth thread can be created to execute the BA after a closed loop. This algorithm is a feature-based approach, which represents the detected points in a three-dimensional MapPoint [14]. Figure 3 shows a MapPoint, which is created using image sequences captured in-house. The MapPoint shows a recognized house in an urban environment using ORB2-SLAM. Various advancements and improvements in terms of accuracy, robustness, etc. can be found in further developments based on this approach of ORB-SLAM (ORB2-SLAM [18] and ORB3-SLAM [14]). While the performance of ORB-SLAM is impressive in well-structured sequences, error conditions can occur in poorly structured sequences or when feature points temporarily disappear, e.g., due to motion blur [26].

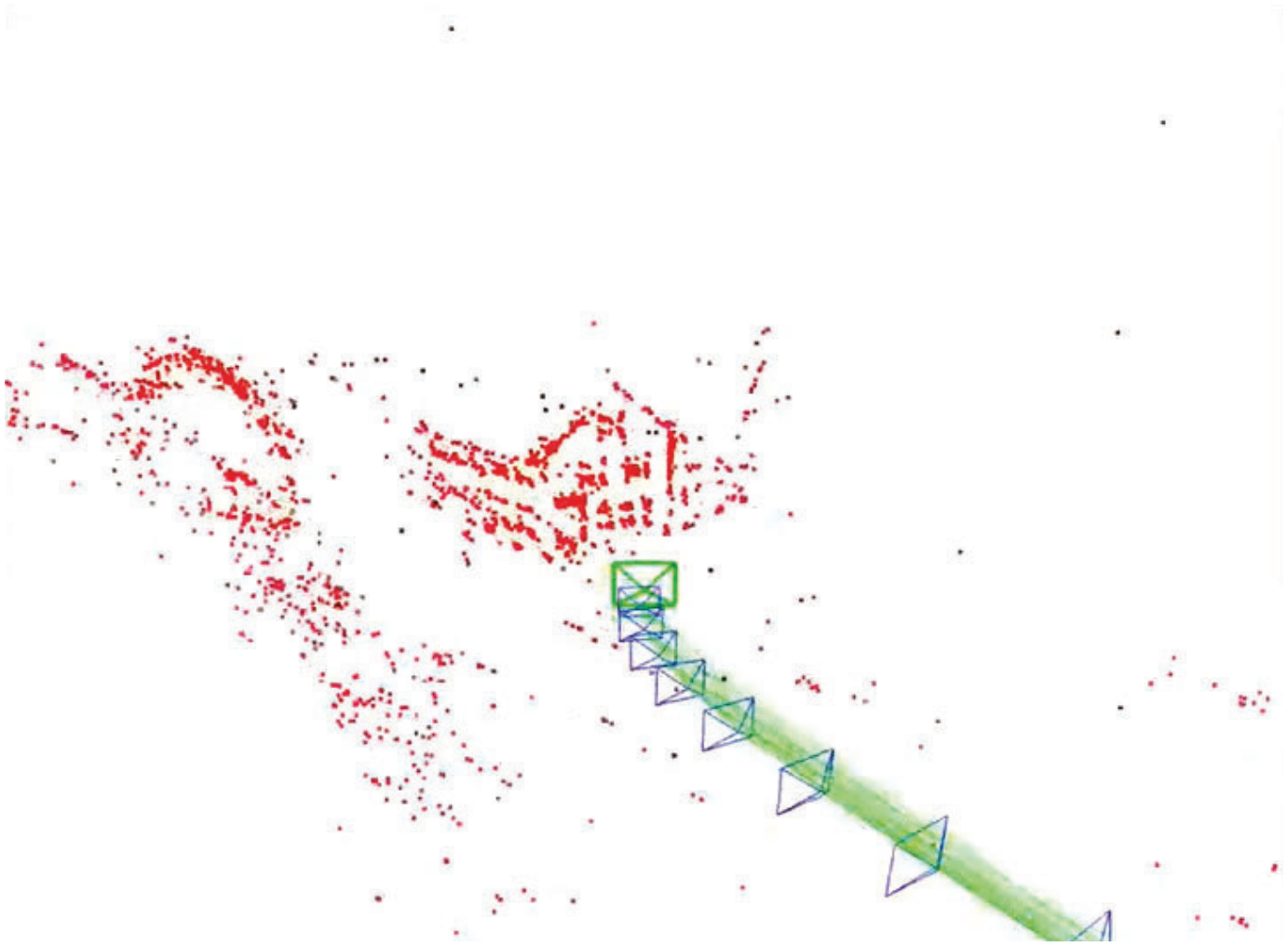

**Fig. 3.** MapPoint with red dots and green line for trajectory based on ORB2 with own image material in an urban environment

DS-SLAM:

ORB-SLAM exhibits excellent performance in most practical situations. However, some problems are not solved by the ORB-SLAM. First, the ORB-SLAM algorithm exhibits weaknesses in exceptionally dynamic and harsh environments. On the other hand, the map model created is based on geometric information like the MapPoint in Figure 3. This MapPoint does not provide a higher-level understanding of the environment. The DS-SLAM approach, firstly presented in 2018, combines the ORB-SLAM with the Semantic Segmentation approach using artificial intelligence to achieve a higher-level understanding of the environment. This approach further intends to increase the robustness of the SLAM system in dynamic environments. Based on ORB2-SLAM the DS-SLAM consists of basic SLAM-Modules like Tracking, Mapping, and Loop-Closing. Furthermore, DS-SLAM has two additional threads like Sementic Segmentation and Dense Map Creation. Using these additional threads improves the localization and mapping concerning robustness and accuracy in dynamic scenarios [26].

PL-SLAM:

Another approach to increase accuracy in poorly textured environments is PL-SLAM (Point and Line Simultaneous Localisation and Mapping), firstly presented in 2017. PL-SLAM extends the point-based approach known from ORB-SLAM with a line-based method. This line-based approach enables an improvement in terms of occlusions and false detections. Besides the improvement in poorly textured environments, this approach also shows increased performance in very well-textured environments, without significantly degrading the efficiency of this algorithm. Like the ORB-SLAM algorithm, the PL-SLAM has the basic SLAM modules for initialization, tracking, mapping, and loop closing. The extension of this approach is to use line-based algorithms in parallel with the point-based algorithms in each SLAM module. This approach ensures that the resulting map is more valuable and more diverse in 3D elements to derive important higher-level scene structures such as planes, voids, ground surfaces, etc. [27].

Based on the presented approaches in this section, in a further step single features of these SLAM-Algorithm are to extend and insert in a new approach for testing of ADAS in automotive vehicles. An introduction for the next steps is presented in the following section.

### 3.2 Use of Object Detection and vSLAM for AR in Automotive

For the use of AR in automotive vehicles, the approach should consist of using a state-of-the-art method and extending the feature point detection with an object detection. This should improve the following criteria:

- Robustness against blurred effects.
- Increase the accuracy of the 3D environment through improved depth information.
- Detect occlusions and improve 3D environment detail.
- Achieve robustness against weather effects.
- Increase realism for the control unit as end device.
- Increase computational speed with improved accuracy.
- Achieve higher-level understanding of the environment.

To achieve these criteria, the following features are to be extracted from the vSLAM approaches presented and examined in more detail for further research investigations:

ORB-SLAM:

- State-of-the-Art method to generating a MapPoint.
- Feature-Point approach should represent the basic framework.
- Selection of which criteria can be used from ORB or the further developments ORB2 and ORB3 based on it.

DS-SLAM:

- Approach of AI and Image Segmentation to generate an Object Detection.
- Creation of a Dense-Map for overlay on MapPoint.
- Achieve a higher-level understanding of the environment.

PL-SLAM:

- Based on edges to improve occlusions and improve object detection.
- Better 3D-reconstruction of objects through the detection of edges, points and lines.
- Improved realism through correct lighting and shadowing of augmented objects.

By cleverly combining the individual elements of the previously known SLAM algorithms, augmented reality in automobiles could be used in high-speed ADAS tests. In addition to the increased computing speed, increased accuracy should be achieved to be able to make a meaningful assessment of the performance of the ADAS tests. Figure 4 shows the original image, the feature point detection of the ORB2 algorithm, and the image segmentation used so far for object detection. The next step is to combine these approaches using AI.

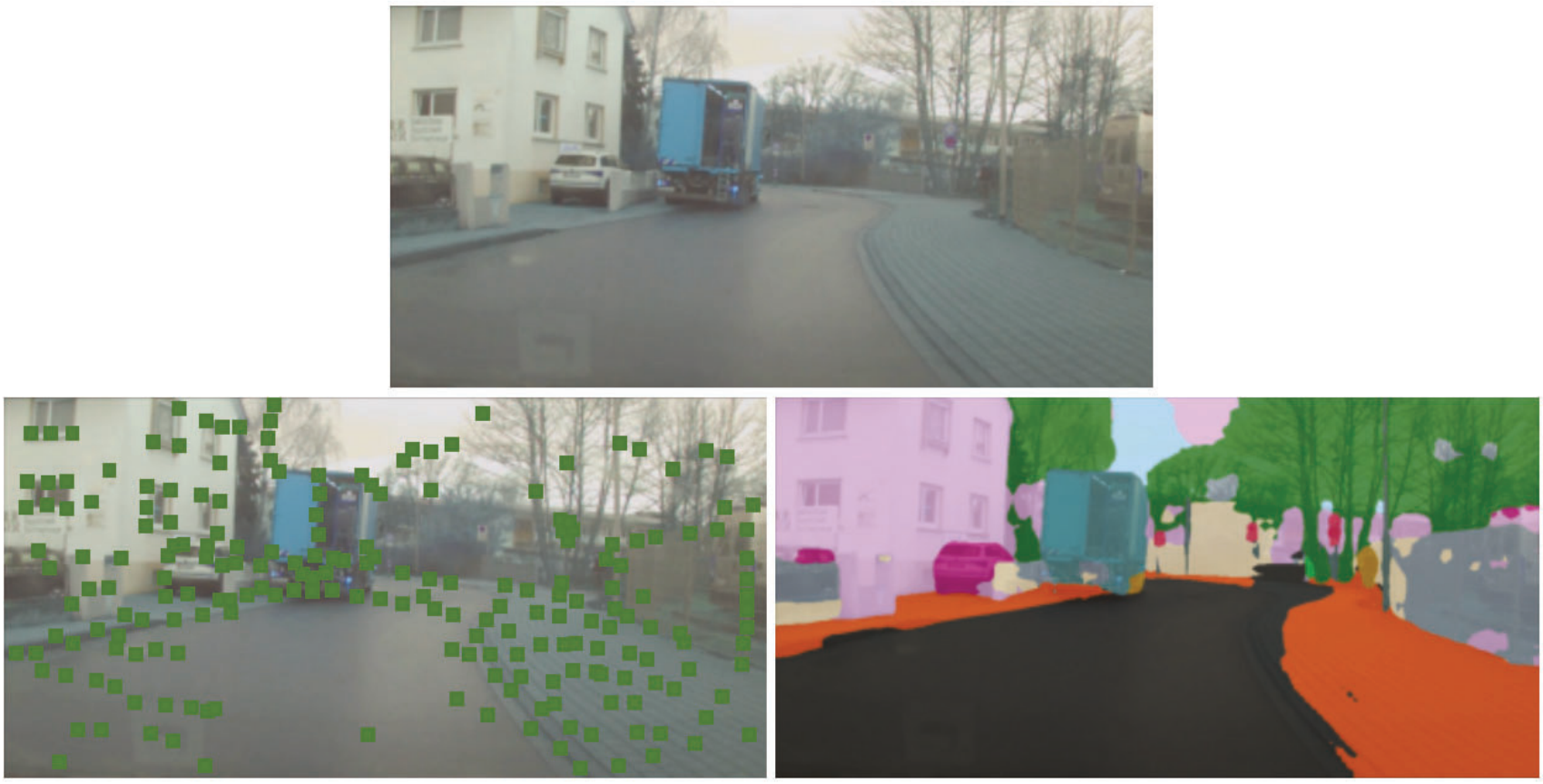

**Fig. 4.** Top: Original Image in an Urban Environment;
bottom left: Feature-Point-Detection using ORB2-SLAM-Algorithm;
bottom right: Object Detection using Image Segmentation

## 4 Conclusion

In this paper, we have proposed an approach to use Augmented Reality in automotive vehicles. We modeled the problem of creating an urban environment to use AR for testing in high-speed ADAS. Our approach is based on a combination of vSLAM-Algorithms like ORB-SLAM, DS-SLAM, and PL-SLAM within the combination of Artificial Intelligence to use Object Detection. This should help to generate a better overall performance concerning computing speed and accuracy.

The creation of a virtual 3D environment with a superior understanding of the individual objects should, in a further step, make it possible to augment other sensors such as the car's radar and lidar with objects in addition to the camera data. This should once again increase the overall performance of the entire system.

# Improving Temporal Consistency in Aerial Based Crowd Monitoring Using Bayes Filters

Jan Calvin Kramer[1], Thomas Golda[2], Jonas Hansert[1], and Thomas Schlegel[1]

[1] Karlsruhe University of Applied Sciences, Institute of Ubiquitous Mobility Systems IUMS
{Jan_Calvin.Kramer, Jonas.Hansert, Thomas.Schlegel}@h-ka.de
[2] Fraunhofer Institute for Optronics, System Technologies and Image Exploitation IOSB
Thomas.Golda@iosb.fraunhofer.de

**Abstract.** In order to monitor mass events, crowd managers continuously require reliable measurements of the crowd count. For this purpose, a variety of deep learning algorithms has been developed. Most of these so-called crowd counting algorithms return good results for still imagery but return oscillating crowd counts for video data. This is because, most crowd counting algorithms evaluate video data frame by frame and ignore the temporal relation between adjacent frames. In this paper, a variety of Bayesian filters is presented that successfully smooth the oscillating counts which in turn can lead crowd managers to trust the system more. The proposed filters work on top of the crowd counting algorithms' estimates. Thus, they can be easily used with any existing crowd counting algorithm that outputs a density map for a given input image.

## 1 Introduction

Mass events take place every day all over the world. Despite the joy these events bring to many people, they always come with the threat of turning into a stampede. To prevent this, crowd managers must be aware of the current crowd count and density and take the right measures in time.

For this purpose, a variety of algorithms have been developed since 2006. These so-called crowd counting algorithms take an image as their input and estimate the amount of people in the picture. Recent crowd counting algorithms go even further. Using Convolutional Neural Networks, they estimate a whole density map for a given input image. The density map itself holds the information of the crowd count that can be obtained by adding up the density map's pixel values. Gao et al. give a good insight into current advances and the huge amount of different crowd counting algorithms that exist [1].

Despite the many advances, current crowd counting algorithms do not consider the temporal relation between adjacent frames. Most crowd counting algorithms evaluate video data frame by frame and ignore the temporal relation. This often leads to fluctuating counts that the crowd manager cannot rely on. Only a few isolated approaches, e.g., the Temporal Aware Network [2], exist, that try to incorporate the temporal relation of adjacent frames. In contrast to their approach of presenting a new architecture that incorporates the temporal relation, we developed a variety of Bayesian filters that work on top of the counts estimated by current crowd counting algorithms. The filters can be used with any modern crowd counting algorithm that outputs a density map for a given input image.

## 2 Related Work

The following section gives a brief overview of the existing technologies this work builds upon. Firstly, two crowd counting algorithms, i.e., the CSRNet and MRCNet, are introduced. The crowd counters are used throughout this work to estimate the crowd counts of different data sets and to reproduce the problem of having oscillating counts. Secondly, Bayesian filters on which the developed concepts are based are briefly discussed.

### 2.1 CSRNet

The Congested Scene Recognition Network (called CSRNet hereafter) is a crowd counting algorithm developed by Li et al. in 2018 [3]. The network is divided into two parts, i.e. the front- and back-end. Both parts exclusively rely on (dilated) convolutional and max pooling layers.

The CSRNet's front-end consists of the VGG-16's first 10 layers. It extracts several features from an input image. The features are stored in a so-called feature map whose resolution is 1/8 of the resolution of the input image. The feature map is further processed by the back-end of the network. The back-end uses several dilated convolutional layers. Thus, it is able to extract even deeper features without further shrinking the resolution of the feature map. In order to output a density map that has the same size as the input image, the CSRNet uses bilinear interpolation with a factor of 8. A more detailed description can be taken from [3].

### 2.2 MRCNet

The Multi-Resolution Crowd Network (called MRCNet hereafter) is a crowd counting algorithm developed by Bahmanyar et al. from the German Aerospace Center [4]. The network relies on an encoder-decoder-structure. Analogous to the front-end of the CSRNet, the encoder of the MRCNet relies on the VGG-16. Yet, it does not only use the first ten layers but the first five CNN blocks of the VGG-16 that consist of 13 convolutional layers and five max-pooling layers in total. Since it uses more pooling layers, the size of the outputted feature map is 1/32 the size of the input image. Such a drastic reduction of the resolution can accidentally lead to people being removed. To prevent this, the feature maps at different stages of the encoder are added element-wise to the feature maps of the decoder.

The decoder of the MRCNet consists of five CNN blocks as well. However, instead of using pooling layers at the end of each CNN block, the decoder uses up-sampling layers. Each up-sampling layer increases the size of the feature map by a factor of two. Thus, at the end of the decoder, the MRCNet outputs a density map that has the same resolution as the input. Prior to outputting the density map, the MRCNet outputs a feature map whose resolution is 1/4 the resolution of the input image. This feature map is used to estimate the overall amount of people in an image. By estimating the amount of people at an early stage of the decoder, the remaining part of the decoder can be further used to output a full-resolution density map that has a higher localization precision. Further information about the MRCNet can be taken from [4].

### 2.3 Bayesian Filters

Bayesian filters have been around for quite a while. In short, they are a set of algorithms that iteratively estimate the hidden state of a system, e.g., the current crowd count, using imprecise measurements, e.g., estimated density maps, and a model of the system state, e.g., a traffic flow model [5].
The Kalman and particle filter are a subset of the Bayesian filters. They differ in that the Kalman filter returns an optimal solution under certain restrictions whereas the particle filter returns a good approximation while being less restrictive. The Kalman filter assumes the underlying model to be linear and discrete in the time domain. Furthermore, the process and measurement noise are assumed to be Gaussian with a zero mean. If these restrictions are not fully complied with, the particle filter is likely to return even better results than the Kalman filter.
A thorough understanding of both filters is necessary to fully understand the concepts that follow. A good insight is given by [5], [6] and [7].

## 3 Concepts

The following chapter explains the developed concepts of this paper.

### 3.1 Kalman Filter

A complete concept of a Kalman filter requires the definition of the state vector, state transition matrix, measurement matrix, process noise variance and measurement noise variance.
To not further increase the computational costs that come with crowd counting, the state vector considers only the crowd count and not the density map as a whole. The crowd count is further expected not to change between two consecutive frames. This assumption leads to a linear state transition matrix that only consists of the value 1.
The accuracy of this assumption depends mainly on two variables, i.e., the area under consideration (denoted by $A$) and the frame rate of the video (denoted by $f$). For either $A$ going towards zero or $f$ going towards infinity, the assumption that the crowd count does not change becomes true. Thus, the concept is supposed to return better results for video data with higher frame rates and scenes with smaller areas. The scene of an image can be artificially reduced by applying the Kalman filter on grids rather than the whole image. Experiments testing this behavior are conducted in the next chapter.
Yet, in practice, neither $A$ becomes zero nor $f$ goes towards infinity. This makes the state transition matrix inaccurate. To model this error, one must define the process noise variance (matrix).

**Data-Driven Process Noise Variance.** To get an estimate for the process noise variance, the information given by the training data is used. Given the annotations of the training data, one can calculate the change of pedestrians between two consecutive frames by subtracting their annotated crowd counts. The calculated pedestrian changes can be fitted to a Gaussian curve in a next step. Assuming that the set of pedestrian changes is normally distributed with a zero mean, the variance of the Gaussian curve corresponds to the actual variance of the process noise when the Kalman filter was applied to the training data. To obtain an estimate that generalises better on data sets with different crowd counts, one must consider the percentage change of pedestrians rather than the absolute change. Let $c_{k-1}$ and $c_k$ be the crowd count of two consecutive frames, then the percentage change of pedestrians ($\Delta c_{rel}$) can be calculated as follows:

$$\Delta c_{rel} = \frac{c_k - c_{k-1}}{c_{k-1}} \tag{1}$$

Fitting the percentage changes to a Gaussian curve results in a relative process noise variance denoted by $\sigma_{rel}$. To retrieve an absolute estimate of the process noise variance, one must iteratively multiply the relative variance with the previous posterior estimate of the Kalman filter and the frame rates' ratio:

$$\sigma_k = \bar{c}_{k-1} \cdot \sigma_{rel} \cdot \frac{f_{training}}{f_{testing}} \quad (2)$$

**Data-Driven Measurement Noise Variance.** To model the measurement noise variance, a data driven approach similar to the one of the process noise variance is used. Yet, instead of computing pedestrian changes of the training data, the performance of the crowd counter on the validation data is used. For this purpose, the percentage differences between the crowd counter's estimate and the corresponding ground truth is calculated. The percentage differences are also fitted to a Gaussian curve. Again, the variance that is obtained using this approach is a relative value that must be iteratively multiplied with the current measurement of the crowd count.

**Data-Driven Observation Matrix.** The final parameter that must be modelled to obtain a complete concept of a Kalman filter is the observation matrix. The observation matrix says how the measurements must be processed before they are further used within the Kalman filter. Therefore, if the measurements that are passed to the Kalman filter do not have an error with a zero mean, the observation matrix can theoretically be used to process the measurements in such a way that their error has a zero mean afterwards. For this purpose, the Gaussian curve, that is obtained during the computation of the variance of the measurement noise, is used. The relative mean denoted by $\mu_{rel}$ expresses the average deviation of the ground truth from the estimates computed by the crowd counter on the validation set. The oversimplified assumption that the error of the crowd counter depends only on the crowd count, lets one use the simple term $1 - \mu_{rel}$ for the observation matrix.

### 3.2 Particle Filter

The developed particle filter only considers the crowd count as well. Yet, a more complex model for the state transition matrix is used. The model of choice is the macroscopic fundamental diagram of traffic flow. Let $Q$ be the flow, $Q^*$ be the flow density, $v_0$ be the velocity of the pedestrians, $w$ be the width through a gateway in meters, $\rho$ and $\rho_{max}$ be the current density and maximum density of pedestrians respectively, then the flow of pedestrians can be calculated as follows:

$$Q^*(\rho) = \rho \cdot v_0 \cdot (1 - \frac{\rho}{\rho_{max}})\ [\frac{pedestrians}{m \cdot s}] \quad (3)$$

$$Q = Q^* \cdot w\ [\frac{pedestrians}{s}] \quad (4)$$

Assuming that $\rho_{max} = 5\frac{pedestrians}{m^2}$ and $v_0 = 1.4\frac{m}{s}$, it only requires the current density and the width of the gateway to estimate the flow of pedestrians [8].
Unfortunately, the traffic flow model comes with a major restriction that it assumes the flow of pedestrians to be unidirectional. To loosen this restriction and obtain a more realistic model, the filter does not assume the overall flow of a scene to be unidirectional.

Flows of open borders that are located at the edge of the image are assumed to be independent from each other and unidirectional. Making this assumption, three questions arise that must be further clarified:

– How does the system detect open borders at the edge of the image?
– What area around an open border must be considered to estimate the density at the open border?
– Given the flow at an open border, how does one estimate the direction of flow?

### 3.3 Detection of Open Borders

The annotation of open borders is assumed to be manually done by the user at the beginning. The user is expected to create a mask for the first frame of a video. Areas where people can possibly walk are supposed to be colored in white, whereas areas where people cannot be, e.g., a frontage, must be colored in black. By reading in the mask and dividing the pixel values by 255, one obtains a 2 dimensional matrix of zeros and ones. The information where the open borders are located can then be easily obtained by looking for non-zero sequences in its outermost rows and columns.
It is further used as an alternative measurement matrix that is multiplied element-wise with incoming density maps. Subsequently, all elements of the resulting matrix are added up to obtain the crowd count that is further processed by the particle filter. It should be noted that this approach might delete some rightfully annotated persons. This is because the ground truth density maps that are used to train the crowd counting algorithms are created by blurring the given head annotations using a Gaussian kernel. Thus, persons standing nearby a frontage might overlay the frontage in a density map. Yet, this approach ensures that areas where pedestrians can impossibly be, e.g., a sea, are not wrongfully labeled by the crowd counter.

### 3.4 Determining the Density at an Open Border

To estimate the density at an open border $j$ at the previous point in time $k-1$, a rectangular cutout of the density map at $k-1$ is used. One side of the rectangle is the open border itself, whereas the length of the other side is determined by the maximum distance pedestrians are assumed to walk between two consecutive frames (called step size hereafter). Let $v_0$ be the pedestrian's estimated velocity and $\Delta t$ be the time between two consecutive frames, then the step size can be calculated as follows:

$$\Delta s_m = \Delta t \cdot v_0 \; [m] \tag{5}$$

$$\Delta s_{pixel} = \Delta t \cdot v_0 \cdot 1/g \; [pixel] \tag{6}$$

where g corresponds to the ground sampling distance in meter per pixel.
Given a cutout of a density map, you can easily calculate the estimated crowd count within the cutout (denoted by $c_{j,k-1}$) by adding up the pixel values of the cutout. Let further $a_j$ and $b_j$ be the lengths of the rectangle in pixels, then the density of the rectangle can be calculated as follows:

$$\rho_{j,k-1} = \frac{c_{j,k-1}}{a_j \cdot b_j \cdot g^2} \; [\frac{pedestrians}{m^2}] \tag{7}$$

where

- $\rho_{j,k-1}$ $[pedestrians/m^2]$ is the density of the j'th rectangle at k-1,
- $g$ $[m/pixel]$ is the ground sampling distance in meter per pixel.

Given $\rho_{j,k-1}$, the estimated flow at the rectangle can be calculated using the equations 3 and 4. It should be noted that $w$ corresponds to the length of the open border in meters.

### 3.5 Determining the Flow of Direction at an Open Border

It is further important to know whether pedestrians are either leaving or entering the scene. To estimate the flow of direction, Gunnar Farneback's algorithm is used. For each open border, the algorithm is given an enlarged rectangular cutout of the previous and current frame (not the density map!). The algorithm estimates the movement of the pixels in the x- and y-direction within the enlarged rectangle [9]. Although the algorithm calculates the movement in both directions, only one direction is relevant. For open borders located on the left or right of the image, the x-direction is of interest, whereas for open borders located at the top or bottom of the image the movement in the y-direction is of interest. By adding up the magnitudes by which the pixels are estimated to move along the relevant direction, you can obtain the magnitude of the pixels' overall movement along the direction. This value is denoted by $m_{j,k-1}$ $[pixel]$ in the following. Whereas negative values of $m_{j,k-1}$ measured on the left side or at the bottom of an image indicate an outflow, they indicate an inflow when they are measured at the top or on the right side of an image.
If you would use a rectangle with the same dimensions as the rectangle used to determine the density, you would probably run into the problem that the algorithm would detect no movement when all pedestrians in the rectangle were to leave the rectangle. Therefore, an enlarged rectangle is used that holds pedestrians that are unable to leave the rectangle between two consecutive frames. One side of the enlarged rectangle is the open border itself. The other side is three times the step size.
Given this information, one can calculate the maximum magnitude the algorithm can detect. Let $w_j$ be the length of the open border in pixel, then the maximum magnitude can be calculated as follows:

$$m_{j,max} = 2 \cdot \Delta s_{pixel}^2 \cdot w_j \cdot \frac{1}{pixel^2} \; [pixel] \tag{8}$$

Whereas $Q$ is an estimate of the pedestrian's flow obtained by the fundamental traffic flow model, the ratio of $m_{j,k-1}/m_{j,max}$ is an actual measurement. To get an estimate of the pedestrian change, both values are combined:

$$\Delta c_{j,k-1} = Q(\rho_{j,k-1}) \cdot \Delta t \cdot \frac{|m_{j,k-1}|}{m_{j,max}} \; [pedestrians] \tag{9}$$

Depending on whether $m_{j,k-1}$ indicates an outflow, $\Delta c_{j,k-1}$ must be further multiplied by $-1$:

$$\Delta c_{j,k-1} = \begin{cases} +\Delta c_{j,k-1} & inflow \\ -\Delta c_{j,k-1} & outflow \end{cases} \tag{10}$$

To get the total change of pedestrians that is expected between the consecutive frames, one must add up all $\Delta c$:

$$\Delta c_{k-1} = \sum_{j=1}^{k=n} \Delta c_{j,k-1} \tag{11}$$

Adding this result to the samples of the posterior distribution at $k-1$ returns the samples of the priori distribution at $k$. Let $\bar{c}_{k-1}^{i}$ be the i'th sample from the posterior distribution at $k-1$, then the priori estimate of the i'th sample can be calculated as follows:

$$\bar{c}_{k|k-1}^{i} = \bar{c}_{k-1}^{i} + \Delta c_{k-1} \tag{12}$$

## 4 Experiments

To train and test the crowd counting algorithms and filters, different data sets are used. Table 1 shows the fundamental differences of the data sets. Tests in the aerial domain are conducted as follows: Firstly, the crowd counters are pre-trained using the DLR-ACD data set [4] to overcome the lack of video data in that domain. Then, the crowd counting algorithms are fine-tuned on the VisDrone-CC2020 data set [10]. Subsequently, tests on the VisDrone-CC2020 and AgoraSet [11] are conducted.
Finally, tests on the WorldExpo'10 data set [12] are conducted. Due to the different domain, the crowd counting algorithms must be trained from scratch prior to testing.

**Table 1.** Comparison of the data sets. N is the number of annotated frames; FPS is the number of annotated frames per second; GSD is the ground sampling distance in cm/pixel; $A_R$ is the average resolution and $I_P$ is the interval in which the annotated crowd counts lay. The GSDs of the AgoraSet and VisDrone-CC2020 are estimated values. All specified values of the AgoraSet refer to the the frames 418-1936 of its first sequence

| | **N** | **FPS** | **Aerial** | **Artificial** | **GSD** | $\mathbf{A_R}$ | $\mathbf{I_P}$ |
|---|---|---|---|---|---|---|---|
| DLR-ACD | 33 | image | true | false | 4.5 - 15 | 3619×5226 | 285-24368 |
| VisDrone-CC2020 | 3360 | 1 | true | false | 0.118 - 0.706 | 1920×1080 | 25-421 |
| AgoraSet | 1519 | 25 | true | true | 4.5 | 640×480 | 1-180 |
| WorldExpo'10 | 3980 | 1/30 | false | false | perspective | 576×720 | 1-253 |

To determine the performance of the filters, the Mean Absolute Error of the estimated crowd counts and the Mean Absolute Error of the estimated crowd counts' slope is calculated. Let $c$ be the crowd count and $\bar{c}$ be the estimated crowd count, then the MAE of the estimated crowd count and its slope can be calculated as follows:

$$MAE_{crowd\,count} = \frac{\sum_{k=1}^{k=n} |c_k - \bar{c}_k|}{n} \tag{13}$$

$$MAE_{slope} = \frac{\sum_{k=2}^{k=n} |c_k - c_{k-1} - (\bar{c}_k - \bar{c}_{k-1})|}{n-1} \tag{14}$$

Both MAEs indicate better performance for values closer to zero and worse performance for higher values. They differ in that the $MAE_{slope}$ shows how good the filters smooth the data and the $MAE_{crowd\,count}$ shows how close the counts are to the actual crowd count. Since the focus of this paper lays on smoothing the counts, it is the overriding goal to reduce the $MAE_{slope}$ without increasing the $MAE_{crowd\,count}$.

## 4.1 VisDrone-CC2020

Tests on the VisDrone data set are conducted to test the behavior of the filters on a data set from the aerial domain the crowd counters are fine-tuned on.
Table 2 shows the results. In general, the filters smooth the temporal courses of the estimated crowd counts. This is shown by a decrease in the $\text{MAE}_{\text{slope}}$. The values of the raw crowd counters are reduced by 56% to 65%.
The Kalman filter applied to the whole frame smooths the data the most. This contradicts the initial assumption that the Kalman filter works better on smaller grids. The problem here is that the pixel values of the estimated density maps are not consistently positive. Therefore, if the density map is divided into smaller grids, some of the grids contain negative crowd counts. If one inputs these negative values into a Kalman filter, the filter outputs arbitrarily high or low numbers over time. To solve this problem, the count is set to zero, the process noise variance is set to 1, and the measurement noise variance is set to 1000 when a negative crowd count occurs.
Although this approach makes the grid-based Kalman filter work, it does not enable the filter to develop its full potential. Yet, it comes with a more than welcome side effect. By setting the negative crowd counts of single grids to zero, the grid-based Kalman filter returns the best results for the $\text{MAE}_{\text{crowd count}}$.
The displayed results of the grid-based Kalman filter are obtained using 1x1 meter grids. Tests with 3x3 and 10x10 meter grids were conducted. Yet, it turned out that on the VisDrone as well as on the other data sets, 1x1 meter grids return the best results.

**Table 2.** Results on the VisDrone-CC2020 data set. The $MAE_{slope}$ is reduced by 56% to 65%. All filters successfully smooth the estimated crowd counts

| | $\textbf{MAE}_{\textbf{crowd count}}$ | | $\textbf{MAE}_{\textbf{slope}}$ | |
|---|---|---|---|---|
| | CSRNet | MRCNet | CSRNet | MRCNet |
| Unfiltered | 29.30 | 21.08 | 4.23 | 3.98 |
| Kalman | 30.89 (+5%) | 18.31 (-13%) | **1.47 (-65%)** | **1.44 (-64%)** |
| $\text{Kalman}_{\text{grid}}$ | **25.78 (-12%)** | **16.52 (-22%)** | 1.60 (-62%) | 1.75 (-56%) |
| Particle | 28.63 (-2%) | 17.00 (-19%) | 1.59 (-62%) | 1.62 (-59%) |

## 4.2 AgoraSet

In order to see how the filters work on data with more realistic frame rates (25 fps), tests on the AgoraSet are conducted. Although its frame rate covers a more realistic use case, its data is artificial. To the best of our knowledge, non-artificial alternatives are not publicly available due to the high effort that comes with annotating such data sets.
When speaking of the AgoraSet only the first sequence of the AgoraSet is meant. This is because, all sequences cover a pretty similar scenario from a crowd counting perspective. The backgrounds are monotonous and do not significantly differ. In addition to that, pretty much all of the sequences' temporal courses of the crowd count resemble a parabola with a downward opening.
The tests were directly conducted after the tests on the VisDrone-CC2020 data set without fine-tuning the crowd counters on the AgoraSet or reconfiguring the Kalman filter's parameters. Nevertheless, the filters are able to smooth the temporal courses of the crowd count even more. The $MAE_{slope}$ is reduced by 68% to 88%. Again, the Kalman filter applied to the whole image smooths the data the best. Yet, the particle and grid-based Kalman filter return good results as well. If only the $MAE_{crowd\ count}$ is considered, the particle filter returns the best results.
The results show that the filters smooth the data even better on data sets with higher frame rates. In addition to that, the tests indicate that the initialization of the Kalman filter's parameters on a previous data set does not impair its performance on another data set.

**Table 3.** Results on the first sequence of the AgoraSet. Although the crowd counting algorithms and filters have not seen data from the AgoraSet before, the filters are able to smooth the estimated crowd counts even more

| | $\mathbf{MAE_{crowd\ count}}$ | | $\mathbf{MAE_{slope}}$ | |
|---|---|---|---|---|
| | CSRNet | MRCNet | CSRNet | MRCNet |
| Unfiltered | 26.06 | 43.72 | 1.5 | 2.01 |
| Kalman | 27.55 (+6%) | 44.45 (+2%) | **0.27 (-82%)** | **0.25 (-88%)** |
| $Kalman_{grid}$ | 25.49 (-2%) | 43.98 (+1%) | 0.45 (-70%) | 0.65 (-68%) |
| Particle | **25.40 (-3%)** | **43.84 (±0%)** | 0.39 (-74%) | 0.42 (-79%) |

### 4.3 WorldExpo'10

Finally, tests are conducted on a perspective data set that has a significantly lower frame rate than the previous ones. Since the WorldExpo'10 data set does not contain aerial images, the particle filter as conceptualised in this paper cannot be applied. The grid-based Kalman filter also runs into the problem that its hard to divide a perspective image into same sized grids. To handle this problem, the face length of a person in the front and back of a random frame is measured. Assuming that the average face length of a person is 23 centimeters, one can calculate two GSDs - one for the front and one for the back of the image. The average of the two GSDs is used to determine the grids. It should be noted that this approach is only a hot-fix to make the filter work and that grids in the back still cover larger areas than grids in the front.
Table 4 shows the results. It can be seen that the $MAE_{slope}$ does not significantly improve. However, this is much more due to the low frame rate and not the perspective of the data set. The time between two frames of the WorldExpo'10 data set is 30 seconds. Due to the large time interval between the frames, it can be said that consecutive frames do not hold any temporal information that could be incorporated by the filters. To sum up, the results stress out the importance of the frame rate when applying the filters. If the time interval between consecutive frames becomes too large, the filters do not improve the estimates of the crowd counters. Yet, since a frame rate of 1/30 fps does not capture a realistic scenario, the results of the tests on the WorldExpo'10 do not contradict with the applicability of the filters in a real-life situation.

**Table 4.** Results on the WorldExpo'10 data set. Due to the low frame rate, adjacent frames do not share temporal information that can be incorporated by the filters

| | $\mathbf{MAE_{crowd\ count}}$ | | $\mathbf{MAE_{slope}}$ | |
|---|---|---|---|---|
| | CSRNet | MRCNet | CSRNet | MRCNet |
| Unfiltered | 17.15 | 18.12 | 8.8 | 8.6 |
| Kalman | 19.74 (+15%) | 18.14 (±0%) | 8.2 (-7%) | 8.4 (-2%) |
| $\text{Kalman}_{\text{grid}}$ | 19.73 (+15%) | 17.95 (-1%) | 9.54 (+8%) | 8.9 (+3%) |

## 5 Conclusions

The paper shows that the oscillating counts estimated by current crowd counters for video data can be smoothed using Bayesian filters. As a measurement of the false oscillation the $\text{MAE}_{\text{slope}}$ is introduced. Provided that the frame rate of a data set is large enough that consecutive frames share temporal information, all of the three filters are able to smooth the temporal course of the crowd count without significantly increasing the $\text{MAE}_{\text{crowd count}}$. Future work may further address the maturation of the concepts. The development of Bayesian filters that smooth the estimated crowd counts is still in its infancy. A variety of traffic flow models exist that can be used to develop new concepts that may smooth the crowd counts even more.

# Potentials of Semantic Image Segmentation Using Visual Attention Networks for People with Dementia

Liane Meßmer and Christoph Reich

Hochschule Furtwangen – University of Applied Sciences
l.messmer@hs-furtwangen.de
christoph.reich@hs-furtwangen.de

**Abstract.** Due to the increasing number of dementia patients, it is time to include the care sector in digitization as well. Digital media, for example, can be used on tablets in memory care and have considerable potential for reminiscence therapy for people with dementia. The time consuming assembly of digital media content has to be automated for the caretakers.
This work analyzes the potentials of semantic image segmentation with Visual Attention Networks for reminiscence therapy sessions. These approaches enable the selection of digital images to satisfy the patients individual experience and biographically. A detailed comparison of various Visual Attention Networks evaluated by the BLEU score is shown. The most promising networks for semantic image segmentation are VGG16 and VGG19.

## 1 Introduction

The German Federal Statistical Office examines the population development until the year 2060. They conclude that the percentage of people older than 67 years will increase from 19% in 2018 to 27% in 2060 [1]. As people live longer than ever before and therefore, the number of dementia patients is also increasing. Today, more than 44 Million people worldwide are living with dementia, 1.5 Million of them live in Germany [2]. Non-pharmacological methods are effective in improving the lives of dementia patients. Reminiscence therapy belongs to the non-pharmacological techniques, used to address the activation process of people with Dementia (pwD) [3].

Nowadays, Reminiscence therapy sessions use digital support systems, which consists of digital content such as images, movies or music. These can be applied for example on mobile devices. Images, movies or music is used for life review and to evoke memories in patients [4], [5]. Thus, demand-oriented and technical solutions cause a valuable contribution to the care of pwD. Their potential is far from exhausted. Yet, a well known challenge is the identification of suitable content as well as the design and evaluation of high-quality reminiscence care services are very labor-intensive. Besides, this task places high demands on the caretakers qualifications.

Currently, the reminiscence session content must be identified and evaluated by caretakers, because the content of a reminiscence session should be suitable and individual for pwD. So, in practice, a very limited pool of standard content is often used.

To support the reminiscence therapies the caretaker would select images according to the following image characteristics such as objects, colors, shapes, number of objects or meaning according to life themes [6]. In particular, automated, individual and biography-related media selection improves the quality of the reminiscence session and reduces the workload of the caretakers by shortening the preparation time. However, this potential of automation relieves the caretaker in terms of reminiscence sessions. It also gives care giving relatives the opportunity to include memory-triggering content in their care.

The approach of this paper is semantic image segmentation, to extract the features of an image. Semantic segmentation is one of the high-level tasks that pave the way to full understanding of a scene [7]. The importance of full scene understanding as a core computer vision problem affirms by the fact, that an increasing number of applications using the knowledge derivation from images. Some of these applications include medical assistance systems or human-computer interactions. With the popularity of Deep Learning, many semantic segmentation problems are addressed with Deep Learning approaches. They far exceed other methods in terms of accuracy and efficiency.

The goal of this work is to analyzes the potentials of semantic image segmentation using different Convolutional Neural Networks (Inceptionv3, VGG16, VGG19 and ResNet101) in combination with a Recurrent Neural Network (LSTM, GRU) related to people with dementia, to generate automatic descriptions from images (also called image captioning).

This work consists of 8 chapters. Chapter 2 deals with the related work. Life themes identified to activate pwD are described chapter 3. In Chapter 4, the concepts of visual attention networks are presented and explained. The data used for training, as well as the training process, are described in Chapters 5 and 6. Finally, the results are presented in Chapter 7 and Chapter 8 describes the conclusion and future work.

## 2 Related Work

### 2.1 Reminiscence Aid Systems

The Computer Interactive Reminiscence and Conversation Aid (CIRCA) project, proposed by Astell et al. [8] was the first project that developed an application for digital reminiscence therapy to support people with dementia. Over the years, it was supplemented by different new technologies, like a specific interface for the interaction with the system [9] or a touch screen computer to enable an easier interaction with the system [10]. Today, CIRCA is an interactive multimedia application. The latest publication of the project "Computer Interactive Reminiscence and Conversation Aid groups - Delivering cognitive stimulation with technology" demonstrates the effectiveness of CIRCA for group interventions [11]. The growing process of the project shows the effectiveness of digital assistance systems in the area of reminiscence therapies. The difference with our work is that we do not want to use random content for a reminiscence session, but individual content that fits the biography of a patient. Therefore, our system should be able to describe images, that match the life themes of pwD, automatically.

Carós et al. [12] presented in their work "Automatic Reminiscence Therapy for Dementia" a solution approach to automate reminiscence therapy, which uses artificial intelligence based systems. Their system is called "Elisabot" and consists of a system that uses personal images of users and generates questions about their lives, using Visual Attention Networks (VATs) In our work we compare different architectures of Visual Attention Networks to find the potentials of these Networks and whether their generated image captions fit to pwDs needs. All encoder models are pretrained on the ImageNet dataset.

In the work “Image Captioning and Comparison of Different Encoders” by Pal et al. [13] a comparison is presented of different encoder implementations as they are used in a Visual Attention Network, for the automatic generation of image captions. As encoder, they compare different convolutional neural networks, these are Inception_v3, VGG16, VGG19 and InceptionResNetV2. The result is that the Inceptionv3 Encoder works best. Since the range of the BLEU score is lowest for the model. Similar to our work, they use BLEU score for result evaluation. We use the MS COCO dataset for training and instead of using the InceptionResNetV2 for model comparison, a ResNet-101 is used. Furthermore, this work additionally presents differences between Long-Short-Term-Memory (LSTM) and Gated Recurrent Unit (GRU) decoder .

### 2.2 Visual Attention Networks

The caption generation in this paper is based on the work “Show, Attend and Tell” proposed by Xu et al. [14]. They describe a mechanism that generates image captions based on Convolutional Neural Network (CNN) encoder and Recurrent Neural Network (RNN) decoder by using an attention layer in the network. The Encoder extracts specific features from an image and generates a set of feature vectors, which were referred to as annotation vectors in this context. The attention layer takes an annotation vector, to focus on a specific part of an image, because every vector is a representation corresponding to a part of an image. This allows the decoder to selectively focus on specific parts of an image by selecting a subset of all annotation vectors [15]. The results are automatic generated, textual image captions in a natural language.

*Encoder* Recent works have represented the successful deployment of Convolutional Encoder. Therefore, our work is focused on image caption generation using a CNN as Encoder [16], [17]. There are different CNN implementations for image feature extraction, which are compared in this work: VGG16, VGG19, ResNet-101 and Inceptionv3.

Simonyan et al. [18] proposed multiple Versions of VGG Networks in their work “Very Deep Convolutional Networks for Large-Scale Image Recognition”. They differ mainly in their depth, i.e. in their number of layers. As the name suggests, one network has 16 layers and the other 19.

In the work “Deep Residual Learning for Image Recognition”, the Residual Network (ResNet) Architecture is described from He et al. [19]. The special feature of a ResNet is that each layer in the network consists of several blocks. As the depth of the network increases, the number of operations in a block increases too, but not the total number of layers. Thus, ResNets solve the problem of vanishing gradients as neural networks become deeper.

The architecture of the Inceptionv3 network emerges from the GoogleNet architecture and was proposed from Szegdy et al. [20] in the work “Rethinking the Inception Architecture for Computer Vision”. The model is a combination of many ideas that have emerged in recent years. The network consists of several symmetric and asymmetric blocks, which can contain different types of layers. For example, convolutions, average pooling, max pooling, etc. In total, the network has 42 layers [21].

*Decoder* As decoder, a Recurrent Neural Network (RNN) is used. The first RNN was published in the work “Finding structure in time” from Jeffrey Elman [22]. This network is capable of reading the annotation vectors extracted by the CNN in previous steps. The important features of RNNs are the memory cells. A normal feed forward network has an input, hidden and output layer. The RNN loops the hidden layer to process sequential data. With this looping mechanism, the RNN allows flowing information from one step to the next step. During training a RNN Model is enrolled (each word acts as layer) and trained with backpropagation trough stochastic gradient descent [23]. RNNs often have a problem known as short term memory. As the number of words increases, so does the depth of the network. The more steps an RNN has to process, the greater the problem of retaining the information from the previous steps. This phenomenon occurs due to the backpropagation used for training the network.

There are two approaches to tackle the problems of RNNs caused by short-Term-Memory: Long-Short-Term-Memory (LSTM) and Gated-Recurrent Unit (GRU). The LSTM Network was proposed from Hochreiter et al. [24] in the work "Long Short-Term Memory" and the GRU Network was proposed from Cho et al. [25] in the work "Learning Phrase Representations using RNN Encoder-Decoder for Statistical Machine Translation". The underlaying structure of these networks is the same as the structure from an RNN extended by a mechanism that can learn long-term dependecies using "gates". LSTMs are using three gates to process the data while GRUs only use one gate for data selection [26].

## 3 People with Dementia Life Themes

The reminiscence sessions should include content that matches the biographical needs of a pwD. So that an activation in reminiscence sessions can take place [27]. Life themes represent a generalized categorization of life stages or events that play a role in a person's life. The goal of these particular subjects is to evoke memories in pwD, that are associated with a life theme. Life themes can be represented by pictures, videos, music or physical objects. This work focuses on picture retrieval. The following table 1 shows the life themes we identified and used for image feature extraction in this work. As the table shows, there are not only positive activation for pwD, but also negative/fearful issues, which should be avoided. If these negative loaded images are also labeled, they can be discarded before using them in a reminiscence session. Anxiety-producing image content should be screened out before sessions. This prevents a session from having a negative impact on a patient's well-being.

**Table 1.** Life themes used in reminiscence sessions for pwD

| Life Themes | | |
|---|---|---|
| Travel | Animals | Professions |
| Hobbies and Activities | Preferences and Habits | Exterior Appearance |
| Religion | Education | Nature |
| Childhood and Youth | Home | Tradition and Culture |
| Literature | Media | Theather |
| Garden | People | Fears and Disease |
| Food | | |

## 4 Visual Attention Networks

Automatic generation of image descriptions is a difficult task in the field of full scene understanding. The model must transform large disparate sets of data into a natural language. To address this problem, Visual Attention Networks (VATs) are used [14].

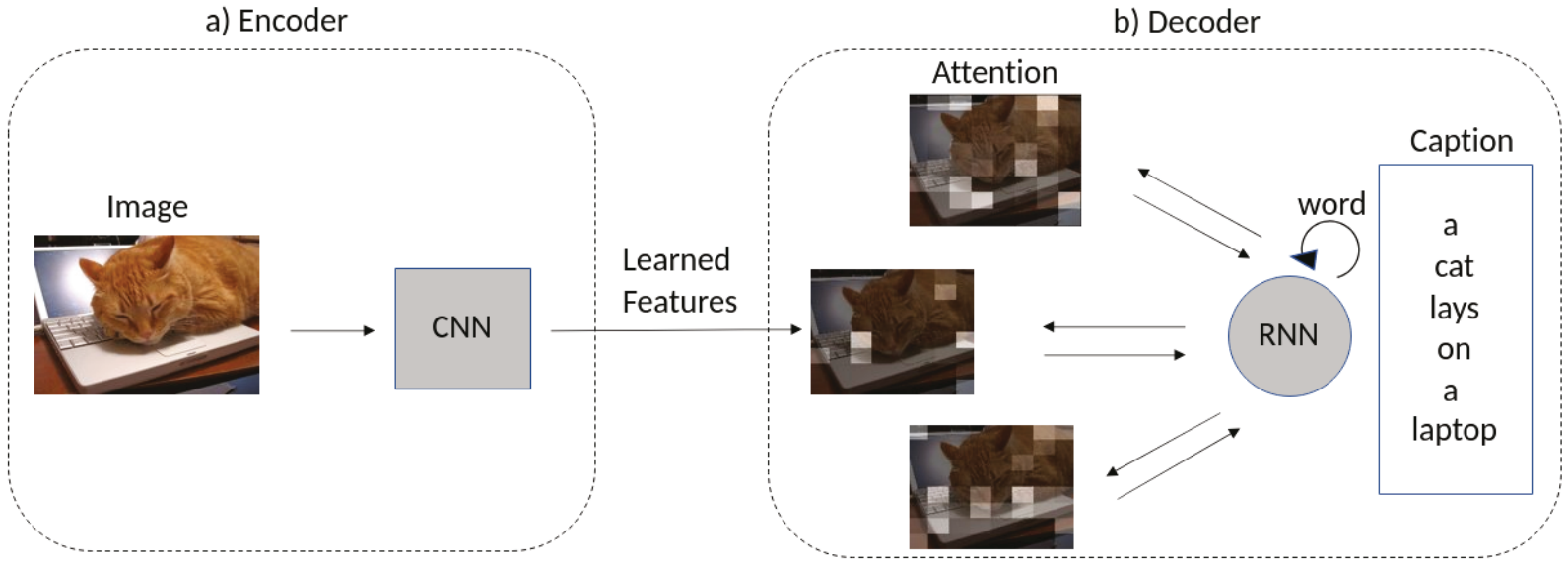

**Fig. 1.** Visual Attention Network Architecture

A VAT consists of two sub-models, as shown in Figure 1. a) Encoder: First, an image is given as input to the CNN to obtain vectorial representation of input images, on the last hidden layer of the network. b) Decoder: The feature vector is used from the RNN decoder as input to generate sentences that contain the objects and the relationships between them [14],[28]. The visual system of a human has the function to pay attention on different parts of an image instead of processing the scene as a whole [29]. Based on this human attention mechanism, an attention layer is integrated into a visual attention network. In this paper we use soft attention mechanisms for training [30]. The areas selected by attention mechanism are captured by the RNN for further processing. Based on this process suitable textual descriptions are generated.

## 5 Dataset Used for Training and Evaluation

The dataset used for training should contain everyday objects, which match the life themes from section 3. There are several datasets that contain labeled content for image captioning, for example MS COCO, Pascal VOC or Flickr30k. Since the MS COCO dataset [31] has the most objects in common with the life themes described in section 1, this dataset was used for training in this work. The matching life themes are for example "Animal", "Person" and "Food" . Each image from the MS COCO Caption dataset is described (labeled) with 5 different sentences. This work primarily targets the description of dogs and cats images, so these categories are filtered from the dataset. Dog images fall into two categories in the context of Reminiscence Therapy: Dog images that activate positive memories in the pwD and dog images that might trigger negative memories. For example dangerous looking dogs, aggressive dogs or snarling dogs. In total, 4298 images of the category "cat" and 4562 images from the categorie "dog", with a total of 43200 image descriptions, are filtered from the dataset.

Since the MS COCO dataset only contains images with friendly looking dogs, the dataset is extended with the category "Angry Dogs" and filled with our own image content. Each image is described with 5 description sentences, similar to the the caption style of MS COCO dataset. The number of images in this category amounts to 360 training images with 1800 descriptions. In total, our dataset consists of 9000 images, with 45000 image descriptions. For training, we use a random 80/20 split on the dataset, to split it into train and validation set.

## 6 Training

This work compares different networks which are used in a Visual Attention Network Architecture. We use ResNet-101, VGG16, VGG19 and Inceptionv3 as encoder and Gated Recurrent Unit (GRU) or Long-Short-Term Memory (LSTM) as decoder, to compare them with each other. The CNNs are all pretrained on the ImageNet [32] dataset. This dataset contains many objects related to the Life themes of people with dementia. Additional we use MS COCO Dataset for training the RNN, as described above.

For training, we use a fixed-length image caption of 9 words per sentence, because the network performes poor on long input or output sequences. This is caused by rolling out the RNN, where each word represents a hidden layer. The longer a caption is, the slower the network is trained. In addition, for our use case of image descriptions for the use in reminiscence therapy, captions of length 9 are sufficient to make sense of the content and at the same time short enough to not impair the performance of the model.

In total, we get 6660 different words as vocabulary, from which we use all words that occur more than three times in the vocabulary. This results in 2206 words. Unknown words, are provided with the token <unk>. We trained the Networks with a batch size of 8 and 100 Epochs.

## 7 Evaluation and Comparison of the VGG16, VGG19, ResNet-101 and Inceptionv3 Model

For evaluation and comparison, a dataset, containing 10 images for each class in the training dataset (cat, dog, angry dog), was created. These images are with 5 captions per image for result evaluation with BLEU score. In the area of image caption generation it's hard to evaluate resulting captions formally. We only have five reference sentences for an image, but there are much more image descriptions that match the image content. Thus, we decided to evaluate the results formally with BLEU score and verify the results with human evaluation. BLEU is a method for automatic evaluation of machine translation [33], it's quickly, inexpensive, and language-independent. The metric correlates highly with human evaluation, because it measures the closeness of the machine translation to human reference translations by taking translation length, word choice and word order into consideration. For our evaluation we use 2-gram BLEU score. This approach does not check the specific order of all words in the generated caption, but only the adjacent ones and compares them with the reference descriptions.

The following Table 2 shows the average calculated BLEU score, using different encoder-decoder combinations, for (a) Inceptionv3, (b) ResNet101, (c) VGG16 and (d) VGG19. Every category contains 10 images for inference. The generated captions of an RNN are not identical, since such a model has no fixed hidden layer size. They always depend on the generated caption length. Therefore, each model was trained five times (GRU-1, GRU-2,...,GRU5 or LSTM-1, LSTM-2,...,LSTM-5). At the end of the tables, the respective average value (ø) is shown, which is used for result comparison. The scores in the tables show that ResNet101 behaves the worst. The BLEU scores are below the scores of the other networks, regardless of which model combination was used. Inceptionv3 has the best average BLEU for the category “Cat”, combined with a GRU model. The best results are produced by VGG16 and VGG19 models, with VGG19 performing best in combination with a GRU decoder and the VGG16 in combination with an LSTM decoder. The Inceptionv3 model outperforms the VGG16 model in the category “Cat” with a BLEU score of 0.02322 compared to 0.2005. The difference is so small that for all classes in general, the VGG16 model perfomes better.

All models provide the worst results for the category “Dog” and the best for the category “Angry Dog”. This is because the images in the angry dog category were self-labeled, specifically tailored to our problem set. The other two categories are labeled with general captions from the COCO dataset. In general, the BLEU scores are stable for each model in each class; no outliers are apparent.

**Table 2.** Calculated BLEU scores

(a) BLEU score for Inceptionv3

| Decoder | Cat | Dog | Angry Dog |
|---|---|---|---|
| GRU-1 | 0.2377 | 0.1018 | 0.6600 |
| GRU-2 | 0.2629 | 0.1619 | 0.6423 |
| GRU-3 | 0.2793 | 0.1650 | 0.6526 |
| GRU-4 | 0.1701 | 0.1846 | 0.5806 |
| GRU-5 | 0.2114 | 0.1467 | 0.6682 |
| ø | **0.2322** | 0.1250 | 0.6407 |
| LSTM-1 | 0.2511 | 0.1126 | 0.5386 |
| LSTM-2 | 0.1620 | 0.0702 | 0.5750 |
| LSTM-3 | 0.2336 | 0.0967 | 0.5910 |
| LSTM-4 | 0.1718 | 0.0845 | 0.6724 |
| LSTM-5 | 0.1616 | 0.0759 | 0.5704 |
| ø | 0.1960 | 0.0879 | 0.5894 |

(b) BLEU score for ResNet101

| Decoder | Cat | Dog | Angry Dog |
|---|---|---|---|
| GRU-1 | 0.0875 | 0.0500 | 0.3789 |
| GRU-2 | 0.1889 | 0.0500 | 0.2951 |
| GRU-3 | 0.0634 | 0.0375 | 0.3361 |
| GRU-4 | 0.0375 | 0.0625 | 0.3020 |
| GRU-5 | 0.1000 | 0.0625 | 0.3896 |
| ø | 0.0954 | 0.0525 | 0.3403 |
| LSTM-1 | 0.1697 | 0.0611 | 0.2402 |
| LSTM-2 | 0.0960 | 0.0500 | 0.2339 |
| LSTM-3 | 0.1625 | 0.1000 | 0.2000 |
| LSTM-4 | 0.1000 | 0.0723 | 0.2978 |
| LSTM-5 | 0.0986 | 0.0611 | 0.2216 |
| ø | 0.1253 | 0.0689 | 0.2387 |

(c) BLEU score for VGG16

| Decoder | Cat | Dog | Angry Dog |
|---|---|---|---|
| GRU-1 | 0.2154 | 0.0666 | 0.5346 |
| GRU-2 | 0.1785 | 0.0767 | 0.5974 |
| GRU-3 | 0.2236 | 0.1142 | 0.6947 |
| GRU-4 | 0.2339 | 0.1077 | 0.5626 |
| GRU-5 | 0.2378 | 0.0583 | 0.6252 |
| ø | 0.2178 | 0.0847 | 0.6029 |
| LSTM-1 | 0.1918 | 0.0875 | 0.6417 |
| LSTM-2 | 0.2430 | 0.0916 | 0.6165 |
| LSTM-3 | 0.2085 | 0.1139 | 0.6167 |
| LSTM-4 | 0.1805 | 0.1111 | 0.6032 |
| LSTM-5 | 0.1791 | 0.0800 | 0.6990 |
| ø | 0.2005 | **0.0968** | **0.6354** |

(d) BLEU score for VGG19

| Decoder | Cat | Dog | Angry Dog |
|---|---|---|---|
| GRU-1 | 0.1932 | 0.0951 | 0.6087 |
| GRU-2 | 0.1883 | 0.1468 | 0.6545 |
| GRU-3 | 0.2231 | 0.1571 | 0.6934 |
| GRU-4 | 0.2261 | 0.1303 | 0.6989 |
| GRU-5 | 0.1723 | 0.1105 | 0.7291 |
| ø | 0.2006 | **0.1279** | **0.6769** |
| LSTM-1 | 0.1684 | 0.0382 | 0.5244 |
| LSTM-2 | 0.2682 | 0.0454 | 0.4502 |
| LSTM-3 | 0.2220 | 0.0737 | 0.6332 |
| LSTM-4 | 0.2316 | 0.0722 | 0.5741 |
| LSTM-5 | 0.1986 | 0.0737 | 0.6888 |
| ø | **0.2177** | 0.0606 | 0.5741 |

The results, calculated by BLEU score are verified manually. The best and worst BLEU scores for eachmodel are taken and the corresponding generated captions were checked. For each model, the best and worst results for all categories are shown in Figure 2. Figure (a) shows the generated captions for cats, (b) shows the captions for dogs and (c) the captions for angry dogs.

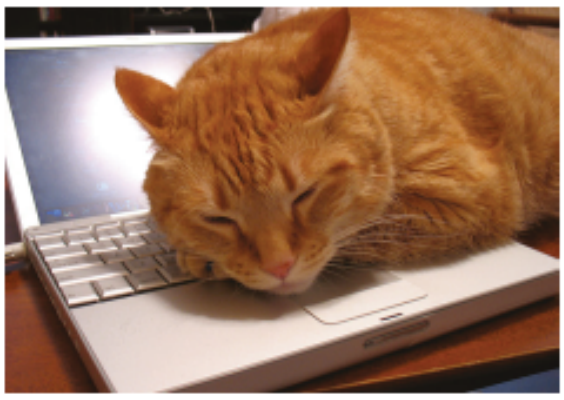

**Best results:**
**Inceptionv3, GRU-3:** a cat is curled up asleep lying on a
**ResNet101, GRU-2:** a cat is using a laptop keyboard
**VGG16, LSTM-2:** cat laying on a laptop computer
**VGG19, GRU-4:** a orange cat sits on a laptop

**Worst results:**
**Inceptionv3, GRU-4:** an orange cat sits resting its head on the
**ResNet101, LSTM-2:** a cat sleeping half lake in an open suitcase
**VGG16, LSTM-5:** an orange cat resting it's camera
**VGG19, GRU-5:** a cat sleeping half on

(a) Resulting Captions "Cat" [31]

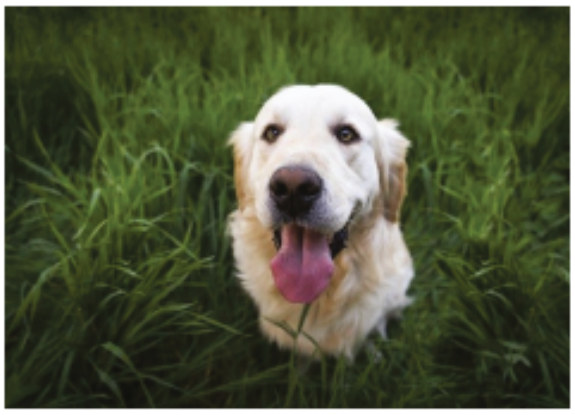

**Best results:**
**Inceptionv3, LSTM-1:** a white dog wearing a green and holds a
**ResNet101, GRU-5:** a dog has a red collar is standing near
**VGG16, LSTM-3:** a black and white dog
**VGG19, LSTM-5:** a dog laying in a grassy field looking grass

**Worst results:**
**Inceptionv3, GRU-1:** 0.1018, a cute hair that its mouth resting while wearing
**ResNet101, LSTM-2:** 0.0500, a dog <unk> in a pink flower and green
**VGG16, GRU-5:** 0.0583, a big panting dog
**VGG19, GRU-1:** 0.6087, the dog is looking to above his mouth

(b) Resulting Captions "Dog" [34]

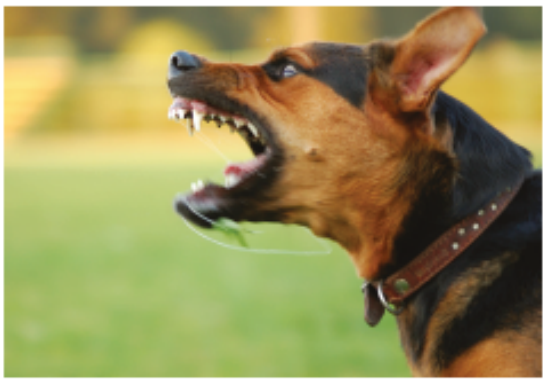

**Best results:**
**Inceptionv3, GRU-5:** a black and brown dog looks angry while baring
**ResNet101, GRU-5:** a black and brown dog baring his theeth on
**VGG16, LSTM-1:** an angry looking black and brown dog shows his
**VGG19, GRU-5:** an angry looking black and brown dog shows his

**Worst results:**
**Inceptionv3, GRU-4:** a black and brown dog shows his theeth on
**ResNet101, GRU-2:** a cat is greeting each other in a chair
**VGG16, LSTM-4:** a brown dog baring his theeth on green grass
**VGG19, GRU-1:** a black and brown dog looks angry while baring

(c) Resulting Captions "Angry Dog" [35]

**Fig. 2.** Resulting Captions

The resulting captions of the category dogs are as good as the BLEU score describes them. Only one bad caption was produced by the ResNet101, GRU-2. The other results coincide with the BLEU score, ResNet101 predictions are generally worse than the others and VGG16, VGG19 generate the best image captions. By comparing the BLEU scores with human evaluation, we came to the conclusion: the better the BLEU score, the better the caption.

## 8 Conclusion

This work reveals that the use of Visual Attention Networks in the context of reminiscence sessions for dementia patients has significant potential. The result of the comparison from different encoder-decoder combinations is that the use of VGG16-LSTM and VGG19-GRU Models generating promising results. This approach allows activation sessions to be simpler, faster and tailored for a patient's needs. Thereby, higher quality and quantity of reminiscence sessions is created. At the same time, the life of a dementia patient is positively influenced.

In the future, it's important to extend the dataset with more categories, to match more life themes from pwD. In addition, it's possible to extend the system with the ability to describe not only images automatically. There are also music and videos that match the life themes of pwD.